Politechnika Rzeszowska

Wydział Elektrotechniki i Informatyki

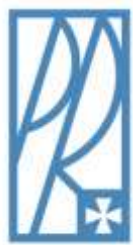

ROZPRAWA DOKTORSKA

# Komputerowe algorytmy ekstrakcji i śledzenia obiektów w czasie rzeczywistym

## mgr inż. Bogusław Rymut

Promotor: dr hab. inż. Bogdan Kwolek, prof. n. AGH

Rzeszów 2016

STRESZCZENIE PRACY DOKTORSKIEJ

## KOMPUTEROWE ALGORYTMY EKSTRAKCJI I ŚLEDZENIA OBIEKTÓW W CZASIE RZECZYWISTYM

Autor: **mgr inż. Bogusław Rymut**

Promotor: dr hab. inż. Bogdan Kwolek, prof. n. AGH

Słowa kluczowe: śledzenie ruchu 3d, obliczenia równoległe, wizja komputerowa

Praca dotyczy bezmarkerowych systemów do śledzenia ruchu 3D w oparciu o obrazy pobierane z kilku skalibrowanych i zsynchronizowanych kamer. Estymacja pozy jest realizowana w oparciu o model 3D, który jest transformowany do przestrzeni obrazów, a następnie renderowany. Śledzenie ruchu 3D całej postaci realizowano w czasie rzeczywistym w oparciu o algorytmy optymalizacji dynamicznej oraz filtracji bayesowskiej. W funkcji celu algorytmu optymalizacyjnego w oparciu o rój cząsteczek oraz model obserwacji filtru cząsteczkowego wykorzystywano dopasowanie między wyrenderowanym modelem 3D w zadanej pozie oraz cechami wydzielonymi na obrazach. Mając na względzie to, że główna część nakładów obliczeniowych związana jest z renderingiem modeli 3D w hipotetycznych pozach, a także wyznaczaniem wartości funkcji celu, w pracy opracowano efektywne metody renderingu modeli 3D w czasie rzeczywistym ze wsparciem sprzętowym OpenGL oraz równoległe metody wyznaczania funkcji celu na GPU. Opracowano i przebadano kilka wariantów funkcji celu do śledzenia ruchu 3D w czasie rzeczywistym w oparciu o rendering programowy i sprzętowy modeli 3D. Zaproponowano efektywne rozwiązania umożliwiające rendering modelu w OpenGL, a także wymianę danych między CUDA i OpenGL. Zaproponowano i przebadano konfigurowalne potoki graficzne OpenGL do renderingu w czasie rzeczywistym znaczącej liczby modeli 3D w zadanej pozie. Opracowano i zaimplementowano metody dekompozycji algorytmu optymalizacji w oparciu o rój cząsteczek oraz równoległego wyznaczania funkcji celu pod kątem efektywnego wykorzystania dostępnych zasobów sprzętowych oraz osiągania najlepszych dokładności i częstotliwości śledzenia. Dzięki opracowanym rozwiązaniom zrealizowano śledzenie ruchu 3D całej postaci w czasie rzeczywistym. Opracowano rozwiązania umożliwiające dobranie liczby cząsteczek i liczby iteracji w algorytmie PSO, od których zależy liczba przetwarzanych klatek na sekundę, która z kolei determinuje zmiany pozy pomiędzy kolejnymi klatkami. Dzięki wspomnianym rozwiązaniom możliwe jest osiągnięcie najmniejszego błędu śledzenia ruchu 3D w czasie rzeczywistym.

DOCTORAL THESIS ABSTRACT

## COMPUTER METHODS FOR 3D MOTION TRACKING IN REAL-TIME

Author: **MSc Bogusław Rymut**

Supervisor: PhD Bogdan Kwolek, Prof. AGH

This thesis is devoted to marker-less 3D human motion tracking in calibrated and synchronized multicamera systems. Pose estimation is based on a 3D model, which is transformed into the image plane and then rendered. Owing to elaborated techniques the tracking of the full body has been achieved in real-time via dynamic optimization or dynamic Bayesian filtering. The objective function of a particle swarm optimization algorithm and the observation model of a particle filter are based on matching between the rendered 3D models in the required poses and image features representing the extracted person. In such an approach the main part of the computational overload is associated with the rendering of 3D models in hypothetical poses as well as determination of value of objective function. Effective methods for rendering of 3D models in real-time with support of OpenGL as well as parallel methods for determining the objective function on the GPU were developed. Several variants of objective function using both software and hardware rendering were proposed and evaluated on real data. Methods for effective rendering of 3D models in OpenGL, as well as data mapping between OpenGL and CUDA were developed and evaluated. Programmable streams in OpenGL were designed and configured to achieve real-time rendering of considerable number of 3D models in desired poses. Methods for parallel execution of particle swarm optimization as well as objective function calculation were developed to achieve effective utilization of hardware resources and the best possible tracking accuracies and frequencies. The elaborated solutions permit 3D tracking of full body motion in real-time. Certain solutions to enable selection of the number of particles and the number of iterations in the PSO algorithm, which determine the number of processed frames per second, and which in turn determines the change in the pose between consecutive frames were investigated and proposed. They make it possible to achieve the lowest errors in real-time 3D motion tracking.

# Spis treści

I

# Wstęp

W ostatnich latach obserwuje się duży wzrost zainteresowania zagadnieniami śledzenia ruchu 3D osób. Wspomniany wzrost wynika z zapotrzebowania na nowe interfejsy dla interaktywnych gier komputerowych, a także interfejsy poszerzonej i wirtualnej rzeczywistości. Wśród innych potencjalnych zastosowań wymienić można interfejsy wspomagające interakcję człowiek-maszyna, systemy: do rehabilitacji, wspomagania treningu oraz rozpoznawania i analizy zachowań ludzi. Jednym z ważniejszych zastosowań systemów do analizy ruchu 3D są rozwiązania śledzące ruch na potrzeby realistycznej syntezy i animacji ruchu w grach komputerowych. Do szczególnie istotnych badań należą badania w kontekście syntezy ruchu dla robotów humanoidalnych. Wprowadzenie sensora Kinect było istotnym przełomem technologicznym, który spowodował dalszy wzrost zainteresowania systemami śledzącymi ruch 3D. Wśród nich znajdują się wielokamerowe systemy do śledzenia ruchu, które dzięki obserwacji osoby przez kilka kamer znacznie lepiej potrafią radzić sobie z przesłonięciami. Inną ważną zaletą omawianych rozwiązań jest to, że śledzenie ruchu może być realizowane z większych odległości i na większej przestrzeni. Konieczność badań nad technologiami bezmarkerowego śledzenia ruchu 3D wynika z zapotrzebowania na tanie rozwiązania, które stanowiłyby alternatywę dla drogich systemów mocap.

Potrzeba usprawnienia systemów bezmarkerowego śledzenia ruchu zwróciła uwagę zespołów badawczych z całego świata. Proponowane w literaturze podejścia można podzielić na rozwiązania dyskryminacyjne, w których modelowana jest zależność między obserwacjami z kamery i konfiguracją ciała człowieka oraz metody generacyjne, które oparte są na znajdowaniu najlepszego dopasowania między modelem 3D zrzutowanym do przestrzeni obrazów i cechami wydzielonymi na obrazach. Zaletą systemów opartych na modelu 3D jest większa precyzja śledzenia, w szczególności w zadaniach śledzenia ruchu na podstawie sekwencji obrazów.

Niniejsza praca dotyczy bezmarkerowych systemów do śledzenia ruchu 3D w oparciu o obrazy pobierane z kilku skalibrowanych i zsynchronizowanych kamer. W porównaniu do systemów markerowych rozwiązania bezmarkerowe nie wymagają stosowania markerów. Śledzenie ruchu 3D bez wykorzystania markerów jest jednym z najbardziej wymagających obliczeniowo problemów w komputerowym przetwarzaniu obrazów z powodu różnorodności w wyglądzie śledzonej postaci, a w szczególności konieczności eksploracji znaczącej przestrzeni poszukiwań. Istotnym problemem są również przesłonięcia poszczególnych części ciała, a także znaczne narzuty obliczeniowe. W omawianych systemach śledzenia ruchu w oparciu o model 3D główna część nakładów obliczeniowych dotyczy renderingu modelu 3D dla rozpatrywanego zbioru hipotez. Jest to operacja, którą można realizować równolegle, w tym także ze wsparciem renderingu sprzętowego.

Powyższe kwestie lokalizują problem śledzenia ruchu 3D człowieka na styku trzech dziedzin informatyki – widzenia komputerowego, uczenia maszynowego i grafiki komputerowej. Zagadnienia ekstrakcji cech obrazu reprezentujących osobę wymagają znajomości technik kalibracji kamery, modelowania i ekstrakcji obiektów pierwszego planu, detekcji krawędzi i wyznaczania map odległości od krawędzi. Estymacja pozy 3D na podstawie dwuwymiarowych obrazów jest problemem źle uwarunkowanym (ang. *ill-posed*) i wymaga użycia technik wnioskowania probabilistycznego, metod predykcji dla wielowymiarowych rozkładów lub użycia technik optymalizacji dynamicznej. Modelowanie wielowymiarowej przestrzeni konfiguracji wymaga zastosowania technik klasteryzacji oraz metod wspierających przeszukiwanie przestrzeni potencjalnych konfiguracji ciała człowieka. Z kolei śledzenie ruchu w czasie rzeczywistym w oparciu o metody generacyjne wymaga znajomości technik renderingu modelu 3D.

Mając na względzie zapotrzebowanie na systemy do bezmarkerowego śledzenia ruchu 3D na scenie o większych wymiarach, a w szczególności zapotrzebowanie na technologie do śledzenia ruchu w czasie rzeczywistym, cele niniejszej pracy sformułowano w następujący sposób:

- zaproponowanie efektywnych rozwiązań do śledzenia ruchu w oparciu o model 3D,
- przygotowanie i przebadanie rozwiązań, które umożliwiłyby skrócenie czasu śledzenia ruchu 3D poprzez zastosowanie renderingu sprzętowego,
- opracowanie konfigurowalnego modelu 3D postaci ludzkiej na potrzeby bezmarkerowego śledzenia ruchu,
- zaproponowanie i przebadanie efektywnych metod renderingu modelu 3D zarówno programowych, jak i sprzętowych oraz porównanie ich,
- opracowanie metod i narzędzi dla śledzenia ruchu z wykorzystaniem CUDA i OpenGL.

W ramach niniejszej pracy opracowano, zaimplementowano i przebadano równoległe algorytmy optymalizacji w oparciu o rój cząsteczek (PSO) i algorytm filtru cząsteczkowego (PF). Opracowano i przebadano kilka wariantów funkcji celu dla śledzenia ruchu 3D w czasie rzeczywistym w oparciu o rendering programowy i sprzętowy modelu 3D. Zaproponowano efektywne rozwiązania umożliwiające rendering modelu w OpenGL, a także wymianę danych między CUDA i OpenGL. Zaproponowano i przebadano konstrukcję konfigurowalnych potoków graficznych OpenGL do renderingu w czasie rzeczywistym znaczącej liczby modeli 3D w zadanej pozie. Opracowano i zaimplementowano metody dekompozycji algorytmu optymalizacji w oparciu o rój cząsteczek oraz równoległego wyznaczania funkcji celu pod kątem efektywnego wykorzystania dostępnych zasobów sprzętowych oraz osiągania najlepszych dokładności i częstotliwości śledzenia. Dzięki opracowanym rozwiązaniom zrealizowano śledzenie ruchu 3D całej postaci w czasie rzeczywistym. Opracowano rozwiązania umożliwiające dobranie liczby cząsteczek i liczby iteracji w algorytmie PSO, od których zależy liczba przetwarzanych klatek na sekundę, która z kolei determinuje zmiany pozy pomiędzy kolejnymi klatkami. Dzięki nim możliwe jest osiągnięcie najmniejszego błędu śledzenia ruchu 3D w czasie rzeczywistym. Określenie wspomnianych zależności nie należy do zadań trywialnych ze względu na charakter obliczeń równoległych CUDA-OpenGL oraz potrzebę pełnego wykorzystania zasobów sprzętowych dla osiągnięcia śledzenia ruchu postaci w czasie rzeczywistym.

Praca składa się z sześciu rozdziałów. W rozdziale pierwszym omówiono problematykę śledzenia ruchu w czasie rzeczywistym w kontekście aktualnie realizowanych prac badawczych. W rozdziale drugim zaprezentowano wykorzystywane metody komputerowego przetwarzania obrazów oraz śledzenia ruchu. Omówiono kalibrację systemu wielokamerowego oraz stanowisko badawcze. W rozdziale trzecim szczegółowo scharakteryzowano metody i narzędzia programowania GPU z wykorzystaniem CUDA, OpenCL oraz OpenGL. W rozdziale czwartym zaprezentowano zaprojektowany model 3D, jego parametryzację, a następnie rasteryzację. Zaprezentowano reprezentacje modelu 3D w postaci siatki i figur płaskich. Omówiono zaproponowane modele obserwacji i funkcje celu. Rozdział piąty poświęcono zagadnieniom zrównoleglenia funkcji celu. Zaprezentowano opracowane metody renderingu modelu 3D z wykorzystaniem CUDA, CUDA-OpenGL, OpenCL-OpenGL, a także wykorzystanie renderingu w równoległej funkcji celu. W rozdziale szóstym omówiono wyniki badań eksperymentalnych. Pracę podsumowano omówieniem uzyskanych wyników oraz kierunków dalszych prac.

# Rozdział 1
# Problematyka śledzenia ruchu w czasie rzeczywistym

W pierwszej części niniejszego rozdziału zamieszczono wprowadzenie do problematyki śledzenia ruchu w systemach komputerowych. W drugim podrozdziale przedstawiono przegląd kluczowych metod oraz prac w obszarze śledzenia ruchu postaci. W ostatnim podrozdziale scharakteryzowano problematykę śledzenia ruchu postaci w czasie rzeczywistym oraz otwarte problemy badawcze.

## 1.1. Problematyka śledzenia ruchu postaci

Celem śledzenia ruchu postaci ludzkiej jest wyznaczenie aktualnej pozy postaci w poszczególnych jednostkach czasu. Wyznaczenie pozy postaci ludzkiej może być zrealizowane na podstawie sekwencji obserwacji pochodzących z różnorakich sensorów. Do obserwacji ruchu najpowszechniej wykorzystuje się kamery wizyjne oraz sensory inercyjne. Systemy do śledzenia ruchu 3D wykorzystywane są dość powszechnie w przemyśle do produkcji gier (ang. *video game industry*) i przemyśle rozrywkowym (ang. *entertainment industry*), medycznych systemach diagnostycznych [165,166] oraz treningu sportowców [83], a także analizie chodu osób [82,84,181]. Do analizy ruchu we wspomnianych rozwiązaniach najczęściej wykorzystuje się markerowe śledzenie ruchu (ang. *marker-based motion capture*), w których dokonuje się obserwacji położenia markerów umieszczonych na kilku częściach ciała. Zasadniczą wadą wspomnianych rozwiązań są znaczące koszty systemów i nakłady potrzebne do realizacji eksperymentów, a w szczególności szereg niedogodności praktycznych wynikających z konieczności przymocowywania markerów, przemieszczania się markerów w trakcie eksperymentów, trudności w realizacji powtarzalnych eksperymentów. Warto także podkreślić, że większość oferowanych obecnie rozwiązań nie pozwala na śledzenie ruchu 3D w czasie rzeczywistym.

Jak już wspomniano, markerowe systemy przechwytywania ruchu działają w oparciu o znaczniki, które umieszczane są na ciele osoby w znanych punktach antropometrycznych, na kościach, w punktach charakteryzujących osie obrotów stawów, czy też na innych miejscach w zależności od potrzeb lub użytego modelu. Działanie tych systemów oparte może być na różnych zjawiskach i właściwościach fizycznych, magnetycznych, inercyjnych, ultradźwiękowych, elektromechanicznych oraz optycznych. Ponieważ markerowe systemy wykorzystywane są głównie do przechwytywania trójwymiarowych ruchów aktorów, przyjęło się nazywać je systemami motion capture lub mocap. Obecnie najczęściej stosuje się systemy pracujące w oparciu o markery optyczne - głównie ze względu na ich zalety: dużą precyzję odwzorowania ruchu oraz możliwość śledzenia wielu osób jednocześnie. Posiadają one jednak wiele wad, z których najistotniejszą jest koszt ich zakupu, co wynika z wykorzystywania specjalistycznego sprzętu oraz technologii, w szczególności z niewielkiej liczby producentów. Jak już wspomniano, markerowe systemy stosowane są głównie do śledzenia ruchu 3D w warunkach laboratoryjnych. Jest to spowodowane nie tylko stosowaniem specjalistycznego sprzętu i wyposażenia, ale także koniecznością noszenia przez śledzoną osobę znacznej liczby markerów, a bardzo często również specjalnego kombinezonu. Ponadto znaczniki mogą w trakcie sesji nagraniowej zostać przesłonięte, odpaść lub ulec przemieszczaniu, co może być przyczyną błędów. Z reguły systemy te nie są przystosowane do śledzenia ruchu 3D w czasie rzeczywistym, gdyż po zarejestrowaniu ruchu wymagane jest dodatkowe przetworzenie danych. Istotnym zagadnieniem w systemach mocap jest kalibracja systemu wielokamerowego, od której znacząco zależy dokładność pozyskiwanych danych.

Mnogość zastosowań systemów analizy ruchu 3D doprowadziła do powstania rynku o znaczącej wartości. Do wiodących firm należy firma Vicon, która dostarcza rozwiązania sprzętowe, programowe oraz kompleksowe rozwiązania do analizy ruchu 3D z wykorzystaniem wielu kamer. Mimo szeregu wad wspomnianych systemów, znaczące zapotrzebowanie rynkowe na tańsze rozwiązania doprowadziło do powstania niszy rynkowej z rozwiązaniami opartymi na podejściu bezmarkerowym (ang. *markerless motion capture systems*) [102,188]. Dzięki znaczącej liczbie grup badawczych zaangażowanych w badania nad bezmarkerowymi systemami do analizy ruchu 3D obserwuje się dynamiczny rozwój alternatywnych systemów do analizy ruchu 3D [102]. W ostatnich latach technologie bezmarkerowe wykorzystywane były do usprawnienia komunikacji człowiek-komputer, a także na potrzeby sterowania robotami mobilnymi [86,139,144]. Popularyzacja systemów bezmarkerowych powiązana jest z rozwojem interaktywnych gier wideo, w których sterowanie realizowane jest za pomocą gestów i ruchów ciała. Do grupy produktów, które znacząco przyczyniły się do popularyzacji technik bezmarkerowego śledzenia ruchu, należą kamery do konsoli Playstation firmy Sony oraz sensory ruchu firmy Microsoft. Do znaczącego spopularyzowania bezmarkerowego śledzenia ruchu przysłużyło się także wprowadzenie na rynek technologii Leap Motion oraz HTC Vive [58].

Wysiłki szeregu firm w kierunku opracowania bezmarkerowych systemów do śledzenia ruchu zostały zwieńczone opracowaniem i skomercjalizowaniem przez firmę Microsoft sensora Kinect [72,150]. Wprowadzenie na rynek sensora ruchu Kinect było znaczącym przełomem technologicznym [184], w wyniku którego powstał szereg nowych technologii i rozwiązań [41]. Sprzedaż kilkuset tysięcy egzemplarzy sensora Kinect w ciągu pierwszych trzech miesięcy dystrybucji była wynikiem wprowadzenia szeregu innowacyjnych rozwiązań z jednej strony, z drugiej strony zaś ogromnego zapotrzebowania rynku na tanie i efektywne rozwiązania do śledzenia ruchu 3D. Mimo że systemy do śledzenia ruchu 3D wykorzystujące kamery głębokości upowszechniły się stosunkowo niedawno, liczba użytecznych aplikacji oraz realizowanych projektów badawczych jest znacząca [31].

Przed wprowadzeniem sensora Kinect, do pozyskania map głębokości stosowane były skanery laserowe [128], kamery stereowizyjne [32] lub kamery ToF (ang. *Time-of-*

*Flight*) [32,77,128]. Cena sensora ruchu Kinect jest bardzo atrakcyjna w porównaniu do wcześniejszych rozwiązań. Wcześniej stosowane rozwiązania są dość drogie i dlatego wykorzystuje się je głównie w baniach laboratoryjnych lub wysoce specjalizowanych systemach. Pierwsza wersja sensora Kinect oparta była na technologiach światła strukturalnego, zaś nowsze rozwiązanie oparte jest na technologii ToF. Zastosowanie światła strukturalnego uniemożliwia estymację pozy 3D w świetle słonecznym, a także w pomieszczeniach z przedmiotami odbijającymi światło. Wadą obydwu technologii jest ograniczony zasięg pomiarów, który zwykle nie przekracza 6,5 metra. Celem przezwyciężenia ograniczeń wynikających ze stosowania pojedynczego sensora, m.in. niewidoczności przesłoniętych fragmentów sceny, podjęto szereg prób polegających na wykorzystaniu kilku sensorów Kinect do obserwacji tego samego obiektu [9]. Realizowane są także prace polegające na polepszeniu dokładności estymacji pozy 3D [26,180]. Estymacja pozy przez sensor Kinect realizowana jest w oparciu o metody lasów losowych (ang. *random forest*), które określają pozę metodami regresji, wykorzystując stosunki głębokości dla znaczącej liczby par punktów na mapach głębi. Oznacza to, że określenie pozy odbywa się bez uwzględnienia konfiguracji ciała w poprzednich klatkach. Model wykorzystywany w sensorze Kinect uczony był na podstawie setek tysięcy póz wygenerowanych komputerowo oraz kilku tysięcy póz przyjętych przez pewną populację osób reprezentujących możliwe postury. Ze względu na przeznaczenie sensora Kinect, który był głównie projektowany na potrzeby interaktywnych gier, model ten uczono pod kątem rozpoznawania póz przydatnych do sterowania przebiegiem akcji. W pracach pokazano doświadczalnie, że w wielu sytuacjach estymaty póz generowane przez sensor Kinect są znacząco różne od rzeczywistych konfiguracji człowieka [26,72]. Inne ograniczenie sensora Kinect dotyczy liczby osób, dla których możliwa jest estymacja pozy 3D. W pierwszej wersji sensora Kinect możliwa była estymacja pozy dla dwóch osób jednocześnie. W obecnej wersji możliwa jest estymacja pozy dla sześciu osób, tym niemniej należy liczyć się z trudnościami w estymacji pozy w przypadku przesłonięć. Warto przy tym wspomnieć, że w systemach z większą liczbą kamer obserwujących scenę o większej powierzchni możliwa jest bardziej efektywna analiza przesłonięć i tym samym estymacja pozy w trakcie naturalnych interakcji między ludźmi, przykładowo podczas przemieszczania się ludzi na scenie.

W wielu laboratoriach badawczych realizuje się pracę nad alternatywnymi rozwiązaniami dla sensora Kinect oraz dla wielokamerowych systemów śledzenia ruchu 3D. Do najbardziej obiecujących technologii należą rozwiązania oparte na zestawach sensorów inercyjnych IMU (ang. *Inertial Measurement Unit*) [26,46]. Cechą wspólną tych rozwiązań jest wykorzystywanie modelu 3D postaci ludzkiej do estymacji pozy. Zasadniczą wadą sensorów inercyjnych jest to, że nie dostarczają one informacji o pozycji sensora IMU względem początku globalnego układu współrzędnych. Z tego też względu w wielu zastosowaniach praktycznych zastąpienie systemu wizyjnego przez system oparty na sensorach IMU nie jest możliwe bądź wymaga zastosowania dodatkowych rozwiązań.

## 1.2. Podział metod śledzenia ruchu 3D

Zagadnieniu śledzenia ruchu człowieka poświęcono kilka prac przeglądowych [101,102]. We wspomnianych pracach zaproponowano kilka podziałów metod śledzenia ruchu 3D. Ze względu na liczbę użytych kamer systemy do śledzenia ruchu człowieka można podzielić na systemy oparte na obrazach z jednej kamery (ang. *monocular*) i systemy oparte na wielu kamerach (ang. *multicamera*) [102].

Ze względu na sposób modelowania zależności między obserwacjami i konfiguracjami ciała wyróżnić można podejście generacyjne (ang. *generative*) i podejście dyskryminacyjne (ang. *discriminative*). W pierwszej z wymienionych metod do modelowania w sposób pośredni zależności między obserwacjami i pozą wykorzystuje się wiedzę *a priori* o budowie ciała człowieka, natomiast w drugiej wyznaczana jest bezpośrednia relacja między obrazem a pozą postaci ludzkiej [153]. Wśród wykorzystywanych modeli znajdują się zarówno modele geometryczne, jak i modele ruchu [102,123,172]. Zależności geometryczne mogą być reprezentowane w oparciu o drzewo kinematyczne (ang. *kinematic tree*) [142,153] lub modelowanie za pomocą luźno powiązanych części (ang. *part-based models*) [22,164], które rozpatrywane są niezależnie.

W podejściu generacyjnym, które jest podejściem od ogółu do szczegółu (ang. *top-down*), stosuje się modele 3D wyrażające geometryczne zależności pomiędzy głównymi częściami ciała. W omawianym podejściu wpierw generowana jest hipotetyczna poza ciała, a następnie ocenia się jej dopasowanie do obrazu z wykorzystaniem modelu obserwacji. Hipotetyczne pozy ciała generowane są w oparciu o model 3D wyrażający geometryczne zależności pomiędzy głównymi częściami ciała. Estymacja pozy odbywa się poprzez porównanie zrzutowanego modelu 3D z obserwacjami na obrazach. W systemach wielokamerowych rzutowanie modelu odbywa się do każdej z kamer celem wyznaczenia wartości funkcji celu, która wyraża dopasowanie sylwetki na obrazach we wszystkich kamerach. Celem wyznaczenia pozy osoby generuje się wiele hipotetycznych póz modelu, które w dalszej części są rzutowane i porównywane z cechami obrazu reprezentującymi wydzieloną postać. Hipotetyczne pozy generowane są w podprzestrzeni konfiguracji, która określana jest na podstawie estymaty pozy człowieka w poprzedniej chwili. Omawiany sposób wyznaczania pozy jest kosztowny obliczeniowo ze względu na operację renderingu modelu dla znaczącej liczby potencjalnych konfiguracji ciała.

Wśród wykorzystywanych modeli geometrycznych znajdują się modele 2D [127], modele 3D [49,152,147] oraz modele 2,5D [56]. Modele geometryczne wykorzystywane są do reprezentacji zależności geometrycznych pomiędzy częściami ciała oraz do modelowania ruchu ciała. W podejściu opartym na kamerach cyfrowych modele te są konfigurowane do zadanych póz, a następnie po zrzutowaniu do przestrzeni obrazu porównywane są z cechami wydzielonymi na obrazie. Modele 2D stosowane są w najprostszych rozwiązaniach do reprezentacji sylwetki. Są one wykorzystywane w systemach opartych na obrazach pobieranych z jednej kamery [2,55,92,133]. W modelach płaskich wykorzystuje się płaskie figury geometryczne do reprezentacji sylwetki postaci oraz ruchu w dwóch wymiarach. Natomiast modele 3D i modele 2,5D charakteryzują się większą złożonością, przez co możliwe jest reprezentowanie śledzonej osoby nie tylko na płaszczyźnie obrazu, ale także na scenie.

Modele 3D wykorzystywane są głównie w systemach wielokamerowych [88,153,179]. W zależności od zastosowania modele mogą być zbudowane z kilku do kilkunastu części, reprezentowanych przez parametryzowane figury geometryczne lub siatki wierzchołków. Szczegółowe zestawienie reprezentacji modeli wykorzystywanych do śledzenia zamieszczono w podrozdziale 4.2. Bardziej złożona reprezentacja kształtu sylwetki umożliwia uzyskanie większych dokładności śledzenia [47,147]. Warto jednak wspomnieć, że rendering bardziej złożonych modeli wymaga użycia wyspecjalizowanych narzędzi lub układów graficznych.

Do renderingu modeli 3D wykorzystuje się rendering programowy [12,62] i rendering sprzętowy [138]. Rendering programowy wykorzystuje biblioteki programistyczne do generowania obrazu. Natomiast rendering sprzętowy wykorzystuje API programistyczne do obsługi dedykowanego sprzętu do generowania obrazu. Wśród systemów markerowych dominują rozwiązania umożliwiające wstępną wizualizację i ocenę estymat ruchu w czasie sesji nagraniowej. W końcowym etapie realizowany jest proces odzyskiwania zgubionych markerów, a następnie realizowany jest ostateczny proces syntezy i wizualizacji ruchu. W wizyjnych systemach bezmarkerowych proces wyznaczania estymat ruchu typowo realizowany jest offline, jedynie nieliczne systemy oparte na modelu umożliwiają śledzenie ruchu 3D w czasie rzeczywistym [24,87,138,140].

## 1.3. Przegląd prac i rozwiązań

Jak już wspomniano, śledzenie ruchu człowieka jest zagadnieniem, któremu obecnie poświęca się znaczną uwagę [26]. Ze względu na liczbę użytych kamer systemy do śledzenia ruchu człowieka można podzielić na systemy oparte na obrazach z jednej kamery i systemy oparte na wielu kamerach [102]. Ze względu na sposób modelowania zależności między obserwacjami i konfiguracjami ciała wyróżnić można podejście generacyjne i podejście dyskryminacyjne. W pierwszej z wymienionych metod wykorzystuje się wiedzę *a priori* o budowie ciała człowieka do modelowania w sposób pośredni zależności między obserwacjami i pozą, natomiast w drugiej wyznaczana jest bezpośrednia relacja między obrazem a pozą postaci ludzkiej.

W metodach generacyjnych zależność między obserwacjami i pozą modelowana jest w sposób pośredni. W omawianej grupie metod wyróżnić można dwa podejścia, które różnią się przyjętą reprezentacją ciała człowieka. W pierwszym z nich zakłada się, że poza może być reprezentowana w oparciu o drzewo kinematyczne [142,153], natomiast w drugim z nich do modelowania konfiguracji człowieka wykorzystuje się modele ciała oparte na luźno powiązanych częściach [164], które rozpatrywane są niezależnie. Z kolei w technikach opartych na drzewie kinematycznym wyróżnić można podejścia znajdowania konfiguracji ciała z użyciem metod optymalizacyjnych, które sprowadzają się do znajdowania konfiguracji ciała z wykorzystaniem estymatorów maksymalnego prawdopodobieństwa *a posteriori* (*Maximum a posteriori*, *MAP*) lub metod Bayesowskich, które szacują rozkłady *a posteriori* i na jego podstawie wyznaczają estymaty konfiguracji ciała. Jako metody optymalizacyjne stosuje się standardowe metody gradientowe [171], algorytmy genetyczne [187], metody stochastyczne [48] oraz algorytmy optymalizacji w oparciu o rój cząsteczek [139]. Z kolei wyznaczanie estymat póz na podstawie rozkładu *a posteriori* odbywa się z użyciem filtru cząsteczkowego (ang. *particie filter*), który także nazywany jest algorytmem kondensacji stanu [61]. Zaletą filtru cząsteczkowego jest jego zdolność do aproksymacji wielomodalnych rozkładów prawdopodobieństwa. Mimo wspomnianych zalet filtru cząsteczkowego, w kilku pracach pokazano, że algorytm optymalizacji w oparciu o rój cząsteczek uzyskuje znacząco lepsze dokładności śledzenia w porównaniu do filtru cząsteczkowego [25,66,79,185].

Jedną z popularnie wykorzystywanych modyfikacji filtru cząsteczkowego jest filtr cząsteczkowy z symulowanym wyżarzaniem (ang. *Annealed Particle Filter*, *APF*) [36]. W filtrze cząsteczkowym APF połączono algorytm symulowanego wyżarzania [75] z filtrem cząsteczkowym. W innym podejściu [143] do wyznaczania lokalnych maksimów rozkładów prawdopodobieństwa wykorzystano algorytm filtru cząsteczkowego z jądrem (ang. *Kernel Particle Filter*, *KPF*). Ze względu na znaczący wymiar przestrzeni konfiguracji ciała opracowano szereg innych technik, które narzucają ograniczenia na ruchomość poszczególnych stawów [21,158], limity kątów [36,37] oraz wspomagają generowanie przewidywanych ułożeń części ciała w oparciu o modele przybliżające rozmaitości (ang. *manifold*), na której rozłożone są potencjalne konfiguracje ciała i zmiany pomiędzy nimi, tzn. zmiany pozy [51,55,91]. Do poprawy eksploracji wielowymiarowych przestrzeni stanu stosuje się także metody oparte o *Variable Length Markov Model* [59,163], hierarchiczne modele Markowa [121] oraz metody oparte o GPLVM (ang. *Gaussian Process Latent Variable Models*) [59] czy GPDM (ang. *Gaussian Process Dynamical Models*) [172]. Ograniczenie przestrzeni poszukiwań może być realizowane także poprzez detekcję kolizji [158].

Jedną z najczęściej wykorzystywanych sekwencji do porównywania dokładności systemów śledzących ruch jest test sekwencja LeeWalk [12]. Na omawianej sekwencji zarejestrowano chód osoby po okręgu. Rejestrację ruchu zrealizowano z wykorzystaniem skalibrowanego i zsynchronizowanego systemu czterokamerowego oraz systemu mocap. W oparciu o hierarchiczny algorytm PSO [66,89] uzyskano na omawianej sekwencji dokładność śledzenia równą $52.5 \pm 11.7$ mm dla częstotliwości 30 Hz oraz $72.45 \pm 16.7$ mm dla częstotliwości 20 Hz.

## 1.4. Śledzenie ruchu w czasie rzeczywistym

W pracy [24] zaprezentowano strategie zrównoleglenia funkcji celu na potrzeby śledzenia ruchu 3D. Postać ludzka reprezentowana jest przez model składający się z dużej liczby wierzchołków, które następnie rzutowane są do przestrzeni 2D. Rozpatrywano modele składające się z 13695, 20544 oraz 27393 wierzchołków. W omawianej pracy uzyskano 8-krotne przyspieszenie obliczeń dla funkcji celu wyznaczanej na CPU wie-lordzeniowym, 30-krotne przyspieszenie dla funkcji celu wyznaczanej na pojedynczym GPU oraz 110-krotne przyspieszenie dla czterech GPU. Mimo znaczących przyspieszeń obliczeń, nakłady obliczeniowe na wyznaczanie funkcji celu wykorzystywanych w omawianej pracy są większe od czasów obliczeń funkcji celu zaproponowanych w niniejszej pracy. W omawianej pracy czas wyznaczania funkcji celu dla 500 konfiguracji modelu na czterech GPU wynosi od 18 ms do 26 ms w zależności od liczby wierzchołków. Do wyznaczenia funkcji celu zaproponowanej w niniejszej pracy, która opisuje 512 konfiguracji modelu 3D, wymagany jest czas nieco większy od 8 ms, zob. podrozdział 6.7. Dla wspomnianej liczby konfiguracji modelu 3D, przyspieszenie obliczeń na pojedynczym GPU wynosi 21. Dzięki opracowanym rozwiązaniom śledzenie ruchu 3D całej postaci ludzkiej realizowane jest w czasie rzeczywistym z częstotliwością 12 Hz.

Jak już wspomniano, celem poprawy eksploracji wielowymiarowej przestrzeni stanu, opracowano szereg metod, które wzbogacają wiedzę aprioryczną o potencjalnych pozach, które może przyjmować ciało człowieka w trakcie śledzenia [21,47]. Dzięki wspomnianym technikom możliwe jest ograniczenie przestrzeni poszukiwań, a w konsekwencji skrócenie czasu estymacji pojedynczej pozy. Warto jednak wspomnieć, że wykorzystanie wspomnianych metod poprzedzone jest budową pewnych modeli w oparciu o skończony zbiór uczący. W konsekwencji metody te najczęściej wykorzystywane są do estymacji dla ograniczonego podzbioru konfiguracji ciała. Innym sposobem ograniczenia przestrzeni poszukiwań jest stosowanie metod hierarchicznych [81,109], które oparte są na wyznaczeniu najpierw dopasowania zrzutowanego tułowia do tułowia osoby na obrazie, a następnie wyznaczeniu dopasowania pozostałych części ciała. Wspomniane metody wymagają segmentacji poszczególnych części ciała na obrazie, co w ogólności nie jest trywialną operacją. A ich wadą jest to, że w razie mało precyzyjnego dopasowania tułowia do obrazu, algorytm może napotkać na trudności w dopasowaniu pozostałych części zrzutowanego modelu do części ciała na obrazie. Opracowano wiele wariantów tego algorytmu [44,66,185]. John i inni [63] zaproponowali śledzenie ruchu w oparciu o hierarchiczny algorytm optymalizacji z wykorzystaniem roju cząsteczek. W omawianym algorytmie proces wyznaczania pozy podzielono na 12 faz, w których określane są konfiguracje poszczególnych części ciała.

Mimo mnogości rozwiązań polegających na ograniczeniu przestrzeni poszukiwań lub opartym na podejściu hierarchicznym, w literaturze przedmiotu dotyczącej śledzenia ruchu w oparciu o model 3D brak jest znaczących prac, w których z powodzeniem zrealizowano by śledzenie ruchu całej postaci ludzkiej w czasie rzeczywistym. W niniejszej pracy postawiono tezę, że możliwe jest opracowanie rozwiązań umożliwiających śledzenie ruchu całej postaci w oparciu o model 3D w czasie rzeczywistym, z częstotliwością większą od 10 Hz.

Jak zauważono w pracach [133,140] operacja renderingu modelu 3D w zadanej pozie wiąże się ze znaczącymi nakładami obliczeniowymi. Zasadnicze nakłady obliczeniowe wynikają z konieczności renderingu dużej liczby modeli, a następnie rzutowania ich do przestrzeni obrazów. Ze względu na konieczność rasteryzacji krawędzi modelu, konieczne jest sprawdzanie kolejności zamalowywania pikseli, co wiąże się z dodatkowymi nakładami obliczeniowymi. W konsekwencji, proces wyznaczania funkcji celu algorytmu śledzącego wiąże się z użyciem szeregu czasochłonnych operacji graficznych. W niniejszej pracy pokazano, że znaczące skrócenie czasu wyznaczania funkcji celu uzyskać można dzięki zrównolegleniu obliczeń oraz wykorzystaniu CUDA i OpenGL. Dzięki opracowanym rozwiązaniom możliwe jest śledzenie ruchu 3D całej postaci w systemie wielokamerowym w czasie rzeczywistym.

# Rozdział 2
# Komputerowe metody przetwarzania obrazów
# i śledzenia ruchu

W pierwszej części niniejszego rozdziału opisano wykorzystywany algorytm ekstrakcji cech osoby na obrazach pobieranych ze statycznych kamer systemu wizyjnego. Opisano także sposób wydzielania krawędzi w obszarze zainteresowania. Omówiono konstrukcję mapy odległości dla wydzielonych krawędzi. W drugim podrozdziale omówiono kalibrację systemu wizyjnego z wykorzystaniem danych mocap. Kolejny podrozdział opisuje sposób wykorzystania algorytmu optymalizacji w oparciu o rój cząsteczek oraz algorytm filtru cząsteczkowego do śledzenia ruchu 3D. W czwartym podrozdziale zaprezentowano stanowisko badawcze. Rozdział zamyka krótkie podsumowanie.

## 2.1. Ekstrakcja cech obiektów zainteresowania

Niniejsza praca dotyczy komputerowych metod śledzenia ruchu 3D w czasie rzeczywistym w oparciu o sekwencje obrazów pobierane z systemu wielokamerowego. W wyniku wieloletnich prac badawczych wypracowano skuteczne metody śledzenia ruchu 3D. W typowym podejściu realizowana jest ekstrakcja cech osoby, wydzielane jest tło oraz krawędzie obrazu, a następnie budowane są mapy odległości. Wspomniane operacje realizowane są z wykorzystaniem typowych algorytmów przetwarzania obrazów. Otwarte problemy badawcze dotyczą budowy i renderingu modeli 3D, ich reprezentacji, inicjalizacji śledzenia, a także zrównoleglenia algorytmów celem uzyskania śledzenia w czasie rzeczywistym. Badania zrealizowane w ramach niniejszej pracy dotyczą wspomnianych problemów śledzenia ruchu 3D. Ekstrakcja cech obiektów zainteresowania realizowana była z wykorzystaniem typowych podejść prezentowanych w literaturze związanej ze śledzeniem ruchu 3D [11,153]. W pozostałej części niniejszego podrozdziału omówiono wykorzystywane metody przetwarzania obrazu i ekstrakcji cech obiektów.

## Detekcja obiektu zainteresowania

W niniejszej pracy detekcja obiektu zainteresowania realizowana jest w oparciu o algorytm mieszaniny rozkładów Gaussa (ang. *Mixture of Gaussians*) [162]. W omawianej metodzie wartość każdego piksela modelowana jest za pomocą kilku rozkładów Gaussa, które są adaptowane w czasie. Niech kolejne wartości danego piksela o współrzędnych $x_0$, $y_0$ do chwili czasu $t$ scharakteryzowane będą przez następujący ciąg ostatnio obserwowanych wartości piksela:

$$\{X_1, \ldots, X_t\} = \{I(x_0, y_0, i) : 1 \leq i \leq t\}, \tag{2.1}$$

gdzie $I()$ jest sekwencją obrazów. Omawiany ciąg modelowany jest za pomocą mieszaniny $K$ rozkładów Gaussa. Prawdopodobieństwo zaobserwowania wartości aktualnego piksela określa się zależnością:

$$P(X_t) = \sum_{i=1}^{K} \omega_{i,t} \cdot \mathcal{N}(X_t, \mu_{i,t}, \Sigma_{i,t}) \,, \tag{2.2}$$

gdzie $K$ oznacza liczbę rozkładów Gaussa, $\omega_{i,t}$ jest oszacowaną wagą $i$-tego rozkładu w czasie $t$, $\mu_{i,t}$ jest średnią wartością $i$-tego Gaussiana w czasie $t$, $\Sigma_{i,t}$ jest macierzą kowariancji $i$-tego rozkładu w mieszance w chwili $t$, zaś $\mathcal{N}$ jest funkcją gęstości prawdopodobieństwa rozkładu Gaussa, która jest określona następującym wzorem:

$$\mathcal{N}(X_t, \mu_t, \Sigma) = \frac{1}{(2\pi)^{\frac{n}{2}} |\Sigma|^{\frac{1}{2}}} \, e^{-\frac{1}{2}(X_t - \mu_t)^T \Sigma^{-1}(X_t - \mu_t)} \,. \tag{2.3}$$

W praktycznych implementacjach algorytmu wartość $K$ przyjmuje wartości w przedziale od 3 do 5. Celem zmniejszenia nakładów obliczeniowych stosuje się macierze kowariancji przyjmujące następującą postać:

$$\Sigma_{i,t} = \sigma_k^2 I \,. \tag{2.4}$$

W opisywanym podejściu zakłada się, że wartości dla poszczególnych kolorów składowych każdego piksela posiadają identyczną wariancję. Dzięki temu unika się odwracania macierzy.

W prezentowanej metodzie rozkład ostatnio obserwowanych wartości dla piksela modelowany jest za pomocą mieszaniny rozkładów Gaussa. Nowa wartość piksela jest na ogół reprezentowana przez jeden z głównych składników mieszaniny. Aktualizacja modelu odbywa się w oparciu o algorytm *on-line K-means* [162]. W omawianym podejściu nowa wartość każdego piksela $X_t$ sprawdzana pod kątem dopasowania z istniejącymi $K$ rozkładami Gaussa. Warunkiem dopasowania jest, by wartość piksela znajdowała się w zakresie do 2,5-krotnej wartości odchylenia standardowego. Jeśli żaden z $K$ rozkładów Gaussa nie jest dopasowany do bieżącej wartości piksela to rozkład z najmniejszą wagą jest zastępowany nowym rozkładem, którego wartość średnia jest równa bieżącej wartości piksela, wariancja jest równa dużej wartości, zaś waga przyjmuje małą wartość. Poprzednie wagi $\omega_{k,t}$ dla $K$ rozkładów w czasie $t$ są adaptowane zgodnie z zależnością:

$$\omega_{k,t} = (1 - \alpha)\omega_{k,t-1} + \alpha(M_{k,t}) \,, \tag{2.5}$$

gdzie $\alpha$ jest współczynnikiem określającym tempo uczenia, $M_{k,t}$ jest równe jeden dla modelu, do którego został dopasowany piksel, zaś dla pozostałych modeli wartość $M_{k,t}$ jest równa zero. Dla niedopasowanych rozkładów, parametry $\mu$ i $\sigma$ pozostają niezmie-nione. Natomiast parametry rozkładów, dla których znaleziono dopasowanie aktualizowane są w następujący sposób:

$$\mu_t = (1 - p)\mu_{t-1} + pX_t \,,$$
$$\sigma_t^2 = (1 - p)\sigma_{t-1}^2 + p(X_t + \mu_t)^T(X_t - \mu_t) \,,$$
$$p = \alpha\eta(X_t|\mu_k, \sigma_k) \,. \tag{2.6}$$

Jedną z zalet omawianego rozwiązania jest to, że jeśli jakiś obiekt zostanie dodany do tła to nie niszczy on istniejącego modelu. Oryginalny kolor tła zostanie zachowany w mieszaninie Gaussianów do czasu, aż stanie się najmniej prawdopodobnym kolorem. Zatem jeśli obiekt pozostanie nieruchomy, a następnie poruszy się, to rozkład opisujący poprzednie tło ciągle będzie istniał z niezmienionymi wartościami $\mu$ i $\sigma^2$, z tym że $\omega$ przyjmie mniejsze wartości, przez co może zostać szybko ponownie dołączony do modelu tła.

## Ekstrakcja krawędzi

Metody ekstrakcji krawędzi [52,96,167] należą do podstawowych technik wykorzystywanych w przetwarzaniu i analizie obrazów. W pracy [161] wykorzystano wyznaczone na obrazie krawędzie do rozpoznawania obiektów 3D, z kolei w pracach [104,105] zaproponowano metodę *shape context*, w której wykorzystuje się wydzielone zawczasu krawędzie do estymacji konfiguracji postaci ludzkiej.

Krawędzie na obrazie cyfrowym występują w miejscach gwałtownej zmiany wartości funkcji jasności, a podstawą większości algorytmów do ekstrakcji krawędzi jest analiza pochodnych obrazu (tzw. operatorów gradientowych). Pierwsza pochodna obrazu może być wykorzystana do detekcji krawędzi i wyznaczenia jej kierunku, natomiast druga pochodna do określenia miejsca wystąpienia krawędzi. Krawędź na obrazie definiowana jest za pomocą kierunku krawędzi oraz modułu gradientu. Gradient obrazu $I$ w punkcie o współrzędnych $(x, y)$ jest wektorem określonym następującą zależnością:

$$\nabla I(x, y) = \begin{bmatrix} G_x \\ G_y \end{bmatrix} = \begin{bmatrix} \dfrac{\partial I(x, y)}{\partial x} \\ \dfrac{\partial I(x, y)}{\partial y} \end{bmatrix}. \tag{2.7}$$

Wektor ten jest skierowany w kierunku największego wzrostu intensywności obrazu $I$ dla punku $(x, y)$. Mając w dyspozycji gradient obrazu, można określić jego moduł $M$:

$$M(x, y) = |\nabla I(x, y)| = \sqrt{G_x^2 + G_y^2} \,, \tag{2.8}$$

który określa tzw. siłę krawędzi oraz kierunek krawędzi $\Psi$:

$$\Psi(x, y) = \tan^{-1}\left(\frac{G_y}{G_x}\right). \tag{2.9}$$

Ze względu na dyskretny charakter obrazu cyfrowego, gradient może być aproksymowany w oparciu o następujące zależności:

$$G_x = \frac{\partial I(x,y)}{\partial x} \approx I(x+1,y) - I(x,y),$$
$$G_y = \frac{\partial I(x,y)}{\partial y} \approx I(x,y+1) - I(x,y).$$

(2.10)

Projektanci algorytmów do przetwarzania obrazów mają do dyspozycji wiele operatorów gradientowych [167]. Najpowszechniej wykorzystywane operatory dają się przedstawić za pomocą masek. Dzięki wspomnianym maskom możliwe jest efektywne wyznaczenie nowej wartości danego piksela z uwzględnieniem wartości pikseli z jego sąsiedztwa. Zaletą podejścia opartego o maski jest to, że wykorzystuje się ten sam algorytm dla różnych wartości w masce. W przypadku operatorów gradientowych najczęściej wykorzystywane maski mają rozmiar 3x3 lub 2x2. Do najprostszych operatorów tego typu należą operatory Robertsa, w których wykorzystuje się maski o rozmiarach 2x2. Przykładowe maski dla omawianego operatora mają następującą postać:

$$\begin{bmatrix} 1 & -1 \\ 0 & 0 \end{bmatrix} \qquad \begin{bmatrix} 1 & 0 \\ -1 & 0 \end{bmatrix} \qquad \begin{bmatrix} 1 & 0 \\ 0 & -1 \end{bmatrix} \qquad \begin{bmatrix} 0 & 1 \\ -1 & 0 \end{bmatrix}$$
$$180° \qquad\quad 270° \qquad\quad 225° \qquad\quad 315°$$

(2.11)

Pod maskami zamieszczono kąty określające kierunki operatorów. Operator Robertsa, mimo szeregu wad, a w szczególności dużej czułości na szumy oraz niewielkimi różnicami pomiędzy odpowiedzią operatora na krawędź i szum, jest często wykorzystywany w praktyce ze względu na prostotę użycia oraz niewielkie wymagania obliczeniowe. W celu zmniejszenia wrażliwości detektora na szum wykorzystuje się maski o wymiarach 3x3. Przykładowymi operatorami tego typu są operatory Prewitta (zob. rys. 2.1d) oraz operatory Sobela (zob. rys. 2.1e).

$$\begin{bmatrix} 1 & 1 & 1 \\ 0 & 0 & 0 \\ -1 & -1 & -1 \end{bmatrix} \quad \begin{bmatrix} 1 & 0 & -1 \\ 1 & 0 & -1 \\ 1 & 0 & -1 \end{bmatrix} \quad \begin{bmatrix} 1 & 1 & 0 \\ 1 & 0 & -1 \\ 0 & -1 & -1 \end{bmatrix} \quad \begin{bmatrix} 0 & 1 & 1 \\ -1 & 0 & 1 \\ -1 & -1 & 0 \end{bmatrix}$$
$$90° \qquad\qquad 180° \qquad\qquad 135° \qquad\qquad 45°$$

(2.12)

$$\begin{bmatrix} 1 & 2 & 1 \\ 0 & 0 & 0 \\ -1 & -2 & -1 \end{bmatrix} \quad \begin{bmatrix} 1 & 0 & -1 \\ 2 & 0 & -2 \\ 1 & 0 & -1 \end{bmatrix} \quad \begin{bmatrix} 2 & 1 & 0 \\ 1 & 0 & -1 \\ 0 & -1 & -2 \end{bmatrix} \quad \begin{bmatrix} 0 & 1 & 2 \\ -1 & 0 & 1 \\ -2 & -1 & 0 \end{bmatrix}$$
$$90° \qquad\qquad 180° \qquad\qquad 135° \qquad\qquad 45°$$

(2.13)

Jak już wspomniano, w wyniku zastosowania masek o większych rozmiarach, operatory Prewitta (2.12) oraz Sobela (2.13) są mniej wrażliwe na szum obrazu w porównaniu do operatorów Robertsa. Ponadto dla identycznych krawędzi, operatory Sobela generują na wyjściu większe wartości. Wartości ujemne występujące w maskach mogą prowadzić do wartości ujemnych na obrazach wynikowych, w związku z tym obraz gradientu należy poddać operacji skalowania lub wyznaczyć bezwzględne wartości pikseli. Operatory gradientowe są wrażliwe na szum, dlatego w praktyce, przed wydzieleniem krawędzi najczęściej realizuje się filtrację obrazu, której celem jest usunięcie występującego szumu. Do najprostszych filtrów eliminujących szumy obrazu możemy zaliczyć filtry uśredniające, filtry medianowe oraz filtry Gaussa. Opracowano także bardziej złożone sposoby filtracji obrazów. Przykładem może być algorytm opisany w pracach [159,160], który w procesie filtracji obrazu wykorzystuje metody probabilistyczne.

Ponieważ algorytmy wykorzystujące maski operatorów gradientowych wykrywają krawędzie na obrazach tylko w jednym kierunku, w praktyce wykorzystuje się filtry nieliniowe (potocznie zwane także filtrami kombinowanymi). Idea działania sprowadza się do zastosowania dwóch gradientów w prostopadłych do siebie kierunkach i połączenia przetwarzanych tym sposobem obrazów. Połączenia wspomnianych obrazów można uzyskać za pomocą modułu gradientu (2.8). W praktyce wykorzystuje się jednak uproszczone zależności (2.14) i (2.15), które umożliwiają uzyskanie zbliżonych wyników, a jednocześnie mają znacząco mniejsze wymagania obliczeniowe.

$$M(x,y) \approx |G_x| + |G_y| \qquad (2.14)$$
$$M(x,y) \approx max\big(|G_x|, |G_y|\big) \qquad (2.15)$$

Jak już wspomniano, do ekstrakcji krawędzi mogą także być wykorzystane drugie pochodne obrazu. Przykładem operatora opartego o drugą pochodną jest operator Laplace'a, nazwany także laplasjanem. Laplasjan dwuwymiarowej funkcji obrazu $I(x, y)$ zdefiniowany jest jako suma drugich pochodnych cząstkowych:

$$L[I(x,y)] = \frac{\partial^2 I(x,y)}{\partial x^2} + \frac{\partial^2 I(x,y)}{\partial y^2}. \qquad (2.16)$$

Laplasjan w dyskretnej dziedzinie obrazu może być aproksymowany w oparciu o maski o rozmiarach 3x3. Przykładowe maski dla operatora Laplace'a mają następującą postać:

$$\begin{bmatrix} 0 & -1 & 0 \\ -1 & 4 & -1 \\ 0 & -1 & 0 \end{bmatrix} \begin{bmatrix} -1 & -1 & -1 \\ -1 & 8 & -1 \\ -1 & -1 & -1 \end{bmatrix} \begin{bmatrix} 1 & -2 & 1 \\ -2 & 4 & -2 \\ 1 & -2 & 1 \end{bmatrix}. \qquad (2.17)$$

Operatory Laplace'a, podobnie jak inne operatory gradientowe, wrażliwe są na szumy obrazu. Do innych wad można zaliczyć to, że oparte o nie algorytmy generują podwójne kontury oraz nie pozwalają na wyznaczenie kierunku brzegu. Zasadniczą zaletą laplasjanów jest ich zdolność do wykrywania krawędzi we wszystkich kierunkach bez dodatkowych operacji, z którymi należy się liczyć w razie wykorzystywania filtrów opartych o operatory gradientowe.

Operatory gradientowe i laplasjany, mimo swoich wad, są dość powszechnie stosowane w komputerowych metodach przetwarzania obrazów. Jednak w zastosowaniach praktycznych, w których wymagana jest lepsza jakość krawędzi, to znaczy wierniejsze odwzorowanie rzeczywistych krawędzi na obrazie, a czas przetwarzania obrazu jest mniej istotny, do detekcji krawędzi obrazu wykorzystuje się bardziej złożone metody, np. algorytm Canny'ego [23,52]. We wspomnianej pracy zaprezentowano trzy kryteria, które określają optymalny algorytm detekcji krawędzi, mianowicie:

- kryterium dobrej lokalizacji – wykryte krawędzie powinny znajdować się możliwie blisko środka rzeczywistej krawędzi na obrazie cyfrowym;
- kryterium dobrej detekcji – metoda powinna zapewnić jak najmniejsze prawdopodobieństwo detekcji fałszywych detekcji krawędzi i jak największe prawdopodobieństwo wykrycia prawdziwych krawędzi;
- kryterium minimalnej odpowiedzi – dla każdej prawdziwej krawędzi, algorytm powinien wygenerować pojedynczą odpowiedź. Algorytm spełniający to kryterium wyznacza krawędzie o grubości jednego piksela.

Wspomniane kryteria spełnia algorytm Canny'ego. W omawianym algorytmie można wyróżnić cztery zasadnicze etapy:

- rozmycie obrazu za pomocą filtru Gaussa – operacja ta wykonywana jest celem usunięcia szumów występujących na obrazie;
- obliczenie modułu gradientu (2.8) i kierunku gradientu (2.9) – do wyznaczenia gradientów $G_x$ oraz $G_y$ można wykorzystać dowolny z zaprezentowanych wcześniej operatorów, jednak najpowszechniej stosuje się operator Sobela. W razie wyznaczenia kierunku krawędzi, wynik otrzymany z zależności (2.9) jest przybliżany do jednego z czterech kątów, tj. 0°, 45°, 90° i 135°;
- usunięcie pikseli o nie maksymalnej jasności (ang. *Non Maxima Suppression*) – etap ten ma za zadanie zmniejszenie grubości krawędzi do jednego piksela i jest realizowany na podstawie informacji uzyskanych w poprzednim kroku, tzn. na podstawie modułu gradientu i kierunku krawędzi;
- usunięcie błędnie wykrytych krawędzi – ostatnią operacją jest usunięcie błędnie wydzielonych krawędzi metodą progowania z histerezą. W tym celu określane są dwa progi (górny i dolny), a następnie sprawdza się, czy wartość gradientu analizowanego piksela znajduje się poniżej dolnego progu, powyżej górnego progu czy też pomiędzy progami. Jeśli zachodzi pierwszy z wymienionych przypadków, rozpatrywany piksel uznaje się za należący do rzeczywistej krawędzi. Jeśli zaś wartość gradientu jest mniejsza od dolnego progu, to rozpatrywany piksel jest odrzucany. Jeśli wartość gradientu dla danego piksela znajduje się pomiędzy progiem dolnym a górnym, to jest on akceptowany jako krawędź jedynie wtedy, gdy wartość gradientu sąsiedniego piksela jest większa od górnego progu.

W porównaniu do wcześniej omówionych metod, algorytm Canny'ego umożliwia bardziej precyzyjne wydzielenie krawędzi kosztem znacząco większych nakładów obli-czeniowych. Na rys. 2.1 przedstawiono przykładowe wyniki ekstrakcji krawędzi za pomocą zaprezentowanych metod. W przypadku operatorów Robertsa, Prewitta, Sobela oraz laplasjanu, obraz wejściowy został wstępnie przetworzony filtrem Gaussa z maską o rozmiarze 3x3. Celem wyznaczenia modułu gradientu wykorzystano zależność (2.8), a ponadto otrzymany tym sposobem obraz został poddany progowaniu. Na prezentowanych obrazach można zaobserwować przewagę algorytmu Canny'ego nad prostymi algorytmami ekstrakcji krawędzi. Na obrazie krawędzi uzyskanym za pomocą omawianego algorytmu (zob. rys. 2.1f) zaobserwować można znaczną liczbę krawędzi, które zostały pominięte przez pozostałe algorytmy. Co więcej, na obrazach prezentujących wyniki ekstrakcji krawędzi w oparciu o proste operatory można zaobserwować znaczną liczbę błędnych detekcji, które w szczególności są widoczne na kapeluszu zaprezentowanej osoby.

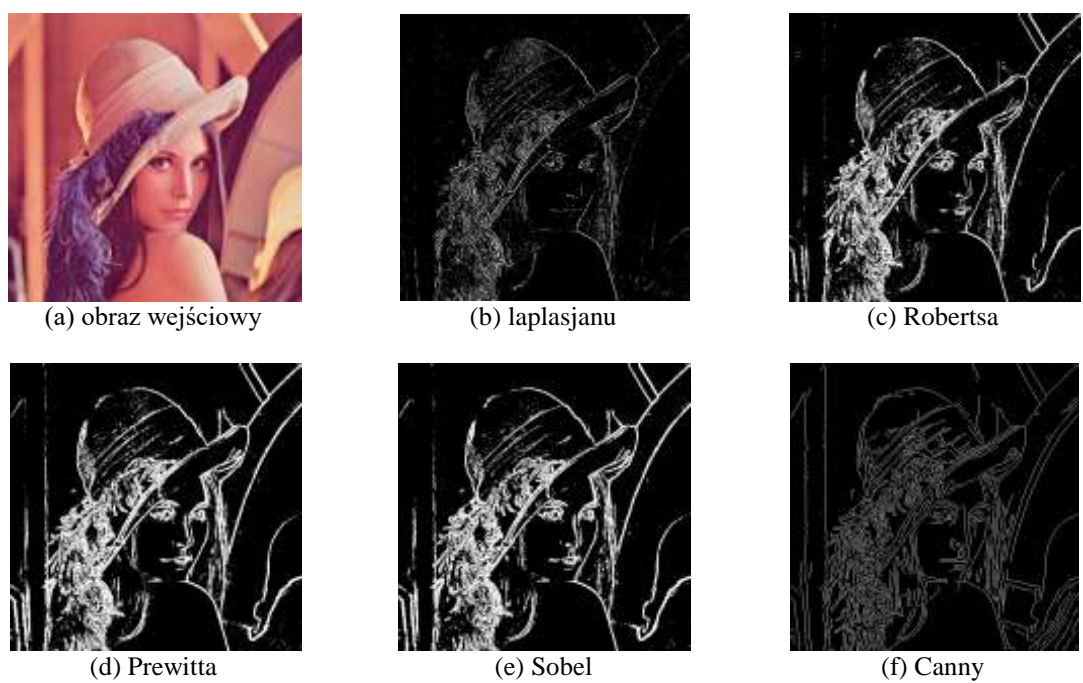

(a) obraz wejściowy            (b) laplasjanu            (c) Robertsa

(d) Prewitta            (e) Sobel            (f) Canny

**Rys. 2.1. Porównanie metod detekcji krawędzi**

Jak już wspomniano, algorytm Canny'ego pozwala na uzyskanie bardzo dobrych wyników ekstrakcji krawędzi. Wyniki badań eksperymentalnych pokazują, że zastosowanie algorytmu Canny nie prowadzi do znacząco lepszych dokładności śledzenia ruchu 3D postaci. Mając na względzie znaczące wymagania obliczeniowe oraz ograniczenia wynikające z potrzeby przetwarzania w czasie rzeczywistym, w niniejszej pracy wykorzystywano prostszą metodę, która opiera się na operatorze Sobela. Ekstrakcja krawędzi na potrzeby śledzenia ruchu 3D postaci ludzkiej realizowana była w oparciu o filtr nieliniowy złożony z dwóch operatorów Sobela, wykrywających krawędzie w prostopadłych do siebie kierunkach. Przed zastosowaniem wspomnianego filtru z obrazów usunięto szumy, wykorzystując w tym celu z filtr Gaussa z maską

o rozmiarze 3x3. Na rys. 2.2b przedstawiono przykładowy obraz uzyskany z użyciem operatora Sobela.

Obraz z wyznaczonymi krawędziami, oprócz tych należących do śledzonej osoby może zawierać znaczną liczbę krawędzi należących do sceny, zob. rys. 2.2b. Wydzielenie krawędzi należących do śledzonej osoby wymaga usunięcia z obrazu krawędzi należących do tła. W celu usunięcia krawędzi należących do tła, obraz z krawędziami jest poddawany operacji iloczynu logicznego z obrazem zawierającym wydzieloną sylwetką postaci. Przed wykonaniem tej operacji, wydzielana tym sposobem sylwetka osoby poddawana jest operacji dylatacji [23,52]. W wyniku zastosowania wspomnianych operacji uzyskuje się obraz zawierający jedynie krawędzie należące do śledzonej osoby, zob. rys. 2.2c. Obraz sylwetki przetworzony tym sposobem jest następnie wykorzystywany do wyznaczenia wartości funkcji dopasowania, a także wykorzystywany jest do wyznaczenia mapy odległości od krawędzi.

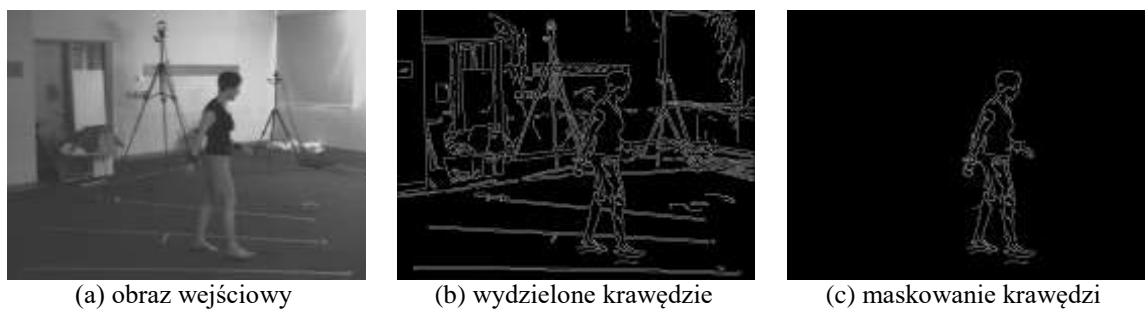

(a) obraz wejściowy      (b) wydzielone krawędzie      (c) maskowanie krawędzi

**Rys. 2.2. Ekstrakcja krawędzi należących do obszaru zainteresowania**

Opisana metoda usunięcia krawędzi należących do sceny ma jednak pewną wadę, mianowicie w pewnych sytuacjach może się zdarzyć, że pewna część ciała osoby nie zostanie prawidłowo wydzielona. W takim przypadku część krawędzi należących do osoby zostanie pominięta. Omawiana sytuacja dotyczy w szczególności rąk i innych części ciała, których barwa jest podobna do barwy podłogi. Warto jednak wspomnieć, że na wykorzystywanych sekwencjach obrazów omawiane zjawisko występowało sporadycznie. Celem poprawy skuteczności ekstrakcji obiektów, których kolor jest zbliżony do koloru tła, należałoby zastosować bardziej zaawansowane metody wykorzystujące klasyfikatory obiektów [33].

# Mapa odległości od krawędzi

Dyskretna mapa odległości od krawędzi zawiera informację o minimalnej wartości odległości danego punktu na obrazie od najbliższej krawędzi na obrazie [34]. Najpowszechniej wykorzystywanymi miarami odległości dla omawianych map są metryki: Euklidesowa, miejska (ang. *city-block*), szachownicy (ang. *chessboard*) oraz Quasi-Euklidesowa. Na rys. 2.3 przedstawiono obrazy ilustrujące mapy odległości uzyskane w oparciu o wspomniane wyżej metryki odległości.

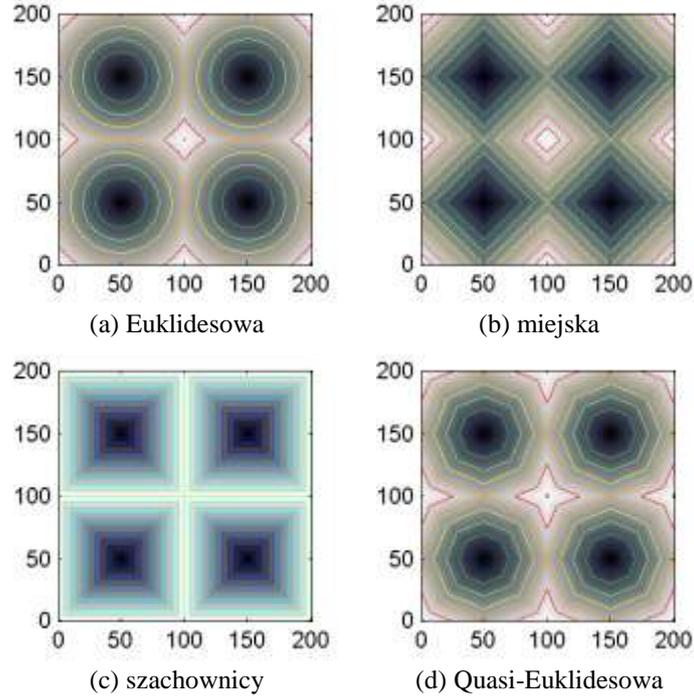

(a) Euklidesowa            (b) miejska

(c) szachownicy         (d) Quasi-Euklidesowa

**Rys. 2.3. Dyskretne mapy odległości dla wybranych metryk**

Jak już wspomniano, odległość od krawędzi można wyznaczyć w oparciu o jedną z wymienionych wyżej metryk. Poniżej zilustrowano sposób wyznaczania odległości między dwoma punktami $p_1$ i $p_2$ o współrzędnych $x_{p_1}, y_{p_1}$, $x_{p_2}, y_{p_2}$ w oparciu o wspomniane metryki:

$$D_{Euclidean}(p_1, p_2) = \sqrt{\left(x_{p_1} - x_{p_2}\right)^2 + \left(y_{p_1} - y_{p_2}\right)^2} \qquad (2.18)$$

$$D_{CityBlock}(p_1, p_2) = \left|x_{p_1} - x_{p_2}\right| + \left|y_{p_1} - y_{p_2}\right| \qquad (2.19)$$

$$D_{ChessBoard}(p_1, p_2) = max\left(\left|x_{p_1} - x_{p_2}\right|, \left|y_{p_1} - y_{p_2}\right|\right) \qquad (2.20)$$

$$D_{Quasi}(p_1, p_2) = \begin{cases} \left|x_{p_1} - x_{p_2}\right| + (\sqrt{2} - 1)\left|y_{p_1} - y_{p_2}\right| & gdy \left|x_{p_1} - x_{p_2}\right| > \left|y_{p_1} - y_{p_2}\right| \\ (\sqrt{2} - 1)\left|x_{p_1} - x_{p_2}\right| + \left|y_{p_1} - y_{p_2}\right| & gdy \left|x_{p_1} - x_{p_2}\right| \leq \left|y_{p_1} - y_{p_2}\right| \end{cases} \qquad (2.21)$$

W niniejszej pracy generowanie mapy odległości realizowano w oparciu o metrykę chessboard. Na rys. 2.4 zamieszczono mapy odległości wyznaczone dla 25-tego obrazu z sekwencji LeeWalk. Mapy odległości zaprezentowane na omawianym rysunku uzyskano w 10-ciu iteracjach, 100 iteracjach oraz 254 iteracjach, posługując się algorytmem [11,88].

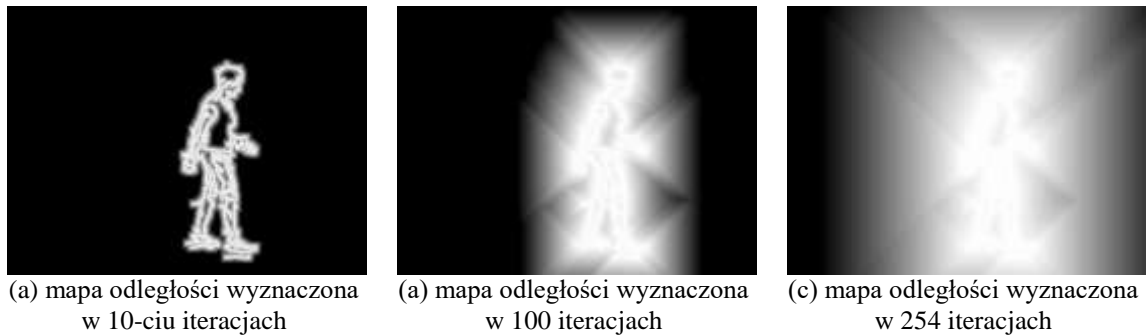

(a) mapa odległości wyznaczona
w 10-ciu iteracjach

(a) mapa odległości wyznaczona
w 100 iteracjach

(c) mapa odległości wyznaczona
w 254 iteracjach

**Rys. 2.4. Mapa odległości od krawędzi dla obrazu z sekwencji LeeWalk**

Mapa odległości od krawędzi należących do osoby zawiera szereg istotnych informacji o aktualnej pozie. Dzięki niej każdy piksel obrazu posiada informację o odległości od najbliższej krawędzi postaci ludzkiej. Jeśli zatem w wyniku rzutowania modelu 3D na obraz, krawędź zrzutowanego modelu znajduje się w pewnej odległości od krawędzi na obrazie, możliwe będzie wyznaczenie dopasowania między krawędziami. W konsekwencji jeśli zrzutowane krawędzie modelu nie pokrywają się z krawędziami na obrazie, co ma miejsce bardzo często w śledzeniu ruchu 3D postaci ludzkiej, możliwe jest oszacowanie dopasowania między pozą zrzutowanego modelu a pozą osoby na obrazie.

## 2.2. Kalibracja systemu wielokamerowego

Model kamery służy do modelowania projekcji z przestrzeni trójwymiarowej do dwuwymiarowej przestrzeni obrazu oraz modelowania projekcji z dwuwymiarowej przestrzeni obrazu do przestrzeni 3D. W prezentowanym podejściu projekcja jest konieczna do wyznaczenia dopasowania między zrzutowanym modelem 3D do przestrzeni obrazu i cechami wydzielonymi na obrazie. Na rys. 2.5 zilustrowano w sposób poglądowy projekcję modelu 3D do dwuwymiarowej przestrzeni obrazu w systemie jednokamerowym.

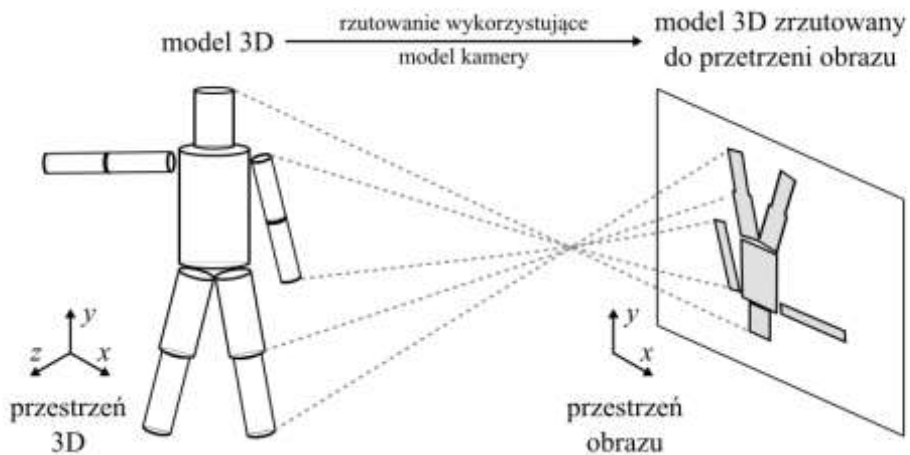

**Rys. 2.5. Rzutowanie modelu 3D do dwuwymiarowej przestrzeni obrazu**

W zastosowaniach wykorzystujących kamery szerokokątne istotne jest wykorzystanie takiego modelu kamery, który uwzględni zniekształcenia na obrazach zarejestrowanych przez rzeczywiste kamery. W niniejszej pracy wykorzystywany jest model kamery Tsai [169,170]. Wspomniany model oparty jest na modelu kamery otworkowej (ang. *pinhole camera model*) i opisuje zniekształcenia radialne. Model kamery Tsai jest parametryzowany przez 11 parametrów:

- $r_x, r_y, r_z$ – kąty obrotu opisujące przejście z globalnego układu współrzędnych do układu współrzędnych kamery,
- $t_x, t_y, t_z$ – translacje zapewniające przejście z globalnego układu współrzędnych do układu współrzędnych kamery,
- $f$ – ogniskowa kamery (ang. *focal length*),
- $k$ – współczynnik radialnych zniekształceń układu optycznego,
- $c_x, c_y$ – współrzędne środka w radialnych równaniach opisujących zniekształcenia toru optycznego,
- $s_x$ – współczynnik skalujący.

Wśród wymienionych parametrów kamery można wyróżnić parametry zewnętrzne $(r_x, r_y, r_z, t_x, t_y, t_z)$ oraz parametry wewnętrzne $(f, k, c_x, c_y, s_x)$. Oprócz wspomnianych parametrów, model kamery Tsai wykorzystuje sześć stałych parametrów, które są podawane przez producenta konkretnego modelu kamery:

- $N_{cx}$ – liczba elementów sensora kamery dla osi x,
- $N_{fx}$ – liczba pikseli obrazu dla osi x (w pikselach),
- $d_x$ – szerokość elementów sensora kamery,
- $d_y$ – wysokość elementów sensora kamery,
- $d_{px}$ – efektywna szerokość piksela obrazu (w mm/piksel),
- $d_{py}$ – efektywna wysokość piksela obrazu (w mm/piksel).

Na rys. 2.6 zilustrowano w sposób poglądowy rzutowanie punktu 3D do dwuwymiarowej przestrzeni obrazu w oparciu o model kamery Tsai. W trakcie rzutowania w pierwszej kolejności następuje przekształcenie współrzędnej punktu 3D z układu współrzędnych świata $(x_w, y_w, z_w)$ do układu współrzędnych kamery $(x_i, y_i, z_i)$. Omawiane przekształcenie realizowane jest z wykorzystaniem parametrów zewnętrznych kamery $(r_x, r_y, r_z, t_x, t_y, t_z)$ na podstawie zależności:

$$p_k = Rp_w + tx, \qquad (2.22)$$

gdzie $R$ oznacza macierz rotacji parametryzowaną przez $r_x, r_y, r_z$, parametr $t$ oznacza wektor translacji $[t_x, t_y, t_z]$, natomiast $p_w$ oraz $p_k$ to punkty w układzie współrzędnych świata oraz w układzie współrzędnych kamery.

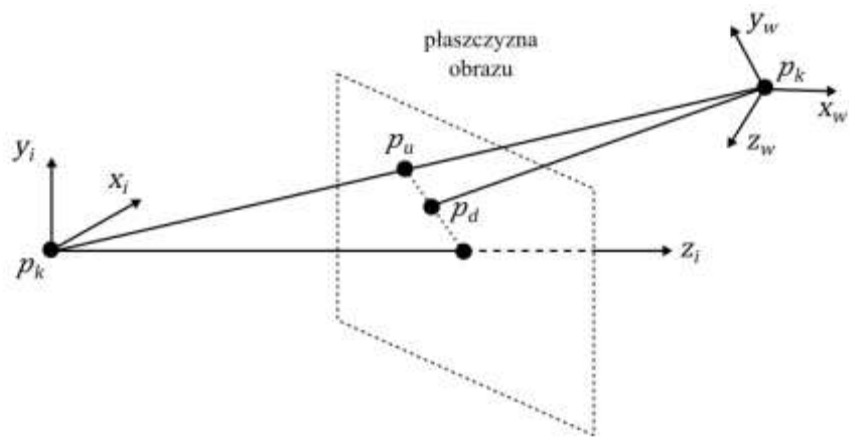

**Rys. 2.6. Projekcja perspektywiczna punktu 3D z wykorzystaniem modelu kamery Tsai**

Po określeniu współrzędnych punktu $p_k(x_k, y_k, z_k)$ w trójwymiarowym układzie współrzędnych kamery, przekształcany punkt jest rzutowany na płaszczyznę obrazu (zob. punkt $p_u(x_u, y_u)$ na rys. 2.6). Współrzędne omawianego punktu $p_u(x_u, y_u)$ wyznaczane są na podstawie równań:

$$x_u = f \frac{x_k}{z_k}\,,$$
$$y_u = f \frac{y_k}{z_k}\,.$$

$$(2.23)$$

Proces rzutowania punktu realizowany jest w oparciu o wspomniany model kamery otworkowej, w którym nie są brane pod uwagę zniekształcenia występujące na obrazach uzyskanych z użyciem rzeczywistej kamery. Celem wyznaczenia współrzędnych punktu na płaszczyźnie obrazu z uwzględnieniem zniekształceń $p_d(x_d, y_d)$ należy dokonać następujących transformacji:

$$x_u = x_d(1 + kr^2)\,,$$
$$y_u = y_d(1 + kr^2)\,,$$

$$(2.24)$$

gdzie $r = \sqrt{x_d^2 + y_d^2}$, natomiast $k$ jest współczynnikiem zniekształceń toru optycznego. W końcowym etapie uwzględnia się współrzędne środka radialnych zniekształceń układu optycznego $(c_x, c_y)$, współczynnik skalujący $s_x$ oraz odległości pomiędzy sąsiednimi pikselami sensora kamery.

Wyznaczenie ostatecznych współrzędnych punktu na obrazie realizowane jest na podstawie następujących równań:

$$x_f = \frac{s_x x_d}{d_x} + c_x\,,$$

$$(2.25)$$

$$y_f = \frac{s_x y_d}{d_y} + c_y \,, \qquad\qquad (2.26)$$

w których $x_f$, $y_f$ oznaczają ostateczne współrzędne punktu na płaszczyźnie obrazu, zaś parametry $d_x$, $d_y$ określają odległości pomiędzy elementami sensora kamery. Stałe parametry kamery $d_x$, $d_y$ zależą jedynie od wielkości sensora kamery oraz rozdzielczości obrazu.

Parametry zewnętrzne i wewnętrzne kamery określa się w procesie kalibracji. Metoda Tsai wymaga podania położeń 3D dla pewnej liczby punktów oraz położeń odpowiadających im punktów na obrazie [169]. Na ich podstawie wyznaczane są poszczególne parametry modelu kamery. Parametry te określa się w procesie optymalizacji nieliniowej Levenberga-Marquardta [103]. Proces wyznaczania parametrów modelu Tsai kamery podzielony jest na dwie główne fazy. W fazie pierwszej określana jest pozycja oraz orientacja kamery, zaś w drugiej fazie określane są parametry wewnętrzne kamery. Omawiana metoda kalibracji kamery dostępna jest w wersji koplanarnej (ang. *coplanar*) i niekoplanarnej (ang. *non-coplanar*). W metodzie koplanarnej wymagane jest, aby współrzędna $z$ punktów 3D, które wykorzystywane są w procesie kalibracji, była równa zero. Dzięki wyzerowaniu współrzędnej $z$ upraszcza się proces optymalizacji, przez co algorytm w podstawowej wersji wymaga jedynie podania współrzędnych pięciu punktów w przestrzeni 3D oraz odpowiadających im punktów na obrazie. W metodzie niekoplanarnej konieczne jest zaś podanie co najmniej siedmiu współrzędnych punktów 3D. W obydwu wersjach algorytmu zaleca się jednak wykorzystanie większej liczby punktów. Wyzerowanie jednej ze współrzędnych punktów 3D w metodzie koplanarnej ogranicza obszar zastosowań metody. Oznacza to, że omawiany algorytm można wykorzystać tylko wtedy, jeśli kamera jest skierowana w kierunku płaszczyzny $xy$, zob. rys 2.6. W innym przypadku wykorzystuje się metodę niekoplanarną. Inna z powszechnie wykorzystywanych metod kalibracji zakłada użycie znanego obiektu (zazwyczaj wykorzystuje się plansze ze wzorem szachownicy) do określenia parametrów wewnętrznych. Gdy parametry wewnętrzne kamery są już określone, wyznaczana jest pozycja oraz orientacja kamery. Oprócz modelu kamery Tsai wykorzystywane są także inne modele [57,183]. Metoda Tsai jest powszechnie wykorzystywana w systemach monitoringu i nadzoru [43]. Porównanie między metodą Tsai i metodą Zhang, która jest dość powszechnie wykorzystywana w systemach wizyjnych, znaleźć można w pracy [94].

Parametry zewnętrzne i wewnętrzne kamer składające się na wielokamerowy system wizyjny określone zostały w oparciu o metodę kalibracji [65]. W omawianej metodzie wykorzystuje się pozycję 3D markerów, które określa się w oparciu o skalibrowany system mocap. W pierwszej kolejności określa się pozycję 3D punktów w układzie współrzędnych systemu mocap wraz z odpowiadającymi im współrzędnymi na obrazach z poszczególnych kamer. W tym celu osoba wykonująca kalibrację kamer wykorzystuje narzędzie MX Calibration Wand i porusza się po całej scenie, zaś system mocap rejestruje i zapisuje współrzędne znaczników w plikach w formacie C3D. W następnej kolejności wyznaczane są współrzędne znaczników na obrazach z poszczególnych kamer. Otrzymany w ten sposób zbiór danych jest wykorzystywany w kalibracji niekoplanarnej, w której określa się parametry modelu kamery Tsai. W niniejszej pracy wykorzystywany był model kamery, który zbudowano w oparciu o zbiór 100 punktów dla każdej z kamer.

## 2.3. Algorytmy śledzenia obiektów

Określenie pozy w śledzeniu ruchu 3D postaci ludzkiej może być zrealizowane wykorzystując zadanie optymalizacji dynamicznej lub przy użyciu metod Bayesowskich. W niniejszym podrozdziale omówiono algorytm optymalizacji w oparciu o rój cząsteczek, a także jego modyfikację dla problemu optymalizacji dynamicznej oraz filtr cząsteczkowy.

## Algorytm optymalizacji w oparciu o rój cząsteczek

Optymalizacja w oparciu o rój cząsteczek (ang. *Particle Swarm Optimization*) jest metodą poszukiwania stochastycznego, która czerpie z reguł zachowań społecznych w populacjach organizmów żywych [27,71]. Algorytm PSO i jego liczne modyfikacje są obecnie dość powszechnie wykorzystywane w wielu dziedzinach nauki do rozwiązywania złożonych problemów optymalizacyjnych [186], m.in. w przetwarzaniu sygnałów, analizie i rozpoznawaniu obrazów oraz sieciach neuronowych [124]. Algorytm ten jest także bardzo często wykorzystywany w problemie śledzenia obiektów [25,85,139] oraz śledzenia ruchu 3D postaci ludzkiej [62,66,79,80,88,108,133]. Poszukiwanie optymalnego rozwiązania zadanej funkcji celu odbywa się w oparciu o populację cząsteczek, które wymieniają się informacją. Każdy z osobników reprezentuje potencjalne rozwiązanie problemu. Istotną cechą tego algorytmu jest to, że współpracujący ze sobą osobnicy dysponują pseudo-gradientową informacją o przestrzeni poszukiwań. W trakcie poszukiwań optymalnego rozwiązania osobnicy zapamiętują najlepsze znalezione rozwiązania, a także zapamiętują rozwiązanie globalne.

W niniejszej pracy wykorzystywany był synchroniczny algorytm optymalizacji w oparciu o rój cząsteczek z globalnym sąsiedztwem. W omawianym wariancie algorytmu z globalnym sąsiedztwem każda cząsteczka o indeksie $i$ przechowuje aktualną pozycję $x^i$, prędkość $v^i$ oraz najlepszą pozycję $p^i$ znalezioną do tej pory. Prędkość $v^i$ każdej cząsteczki jest wyznaczona na podstawie następującej zależności [85]:

$$v^{i,k+1} = \omega v^{i,k} + c_1 r_1 \left( p^i - x^{i,k} \right) + c_2 r_2 \left( g - x^{i,k} \right), \qquad (2.27)$$

w której $g$ oznacza najlepszą pozycję znalezioną przez wszystkie osobniki roju, parametr $\omega$ jest współczynnikiem inercyjnym (ang. *inertial factor*), $r_1$ i $r_2$ są liczbami losowymi z rozkładu jednorodnego, którego wartości zawierają się w przedziale $< 0, 1 >$, zaś parametry $c_1$ oraz $c_2$ są mnożnikami wagowymi.

Pozycja osobnika w przestrzeni poszukiwań aktualizowana jest w oparciu o prędkość (2.27) zgodnie z następującym równaniem:

$$x^{i,k+1} = x^{i,k} + v^{i,k+1} \,.$$ (2.28)

Aktualizacja wartości $g$ oraz $p^i$ realizowana jest w oparciu o wartość funkcji celu $f(x)$. Wartość $p^i$ aktualizowana jest na podstawie następującej zależności:

$$p^i = \begin{cases} x^{i,k}, & gdy\, f(x^{i,k}) > f(p^i) \\ p^i, & gdy\, f(x^{i,k}) \leq f(p^i) \end{cases} \,.$$ (2.29)

W oparciu o wyznaczone tym sposobem wartości $p^i$ aktualizuje się wartość $g$ z wykorzystaniem następującej zależności:

$$g = \begin{cases} p^i, & gdy\, f(p^i) > f(g) \\ g, & gdy\, f(p^i) \leq f(g) \end{cases} \,.$$ (2.30)

Jak już wspomniano, śledzenie ruchu odbywa się w oparciu o optymalizację dynamiczną, w której po wyznaczeniu optymalnej pozy dla danej klatki następuję rozproszenie cząsteczek celem uwzględnienia możliwej zmiany pozy ludzkiej w następnej klatce. W niniejszej pracy, rozproszenie (inicjalizacja) cząsteczek dla klatki pobranej w czasie $t$ odbywa się w oparciu o rozkład normalny:

$$x_t^i = \mathcal{N}(x_{t-1}, \Sigma_N) \,,$$ (2.31)

gdzie $x_{t-1}$ oznacza pozycję najlepszej cząsteczki, zaś $\Sigma_N$ oznacza macierz kowariancji. Pozycja najlepszej cząsteczki, tzn. poza postaci ludzkiej, wyznaczana jest na podstawie wartości $g$, która została określona w ostatniej iteracji algorytmu optymalizacji w oparciu o rój cząsteczek.

# Filtracja bayesowska

Metody określane mianem filtrów Bayesa umożliwiają probabilistyczną estymację ukrytego stanu systemu dynamicznego w oparciu o pojawiające się na bieżąco dane pomiarowe. W filtracji Bayesowskiej zakłada się, że proces stochastyczny może być opisany jako proces Markowa [45,153]. W procesie Markowa stan obiektu $x_t$ w chwili $t$ zależy bezpośrednio od aktualnej obserwacji $z_t$ i stanu procesu w poprzedniej jednostce czasu $x_{t-1}$, zob. rys. 2.7.

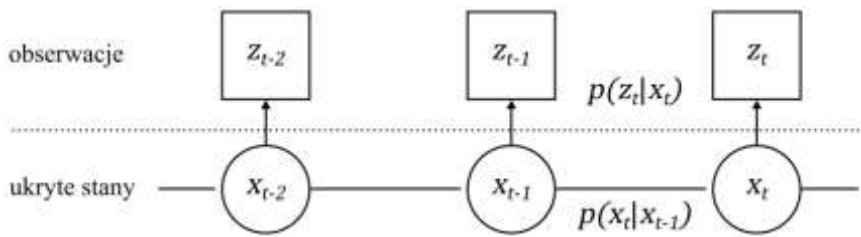

**Rys. 2.7. Schemat filtracji Bayesowskiej**

W filtracji Bayesowskiej wiedza o aktualnym stanie systemu dynamicznego reprezentowana jest za pomocą warunkowego rozkładu prawdopodobieństwa *a posteriori* $p(x_t, z_t)$ zmiennej stanu $x_t$. Rozkład prawdopodobieństwa *a posteriori* jest wyznaczany zgodnie z równaniem:

$$p(x_t|z_{1:t}) \propto p(z_t|x_t) \int p(x_t|x_{t-1}) p(x_{t-1}|z_{1:t-1}) dx_{t-1}, \tag{2.32}$$

w którym $p(z_t|x_t)$ oznacza probabilistyczny model obserwacji, $p(x_t|x_{t-1})$ reprezentuje probabilistyczny model lokomocji, zaś $p(x_{t-1}|z_{1:t-1})$ reprezentuje rozkład prawdopodobieństwa *a posteriori*. Model obserwacji $p(z_t|x_t)$ aproksymuje prawdopodobieństwo dokonania obserwacji $z_t$ w systemie dynamicznym znajdującym się w stanie $x_t$. Model lokomocji $p(x_t|x_{t-1})$ określa prawdopodobieństwo, że obiekt znajdzie się w stanie $x_t$, jeśli w poprzednim kwancie czasu znajdował się w stanie $x_{t-1}$.

W filtracji Bayesowskiej po dokonaniu obserwacji $z_t$ realizowana jest korekcja rozkładu prawdopodobieństwa *a priori* w oparciu o następującą zależność:

$$p(x_t|z_{1:t}) \propto p(z_t|x_t) p(x_t|z_{1:t-1}). \tag{2.33}$$

Natomiast rozkład prawdopodobieństwa $p(x_t|z_{1:t-1})$ korygowany jest zgodnie z następującym modelem lokomocji:

$$p(x_t|z_{1:t-1}) = \int p(x_t|x_{t-1}) p(x_{t-1}|z_{1:t-1}) dx_{t-1}. \tag{2.34}$$

W oparciu o równania (2.33) i (2.34), po dokonaniu kolejnego pomiaru możliwe jest wyznaczenie $p(x_t|z_{1:t})$ [61]. Popularnymi implementacjami filtru Bayesa są: filtr Kalmana (ang. *Kalman Filter*), rozszerzony filtr Kalmana (ang. *Extended Kalman Filter*) [10,95] oraz filtr cząsteczkowy (ang. *Particle Filter, PF*) [8,53,61]. Filtry cząsteczkowe, nazywane także algorytmami kondensacji stanu, są powszechnie wykorzystywane są w algorytmach wizyjnych do śledzenia obiektów [61,86,132,136], rozpoznawania gestów wykonywanych rękami [69] i lokalizacji robotów mobilnych [45].

## Algorytm filtru cząsteczkowego

Filtr cząsteczkowy, nazywany również sekwencyjną metodą Monte Carlo, aproksymuje rozkład prawdopodobieństwa za pomocą cząsteczek. W omawianym algorytmie zakłada się, że zmienna losowa $x$ ma rozkład o gęstości $p(x)$, zaś estymatorem miary prawdopodobieństwa $p$ jest rozkład empiryczny o następującej postaci:

$$p(x) = \frac{1}{N} \sum_{i=1}^{N} \delta(x - x^{(i)}), \tag{2.35}$$

gdzie $\delta$ oznacza funkcję delty Diraca, zaś $\{x^{(i)}\}_{i=1}^{N}$ oznacza ciąg niezależnych próbek wygenerowanych na podstawie rozkładu $p$.

Wygenerowanie zbioru niezależnych próbek na podstawie rozkładu $p$ nie jest trywialnym zadaniem i dlatego do wygenerowania próbek z rozkładu empirycznego stosuje się metody Monte Carlo lub rozkład proponowany, nazywany także funkcją ważności. Przy wykorzystaniu funkcji ważności rozkład $p$ zastępuje się rozkładem $q$ o przybliżonych właściwościach. Omawiany rozkład $q$ wykorzystywany jest do generowania niezależnych próbek, ważonych zmiennych losowych $\{x^{(i)}, w^{(i)}\}_{i=1}^{N}$, gdzie $w^{(i)}$ są wagami odzwierciedlającymi prawdopodobieństwo wyznaczenia próbki $x^{(i)}$ z rozkładu $p$ opisanego następującym równaniem:

$$p(x) = \sum_{i=1}^{N} w^{(i)} \delta\big(x - x^{(i)}\big), \tag{2.36}$$

gdzie $\sum_{i=1}^{N} w^{(i)} = 1$ oraz $w^{(i)} \geq 0 \; \forall \; i = 0, \dots, N$, zaś $x^{(i)}$ jest próbką z rozkładu $q$. Najprostszy filtr cząsteczkowy wykorzystuje sekwencyjne próbkowanie SIS (ang. *Sequential Importance Sampling*). W takim podejściu rozkład prawdopodobieństwa aproksymowany jest za pomocą zbioru cząsteczek z określonymi wagami. Wagi cząsteczek wyznaczane są na podstawie następującego równania [8]:

$$w_t^{(i)} = \frac{p\big(x_{0:t}^{(i)} \big| z_{1:t}\big)}{q\big(x_{0:t}^{(i)} \big| z_{1:t}\big)} \propto w_{t-1}^{(i)} \frac{p\big(z_t \big| x_t^{(i)}\big) p\big(x_t^{(i)} \big| x_{t-1}^{(i)}\big)}{q\big(x_t^{(i)} \big| x_{t-1}^{(i)}, z_t\big)} . \tag{2.37}$$

Efektywność filtru cząsteczkowego zależy od doboru funkcji ważności. W algorytmie *bootstrap* [53], funkcji ważności $q\big(x_t^{(i)} \big| x_{t-1}^{(i)}, z_t\big)$ przyporządkowuje się prawdopodobieństwo przejścia:

$$q\big(x_t^{(i)} \big| x_{t-1}^{(i)}, z_t\big) = p\big(x_t^{(i)} \big| x_{t-1}^{(i)}\big). \tag{2.38}$$

W omawianym filtrze wagi cząsteczek określa się w oparciu o następujące równanie:

$$w_t^{(i)} \approx p\big(z_t \big| x_t^{(i)}\big). \tag{2.39}$$

Konsekwencją przyjęcia wspomnianej funkcji ważności jest mała sprawność filtru cząsteczkowego, w szczególności w przypadku dynamicznych zmian stanu śledzonego obiektu. Niekorzystnym zjawiskiem jest także degeneracja próbek, która polega na zaniedbaniu wagi większości próbek w wyniku wielokrotnego aktualizowania wagi danej cząsteczki. Celem uniknięcia degeneracji cząsteczek w procesie śledzenia, stosuje się próbkowanie cząsteczek (ang. *resampling*), które polega na odrzuceniu cząsteczek o nieistotnych wagach, w miejsce których generuje się cząsteczki z istotnymi wagami. Mechanizm próbkowania realizowany jest przez odwzorowanie zbioru $N$ cząsteczek z różnymi wagami przez zbiór cząsteczek z identycznymi wagami:

$$\left\{ x_t^{(i)}, w_t^{(i)} \right\} = \left\{ x_t^{(i)}, \frac{1}{N} \right\}. \tag{2.40}$$

Wspomniany zabieg zapobiega propagowaniu nieistotnych cząsteczek do następnej iteracji. Algorytm filtru cząsteczkowego wykorzystywany w niniejszej pracy jest rozszerzeniem algorytmu SIS, który nazywany jest także algorytmem SIR (ang. *Sampling Importance Resampling*). Algorytm ten składa się z następujących po sobie etapów: predykcji, korekty i przepróbowania, zob. rys 2.8. W danej chwili czasu $t$ zbiór cząsteczek z jednakowymi wagami aproksymuje rozkład $p\left( x_t^{(i)} \middle| z_{1:t-1} \right)$. Wagi $w_t^{(i)}$ wyznaczane są na podstawie modelu obserwacji $p(z_t | x_t)$, natomiast nowy zbiór cząsteczek z jednakowymi wagami wyznaczany jest w etapie przepróbowania.

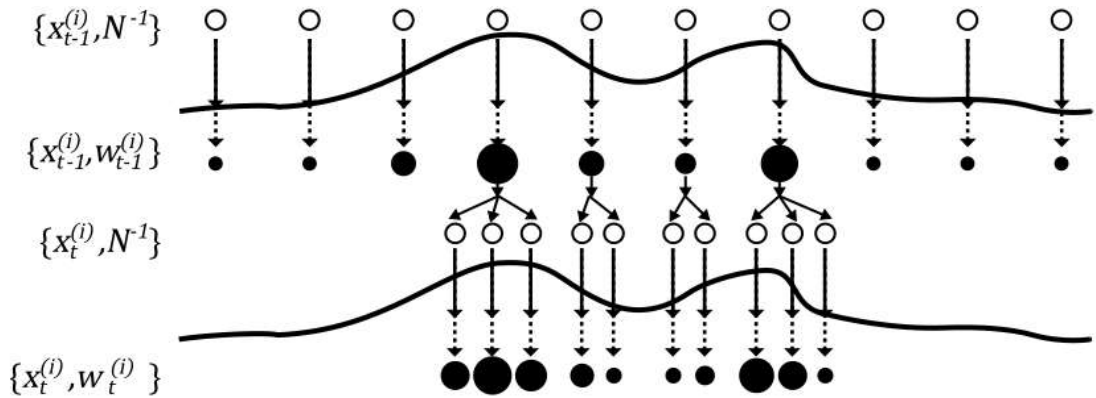

**Rys. 2.8. Schemat działania filtru cząsteczkowego**

W algorytmie filtru cząsteczkowego SIR w wersji *bootstrap*, cykl etapów predykcji, korekty i próbkowania powtarzany jest dla każdej chwili $t$. Operacje te można przedstawić za pomocą następującego pseudokodu:

**Algorytm 1**

1 $\left\{ x_t^{(i)}, w_t^{(i)} \right\} = pf\left( \left\{ x_{t-1}^{(i)}, w_{t-1}^{(i)} \right\}_{i=1}^{N}, z_t \right)$

2    Dla $i = 0, 1, \dots, N$

3      $x_t^{(i)} \sim p\left( x_t^{(i)} \middle| x_{t-1}^{(i)} \right)$

4      $w_t^{(i)} \sim p\left( z_t \middle| x_t^{(i)} \right)$

5      $norm = \sum_{i=1}^{N} w_t^{(i)}$

6    Dla $i = 0, 1, \dots, N$

7      $w_t^{(i)} = \frac{w_t^{(i)}}{norm}$

8    $\left\{ x_t^{(i)}, w_t^{(i)} \right\}_{i=1}^{N} = resample\left( \left\{ x_t^{(i)}, w_t^{(i)} \right\}_{i=1}^{N} \right)$

Parametrami wejściowym algorytmu jest zbiór cząsteczek w chwili $t-1$ wraz z ich wagami, zaś parametrem wyjściowym jest zbiór cząsteczek w chwili $t$ wraz z ich wagami. W pierwszym kroku algorytmu następuje predykcja położenia cząsteczek, zob. linia 3, określenie ich wag, zob. linia 4, a następnie wyznaczana jest suma wag, zob. linia 5. W drugim kroku realizowana jest normalizacja wag wszystkich cząsteczek, zob. linia 7. W ostatnim kroku realizowana jest operacja przepróbowania. Operację przepróbowania przedstawić można za pomocą następującego pseudokodu:

**Algorytm 2**

1 $\left\{x_t^{(i)}, w_t^{(i)}\right\}_{i=1}^{N} = resample\left(\left\{x_t^{(i)}, w_t^{(i)}\right\}_{i=1}^{N}\right)$

2     $c_i = 0$

3     Dla $i = 2,3, \dots, N$

4        $c_i = c_{i+1} + w_t^{(i)}$

5     $i = 1$

6     $u_i \sim U\left[0, \frac{1}{N}\right]$

7     Dla $j = 1,2, \dots, N$

8        $u_j = u_i + \frac{1}{N}(j-1)$

9        Dopóki $u_j > c_i$

10          $i = i + 1$

11        $x_t^{(i)} = x_t^{(i)}$

12        $w_t^{(i)} = \frac{1}{N}$

Implementacja filtru cząsteczkowego wymaga określenia probabilistycznego modelu obserwacji i lokomocji. W niniejszej pracy wykorzystywany jest rozkład normalny do propagacji cząsteczek w przestrzeni:

$$x_t^{(i)} = \mathcal{N}\left(x_{t-1}^{(i)}, \Sigma_N\right), \tag{2.41}$$

w którym $\Sigma_N$ jest macierzą kowariancji rozkładu normalnego. Elementy diagonalne macierzy kowariancji są proporcjonalne do maksymalnej prędkości śledzonego obiektu.

W niniejszej pracy wagi cząsteczek wyznaczane były w oparciu o równanie:

$$w_t^{(i)} = \frac{1}{\sigma_z\sqrt{2\pi}} exp\left(-\frac{\left(1 - f\left(x_t^{(i)}\right)\right)^2}{\sigma_z^2}\right), \tag{2.42}$$

gdzie $f\left(x_t^{(i)}\right)$ określa funkcję celu, natomiast współczynnik $\sigma_z$ wyznaczany jest eksperymentalnie.

## 2.4. Stanowisko badawcze do badań eksperymentalnych

W niniejszej pracy śledzenie ruchu 3D postaci ludzkiej realizowano w systemie czterokamerowym. Badania realizowane były w oparciu o sekwencje obrazów nagrane offline wraz z danymi *ground-truth*. Sekwencje zostały nagrane w laboratoriach PJWSTK w oparciu o cztery skalibrowane i zsynchronizowane kolorowe kamery HD [65][1]. Kolorowe obrazy RGB o rozdzielczości 1920x1080 przeskalowano do rozdzielczości 480x270 pikseli. Obrazy zapisywane były z częstotliwością 25 klatek na sekundę, natomiast dane odniesienia zapisywane były z częstotliwością 100 Hz. Badania zrealizowano na czterech sekwencjach P1S, P2S, P1D, P2D, które były wykorzystywane we wcześniejszych pracach [81,87,88,138,141]. Dwie pary kamer umieszczone na wprost obserwowały obszar o wymiarach 6,5x4,5 metra. Dane odniesienia zarejestrowano przy pomocy systemu mocap firmy Vicon, złożonego z 10 kamer. Błędy śledzenia wyznaczano w oparciu o 39 markerów. Na czterech sekwencjach zarejestrowano ruch dwóch osób poruszających się na wprost kamer i po przekątnej sceny. Badania eksperymentalne realizowano także na dostępnej i powszechnie wykorzystywanej sekwencji LeeWalk [11]. Wspomniana sekwencja została nagrana przy pomocy kamer monochromatycznych o rozdzielczości 640x480 z częstotliwością 60 Hz. Dane odniesienia są zsynchronizowane z danymi systemu wizyjnego. Dane odniesienia zawierają pozycje 3D 15 markerów. Sekwencja składa się z 450 klatek i przedstawia ruch osoby po okręgu. Dzięki wynikom badań zamieszczonym w pokrewnych pracach możliwe było porównanie dokładności uzyskanej w niniejszej pracy z dokładnościami uzyskiwanymi przez inne zespoły badawcze, a także częstotliwości pracy porównywanych systemów. W obydwu sekwencjach dane odniesienia zapisane były w plikach C3D.

Do osiągnięcia celu pracy, polegającego na śledzeniu ruchu 3D postaci ludzkiej w systemie czterokamerowym w czasie rzeczywistym, zbudowano specjalistyczne stanowisko badawcze. Stanowisko badawcze składa się z dwuprocesorowego komputera HP Z800, wyposażonego w procesory Intel Xeon X5690 taktowane z częstotliwością 3,46 GHz oraz 16 GB pamięci operacyjnej. Każdy z procesorów posiada 6 rdzeni z funkcją *hyper-threading*. Komputer wyposażono w dwie karty graficzne Nvidia GTX 590 oraz zamiennie w kartę graficzną Nvidia GTX 780 Ti. Karta GTX 590 zawiera dwa układy graficzne, z których każdy posiada 8 wieloprocesorów z 64 rdzeniami na wieloprocesor. Każdy wieloprocesor zawiera 48 KB zintegrowanej pamięci współdzielonej. Karta GTX 780 posiada 12 wieloprocesorów, z których każdy dysponuje 240 rdzeniami i 64 KB zintegrowanej pamięci wspólnej. Dostępna moc obliczeniowa dwóch kart GTX 590 wynosi 4976 GFLOPS, natomiast moc obliczeniowa karty GTX 780 wynosi 5045 GFLOPS. Wymiana danych między komputerem PC a kartami graficznymi odbywa się przez magistralę PCI-Express 3.0 i PCI-Express 2.0 x16 o przepustowości 16 GB/s i 8 GB/s.

---

[1] Prace realizowane w ramach projektu O R00 0021 11: „Zastosowanie systemów nadzoru wizyjnego do identyfikacji zachowań i osób oraz detekcji sytuacji niebezpiecznych przy pomocy technik biometrycznych i inferencji postaci w 3D z wideo".

Na rys. 2.9 zamieszczono schemat stanowiska badawczego. Na komputerze PC dokonywano ekstrakcji postaci ludzkiej, wyznaczano krawędzie, a także wyznaczano mapę odległości od krawędzi. Wspomniane obliczenia realizowano w środowisku wielowątkowym OpenMP ze wsparciem biblioteki OpenCV. Badania nad algorytmami do śledzenia ruchu realizowano w oparciu o oprogramowanie dla CPU oraz GPU. Oprogramowanie przygotowano w środowisku Microsoft Visual Studio 2013. Oprogramowanie dla GPU przygotowano z wykorzystaniem CUDA SDK 6.0, która dostarcza także implementację API OpenCL 1.1. Do zarządzania kontekstem OpenGL wykorzystano biblioteki GLFW i GLES. Operację renderingu, zob. blok rendering na rys. 2.10, realizowano zarówno programowo, jak i sprzętowo z wykorzystaniem CPU i GPU. Model 3D w zadanej pozie renderowano programowo na CPU i Nvidia CUDA. Rendering sprzętowy realizowano w oparciu o OpenGL z wykorzystaniem programów cieniujących. Śledzenie ruchu w czasie rzeczywistym realizowano w oparciu o oprogramowanie CUDA, CUDA-OpenGL i OpenCL-OpenGL.

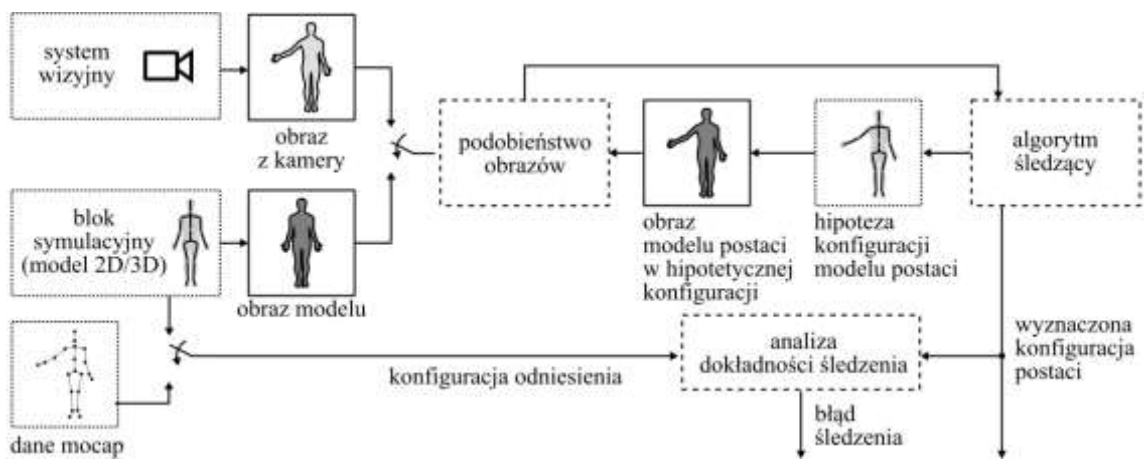

**Rys. 2.9. Schemat stanowiska do badań eksperymentalnych**

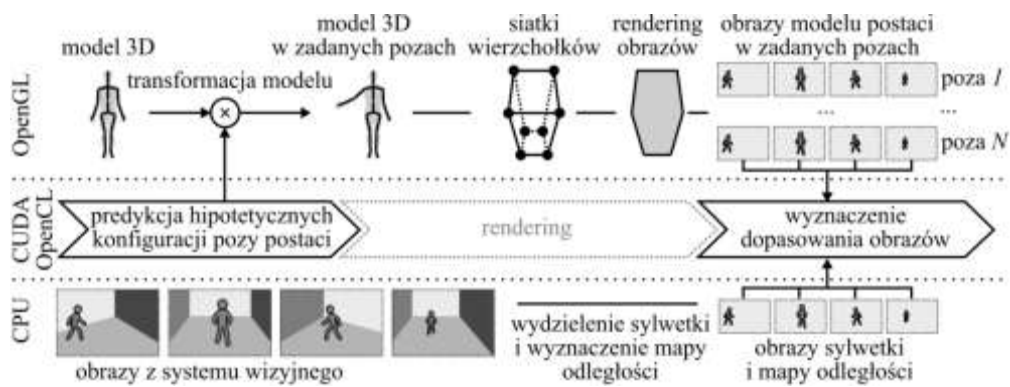

**Rys. 2.10. Schemat metody śledzenia ruchu w oparciu o model 3D**

## 2.5. Podsumowanie

W niniejszym rozdziale omówiono metody ekstrakcji cech na potrzeby śledzenia ruchu w oparciu o model 3D. Omówiono sposób kalibracji systemu wizyjnego w powiązaniu z układem współrzędnych systemu mocap. Zaprezentowano algorytm optymalizacji w oparciu o rój cząsteczek i algorytm filtru cząsteczkowego w kontekście ich wykorzystania do śledzenia ruchu 3D. Omówiono zbudowane stanowisko badawcze na potrzeby śledzenia ruchu 3D w czasie rzeczywistym.

# Rozdział 3
# Metody i narzędzia programowania GPU

W niniejszym rozdziale omówiono metody i narzędzia programowania GPU (ang. *Graphics Processing Unit*) pod kątem potencjalnych zastosowań w śledzeniu ruchu postaci ludzkiej. Na wstępie zdefiniowano i omówiono miary do szacowania wydajności obliczeń równoległych oraz architektury systemów do obliczeń równoległych. W dalszej części rozdziału omówiono platformy obliczeń równoległych CUDA, a także platformy OpenCL i OpenGL. W końcowej części rozdziału omówiono hierarchię wątków i pamięci w CUDA i OpenCL.

## 3.1. Wydajność obliczeń równoległych

Obliczeniami równoległymi nazywamy obliczenia, w których przetwarzanych jest jednocześnie wiele danych, a sam proces zamiany sekwencyjnego algorytmu na algorytm równoległy nazywany jest zrównolegleniem. Oczekiwanym efektem zrównoleglenia algorytmu jest przyspieszenie $S$, które definiowane jest jako stosunek czasu potrzebnego na realizację zadania przez algorytm sekwencyjny do czasu realizacji tego samego zadania przez algorytm równoległy. Przyspieszenie obliczeń przedstawić można za pomocą równania [110,117]:

$$S = \frac{T_s}{T_p} \, ,$$ 
(3.1)

gdzie $T_s$ oznacza czas przetwarzania zadania przez najszybszy algorytm sekwencyjny, zaś $T_p$ oznacza czas przetwarzania zadania przez algorytm równoległy na $p$ procesorach. W systemach wieloprocesorowych realna wartość przyspieszenia jest mniejsza od wartości wynikającej z liczby użytych procesorów ze względu na narzuty komunikacyjne oraz konieczność współdzielenia zasobów sprzętowych. Miarą wykorzystania zasobów współbieżnych jest efektywność, którą opisać można następującą zależnością [110,117]:

$$E = \frac{S}{p} \, ,$$ 
(3.2)

gdzie $p$ oznacza liczbę procesorów. Implementacja algorytmów równoległych nie jest zadaniem trywialnym i wymaga sporych nakładów czasowych oraz znajomości docelowej architektury sprzętowej, na której realizowane będą obliczenia. Ze względu na sposób organizacji obliczeń równoległych wyróżnia się szereg architektur komputerowych do obliczeń równoległych [117]. Wybrane architektury komputerowe do obliczeń równoległych przybliżono w dalszej części podrozdziału.

## 3.2. Szacowanie przyspieszenia obliczeń

W niniejszej części podrozdziału przedstawiono prawa Amdahla, Gustafsona, Moore'a oraz prawo Little'a [117], które wykorzystywane są do oszacowania przyspieszenia obliczeń równoległych.

Prawo Amdahla zakłada, że fragmenty programu komputerowego można podzielić na cześć równoległą i sekwencyjną [110,117]. Sekwencyjne fragmenty kodu wykonywane są na jednej jednostce arytmetyczno-logicznej ALU (ang. *Arithmetic Logic Unit*) w postaci ciągu instrukcji realizowanych w kolejności pierwszy na wejściu, pierwszy na wyjściu FIFO (ang. *First In, First Out*). Fragmenty kodu wykonywane równolegle realizują operacje jednocześnie na wielu ALU. Zgodnie z regułą (3.1) teoretyczne przyspieszenie po zastosowaniu obliczeń równoległych zależne jest od części programu, które wykonywane są sekwencyjnie:

$$S = \frac{T(1)}{T(p)} = \frac{1}{s + \frac{r}{p}} \ , \tag{3.3}$$

gdzie $s$ oznacza czas wykonywania części sekwencyjnej, zaś $r$ oznacza czas wykonywania części równoległej, przy czym $r$ i $s$ są znormalizowane tak, aby był spełniony warunek $s + r = 1$. Zakładając, że liczba procesorów $p$ dąży do nieskończoności, możemy wyznaczyć górną granicę przyspieszenia:

$$\lim_{p \to \infty} S = \lim_{p \to \infty} \frac{1}{s + \frac{r}{p}} = \frac{1}{s}. \tag{3.4}$$

Z prawem Amdahla ściśle powiązane jest prawo Gustafsona [177], które określa górną granicę przyspieszenia algorytmu realizowanego w środowisku równoległym. Przyjmując, że przyspieszenie obliczeń na $p$ procesorach uzależnione jest wyłącznie od części kodu $s$, których nie można zrównoleglić, to przyspieszenie opisać można za pomocą równania, które jest znane jako prawo Gustafsona:

$$S_{gustafson}(p) = s + p(1 - s) \ . \tag{3.5}$$

Przy znajomości przyspieszenia obliczeń w środowisku równoległym, możliwe jest wyznaczenie miary równoległości kodu, nazywanej miarą Karpa-Flatta. Metryka wyznaczana jest zgodnie z równaniem:

$$e(p) = \frac{\frac{1}{S(p)} - \frac{1}{p}}{1 - \frac{1}{p}} \ , \tag{3.6}$$

w którym algorytm równoległy na $p$ procesorach osiąga przyspieszenie $S(p)$. Im mniejsza jest wartość miary $e(p)$, tym większy jest stopień współbieżności.

Prawo Little'a mówi, że średnia liczba klientów $L$ w systemie jest równa iloczynowi średniego czasu przebywania $\lambda$ w systemie oraz średniego tempa ich napływania $W$ [177]:

$$L = \lambda \cdot W. \tag{3.7}$$

Omawiane prawo przytaczane jest w kontekście obliczeń równoległych ze względu na możliwość oszacowania wymaganego stopnia zrównoleglenia systemu. Przez powiązanie czasu przebywania $\lambda$, który w obliczeniach równoległych oznacza opóźnienie (ang. *latency*) oraz średniego tempa napływania klientów $W$, które w obliczeniach równoległych jest miarą przepustowości obliczeń (ang. *throughput*) możliwe jest określenie wymaganego stopnia zrównoleglenia systemu [117,174,177]:

$$parallelism = latency \cdot throughput. \tag{3.8}$$

Prawo Moore'a jest prawem empirycznym, mówiącym, że liczba tranzystorów w zintegrowanym układzie scalonym podwaja się co dwa lata. Prawo to można odnieść zarówno do rozmiaru pamięci, mocy obliczeniowej procesorów, jak i do liczby rdzeni dostępnych w procesorze [110,177].

## 3.3. Architektury i mechanizmy obliczeń równoległych

### Architektury komputerowe w taksonomii Flynna

Najpopularniejsza klasyfikacja architektur komputerowych, nazywana taksonomią Flynna, została zaproponowana w 1966 roku, a następnie rozszerzona w 1972 roku [117]. W omawianej taksonomii brane są pod uwagę dwa kryteria:

- liczba strumieni instrukcji,
- liczba strumieni danych.

Przy wzięciu pod uwagę wspomnianych kryteriów, możliwe jest wyróżnienie czterech architektur komputerowych (zob. także tabele 3.1):

- SISD (ang. *Single Instruction Single Data*),
- SIMD (ang. *Single Instruction Multiple Data*),
- MISD (ang. *Multiple Instruction Single Data*),
- MIMD (ang. *Multiple Instruction Multiple Data*).

Relację między architekturami przedstawić można w sposób poglądowy za pomocą dwuwymiarowej tablicy, mając na względzie liczbę strumieni instrukcji, które realizowane są na określonej liczbie jednostek obliczeniowych (ang. *Processing Unit*, *PU*) z jednej strony oraz liczbę strumieni danych z drugiej strony.

**Tabela 3.1. Taksonomia Flynna**

|  | Jeden strumień danych | Wiele strumieni danych |
|---|---|---|
| Jeden strumień instrukcji | SISD | MISD |
| Wiele strumieni instrukcji | SIMD | MIMD |

Maszyny wykorzystujące architekturę SISD są tradycyjnymi maszynami jednoprocesorowymi, w których jeden strumień danych przetwarzany jest przez jeden strumień instrukcji, zob. rys. 3.1a. Omawiany mechanizm przetwarzania danych jest zgodny z architekturą von Neumanna [117]. Na maszynie zgodnej z architekturą SISD można realizować zadania równolegle przy wykorzystaniu przetwarzania potokowego oraz superskalarnego [117].

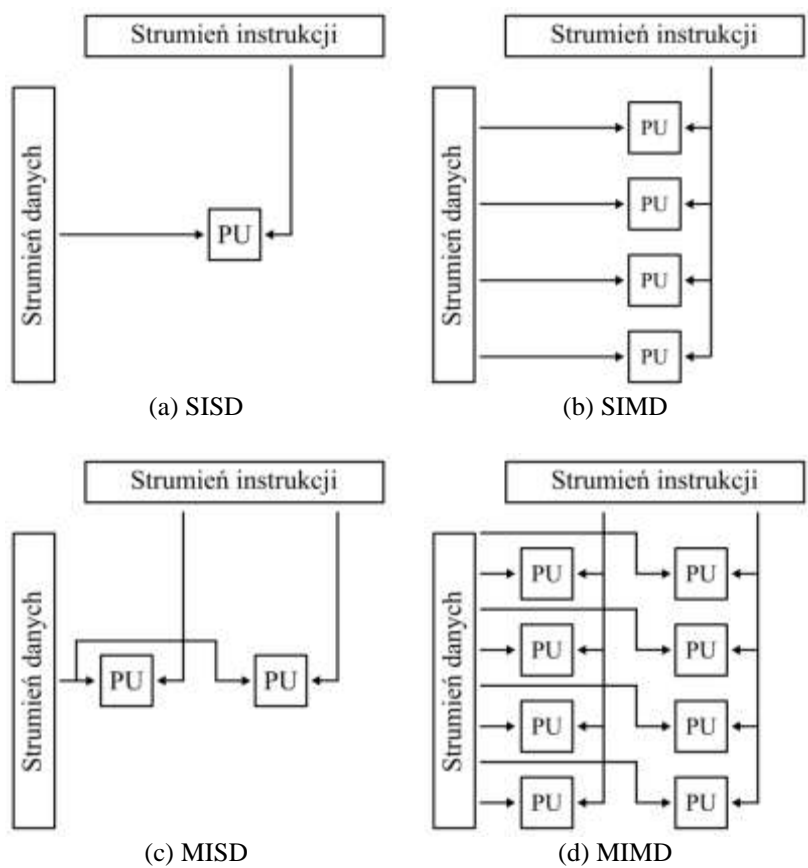

**Rys. 3.1. Architektury komputerowe według klasyfikacji Flynna**

W architekturze SIMD ma miejsce sprzętowe zrównoleglenie danych, dzięki czemu jeden strumień instrukcji realizuje operacje na wielu strumieniach danych, zob. rys. 3.1b. Komputery wektorowe i macierzowe są typowymi reprezentantami architektury SIMD. Rozwiązania zgodne z architekturą SIMD dostępne są w większości współczesnych jednostek centralnych dzięki dedykowanym zestawom instrukcji, w szczególności instrukcji SSE i AVX [117].

Komputery wysokiej dostępności wykorzystują rozwiązania zaczerpnięte z architektury MISD, w której jednostki realizują niezależne strumienie instrukcji, zaś obliczenia realizowane są na identycznym zbiorze danych, zob. rys. 3.1c. Komputerami wysokiej dostępności są m.in. komputery przeznaczone do sterowania lotem promu kosmicznego oraz tablice systoliczne (ang. *systolic arrays*) [117].

Ostatnią z architektur w klasyfikacji Flynna jest architektura MIMD, w której wiele strumieni instrukcji przetwarzać może wiele strumieni danych, zob. rys. 3.1d. Z tego powodu jest ona powszechnie stosowana m.in. w stacjach roboczych i komputerach osobistych. W omawianej architekturze MIMD wykorzystuje się wiele wzajemnie połączonych procesorów, które w zależności od przyjętego rozwiązania przetwarzają dane synchronicznie bądź asynchronicznie.

## Mechanizmy współdzielenia danych

W architekturze MIMD wymagany jest podział zadania na mniejsze fragmenty nazywane podzadaniami. W omawianej architekturze każde podzadanie operuje na własnym strumieniu danych. Zadania niewymagające komunikacji z pozostałymi podzadaniami określane są mianem podzadań niezależnych, zaś zadaniami zależnymi nazywane są podzadania, które wymagają komunikacji z pozostałymi podzadaniami. W zależności od stopnia komunikacji pomiędzy podzadaniami wyszczególnia się klasy:

- równoległości drobnoziarnistej (ang. *fine-grained parallelism*),
- równoległości gruboziarnistej (ang. *coarse-grained parallelism*),
- równoległości zawstydzającej (ang. *embarrassing parallelism*).

Pierwsza grupa równoległości charakteryzuje się znaczącym udziałem komunikacji między podzadaniami oraz stosunkowo niewielkim rozmiarem podzadań. Równoległość gruboziarnista cechuje się dużym rozmiarem podzadania oraz niewielkim udziałem komunikacji między podzadaniami. Ostatnia grupa równoległości dotyczy obliczeń, w których podzadania nie komunikują się ze sobą lub komunikacja jest sporadyczna [117]. Termin równoległość zawstydzająca dotyczy klasy problemów określanych także mianem *embarrassingly parallel* (*perfectly parallel* lub *pleasingly parallel*), które mogą być łatwo zrównoleglone, szczególnie wtedy, gdy nie zachodzi komunikacja między zadaniami lub udział komunikacji jest pomijalny.

## Mechanizmy zrównoleglenia obliczeń

Wzrost liczby tranzystorów w układach scalonych umożliwił wprowadzenie mechanizmów pozwalających na zrównoleglenie obliczeń. Mechanizmy zrównoleglenia obliczeń można podzielić na cztery klasy:

- zrównoleglenie bitowe (ang. *bit-level parallelism*),
- zrównoleglenie instrukcji (ang. *instruction-level parallelism*),

- zrównoleglenie danych (ang. *data parallelism*),
- zrównoleglenie zadań (ang. *task parallelism*).

Jednym z najbardziej widocznych efektów rozwoju architektury sprzętowej jest zwiększenie długości słowa maszynowego, które określa liczbę bitów przetwarzanych w tym samym czasie przez jednostkę arytmetyczno-logiczną. Współczesne ALU mogą przetwarzać zarówno słowa dłuższe, jak i krótsze niż natywnie wspierana długość słowa maszynowego. Podczas przetwarzania słowa dłuższego niż słowo maszynowe wymagane jest wywołanie ciągu instrukcji przetwarzających fragmenty słowa wejściowego, które budują pojedyncze słowa maszynowe. Słowa krótsze niż słowo maszynowe uzupełniane są zerami, zaś sam proces nazywany jest dopełnieniem. W wyniku dopełnienia krótszych słów zerami dostępna moc obliczeniowa ALU nie jest w pełni wykorzystywana. Z tego powodu we współczesnych procesorach stosowane jest zrównoleglenie bitowe, które zamiast dopełniania słowa wejściowego zerami, umieszcza w jednym słowie maszynowym kilka krótszych słów wejściowych. Dzięki temu następuje lepsze wykorzystanie zasobów ALU [117].

Innym rozwiązaniem, które jest powszechnie wykorzystywane w architekturach równoległych jest zrównoleglenie instrukcji [117]. Zrównoleglenie to wykorzystywane jest w trakcie wywoływania złożonych funkcji, które realizowane są w postaci potoku prostych instrukcji. Ponieważ instrukcje uruchamiane przez potok realizują inne zadania, możliwe jest uruchomienie kilku potoków jednocześnie, a co za tym idzie realizacja funkcji złożonej.

Pozostałe dwa mechanizmy, tzn. zrównoleglenie danych i zrównoleglenie instrukcji mają zastosowanie w układach scalonych wyposażonych w więcej niż jeden procesor i ALU. W razie zrównoleglenia zadań każda z jednostek może realizować własny potok instrukcji. Natomiast w przypadku zrównoleglenia danych, jednostki arytmetyczno-logiczne wykorzystują wspólny potok instrukcji. W omawianym rozwiązaniu każda jednostka arytmetyczno-logiczna operuje na innym podzbiorze danych.

## 3.4. Obliczenia równoległe na GPU

Masowe przetwarzanie równoległe MPP (ang. *massively parallel processing*) jest wykorzystywane do skrócenia czasu realizacji złożonych zadań obliczeniowych. Uproszczenie procesu tworzenia i implementacji równoległych algorytmów dla układów wielordzeniowych i wieloprocesorowych, które miało miejsce w ciągu ostatniego dziesięciolecia, spopularyzowało obliczenia równoległe. Zapotrzebowanie na moc obliczeniową doprowadziło do wykorzystania dedykowanych układów graficznych jako efektywnych procesorów obliczeniowych. Podejście to nazwano mianem obliczeń ogólnego przeznaczenia na układach graficznych GPGPU (ang. *General-Purpose computing/computation on Graphics Processor Units*). W odpowiedzi na zapotrzebowanie rynku, producenci układów graficznych dostarczają szeroką gamę produktów przeznaczonych nie tylko do generowania grafiki komputerowej, ale także do realizacji obli-czeń GPGPU. Układy graficzne oferowane przez korporację Nvidia są wspierane zarówno przez architekturę CUDA (ang. *Compute Unified Device Architecture*), jak i API OpenCL [30], podczas gdy karty graficzne firmy ATI wspierane są jedynie przez API OpenCL [74].

W obliczeniach GPGPU fragmenty kodu wykonywane są na układzie graficznym, który nazywany jest urządzeniem (ang. *device*). W trakcie wykonywania zadań procesor centralny, który nazywany jest gospodarzem (ang. *host*) odpowiedzialny jest za nadzorowanie wykonania kodu [30]. Funkcja uruchamiana na układzie graficznym nazywana jest funkcją jądra (ang. *kernel function*), która uruchamia lekkie wątki (ang. *extremely lightweight threads*). Wątki lekkie charakteryzują się niskim kosztem tworzenia oraz minimalnym czasem przełączenia. W typowym rozwiązaniu gospodarz odpowiedzialny jest za dostarczenie danych wejściowych, nadzór nad realizacją zadania oraz odczyt wyników z urządzenia [30]. W takim rozwiązaniu urządzenie odgrywa rolę koprocesora z dedykowaną pamięcią (zob. rys. 3.2).

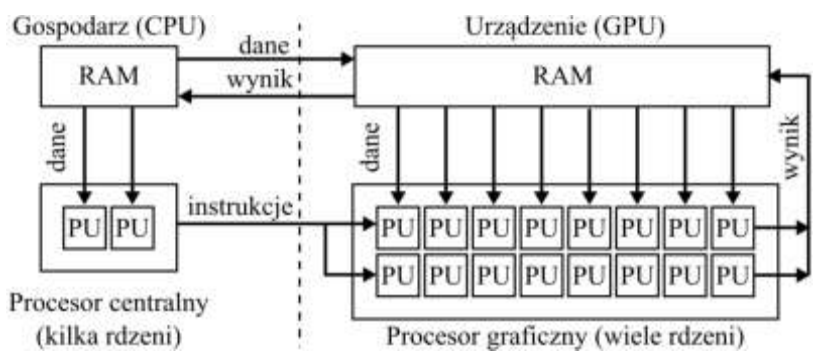

**Rys. 3.2. Przepływ danych w urządzeniach GPGPU**

Funkcjonalność architektury CUDA zależy od wspieranej zgodności obliczeniowej (ang. *Compute Capability*). Zgodność obliczeniowa określa fizyczne cechy układu graficznego firmy Nvidia oraz dostępne funkcje w interfejsie programistycznym CUDA. Zgodność obliczeniowa zapewnia wsteczną kompatybilność kodu, jak również gwarantuje wsparcie interfejsu programistycznego OpenCL [74,117].

## 3.5. Platforma obliczeń równoległych CUDA

Platforma obliczeń równoległych CUDA jest środowiskiem programistycznym umożliwiającym tworzenie aplikacji dla procesorów graficznych firmy Nvidia. Technologia CUDA znalazła zastosowanie w wielu rozwiązaniach wykorzystywanych przez szeroką grupę odbiorców, począwszy od rynku urządzeń mobilnych, a skończywszy na instytucjach badawczych. Układy graficzne umożliwiają realizację niemal dowolnych zadań obliczeniowych, jednakże ich specyficzna budowa ogranicza zakres ich wykorzystania. Wyszczególnić można kilka cech, dzięki którym zadanie może być efektywnie realizowane na układzie graficznym. Najważniejszymi z nich są znaczące nakłady obliczeniowe (ang. *computational intensity*) oraz możliwość zrównoleglenia znaczącej części za-dania, zob. podrozdział 3.1. Grafika komputerowa jest typową reprezentantką zadań, które charakteryzują się wspomnianymi cechami. Zadania posiadające duży udział części równoległej umożliwiają uzyskanie znaczących przyspieszeń obliczeń oraz efektywne wykorzystanie rdzeni układu graficznego.

Ze względu na budowę układy graficzne różnią się znacząco od klasycznych jednostek centralnych. Warto przy tym wspomnieć, że podobnie jak to ma miejsce w wieloprocesorowych komputerach osobistych, jedna karta graficzna może zawierać więcej niż jeden układ graficzny [30]. Współczesne układy graficzne firmy Nvidia zbudowane są z wielu równoległych procesorów wielordzeniowych. W omawianych układach każdy z procesorów może przetwarzać inny strumień danych, tak jak ma to miejsce w architekturze MIMD. Procesorom równoległym odpowiadają wieloprocesory strumieniowe SM (ang. *Streaming Multiprocessor*), które w najnowszych układach graficznych nazywane są wieloprocesorami strumieniowymi nowej generacji SMX (ang. *Next Generation Streaming Multiprocessor*) i SMM (ang. *Streaming Multiprocessor for the Maxwell*). Wieloprocesor zawiera wiele rdzeni strumieniowych (ang. *Streaming Cores*), które w nomenklaturze CUDA nazywane są rdzeniami CUDA (ang. *CUDA Cores*) [30]. Grupy rdzeni w wieloprocesorze mają odpowiedniki w architekturze SIMD, tzn. wiele rdzeni wykonuje tę samą instrukcję na pewnym zbiorze danych. Ze względu na możliwość realizacji zadań obliczeniowych przez wątki oraz ze względu na możliwość realizacji przez grupy wątków różnych zadań, architektura układów graficznych Nvidia określana jest mianem SIMT (ang. *Single Instruction Multiple Threads*) [30].

Podstawowym elementem wieloprocesora strumieniowego są rdzenie CUDA w wariantach 32-bitowych realizujących operacje I32 (ang. *Integer 32-bit*) i FP32 (ang. *Floating Point 32-bit*) oraz 64-bitowych FP64 (ang. *Floating Point 64-bit*). Jeśli procesor nie posiada rdzeni 64-bitowych, wówczas wszystkie obliczenia realizowane są na rdzeniach 32-bitowych, co może jednak skutkować spadkiem wydajności obliczeń równoległych (zob. podrozdział 3.1). Spadek wydajności spowodowany jest koniecznością realizacji operacji na słowie maszynowym dłuższym niż rozmiar 32-bitowego słowa maszynowego rdzenia CUDA. Ponadto Nvidia nakłada dodatkowe ograniczenia sprzętowe na obliczenia. Przykładem może być karta graficzna Nvidia GTX 780 Ti, dla której czas obliczeń zadania operującego na liczbach zmiennoprzecinkowych podwójnej precyzji jest 24-krotnie dłuższy w porównaniu do czasu obliczeń zadania wykorzystującego liczby rzeczywiste pojedynczej precyzji (FP64 = 1/24 FP32) [112]. Karta GTX Titan Black składa się z tej samej liczby rdzeni pojedynczej i podwójnej precyzji co wspomniana wcześniej karta GTX 780Ti. W porównaniu do karty Nvidia GTX 780Ti, karta ta zawiera dwukrotnie więcej pamięci operacyjnej oraz taktowana jest zegarem o minimalnie większej częstotliwości (889 MHz w porównaniu do 876 MHz karty Nvidia GTX 780Ti). Jednakże sterownik karty GTX Titan Black umożliwia ominięcie ograniczenia prędkości, dzięki czemu czas realizacji obliczeń na liczbach podwójnej precyzji jest jedynie trzykrotnie dłuższy w porównaniu do obliczeń z wykorzystaniem reprezentacji zmiennoprzecinkowych pojedynczej precyzji (FP64 = 1/3 FP32) [112].

Współczesne układy graficzne zawierają także:

- wyjściowe jednostki renderujące ROU (ang. *Render Output Units*),
- jednostki mapowania tekstury TMU (ang. *Texture Mapping Units*) wykorzystywane przez ROU,
- jednostki funkcji specjalnych SFT (ang. *Special Function Units*),
- układy wyszukujące jednostki wykonawcze i rejestry dla rozkazu DU (ang. *dispatch units*),
- układy szeregujące (ang. *warp schedulers*).

Tabela 3.2 przedstawia elementy wchodzące w skład wieloprocesora strumienio-wego układu graficznego w zależności od wspieranej zgodności obliczeniowej CUDA. Do wersji 3.0 zgodności obliczeniowej CUDA projektanci układów graficznych koncentrowali się na zwiększeniu liczby rdzeni oraz jednostek obliczeniowych (zob. kolumny 1–4 w tabeli 3.2). W późniejszych wersjach większa część uwagi poświęcona została zwiększeniu wydajności obliczeniowej układów i zmniejszeniu poboru energii. W wyniku kompromisu pomiędzy wydajnością obliczeniową i efektywnością energetyczną (ang. *energy efficiency*) liczba rdzeni została zmniejszona przy zachowaniu liczby układów szeregujących i jednostek DU, zob. kolumny 6–7 w tabeli 3.2. Typowy wieloprocesor zawiera także kilka typów dedykowanej pamięci, które zostaną omówione w podrozdziale 3.7 poświęconym hierarchii wątków i pamięci.

**Tabela 3.2. Specyfikacja wieloprocesora strumieniującego w zależności od wspieranej zgodności obliczeniowej**

| | Zgodność obliczeniowa CUDA | | | | | | |
|---|---|---|---|---|---|---|---|
| | 1.0–3.0 | 2.0 | 2.1 | 3.0 | 3.5 | 5.0 | 5.2 |
| **ALU (rdzenie) I32/FP32** | 8 | 32 | 48 | 192 | 192 | 128 | 128 |
| **Jednostki SFU** | 2 | 4 | 8 | 32 | 32 | 32 | 32 |
| **Jednostki TMU na ROP** | 2 | 4 | 8 | 16 | 16 | 8 | 8 |
| **Układy szeregujące** | 1 | 2 | 2 | 4 | 4 | 4 | 4 |
| **Jednostki DU** | 1 | 1 | 1–2 | 2 | 2 | 2 | 2 |

Wersja zgodności obliczeniowej CUDA dla wybranego procesora graficznego zależy od mikroarchitektury sprzętowej układu, zob. tabela 3.3. Numer wersji zgodności obliczeniowej określa liczbę dostępnych rdzeni, zob. tabela 3.2, oraz inne kluczowe parametry układu takie jak liczba rejestrów, liczba układów szeregujących itp. [30]. Począwszy od mikroarchitektury Tesla, układy graficzne firmy Nvidia wspierają technologię CUDA.

**Tabela 3.3. Wersje zgodności obliczeniowej dla mikroarchitektur Nvidia**

| Mikroarchitektura | Zgodność obliczeniowa CUDA | | | | | | | | |
|---|---|---|---|---|---|---|---|---|---|
| Tesla | 1.0 | 1.1 | 1.2 | 1.3 | | | | | |
| Fermi | | | | | 2.0 | 2.1 | | | |
| Kepler | | | | | | | 3.0 | 3.5 | 3.7 | |
| Maxwell | | | | | | | | | | 5.0 | 5.2 |

Mikroarchitektura Tesla wprowadziła ujednolicony model programów cieniujących (ang. *unified shader model*), który umożliwia realizację przetwarzania wierzchołków i fragmentów na tych samych jednostkach obliczeniowych. Poprzednio stosowany dedykowany model programów cieniujących posiadał osobne jednostki przetwarzania wierzchołków i fragmentów [131]. W praktyce oznaczało to, że w procesie renderingu mogło dochodzić do sytuacji, w których część jednostek obliczeniowych nie była wykorzystywana w czasie całego procesu generowania grafiki, zob. rys. 3.3a. W modelu ujednoliconym jednostki przetwarzające mogą być wykorzystywane zarówno przez programy cieniowania wierzchołków, jak i jednostki cieniowania fragmentów, zob. rys. 3.3b. Dzięki temu osiągnięto wzrost wydajności układów graficznych oraz zmniejszono opóźnienia w renderingu grafiki. Jak można zauważyć na wspomnianym rysunku, wzrost wydajności osiągany jest przez bardziej równomierny przydział jednostek cieniujących dla zadań cieniowania. W mikroarchitekturze Tesla wprowadzono także operacje atomowe oraz możliwość realizacji obliczeń na reprezentacji zmiennoprzecinkowej podwójnej precyzji.

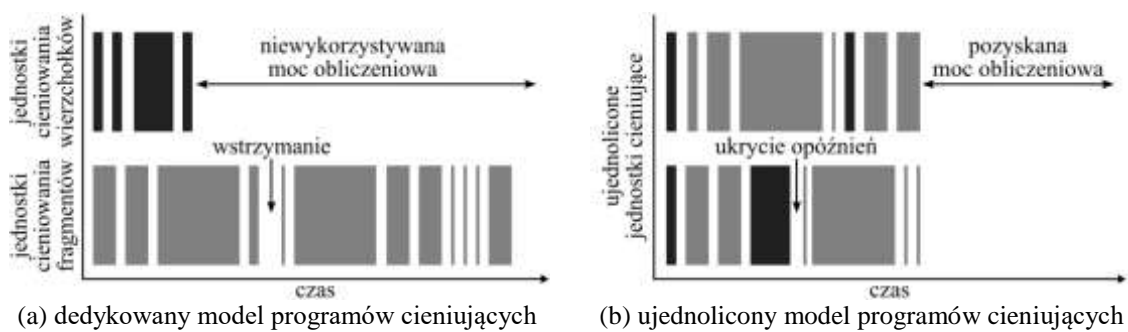

(a) dedykowany model programów cieniujących    (b) ujednolicony model programów cieniujących

**Rys. 3.3. Porównanie dedykowanego i ujednoliconego modelu programów cieniujących**

Następcą Tesli była mikroarchitektura Fermi [30], która posiadała lepszą wydajność. Dzięki optymalizacji instrukcji oraz zgodności ze standardem *IEEE 754-2008* polepszono dokładność obliczeń na liczbach zmiennoprzecinkowych. Ogólna wydajność układu zwiększona została przez zastosowanie złącza PCI Express. Wprowadzenie 64-bitowego adresowania pamięci operacyjnej DDR pozwoliło na znaczne zwiększenie dostępnej pamięci. W układach graficznych wykorzystujących mikroarchitekturę Fermi wprowadzono także jednostki odczytu i zapisu umożliwiające szybszy dostęp do pamię-ci operacyjnej. Wspomniana mikroarchitektura wprowadziła do technologii CUDA także nowy zestaw instrukcji, m.in. rozszerzający zestaw operacji atomowych [30].

Wraz z wprowadzeniem mikroarchitektury Kepler [30], która zastąpiła mikroarchitekturę Fermi, wprowadzono nowy wieloprocesor nazywany SMX. W wieloprocesorze wprowadzono nowy zbiór instrukcji dla układu szeregującego oraz równoległość dynamiczną (ang. *Dynamic Parallelism*). Wprowadzono także technologię Hyper-Q oraz technologię Nvidia GPUDirect [30]. Wieloprocesory układów Kepler są znacząco efektywniejsze, jeśli chodzi o zużycie energii. Obniżenie poboru energii uzyskano przez wprowadzenie ujednoliconego zegara, który zastąpił zegar rdzenia i zegar jednostek obliczeniowych programów cieniujących. Mniejsze zużycie energii pozwoliło na zwiększenie rozmiaru pamięci oraz podwojenie liczby rdzeni realizujących obliczenia na liczbach całkowitych i zmiennoprzecinkowych pojedynczej precyzji. Wprowadzono także dedykowane rdzenie do obliczeń na liczbach zmiennoprzecinkowych podwójnej precyzji. W rdzeniach układu wieloprocesora zastosowano także zrównoleglenie instrukcji, dzięki któremu uzyskano stałe opóźnienie w trakcie wyznaczaniu wartości funkcji matematycznych [30].

Aktualnie w układach graficznych Nvidia stosowana jest mikroarchitektura Maxwell [30], wykorzystująca wieloprocesor SMM. W omawianym wieloprocesorze liczba rdzeni została zredukowana o jedną trzecią, zob. tabela 3.2. Pomimo zredukowania liczby rdzeni, wieloprocesor zapewnia 90% wydajności obliczeniowej swojego poprzednika. Wprowadzone zostało także natywne wsparcie operacji atomowych na pamięci współdzielonej. Wspomniane zagadnienie przybliżono w podrozdziale 3.7.

Mikroarchitektura Pascal, nazywana wcześniej mikroarchitekturą Volta, powinna do 2016 roku zastąpić mikroarchitekturę Maxwell. Wprowadzi ona ujednolicony model pamięci (ang. *unified memory model*), pamięć 3D oraz technologię NVLink umożliwiającą szybszą wymianę danych pomiędzy układem graficznym a procesorem centralnym [112]. Istotną różnicą układu graficznego w stosunku do jednostki centralnej będzie liczba dostępnych rdzeni oraz liczba wątków realizowanych jednocześnie.

## 3.6. Platformy OpenCL i OpenGL

Platforma Nvidia CUDA wykorzystywana może być do programowania aplikacji dla kart firmy Nvidia. Oznacza to, że aplikacje CUDA nie są przenośne na inne architektury GPU. Lukę tę wypełnia API OpenCL (ang. *Open Computing Language*) i OpenGL.

OpenCL wykorzystywany może być do programowania aplikacji uruchamianych na CPU, GPU, FPGA i DSP. OpenCL jest platformą programistyczną do programowania aplikacji z wykorzystaniem równoległości zadań i danych, zob. podrozdział 3.3. W odróżnieniu od Nvidia CUDA, OpenCL zapewnia przenośność aplikacji między układami graficznymi różnych producentów, o ile producent wspiera określony standard OpenCL i dostarcza stosowną bibliotekę programistyczną. OpenCL wspierany jest przez wiele języków programowania, jednak w każdym z nich funkcje jądra przygotowywane są w języku OpenCL, którego składnia oparta jest na standardzie C99 języka C.

W odróżnieniu od CUDA, funkcje jądra są dostarczane w postaci kodu źródłowego, kompilowanego w czasie wykonania [74]. Obliczenia realizowane są zgodnie z abstrakcyjną architekturą OpenCL, zob. rys. 3.4. W omawianej architekturze wyróżnić można urządzenie obliczeniowe CD (ang. *Compute Device*) zbudowane z wielu jednostek obliczeniowych CU (ang. *Compute Unit*). Jednostki obliczeniowe mogą współpracować ze sobą, wykorzystując pamięć współdzieloną. Współpraca jest możliwa jedynie pomiędzy jednostkami obliczeniowymi pracującymi w ramach jednego układu obliczeniowego. Natomiast wszystkie układy obliczeniowe posiadają dostęp do wspólnej pamięci operacyjnej. Obsługa pamięci współdzielonej i pamięć operacyjna zostaną przybliżone w podrozdziale 3.7.

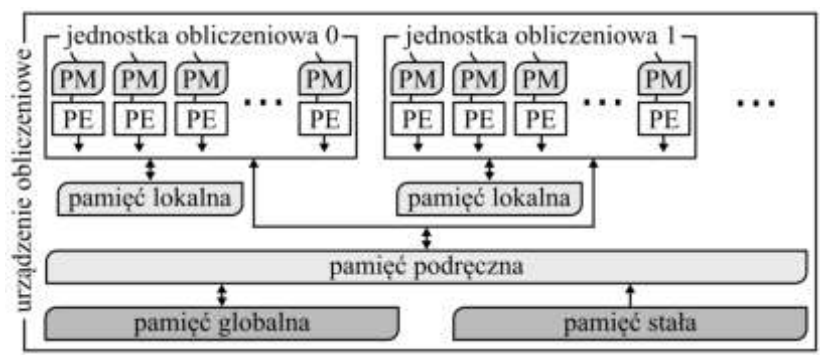

**Rys. 3.4. Schemat abstrakcyjnej architektury OpenCL**

Podstawowym zastosowaniem OpenGL jest tworzenie aplikacji graficznych, w szczególności aplikacji typu CAD. Wraz z wprowadzeniem programowalnego potoku graficznego pojawiła się możliwość zastosowania OpenGL do realizacji obliczeń. Pojawiła się także sposobność tworzenia dedykowanych programów cieniujących umożliwiających realizację obliczeń. Program cieniowania obliczeń (ang. *computer shader*), wykorzystuje dedykowany potok obliczeń. Omawiany potok wykorzystywany jest do wykonywania krótkich programów obliczeniowych. Funkcjonalność programów cieniowania obliczeń zbliżona jest do funkcjonalności dostępnej w standardzie OpenCL.

## 3.7. Hierarchie wątków i pamięci

### Hierarchia wątków w CUDA

Hierarchią wątków na platformie CUDA nazywana jest struktura hierarchiczna, w której grupowane są wszystkie wątki realizujące dane zadanie. Najniższym elementem hierarchii jest *warp*, stanowiący grupę jednocześnie uruchamianych wątków [30,112]. Wątki pracujące w ramach *warp* zgrupowane są w bloki (ang. *blocks*), które z kolei budują siatkę (ang. *grid*), będącą najwyższym elementem w hierarchii. *Warp* stanowi grupę 32 sekwencyjnych wątków uruchamianych przez wieloprocesor [30,112]. Wątki zgrupowane w *warp* dzielone są na dwie klasy, nazywane wątkami aktywnymi (ang. *active threads*) i wątkami nieaktywnymi (ang. *inactive threads*). Wątkami aktyw-nymi są te, które realizują tę samą instrukcję w *warp*, natomiast wątkami nieaktywnymi nazywamy wątki, które oczekują na wykonanie instrukcji. *Warp* wykorzystywany jest do organizacji dostępu do pamięci. Ma to miejsce wtedy, gdy wątki odczytują sekwencyjną pamięć. Przyspieszenie dostępu do pamięci ma miejsce nawet wtedy, gdy operacje realizuje jedynie połowa wątków *warp* (ang. *half-warp*) lub ćwierć wątków w *warp* (ang. *quarter-warp*). Maksymalna wydajność jest uzyskiwana, kiedy ścieżka wykonywania wszystkich wątków w *warp* jest identyczna [30,112,116].

Blok określony jest przez zbiór wątków, podzielonych na grupy *warp*, które przydzielone są do jednego wieloprocesora. Każdy wątek w bloku identyfikowany jest przez pozycję $(x_w, y_w, z_w)$, której wartości mieszczą się w przedziale $x_w \in \langle 0, X_w \rangle$, $y_w \in \langle 0, Y_w \rangle$ i $z_w \in \langle 0, Z_w \rangle$. Wartości $X_w, Y_w, Z_w$ określają rozmiar bloku i są definiowane przy uruchomieniu funkcji jądra. Tabela 3.4 przedstawia ograniczenia nakładane na liczbę wątków budujących blok oraz liczbę wątków w zadanym wymiarze w zależności od wykorzystywanej wersji zgodności obliczeniowej CUDA.

**Tabela 3.4. Maksymalne wymiary bloku wątków CUDA**

| Zgodność obliczeniowa CUDA | 1.0–1.3 | 2.0–5.2 |
|---|---|---|
| Maksymalna liczba wątków w bloku | 512 | 1024 |
| Górna granica wartości $X_w$ i $Y_w$ | 512 | 1024 |
| Górna granica wartości $Z_w$ | 64 | 64 |

Wymiary $X_w, Y_w, Z_w$ służą do organizacji zadania realizowanego w funkcji jądra. Liczba wątków składających się na blok nie może przekroczyć maksymalnej liczby wątków możliwych do uruchomienia na wieloprocesorze. Liczba ta uzależniona jest od zgodności obliczeniowej, która określa także maksymalną liczbę rejestrów oraz rozmiar pamięci lokalnej przydzielanej do wątku (zob. tabela 3.5).

**Tabela 3.5. Ograniczenia liczby wątków uruchamianych w bloku CUDA**

| Zgodność obliczeniowa CUDA | 1.0–1.1 | 1.2–1.3 | 2.0–2.1 | 3.0 | 3.5–5.2 |
|---|---|---|---|---|---|
| Liczba wątków na blok | 768 | 1024 | 1536 | 2048 | 2048 |
| Liczba rejestrów na wątek | 128 | 128 | 63 | 63 | 255 |
| Pamięć lokalna | 16 KB | 16 KB | 512 KB | 512 KB | 512 KB |

Liczbę *warp* uruchamianych przez blok wyznaczyć można z równania:

$$W_{count} = \left\lceil \frac{X_w \cdot Y_w \cdot Z_w}{32} \right\rceil. \tag{3.9}$$

Warto wspomnieć, że jeśli liczba wątków uruchamianych w bloku, określona przez wymiary $X_w, Y_w, Z_w$, nie będzie wielokrotnością 32, to w ostatnim *warp* zostaną zarezerwowane zasoby dla wątków nieaktywnych. Ze względu na to, że wątki w *warp* muszą być wątkami sekwencyjnymi, określić należy transformację pozycji wątku w bloku na jego sekwencyjny identyfikator wątku TID (ang. *Thread Identification*). Identyfikator ten wyznacza się na podstawie równania:

$$TID(x_w, y_w, z_w) = x_w + y_w \cdot X_w + z_w \cdot X_w \cdot Y_w \,. \tag{3.10}$$

Siatka podobnie jak blok służy do grupowania elementów niższego poziomu z hierarchii wątków. Jak już wcześniej wspomniano, bloki służą do grupowania wątków, natomiast siatki spełniają tę samą rolę, z tym że dla bloków. Blok w siatce identyfikowany jest przez pozycję $(x_b, y_b, z_b)$. Podobnie jak dla bloków, przy uruchomieniu funkcji jądra definiowane są maksymalne wartości $X_b, Y_b, Z_b$ (zob. tabela 3.6), które określają górne wartości: $x_b \in \langle 0, X_b \rangle$, $y_b \in \langle 0, Y_b \rangle$ i $z_b \in \langle 0, Z_b \rangle$. Warto podkreślić, że wymiar $z$ nie był dostępny w pierwszych wersjach zgodności obliczeniowej CUDA. Ograniczało to możliwości organizacji zadania funkcji jądra do dwóch wymiarów.

**Tabela 3.6. Ograniczenia liczby wątków w siatce**

| Zgodność obliczeniowa CUDA | 1.0–1.3 | 2.0–2.1 | 3.0–5.2 |
|---|---|---|---|
| **Maksymalny wymiar siatki** | 2 | 3 | 3 |
| **Górna granica wartości $X_b$** | $2^{16} - 1$ | $2^{31} - 1$ | $2^{31} - 1$ |
| **Górna granica wartości $Y_b$ i $Z_b$** | $2^{15} - 1$ | $2^{15} - 1$ | $2^{15} - 1$ |

# Hierarchia pamięci w CUDA

Układ graficzny posiada dostęp do specjalizowanej pamięci operacyjnej nazywanej pamięcią VRAM (ang. *Video Random Access Memory*), która jest niezależna od pamięci procesora centralnego (zob. także rys. 3.2). W trakcie przetwarzania grafiki trójwymiarowej, mechanizmy wbudowane w sterownik układu i sam układ graficzny podejmują decyzje, w jakim rodzaju pamięci dane będą przechowywane. Technologia CUDA dostarcza projektantowi aplikacji mechanizm wspierający zarządzanie pamięcią, dzięki któremu możliwy jest wybór typu pamięci, w której przechowywane będą dane.

Organizację pamięci w układach GPU nazywamy hierarchią pamięci. Na hierarchię składają się:

- pamięć rejestrów (ang. *register memory*),
- pamięć lokalna (ang. *local memory*),
- pamięć współdzielona (ang. *shared memory*),
- pamięć podręczna (ang. *cache memory*),
- pamięć globalna (ang. *global memory*),
- pamięć stała (ang. *constant memory*),
- pamięć tekstur (ang. *texture memory*).

Hierarchia pamięci zbudowana jest wokół cyklu życia elementów pamięci [30,112] oraz zasięgu danej zmiennej (zob. tabela 3.7). Jak można zauważyć w tabeli 3.7, jednostka centralna gospodarza posiada dostęp jedynie do pamięci globalnej, stałej i tekstury, w której przechowywane są dane wejściowe i wyjściowe (zob. także rys. 3.2). Pozostałe typy pamięci mogą być wykorzystywane jedynie przez urządzenie i uruchomioną funkcję jądra.

**Tabela 3.7. Lokalizacja i funkcjonalność pamięci GPU**

| Rodzaj pamięci | Lokalizacja | Buforowanie | Odczyt | Zapis | Zasięg | Cykl życia |
|---|---|---|---|---|---|---|
| Rejestrów | wieloprocesor | - | tak | tak | wątek | wątek |
| Lokalna | układ graficzny | nie | tak | tak | wątek | wątek |
| Współdzielona | wieloprocesor | - | tak | tak | blok | blok |
| Podręczna | wieloprocesor | - | - | - | - | - |
| Globalna | układ graficzny | częściowe | tak | tak | siatka i gospodarz | aplikacja |
| Stała | układ graficzny | tak | tak | nie | siatka i gospodarz | aplikacja |
| Tekstury | układ graficzny | tak | tak | nie | siatka i gospodarz | aplikacja |

Dane przechowywane w pamięci układu graficznego są dostępne dla procedur i funkcji uruchamianych na układzie graficznym oraz funkcji bibliotecznych CUDA. Definiowanie zmiennych, które przechowywane będą w pamięci układu graficznego, odbywa się przez poprzedzenie typu zmiennej odpowiednim specyfikatorem przedstawionym w tabeli 3.8. Wyjątkiem od tej zasady są zmienne alokowane w pamięci lokalnej, pamięci tekstury oraz w szczególnych przypadkach zmienne alokowane w pamięci globalnej. Wspomniane odstępstwa od tej zasady omówiono w dalszej części podrozdziału.

**Tabela 3.8. Specyfikatory pamięci**

| Pamięć | Specyfikator | Obiekt | Metody alokacji | Miejsce alokacji |
|---|---|---|---|---|
| Lokalna | - | - | statyczna, dynamiczna | wątek |
| Współdzielona | __shared__ | - | statyczna, dynamiczna | blok (statyczna), aplikacja (dynamiczna) |
| Globalna | __device__ - | - | statyczna, dynamiczna | aplikacja |
| Stała | __constant__ | - | statyczna | aplikacja |
| Tekstury | - | texture<...> | - | - |

Bezpośredni dostęp do pamięci globalnej, bez wykorzystania pamięci podręcznej, wiąże się z długim czasem dostępu. Wynosi on średnio 400–800 cykli dla układów wspierających zgodność obliczeniową niższą niż 3.0 oraz 200–400 cykli dla zgodności obliczeniowej wyższej lub równej 3.0, zob. tabela 3.9. Czas opóźnienia można zredukować przez umiejętne wykorzystanie pamięci podręcznej oraz zapewnienie odpowiednich mechanizmów dostępu, np. wyrównania pamięci [30,112,177].

Dostęp do pamięci współdzielonej oraz pamięci podręcznej jest znacznie szybszy w porównaniu do pamięci globalnej i wynosi średnio 22 cykle dla zgodności oblicze-niowej niższej niż 3.0 [174], a dla nowszych wersji zgodności obliczeniowej wynosi średnio 11 cykli, zob. tabela 3.9.

**Tabela 3.9. Opóźnienia dostępu do pamięci CUDA**

| Pamięć | Zgodność obliczeniowa CUDA | |
|---|---|---|
| | 1.0–2.1 | 3.0–5.2 |
| Rejestry, Współdzielona | ~22 cykle | ~11 cykli |
| Globalna, Stała, Tekstury | ~400–800 cykli | ~200–400 cykli |

Każdy z prezentowanych typów pamięci projektowany był z myślą o konkretnym przeznaczeniu. Rozmiar pamięci danego typu określany jest przez wspieraną wersję zgodności obliczeniowej, zob. tabela 3.10. Pamięć globalna jest pamięcią ogólnego przeznaczenia i jest ona najczęściej wykorzystywana do przechowywania danych wejściowych oraz wyjściowych. Pamięć globalna układów graficznych w odróżnieniu od klasycznych maszyn obliczeniowych jest stosunkowo niewielka. Typowo, maksymalny rozmiar pamięci VRAM nie przekracza 6 GB. Natomiast rozmiar pozostałych typów pamięci uzależniony jest od wykorzystywanej zgodności obliczeniowej, zob. tabela 3.10. Jak można zauważyć, rozmiar dostępnej pamięci rejestrów oraz współdzielonej wzrasta wraz z wersją zgodności obliczeniowej.

**Tabela 3.10. Rozmiary pamięci wieloprocesora**

| Zgodność obliczeniowa CUDA | 1.0–1.1 | 1.2–1.3 | 2.0–2.1 | 3.0–3.5 | 5.0 | 5.2 |
|---|---|---|---|---|---|---|
| Liczba 32-bitowych rejestrów | 8192 | 16384 | 32768 | 65536 | 65536 | 65536 |
| Rozmiar pamięci współdzielonej [KB] | 16 | 48 | 48 | 48 | 64 | 96 |
| Rozmiar pamięci lokalnej [KB] | 16 | 512 | 512 | 512 | 512 | 512 |
| Rozmiar pamięci stałej [KB] | 64 | 64 | 64 | 64 | 64 | 64 |
| Rozmiar pamięci podręcznej dla pamięci stałej [KB] | 8 | 8 | 8 | 8 | 8 | 10 |
| Rozmiar pamięci podręcznej dla pamięci tekstury [KB] | 6–8 | 6–8 | 12 | 12–48 | 12–48 | 12–48 |

Podstawowym typem pamięci wykorzystywanej przez CUDA jest pamięć współdzielona. Pamięć ta odgrywa dwie ważne role. Pierwszą z nich jest synchronizacja pomiędzy wątkami, które działają w obrębie jednego bloku. Natomiast drugą istotną rolą jest spełnianie funkcji pamięci podręcznej zarządzanej przez programistę. Dzięki takiemu rozwiązaniu uzyskuje się szybszy dostęp do pamięci. Wykorzystanie pamięci współdzielonej jako pamięci podręcznej jest bardzo proste i sprowadza się do:

- wczytania wartości z pamięci globalnej do pamięci współdzielonej,
- synchronizacji pamięci współdzielonej,
- upewnienia się, czy wszystkie elementy zostały wczytane,
- przetworzenia danych w pamięci współdzielonej,
- synchronizacji pamięci współdzielonej,

- upewnienia się, czy wszystkie wątki zapisały wyniki do pamięci,
- zapisania danych z pamięci współdzielonej do pamięci globalnej.

Głównymi zaletami wynikającymi z wykorzystania pamięci współdzielonej jest zmniejszenie liczby odwołań do pamięci globalnej, a w konsekwencji zwiększenie przepustowości pamięci oraz możliwość jednoczesnego odczytu całych bloków pamięci (ang. *coalesced memory access*). Zmienna zdefiniowana w pamięci współdzielonej nie może być zainicjowana w deklaracji zmiennej. Inicjalizacja wartości musi być przeprowadzona przez dany wątek lub grupę wątków. Rozmiar pamięci współdzielonej można deklarować statycznie, przez określenie rozmiaru zmiennej współdzielonej, lub dynamicznie. W gestii programisty jest utworzenie zmiennych lokalnych odwołujących się do odpowiedniego adresu pamięci współdzielonej.

Dostęp do pamięci lokalnej występuje tylko w określonych przypadkach. Decyzję o alokacji zmiennej w pamięci lokalnej podejmuje kompilator, który z reguły umieszcza w tej pamięci zmienne niemieszczące się w rejestrach lub deklarowane dynamicznie. Przykładami takich zmiennych są tablice dynamiczne oraz struktury, które nie mieszczą się w rejestrach.

Pamięć stała, jak nazwa wskazuje, nie zmienia zawartości w trakcie wykonywania funkcji jądra. Oznacza to, że wartości przechowywane w tej pamięci nie mogą ulegać zmianie. Jest to pamięć tylko do odczytu, a jej modyfikacja może odbywać się wyłącznie przez gospodarza. Rozmiar dostępnej pamięci stałej jest stosunkowo niewielki i zazwyczaj wynosi 64 KB. Kopiowanie danych do tej pamięci odbywa się analogicznie jak w przypadku kopiowania danych do pamięci globalnej.

Technologia CUDA wspiera podzbiór funkcji do obsługi tekstur GPU. Tekstura umożliwia odczyt danych z pamięci wykorzystujący mechanizmy pamięci podręcznej. Definicja tekstury może odbywać się przez wysokopoziomowe API. Ponadto do pamięci globalnej możemy odwoływać się pośrednio za pomocą tekstur. Różnią się one znacząco od tekstur wykorzystywanych w grafice trójwymiarowej. Zasadnicza różnica polega na tym, że tekstury CUDA mapują fragment pamięci globalnej, określając jednocześnie, że jest to pamięć wyłącznie do odczytu. Dzięki temu odczyt danych z tego bloku pamięci może odbywać się poprzez użycie pamięci podręcznej i w konsekwencji uzyskuje się skrócenie czasu dostępu do pamięci w porównaniu do czasu dostępu do pamięci globalnej. Podstawową jednostką tekstury jest teksel (ang. *texture element, texel; texture pixel*). Dostęp do tekseli może odbywać się przez koordynaty zmiennoprzecinkowe lub koordynaty znormalizowane. Wykorzystanie koordynat znormalizowanych dla tekstur dwuwymiarowych nazywane jest mapowaniem UV (ang. *UV mapping*), zaś wykorzystanie koordynat tekstur trójwymiarowych nazywane jest mapowaniem UVW (ang. *UVW mapping*). Szerokość, wysokość i głębia tekstury odwołują się do wymiaru tablicy w zadanej osi. Wartość teksela może być reprezentowana wyłącznie przez wartości całkowite lub zmiennoprzecinkowe, lub wektory jedno-, dwu- lub czterowymiarowe tych typów. W przypadku, gdy teksel przechowuje wartości całkowite, możliwe jest przeprowadzenie automatycznej normalizacji. Odczytywane elementy mogą być normalizowane do wartości rzeczywistej z zakresu od zero do jeden dla wartości bez znaku oraz od minus jeden do jeden dla wartości ze znakiem.

## Hierarchia wątków i pamięci w OpenCL

Hierarchia wątków i pamięci OpenCL jest analogiczna do hierarchii wątków i pamięci CUDA. Różnice dotyczą nomenklatury oraz szczegółowości hierarchii. Ponieważ architektura OpenCL jest abstrakcyjna, nie wyróżnia się w niej elementów niskopoziomowych, takich jak pamięć rejestrów, pamięć podręczna. Nie definiuje się także *warp*. W przypadku braku fizycznego elementu architektury jego obsługa realizowana jest programowo. Architektura OpenCL narzuca także ograniczenia na rozmiar dostępnej pamięci danego typu oraz liczbę uruchamianych wątków inne niż architektura CUDA. W konsekwencji część zasobów sprzętowych może nie być w pełni wykorzystywana.

Zadania realizowane w OpenCL wykonywane są przez elementy robocze (ang. *work-items*), które odpowiadają wątkom w technologii CUDA. Wspomniane elementy grupowane są w grupy robocze (ang. *work-groups*) odpowiadające blokom w technologii CUDA [116]. W odróżnieniu od CUDA, nie definiuje się siatki grupującej bloki, zamiast tego definiuje się element NDRange, który określa liczbę uruchamianych elementów roboczych realizujących zadanie funkcji jądra.

Zadania uruchamiane w OpenCL mogą wykorzystywać jeden z czterech typów pamięci: pamięć globalną, pamięć stałą, pamięć lokalną i pamięć prywatną [74]. Pamięć globalna i pamięć stała w OpenCL jest tożsama z pamięcią globalną i pamięcią stałą CUDA. Natomiast pamięć lokalna w OpenCL realizuje zadania pamięci współdzielonej w CUDA. Pamięć prywatna w OpenCL jest natomiast odpowiednikiem pamięci rejestrów i lokalnej w CUDA.

## 3.8. Podsumowanie

W niniejszym rozdziale omówiono metody i narzędzia programowania równoległego. Omówiono miary wydajności obliczeń równoległych, które zastosowane zostały w dalszej części pracy pod kątem lepszego wykorzystania zasobów sprzętowych z jednej strony, a z drugiej strony do osiągnięcia zadanych parametrów śledzenia ruchu 3D. Scharakteryzowano krótko narzędzia i technologie programowania GPU z wykorzystaniem CUDA i OpenCL. Scharakteryzowano układy graficzne w kontekście dostępnych zasobów sprzętowych i możliwości obliczeniowych.

# Rozdział 4
# Model 3D i funkcja celu

Niniejszy rozdział poświęcono omówieniu i przebadaniu modelu 3D oraz funkcji celu. Rozdział składa się z sześciu podrozdziałów, w których przedstawiono najistotniejsze zagadnienia związane z modelowaniem ruchu 3D oraz jego reprezentacją. Pierwszy podrozdział przybliża sposób wyznaczania funkcji celu w oparciu o model 3D. W podrozdziale drugim omówiono szczegółowo trójwymiarowy model postaci wraz z jego graficzną reprezentacją. Trzeci podrozdział poświęcono parametryzacji modelu 3D. W następnym podrozdziale zamieszczono szczegółowy opis rasteryzacji modelu 3D. W podrozdziale piątym zaprezentowano model obserwacji i funkcję celu. Ostatnia część rozdziału zawiera krótkie podsumowanie.

## 4.1. Wprowadzenie

W algorytmach śledzenia ruchu 3D postaci ludzkiej funkcja celu określa, w jakim stopniu rozpatrywana konfiguracja modelu odpowiada konfiguracji obserwowanej sylwetki postaci. W typowym podejściu w funkcji celu odbywa się porównanie cech obserwowanej sylwetki pozyskanych z systemu wizyjnego z cechami sylwetki modelu o zadanej konfiguracji [37,66,153]. Funkcja celu zaproponowana w niniejszej pracy wykorzystuje informacje o sylwetce i krawędziach postaci wydzielonych na obrazach z systemu wizyjnego oraz na obrazach z wyrenderowanym modelem postaci w zadanej konfiguracji, zob. rys. 4.1. Na wspomnianym rysunku zaprezentowano w sposób schematyczny metodę wyznaczania funkcji celu w oparciu o wydzieloną sylwetkę, krawędzie oraz wyrenderowany model 3D. Dla otrzymanych krawędzi postaci ludzkiej wyznaczana jest mapa odległości, która maskowana jest przez krawędzie wyrysowanego modelu, umożliwiając tym samym obliczenia dopasowania krawędzi sylwetki i modelu. Natomiast binarny obraz sylwetki nakładany jest na binarny obraz wyrenderowanego modelu. Obydwa składniki, które reprezentują stopień dopasowania, wykorzystywane są następnie do wyznaczenia wartości funkcji celu.

Obraz sylwetki i krawędzie obserwowanej postaci wyznaczane są w wyniku przetwarzania obrazów wejściowych w oparciu o metody omówione w podrozdziale 2.1. Obraz sylwetki obserwowanej postaci oznaczany jest przez $S^{(c)}$, zaś obraz krawędzi sylwetki oznaczany jest przez $E^{(c)}$, natomiast obraz mapy odległości od krawędzi oznaczany jest przez $D^{(c)}$, gdzie indeks $c \in \langle 0, C \rangle$ identyfikuje numer kamery systemu wizyjnego, zaś $C$ określa liczbę kamer wchodzących w skład systemu wizyjnego. Dla zadanej konfiguracji modelu 3D renderowane są obrazy sylwetki $S^{(c')}$ i krawędzi $E^{(c')}$. Omawiane obrazy generowane są przez rzutowanie modelu 3D na płaszczyznę obrazu w oparciu o wirtualny model kamery parametryzowany przez $c'$ (zob. podrozdział 4.4).

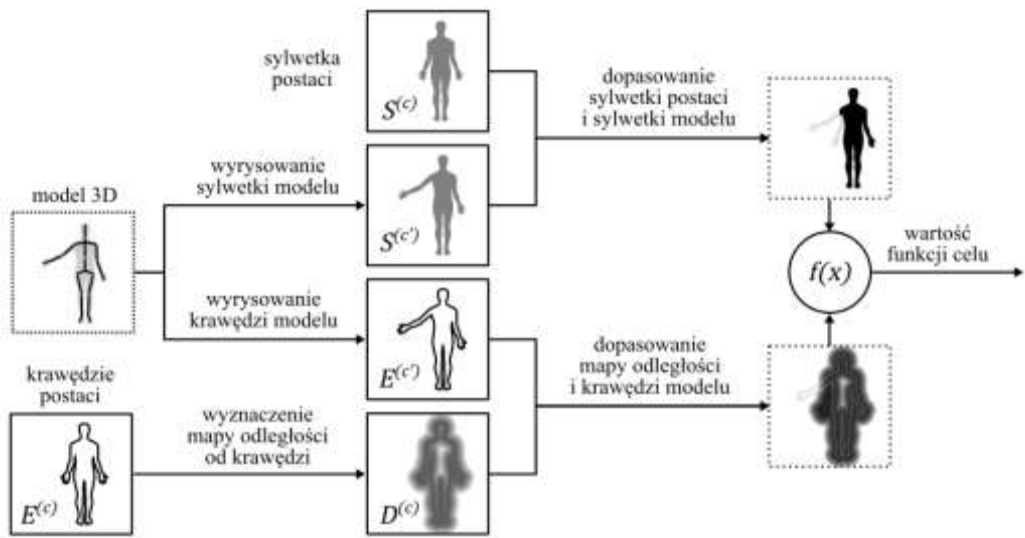

**Rys. 4.1. Wyznaczanie funkcji celu**

W niniejszym podrozdziale zakłada się także, że wszystkie kamery systemu wizyjnego są zsynchronizowane, pracują z tą samą częstotliwością i rozdzielczością, a także posiadają tę samą szerokość $w$ i wysokość $h$ obrazu. Do reprezentacji obrazów wykorzystano notację macierzową. Obraz o szerokości $w$ pikseli i wysokości $h$ pikseli reprezentowany jest przez macierz $I_{w,h} = [i_{x,y}]$, gdzie $i_{x,y}$ jest pikselem obrazu we współrzędnych $x$, y.

## 4.2. Trójwymiarowy model postaci ludzkiej

W niniejszym podrozdziale przedstawiono trójwymiarowy model postaci. Składa się on z modelu szkieletowego postaci ludzkiej (ang. *skeleton model*) opisującego budowę i ruch obiektu w przestrzeni trójwymiarowej oraz modelu kształtu (ang. *skin model*) opisującego wygląd śledzonej postaci [13,119].

### Model szkieletowy

Model szkieletowy jest hierarchiczną strukturą drzewiastą [13,119], która reprezentuje budowę i ruch obiektu. W strukturze modelu wyróżnia się dwa elementy, mianowicie kości (ang. *bones*) i stawy (ang. *joints*), które odpowiadają krawędziom i wierzchołkom klasycznej struktury drzewiastej. Kość w modelu szkieletowym jest elementem posiadającym określoną długość oraz określoną liczbę stopni swobody DoF (ang. *degrees of freedom*). Liczba stopni swobody określa przestrzeń, w której jest realizowany ruch, natomiast staw jest miejscem połączenia co najmniej dwóch kości [119].

Ruch modelu 3D realizowany jest w oparciu o transformacje geometryczne. Ze względu na szereg konwencji wykorzystywanych do opisu ruchu, a także ze względu na obszerność tematyki, problem ten został omówiony bardziej szczegółowo w podrozdziale 4.3. Wyznaczanie transformacji geometrycznych poszczególnych kości modelu realizowane jest w oparciu o zasady kinematyki prostej (ang. *forward kinematics*) [119]. Kinematyka prosta jest jednym z najprostszych sposobów animacji ruchu, ponieważ wykorzystywany model obiektu nie uwzględnia cech fizycznych rzeczywistego obiektu takich jak: masa, moment ruchu oraz tor ruchu [21]. W nomenklaturze kinematyki prostej hierarchiczny model obiektu nazywany jest łańcuchem kinematycznym (ang. *kinematic chain*) [182]. Pojęcie łańcucha kinematycznego i modelu szkieletowego w niniejszej pracy są jednoznaczne. Model szkieletowy zaprezentowany na rys. 4.2b zbudowany jest z 15 kości. Wspomniany model wykorzystywany był we wcześniejszych pracach [138,140,141]. Jest on w pełni konfigurowalny, dzięki czemu może być wykorzystywany do śledzenia zarówno górnych, zob. rys. 4.2a, jak i dolnych części ciała, zob. rys. 4.2c, a także całej postaci. Wspomniany model był wzorowany na modelach wykorzystywanych w pracach [11,78,153].

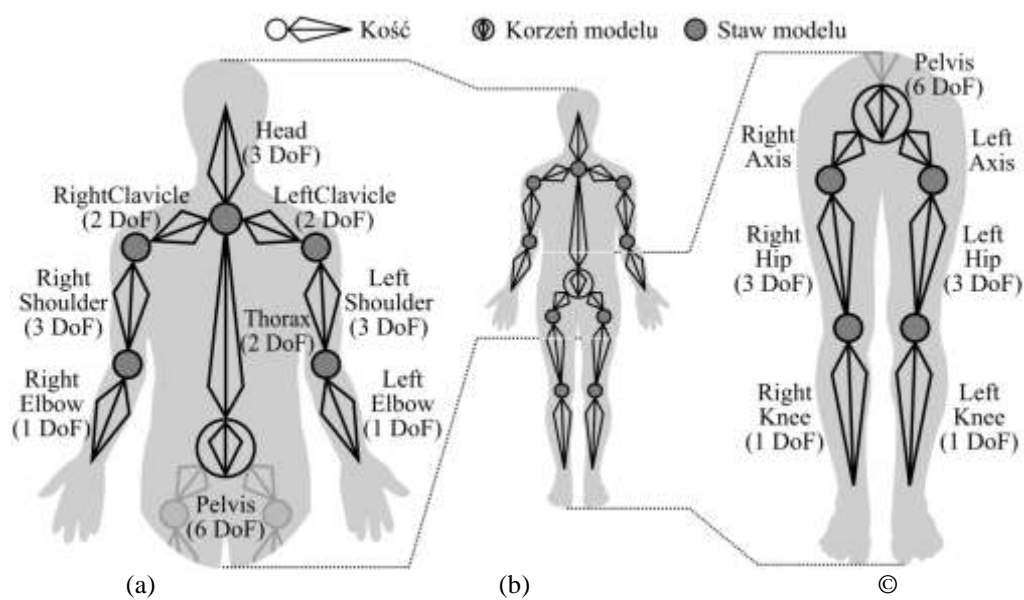

**Rys. 4.2. Model 3D postaci ludzkiej**

W modelu szkieletowym pozycja każdego elementu uzależniona jest bezpośrednio od transformacji obiektu nadrzędnego [13,119]. Wyjątkiem od tej reguły jest korzeń (ang. *root*) struktury hierarchicznej modelu, który nie posiada części nadrzędnej. Transformacja, która nie uwzględnia przekształcenia nadrzędnego elementu nazywana jest transformacją w lokalnym układzie współrzędnych (ang. *local transformation*) bądź transformacją lokalną, zob. rys. 4.3. Reprezentacja uwzględniająca przekształcenia geometryczne części nadrzędnej nazywana jest transformacją w globalnym układzie współrzędnych (ang. *global transformation*) lub skrótowo transformacją globalną, zob. rys. 4.3. Proces wyznaczania transformacji globalnych realizowany z uwzględnieniem struktury hierarchicznej nazywany jest aktualizacją pozy modelu.

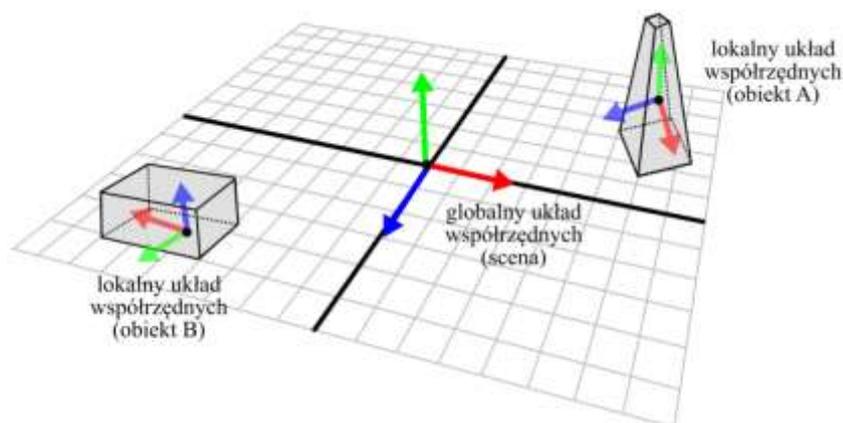

**Rys. 4.3. Globalny układ współrzędnych wraz z układami lokalnymi obiektów**

W trójwymiarowym modelu szkieletowym każdy stopień swobody określa możliwość rotacji wokół zadanej osi lub translacji w osi określonej w lokalnym układzie współrzędnych, zob. rys. 4.3. Tabela 4.1 przedstawia przykładową strukturę hierarchiczną modelu postaci ludzkiej. Prezentowany model postaci posiada 31 stopni swobody i składa się z 15 kości. Identyfikator kości określa relację pomiędzy modelem szkieletowym a modelem kształtu. Warto podkreślić, że jedyną kością na której może być realizowana operacja translacji, jest korzeń modelu. Dzięki temu możliwe jest umieszczenie modelu w zadanej pozycji na scenie. Ograniczenia liczby stopni swobody w poszczególnych kościach eliminują niemożliwe anatomicznie konfiguracje modelu szkieletowego. Ograniczenie zakresu kątów dla ruchu stawów umożliwia jeszcze większą dyskryminację anatomicznie niemożliwych póz. Liczba stopni swobody danej kości określa rozmiar wektora przechowującego wartości rotacji i translacji. Wartości te wykorzystywane są z kolei do wyznaczenia transformacji lokalnej. Transformacje lokalne kości służą do wyznaczenia transformacji globalnych wszystkich elementów modelu. Określone w ten sposób transformacje globalne całego modelu tworzą wektor transformacji globalnych, którego liczba elementów jest równa liczbie kości modelu szkieletowego. Wektor transformacji globalnych wykorzystywany jest przez procedurę rasteryzacji modelu 3D, która omówiona jest bardziej szczegółowo w następnych podrozdziałach.

**Tabela 4.1. Model szkieletowy postaci ludzkiej (31 DoF)**

| Id | Nazwa kości | Element nadrzędny | DoF | Transformacje kości |
|----|-------------|-------------------|-----|---------------------|
| 0 | Pelvis (root) | - | 6 | $t_x, t_y, t_z, r_x, r_y, r_z$ |
| 1 | Thorax | Pelvis | 2 | $r_y, r_z$ |
| 2 | Left Axis | Pelvis | 0 | - |
| 3 | Left Hip | Left Axis | 3 | $r_x, r_y, r_z$ |
| 4 | Left Knee | Left Hip | 1 | $r_y$ |
| 5 | Right Axis | Pelvis | 0 | - |
| 6 | Right Hip | Right Axis | 3 | $r_x, r_y, r_z$ |

**Tabela 4.1 (kontynuacja). Model szkieletowy postaci ludzkiej (31 DoF)**

| Id | Nazwa kości | Element nadrzędny | DoF | Transformacje kości |
|----|-------------|-------------------|-----|---------------------|
| 7 | Right Knee | Right Hip | 1 | $r_y$ |
| 8 | Head | Thorax | 3 | $r_x, r_y, r_z$ |
| 9 | Left Clavicle | Thorax | 2 | $r_x, r_y$ |
| 10 | Left Shoulder | Left Clavicle | 3 | $r_x, r_y, r_z$ |
| 11 | Left Elbow | Left Shoulder | 1 | $r_y$ |
| 12 | Right Clavicle | Thorax | 2 | $r_x, r_y$ |
| 13 | Right Shoulder | Right Clavicle | 3 | $r_x, r_y, r_z$ |
| 14 | Right Elbow | Right Shoulder | 1 | $r_y$ |

# Model kształtu

Model szkieletowy wykorzystywany jest wyłącznie do modelowania ruchu obiektu, zaś wyrysowana struktura szkieletu nazywana jest bardzo często modelem prostym (ang. *stick figure*) [4]. Wyrysowanie bardziej szczegółowej reprezentacji wymaga opisu wyglądu postaci za pomocą struktury, która nazywana jest modelem kształtu [12,13,119]. Omawiany model służy do formalnego opisu wyglądu rzeczywistego obiektu, z kolei jego ruch opisany jest przez omówiony wcześniej model szkieletowy. Model kształtu i model szkieletowy połączone są ze sobą relacją nazywaną mapowaniem kształtu do modelu szkieletowego (ang. *skinning*) [64]. Przyjęta parametryzacja modelu 3D została omówiona bardziej szczegółowo w podrozdziale 4.3.

Tabela 4.2 przedstawia wybrane reprezentacje modelu kształtu, które wykorzystywane były przez inne zespoły badawcze w pracach poświęconych śledzeniu ruchu postaci ludzkiej. W pracach wyszczególnionych w omawianej tabeli modele kształtu reprezentowały wygląd postaci zarówno w przestrzeni dwuwymiarowej (2D), jak i w przestrzeni trójwymiarowej (3D). W modelu 2D wykorzystuje się płaskie figury geometryczne do reprezentacji poszczególnych części ciała modelu, natomiast mianem modelu 3D określamy reprezentację, w której elementy reprezentowane są przez figury przestrzenne. Szczególnym przypadkiem modelu 3D jest reprezentacja części ciała postaci przez figury płaskie umieszczone w przestrzeni trójwymiarowej. Model taki nazywany jest modelem 2,5D [56]. Z prezentowanego zestawienia wynika, że do reprezentacji modeli 2D wykorzystywane były odcinki liniowe, czworokąty, zaokrąglone czworokąty oraz SPM (ang. *Scaled Prismatic Model*). Do reprezentacji modeli 3D wykorzystywane były typowe bryły geometryczne oraz w szczególnych przypadkach figury płaskie.

**Tabela 4.2. Zestawienie reprezentacji modelu kształtu**

| pierwszy autor | rok | przestrzeń | typ modelu |
|----------------|-----|------------|------------|
| Downton [38] | 1992 | 3D | cylindry |
| Gavrila [49] | 1996 | 3D | stożkowe superkwadryki |
| Ju [67] | 1996 | 2D | czworokąty |
| Kakadiaris [68] | 1996 | 3D | siatka wielokątów |

**Tabela 4.2 (kontynuacja). Zestawienie reprezentacji modelu kształtu**

| pierwszy autor | rok | przestrzeń | typ modelu |
|---|---|---|---|
| Bregler [19] | 1998 | 3D | elipsoidy |
| Deutscher [36] | 2000 | 3D | stożki eliptyczne |
| Kakadiaris [68] | 2000 | 3D | siatka wielokątów |
| Rosales [130] | 2000 | 3D | zbudowany z odcinków liniowych (ang. *stick-figure*) |
| Delamarre [35] | 2001 | 3D | stożki ścięte, sfery, prostopadłościany |
| Sidenbladh [151,152] | 2001 | 3D | cylindry |
| Ronfard [129] | 2002 | 2D | czworokąty |
| Grauman [54] | 2003 | 3D | kształtu |
| Shakhnarovich [148] | 2003 | 3D | siatka wielokątów |
| Sminchisescu [157,158] | 2003 | 3D | elipsoidalne superkwadryki |
| Agarwal [3] | 2004 | 2D | zaokrąglone czworokąty |
| Agarwal [1] | 2004 | 3D | siatka wielokątów |
| Sigal [154] | 2004 | 3D | stożki eliptyczne |
| Balan [11] | 2005 | 3D | stożki ścięte |
| Deutscher [36] | 2005 | 3D | stożki eliptyczne |
| Ren [127] | 2005 | 2D | zbudowany z odcinków liniowych (ang. *stick-figure*) |
| Gal [47] | 2006 | 3D | siatka wielokątów |
| Lee [90] | 2006 | 3D | stożki eliptyczne |
| Schmidt [143] | 2006 | 3D | cylindry |
| Sigal [155] | 2006 | 2D | czworokąty |
| Wang [175] | 2006 | 2D | SPM |
| Balan [12] | 2007 | 3D | siatka wielokątów (SCAPE[2]) |
| Mundermann [107] | 2007 | 3D | siatka wielokątów (SCAPE) |
| Peursum [122] | 2007 | 3D | cylindry |
| Ivekovic [62] | 2008 | 3D | siatka wielokątów |
| Raskin [125,126] | 2008 | 3D | stożki eliptyczne |
| Shaheen [147] | 2009 | 3D | siatka wielokątów |
| Krzeszowski [79,80] | 2010 | 3D | prostopadłościany |
| John [66] | 2010 | 3D | stożki ścięte |
| Sigal [153] | 2010 | 3D | stożki ścięte |
| Zhang [185] | 2010 | 3D | stożki ścięte |
| Krzeszowski [81,78] | 2011 | 3D | stożki ścięte |

Typowymi figurami wykorzystywanymi do reprezentacji modelu 2D, nazywanego także modelem płaskim, są czworokąty, elipsy oraz okręgi. Model 3D, w którym każda część ciała reprezentowana jest przez niezależne od siebie obiekty trójwymiarowe, nazywamy modelem 3D zbudowanym z siatek (ang. *mesh model*) [13,119]. Obiekty przestrzenne w grafice komputerowej reprezentowane są z reguły przez siatki (ang. *mesh*). Siatkę tworzą wierzchołki (ang. *vertices*) oraz krawędzie (ang. *edges*). Zbiór krawędzi służy do definicji ścian (ang. *faces*) obiektu, które utożsamiane są z wielokątami (ang. *polygons*) budującymi powierzchnie (ang. *surfaces*), zob. rys. 4.4. Na wspomnianym rysunku przedstawiono przykładową reprezentację siatki obiektu, która zbudowana została z ośmiu trójkątów. Alternatywnie omawiany obiekt przestrzenny można przedstawić za pomocą czterech czworokątów.

---

[2] SCAPE – *Shape Completion and Animation of People*

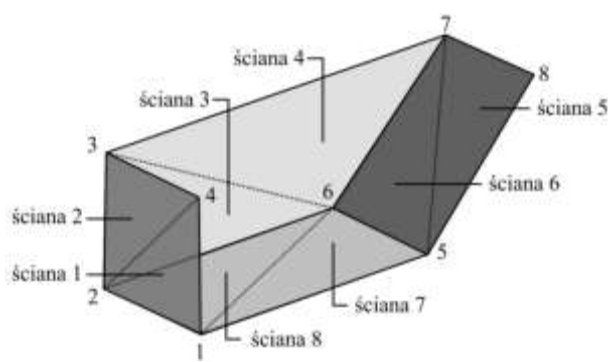

**Rys. 4.4. Reprezentacja obiektu przestrzennego w postaci siatki**

Do reprezentacji poszczególnych elementów postaci wykorzystywane są często stożki ścięte lub prostopadłościany [11,36,78,90,144]. Model 3D, który reprezentuje całą postać przez jedną zamkniętą bryłę, nazywamy modelem siatkowym z pozą ustaloną [106]. Omawiane modele są ze sobą wzajemnie powiązane: model siatkowy z pozą ustaloną po rozłożeniu na poszczególne części reprezentujące pojedyncze kości przyjmie postać model kształtu zbudowanego z siatek. Natomiast uproszczenie modelu zbudowanego z siatek przez rzuty wybranych części na płaszczyznę obrazu przyjmie postać modelu płaskiego [140]. Wspomniane modele wykorzystywane są także w grafice komputerowej do różnych celów, m.in. do renderingu postaci wykorzystywany jest model siatkowy z pozą ustaloną [98]. Model złożony z siatek wykorzystywany jest do reprezentacji powierzchni podczas detekcji kolizji w przestrzeni trójwymiarowej [5], natomiast model płaski wykorzystywany jest do detekcji kolizji w przestrzeni dwuwymiarowej [5].

Modelem płaskim nazywamy model, w którym każda część ciała jest reprezentowana na płaszczyźnie dwuwymiarowej przez figury płaskie. Kształt omawianych figur powinien być zbliżony do kształtu modelowanych części ciała. Ruch modelu może odbywać się zarówno w przestrzeni dwuwymiarowej, jak i trójwymiarowej (zob. tabela 4.2). Główną zaletą omawianego modelu jest przejrzystość oraz znacznie prostsza rasteryzacja w porównaniu do rasteryzacji modelu 3D. Bardziej szczegółowe omówienie budowy modelu płaskiego znajduje się na stronie 69.

## 4.3. Parametryzacja modelu 3D

Parametryzacją modelu 3D nazywamy sposób reprezentacji orientacji i pozycji obiektów w przestrzeni trójwymiarowej oraz sposób zapisu modelu kształtu w pamięci. Do określenia orientacji i pozycji modelu wykorzystuje się układ kartezjański. Miejsce przecięcia trzech wzajemnie prostopadłych osi układu kartezjańskiego określa środek układu współrzędnych świata, który reprezentuje także punkt referencyjny sceny. W zależności od wykorzystywanej reprezentacji stosowany jest układ prawoskrętny (ang. *right-handed*), zob. rys. 4.5a, w którym rotacja realizowana jest zgodnie z regułą prawej ręki lub układ lewoskrętny (ang. *left-handed*), zob. rys. 4.5b, w którym rotacja wykonywana jest zgodnie z regułą lewej ręki [173].

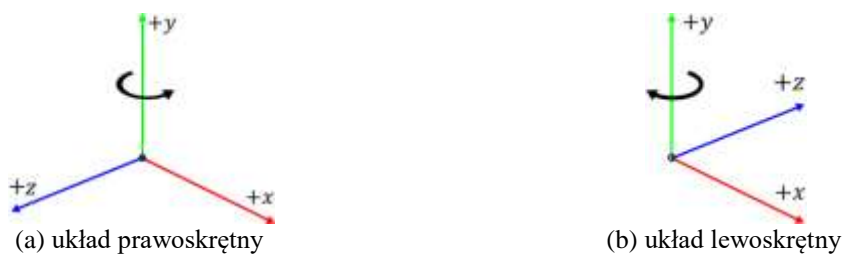

(a) układ prawoskrętny        (b) układ lewoskrętny

**Rys. 4.5. Prawoskrętny i lewoskrętny układ współrzędnych**

W niniejszej pracy przekształcenia geometryczne realizowane są w prawoskrętnym kartezjańskim układzie współrzędnych, którego oś $x$ skierowana jest w prawo, natomiast oś $y$ skierowana jest w górę. Wspomniana konwencja wykorzystywana jest przez markerowy system mocap [65]. Celem uniknięcia zbędnych transformacji między układami współrzędnych, w niniejszej pracy również wykorzystywany został układ prawoskrętny. Spotykane są także układy współrzędnych, w których kierunek osi odbiega od przyjętej w pracy konwencji. Przykładem może być interfejs OpenGL, w którym oś $y$ układu współrzędnych okna skierowana jest ku dołowi [113].

## Macierze jednorodne

Najczęściej wykorzystywaną reprezentacją transformacji geometrycznych są macierze jednorodne o wymiarze 4x4. Macierz jednorodna (4.1) umożliwia realizację transformacji geometrycznych takich jak przesunięcie w danej osi układu współrzędnych ($t_{1\ldots3}$), zob. rys. 4.6b, rotacja wokół osi układu współrzędnych ($r_{1\ldots3,1\ldots3}$), zob. rys. 4.6c, a także skalowanie, zob. rys. 4.6d, i skręcenie ($s_{1\ldots3}$).

$$M = \begin{bmatrix} r_{1,1} & r_{1,2} & r_{1,3} & t_1 \\ r_{2,1} & r_{2,2} & r_{2,3} & t_2 \\ r_{3,1} & r_{3,2} & r_{3,3} & t_3 \\ s_1 & s_2 & s_3 & 1 \end{bmatrix} \tag{4.1}$$

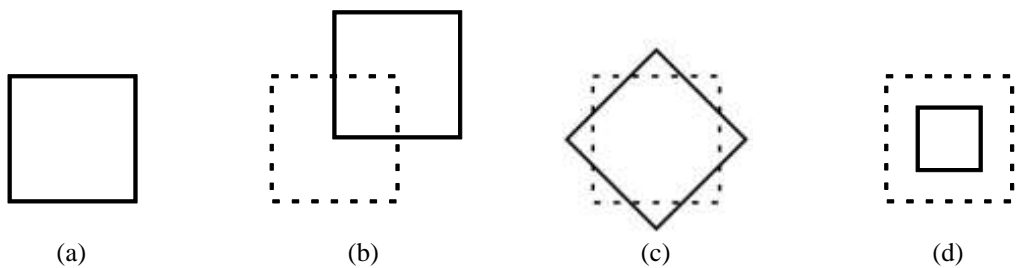

(a)      (b)      (c)      (d)

**Rys. 4.6. Podstawowe transformacje geometryczne**

Jednostkowa macierz jednorodna $M_I$ (4.2) służy do reprezentacji transformacji obiektu, który nie został poddany żadnemu przekształceniu geometrycznemu, zob. rys. 4.6a.

$$M_I = \begin{bmatrix} 1 & 0 & 0 & 0 \\ 0 & 1 & 0 & 0 \\ 0 & 0 & 1 & 0 \\ 0 & 0 & 0 & 1 \end{bmatrix} \tag{4.2}$$

Macierz jednorodna reprezentuje operację geometryczną rotacji o zadany kąt α wokół osi $x$ układu współrzędnych. Macierz ta określona jest za pomocą równania (4.3) dla układu prawoskrętnego i równania (4.4) dla układu lewoskrętnego.

$$R_x^p = \begin{bmatrix} 1 & 0 & 0 & 0 \\ 0 & cos(\alpha) & -sin(\alpha) & 0 \\ 0 & sin(\alpha) & cos(\alpha) & 0 \\ 0 & 0 & 0 & 1 \end{bmatrix} \tag{4.3}$$

$$R_x^l = \begin{bmatrix} 1 & 0 & 0 & 0 \\ 0 & cos(\alpha) & sin(\alpha) & 0 \\ 0 & -sin(\alpha) & cos(\alpha) & 0 \\ 0 & 0 & 0 & 1 \end{bmatrix} \tag{4.4}$$

Rotacja o zadany kąt $\beta$ wokół osi $y$ układu współrzędnych reprezentowana jest przez macierz jednorodną, która przyjmuje postać (4.5) dla układu prawoskrętnego oraz postać (4.6) dla układu lewoskrętnego.

$$R_y^p = \begin{bmatrix} cos(\beta) & 0 & sin(\beta) & 0 \\ 0 & 1 & 0 & 0 \\ -sin(\beta) & 0 & cos(\beta) & 0 \\ 0 & 0 & 0 & 1 \end{bmatrix} \tag{4.5}$$

$$R_y^l = \begin{bmatrix} cos(\beta) & 0 & -sin(\beta) & 0 \\ 0 & 1 & 0 & 0 \\ sin(\beta) & 0 & cos(\beta) & 0 \\ 0 & 0 & 0 & 1 \end{bmatrix} \tag{4.6}$$

Rotacja o kąt $\gamma$ wokół osi $z$ układu współrzędnych reprezentowana jest przez macierz (4.7) dla układu prawoskrętnego i macierz (4.8) dla układu lewoskrętnego.

$$R_z^p = \begin{bmatrix} cos(\gamma) & -sin(\gamma) & 0 & 0 \\ sin(\gamma) & cos(\gamma) & 0 & 0 \\ 0 & 0 & 1 & 0 \\ 0 & 0 & 0 & 1 \end{bmatrix} \tag{4.7}$$

$$R_z^l = \begin{bmatrix} cos(\gamma) & sin(\gamma) & 0 & 0 \\ -sin(\gamma) & cos(\gamma) & 0 & 0 \\ 0 & 0 & 1 & 0 \\ 0 & 0 & 0 & 1 \end{bmatrix} \tag{4.8}$$

Macierzą transponowaną nazywamy macierz, która powstała przez zamianę elementów macierzy $m_{i,j}$ $i$-tego wiersza i $j$-tej kolumny na elementy $m_{j,i}$ $j$-wiersza i $i$-tej kolumny (4.9). Przy wykorzystaniu operacji transpozycji macierzy, możliwe jest prze-kształcenie macierzy rotacji z układu prawoskrętnego do macierzy układu lewoskrętne-go $R^p = (R^l)'$, jak również przekształcenie macierzy rotacji z układu lewoskrętnego do macierzy układu prawoskrętnego $R^l = (R^p)'$.

$$\begin{bmatrix} m_{1,1} & m_{1,2} & m_{1,3} & m_{1,4} \\ m_{2,1} & m_{2,2} & m_{2,3} & m_{2,4} \\ m_{3,1} & m_{3,2} & m_{3,3} & m_{3,4} \\ m_{4,1} & m_{4,2} & m_{4,3} & m_{4,4} \end{bmatrix}' = \begin{bmatrix} m_{1,1} & m_{2,1} & m_{3,1} & m_{4,1} \\ m_{1,2} & m_{2,2} & m_{3,2} & m_{4,2} \\ m_{1,3} & m_{2,3} & m_{3,3} & m_{4,3} \\ m_{1,4} & m_{2,4} & m_{3,4} & m_{4,4} \end{bmatrix} \tag{4.9}$$

Realizacja przesunięcia o wartość $t_x$ dla osi $x$, $t_y$ dla osi $y$, $t_z$ dla osi $z$, która jest także określana jako przesunięcie obiektu o wektor $t = [t_x \ t_y \ t_z]$, jest operacją nieza-leżną od kierunku skrętu układu współrzędnych i przyjmuje postać macierzy jednorod-nej $T$, przedstawionej za pomocą równania:

$$T = \begin{bmatrix} 1 & 0 & 0 & t_x \\ 0 & 1 & 0 & t_y \\ 0 & 0 & 1 & t_z \\ 0 & 0 & 0 & 1 \end{bmatrix}. \tag{4.10}$$

Operacja skalowania obiektu przez współczynnik $s_x$ w osi $x$, $s_y$ w osi $y$ oraz $s_z$ w osi $z$ reprezentowana jest przez macierz jednorodną $S$, opisaną za pomocą równania:

$$S = \begin{bmatrix} s_x & 0 & 0 & 0 \\ 0 & s_y & 0 & 0 \\ 0 & 0 & s_z & 0 \\ 0 & 0 & 0 & 1 \end{bmatrix}. \tag{4.11}$$

Sekwencja przekształceń geometrycznych złożona z dowolnej liczby transformacji określona jest przez operację mnożenia macierzy. Czynniki występujące w sekwencji przekształceń muszą przedstawiać transformacje geometryczne w tym samym układzie współrzędnych. Złożenie sekwencji trzech rotacji wokół osi $z$, $x$, $y$ dla układu prawo-skrętnego $p$ reprezentowane jest przez macierz określoną równaniem:

$$R^p_{zxy} = R^p_z \cdot R^p_x \cdot R^p_y \ . \tag{4.12}$$

Mnożenie macierzy nie jest operacją przemienną, wskutek czego łączenie transfor-macji geometrycznych także nią nie jest [173]. Oznacza to, że macierz transformacji $R_{zxy}$ nie jest równa macierzy rotacji $R_{yxz}$. Transformacją odwrotną dowolnego złożenia przekształceń geometrycznych reprezentowanych przez macierz $M$ jest macierz od-wrotna $M^{-1}$ . W przypadku, gdy sekwencja transformacji składa się wyłącznie z przesunięcia i rotacji, macierz transformacji odwrotnej wyznacza się zgodnie z poniższym równaniem [98]:

$$\begin{bmatrix} r_{1,1} & r_{1,2} & r_{1,3} & t_1 \\ r_{2,1} & r_{2,2} & r_{2,3} & t_2 \\ r_{3,1} & r_{3,2} & r_{3,3} & t_3 \\ 0 & 0 & 0 & 1 \end{bmatrix}^{-1} = \begin{bmatrix} R & t \\ 0 & 1 \end{bmatrix}^{-1} = \begin{bmatrix} R' & -R't \\ 0 & 1 \end{bmatrix} . \qquad (4.13)$$

W systemach komputerowych macierze przechowywane są w pamięci zorganizowanej w sposób liniowy. Kolejność przechowywania elementów macierzy w pamięci nazywana jest notacją macierzową. Wyróżnia się dwie notacje: wierszową (ang. *row-major*) oraz kolumnową (ang. *column-major*). W OpenGL stosowana jest notacja kolumnowa. Wzory (4.14) i (4.15) przedstawiają reprezentację macierzy $M$ (4.1) w notacji wierszowej $M_{row}$ i w notacji kolumnowej $M_{col}$.

$$M_{row} = [r_{1,1}, r_{1,2}, r_{1,3}, t_1, r_{2,1}, r_{2,2}, r_{2,3}, t_2, r_{3,1}, r_{3,2}, r_{3,3}, t_3, s_1, s_2, s_3, 1] \qquad (4.14)$$

$$M_{row} = [r_{1,1}, r_{2,1}, r_{3,1}, s_1, r_{1,2}, r_{2,2}, r_{3,2}, s_2, r_{1,3}, r_{2,3}, r_{3,3}, s_3, t_1, t_2, t_3, 1] \qquad (4.15)$$

Analizując równania (4.14) i (4.15), można zaobserwować, że w notacji wierszowej macierz zapisywana jest wiersz po wierszu, natomiast w reprezentacji kolumnowej elementy macierzy zapisywane są kolumna po kolumnie. Oznacza to, że w celu konwersji notacji można wykorzystać operację transpozycji macierzy:

$$\begin{aligned} M_{row} &= M'_{col} , \\ M_{col} &= M'_{row} . \end{aligned} \qquad (4.16)$$

Relacje pomiędzy notacjami można też rozszerzyć o operacje łączenia transformacji geometrycznych. Połączenie transformacji $A$ i $B$ w notacji kolumnowej realizowane jest przez operację mnożenia lewostronnego (ang. *post-multiplication*), zaś w notacji kolumnowej przez wykorzystanie operacji mnożenia prawostronnego (ang. *pre-multiplication*):

$$A_{col} B_{col} = B_{row} A_{row} . \qquad (4.17)$$

Do parametryzacji modelu i wyznaczenia macierzy transformacji wykorzystywany jest wektor stanu modelu 3D, którego liczba elementów równa jest liczbie stopni swobody całego modelu.

## Reprezentacja wektora stanu modelu 3D

W modelu 3D postaci ludzkiej przedstawionym wcześniej w tabeli 4.1, zauważyć można, że poszczególne kości mają różne stopnie swobody. Jeśli dana kość nie posiada żadnego stopnia swobody oznacza to, że jej pozycja zależy wyłącznie od transformacji elementu nadrzędnego. Jak już wspomniano, wektor o długości równej liczbie stopni swobody modelu i zawierający wartości określające konfigurację modelu nazywamy wektorem stanu modelu. Wektor, w którym dodano zera w miejscach, gdzie nie wystę-pują stopnie swobody danej kości nazywamy pełnym wektorem stanu modelu. Wektor stanu modelu przedstawiony w tabeli 4.1 zawiera 31 elementów, tzn. liczba jego elementów jest równa liczbie stopni swobody modelu składającego się z 15 elementów. Pełny wektor stanu składa się z 90 elementów, zaś liczba elementów pełnego wektora stanu jest równa iloczynowi liczby kości występujących w modelu oraz dopuszczalnemu wymiarowi stopni swobody dla każdej kości. Oznacza to, że uwzględnienie dodatkowych kości, nieposiadających żadnych stopni swobody nie spowoduje zwiększenia liczby elementów wektora stanu, natomiast spowoduje wzrost liczby elementów pełnego wektora stanu modelu. Dzięki takiej reprezentacji możliwa jest parametryzacja modelu przez zbiór kątów Eulera wyrażonych w radianach oraz przesunięć wyrażonych w milimetrach. Kąty Eulera [37,173] wykorzystywane są do reprezentacji rotacji wokół osi układu współrzędnych. Sama nazwa notacji pochodzi od nazwiska Leonharda Eulera, matematyka pochodzenia szwajcarskiego, który dowiódł, że dowolny obrót może zostać uzyskany za pomocą sekwencji obrotów wykonywanych względem osi przyjętego układu współrzędnych [119]. Kolejność sekwencji obrotów, jak i przyjęty układ współrzędnych określany jest mianem notacji Eulera. W poszczególnych konwencjach określa się także kierunek obrotu oraz przyjęte nazewnictwo kątów reprezentujących rotację wokół danej osi [173]. Wadami kątów Eulera są: zależność sposobu interpolacji od przyjętej konwencji, brak jednoznaczności reprezentacji oraz brak prostej metody łączenia sekwencji obrotów. Jednak notacja Eulera posiada także wiele zalet. Do największej z nich należy łatwość interpretacji wartości kątów przez człowieka. Z tego powodu wykorzystywana jest ona do reprezentacji danych wejściowych i wyjściowych opisujących stan modelu szkieletowego. Zazwyczaj dane wejściowe są przekształcane do wewnętrznej reprezentacji transformacji geometrycznej modelu. Tabela 4.3 przedstawia porównanie reprezentacji transformacji modelu wykorzystujących macierze jednorodne, kąty Eulera, kwaterniony (ang. *quaternions*) i podwójne kwaterniony (ang. *dual quaternions*). Kwaterniony są strukturami algebraicznymi, będącymi rozszerzeniem ciała liczby zespolonej [6,119,173], zaś podwójne kwaterniony [70,149] są strukturą algebraiczną, będącą rozszerzeniem kwaternionów. Wspomniane kwaterniony, a także inne reprezentacje wykorzystywane są często w praktyce, ponieważ w reprezentacji rotacji opartej na kątach Eulera występuje zjawisko nazywane blokowaniem przegubów (ang. *gimbal lock*) [92,173].

**Tabela 4.3. Alternatywne metody reprezentacji transformacji geometrycznych**

| Reprezentacja | Macierze jednorodne | Kąty Eulera | Kwaterniony | Podwójne kwaterniony |
|---|---|---|---|---|
| Liczba parametrów | 16 | 3 | 4 | 8 |
| Blokada przegubu | nie | tak | nie | nie |
| Łączenie transformacji (mnożenie) | tak | nie | tak | tak |
| Interpolacja obrotów | tak | nie | tak | tak |
| Reprezentacja rotacji i translacji jednocześnie | tak | nie | nie | tak |
| Nakładanie ograniczeń kątowych | trudne | łatwe | trudne | trudne |

Zjawisko blokowania przegubów występuje, gdy jeden z kolejnych obrotów w osiach układu wykonywany jest o kąt będący wielokrotnością 90 stopni, co w konsekwencji prowadzi do sytuacji, w której jedna lub kilka osi zaczyna się na siebie nakładać, zob. rys. 4.7. W praktyce występowanie blokady przegubów powoduje utratę jednego lub większej liczby stopni swobody [173].

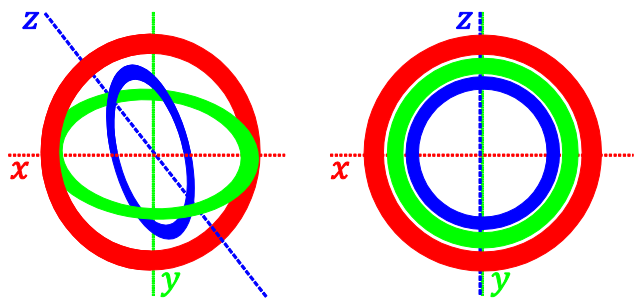

**Rys. 4.7. Ilustracja zjawiska blokady przegubów**

# Wyznaczanie macierzy przekształceń lokalnych

Macierz przekształceń lokalnych wykorzystywana jest do transformacji wierzchołka w lokalnym układzie współrzędnych. Wyznaczanie macierzy transformacji lokalnych danej kości za pomocą macierzy jednorodnych realizowane jest zgodnie z równaniem:

$$L = T \cdot R \, , \tag{4.18}$$

gdzie $T$ jest macierzą jednorodną reprezentującą przesunięcie w przypadku korzenia modelu i długość kości w pozostałych przypadkach, natomiast $R$ jest macierzą złożenia sekwencji rotacji.

Sekwencja rotacji zależy od przyjętego modelu oraz danych odniesienia (ang. *ground truth*) systemu mocap. Dla przyjętej sekwencji rotacji $R_{xyz} = R_x R_y R_z$ dla układu prawoskrętnego lokalna macierz transformacji opisana jest równaniem:

$$TR_{xyz} = TR_x R_y R_z =$$
$$= \begin{bmatrix} cos(\beta)cos(\gamma) & sin(\gamma)cos(\alpha) + sin(\beta)sin(\alpha)cos(\gamma) & sin(\gamma)sin(\alpha) - sin(\beta)cos(\alpha)cos(\gamma) & t_x \\ -cos(\beta)sin(\gamma) & cos(\gamma)cos(\alpha) - sin(\beta)sin(\alpha)sin(\gamma) & cos(\gamma)sin(\alpha) + sin(\beta)cos(\alpha)sin(\gamma) & t_y \\ sin(\beta) & -cos(\beta)sin(\alpha) & cos(\beta)cos(\alpha) & t_z \\ 0 & 0 & 0 & 1 \end{bmatrix}. \tag{4.19}$$

Wykorzystanie rozwinięcia sekwencji transformacji $TR_x R_y R_z$ pozwala na uniknięcie kosztownych obliczeniowo operacji mnożenia macierzy. Rozwinięcie sekwencji mnożeń macierzy może być stosowane dla dowolnej sekwencji transformacji. Warto przy tym wspomnieć, że wymiar macierzy można ograniczyć do trzech wierszy i czterech kolumn. Jest to możliwe, gdyż w trakcie składania macierzy jednorodnych reprezentujących wyłącznie operację rotacji i translacji, ostatni wiersz macierzy zawsze będzie wektorem $[0 \quad 0 \quad 0 \quad 1]$. W tabeli 4.4 przedstawiono postacie macierzy lokalnych dla elementów wykorzystywanego modelu.

**Tabela 4.4. Odwzorowanie transformacji kości na macierze transformacji lokalnych**

| Nazwa kości | Transformacje | Macierz lokalna |
|---|---|---|
| Pelvis | $t_x, t_y, t_z, r_x, r_y, r_z$ | $L_{Pelvis} = T_{xyz}R_{xyz}$ |
| Thorax | $r_y, r_z$ | $L_{Thorax} = R_{yz}$ |
| Left Axis | $-$ | $L_{LeftAxis} = I$ |
| Left Hip | $r_x, r_y, r_z$ | $L_{LeftHip} = R_{xz}$ |
| Left Knee | $r_y$ | $L_{LeftKnee} = R_y$ |
| Right Axis | $-$ | $L_{RightAxis} = I$ |
| Right Hip | $r_x, r_y, r_z$ | $L_{RightHip} = R_{xz}$ |
| Right Knee | $r_y$ | $L_{RightKnee} = R_y$ |
| Head | $r_x, r_y, r_z$ | $L_{Head} = R_{xyz}$ |
| Left Clavicle | $r_x, r_y$ | $L_{LeftClavicle} = R_{xy}$ |
| Left Shoulder | $r_x, r_y, r_z$ | $L_{LeftShoulder} = R_{xyz}$ |
| Left Elbow | $r_y$ | $L_{LeftElbow} = R_y$ |
| Right Clavicle | $r_x, r_y$ | $L_{RightClavicle} = R_{xy}$ |
| Right Shoulder | $r_x, r_y, r_z$ | $L_{RightShoulder} = R_{xyz}$ |
| Right Elbow | $r_y$ | $L_{RightElbow} = R_y$ |

# Wyznaczanie macierzy transformacji globalnych

Macierz przekształceń globalnych służy do przekształcenia koordynat wierzchołka w lokalnym układzie współrzędnych do odpowiadającej mu pozycji w globalnym układzie współrzędnych. Wyznaczanie macierzy transformacji globalnych realizowane jest hierarchicznie, począwszy od korzenia modelu. Operacja wyznaczenia macierzy transformacji globalnych wymaga uprzedniego wyznaczenia macierzy transformacji lokalnych wszystkich kości modelu. Ponieważ korzeń modelu nie posiada elementu nadrzędnego, jego macierz transformacji globalnej jest równa macierzy transformacji lokalnej:

$$W_{root} = L_{root} .$$
(4.20)

Transformacja globalna elementu posiadającego element nadrzędny wyznaczana jest jako iloczyn macierzy transformacji globalnej elementu nadrzędnego i macierzy transformacji lokalnej elementu:

$$W = W_{parent} \cdot L .$$
(4.21)

Tabela 4.5 przedstawia postać macierzy transformacji globalnej dla wszystkich kości modelu szkieletowego, zaprezentowanego w tabeli 4.1.

**Tabela 4.5. Macierze transformacji globalnej modelu postaci ludzkiej**

| Nazwa kości | Element nadrzędny | Macierz transformacji globalnej |
|---|---|---|
| Pelvis | – | $W_{Pelvis} = L_{Pelvis}$ |
| Thorax | Pelvis | $W_{Thorax} = W_{Pelvis} L_{Thorax}$ |
| Left Axis | Pelvis | $W_{LeftAxis} = W_{Pelvis} L_{LeftAxis}$ |
| Left Hip | Left Axis | $W_{LeftHip} = W_{LeftAxis} L_{LeftHip}$ |
| Left Knee | Left Hip | $W_{LeftKnee} = W_{LeftHip} L_{LeftKnee}$ |
| Right Axis | Pelvis | $W_{RightAxis} = W_{Pelvis} L_{RightAxis}$ |
| Right Hip | Right Axis | $W_{RightHip} = W_{RightAxis} L_{RightHip}$ |
| Right Knee | Right Hip | $W_{RightKnee} = W_{RightHip} L_{RightKnee}$ |
| Head | Thorax | $W_{Head} = W_{Thorax} L_{Head}$ |
| Left Clavicle | Thorax | $W_{LeftClavicle} = W_{Thorax} L_{LeftClavicle}$ |
| Left Shoulder | Left Clavicle | $W_{LeftShoulder} = W_{LeftClavicle} L_{LeftShoulder}$ |
| Left Elbow | Left Shoulder | $W_{LeftElbow} = W_{LeftShoulder} L_{LeftElbow}$ |
| Right Clavicle | Thorax | $W_{RightClavicle} = W_{Thorax} L_{RightClavicle}$ |
| Right Shoulder | Right Clavicle | $W_{RightShoulder} = W_{RightClavicle} L_{RightShoulder}$ |
| Right Elbow | Right Shoulder | $W_{RightElbow} = W_{RightShoulder} L_{RightElbow}$ |

Wyznaczanie macierzy transformacji globalnej jest zadaniem trudnym do zrównoleglenia, w szczególności gdy model szkieletu nie jest z góry znany. W takim przypadku wyznaczanie macierzy transformacji globalnej realizowane jest zgodnie z hierarchią modelu szkieletowego.

## Połączenie modelu kształtu i modelu szkieletowego

Wierzchołki modelu kształtu mogą być zdefiniowane w układzie lokalnym lub w układzie globalnym. Model kształtu, w którym współrzędne wierzchołków określone są w lokalnym układzie współrzędnych, nazywamy modelem kształtu w układzie kości modelu szkieletowego [106]. Model kształtu, w którym współrzędne wierzchołków określone są w globalnym układzie współrzędnych, nazywamy modelem kształtu w układzie globalnym [106]. Wierzchołki w układzie lokalnym oznaczane są jako $v_{local}$, zaś wierzchołki modelu w układzie globalnym oznaczane są jako $v'_{global}$, gdzie $v_{global}$ jest pozycją w układzie globalnym odpowiadającą wierzchołkowi modelu w pozie określonej przez konfigurację modelu szkieletowego.

Zaletą stosowania układu lokalnego jest prostota wyznaczania współrzędnych wierzchołka w układzie globalnym (4.22). Transformacja wierzchołka do układu globalnego realizowana jest przez wymnożenie macierzy transformacji globalnej $M_{local \rightarrow global}$ i współrzędnych wierzchołka w układzie lokalnym:

$$\begin{bmatrix} v_{global} \\ 1 \end{bmatrix} = M_{local \rightarrow global} \begin{bmatrix} v_{local} \\ 1 \end{bmatrix} \qquad (4.22)$$

Macierz $M_{local \rightarrow global}$ jest macierzą transformacji świata modelu go $W$ dla zadanej kości, wyznaczającą pozycję wierzchołka obiektu w globalnym ukła-dzie współrzędnych na podstawie pozycji wierzchołka w lokalnym układzie współrzędnych kości. Wadą modelu zbudowanego z wierzchołków we współrzędnych lokalnych jest konieczność wymnożenia wszystkich wierzchołków przez macierze odpowiadających im transformacji. Efektem wyrysowania modelu bez wspomnianej operacji będzie rzut modelu, którego wszystkie części ciała będą miały swój początek w środku globalnego układu współrzędnych.

Model kształtu, którego współrzędne wierzchołków określone są w układzie globalnym, jest pozbawiony tej wady. Dzięki temu model może zostać wyrenderowany poprawnie bez konieczności wymnożenia wierzchołków przez odpowiadające im macierze transformacji globalnej. Wyrenderowany w ten sposób model będzie znajdował się w pozie ustalonej (ang. *bind pose*) [13]. W rozwiązaniu opracowanym w ramach niniejszej pracy wykorzystywana jest poza ustalona A, zob. rys. 4.8a, której alternatywą jest poza ustalona T, zob. rys. 4.8b.

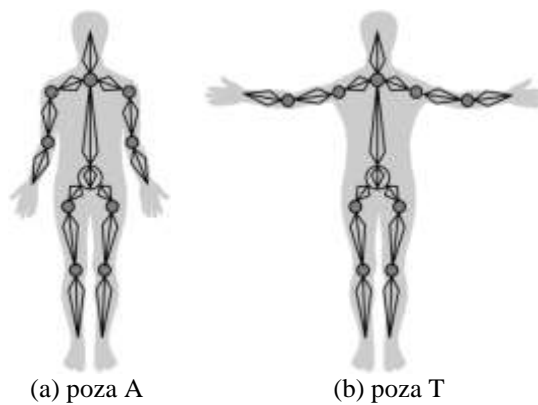

(a) poza A                    (b) poza T

**Rys. 4.8. Poza ustalona modelu**

Wykorzystanie pozy ustalonej pociąga za sobą konieczność transformacji wierzchołka modelu oznaczanego jako $v'_{global}$ do lokalnego układu współrzędnych. Transformacja ta realizowana jest zgodnie z równaniem (4.23), w którym $B_{local \to global}$ jest stałą macierzą transformacji kości szkieletu w pozie ustalonej. Jej macierz odwrotna $B^{-1}_{local \to global}$ umożliwia wyznaczenie wierzchołka modelu $v_{local}$ we współrzędnych lokalnych. Wspomniana macierz wyznaczana jest w sposób identyczny do macierzy transformacji globalnej $W$, przyjmując jako wektor stanu konfigurację modelu w pozie ustalonej:

$$\begin{bmatrix} v_{local} \\ 1 \end{bmatrix} = \frac{1}{B_{local \to global}} \begin{bmatrix} v'_{global} \\ 1 \end{bmatrix} .$$ (4.23)

Wykorzystując zależność pomiędzy równaniami (4.22) i (4.23), można wyznaczyć transformacje pomiędzy wierzchołkiem modelu w pozie ustalonej $v'_{global}$, a wierzchołkiem modelu $v_{global}$ we współrzędnych świata:

$$\begin{bmatrix} v_{global} \\ 1 \end{bmatrix} = M_{local \to global} \frac{1}{B_{local \to global}} \begin{bmatrix} v'_{global} \\ 1 \end{bmatrix} . \tag{4.24}$$

Zastosowanie modelu wykorzystującego pozę ustaloną wymaga większych nakładów obliczeniowych, ponieważ przy każdej transformacji modelu wymagane jest wyznaczenie pozycji wierzchołka w lokalnym układzie współrzędnych, zob. zależność (4.23). Alternatywnym podejściem jest przekształcenie wierzchołków modelu do koordynat lokalnych. Jednakże jest to możliwe jedynie wtedy, gdy pozycja wierzchołka w globalnym układzie współrzędnych zależy wyłącznie od jednej macierzy transformacji globalnej. Relacja taka nazywana jest mapowaniem sztywnym (ang. *rigid skinning*) [13]. W przypadku, gdy pozycja wierzchołka modelu zależy od wielu kości, mówimy o mapowaniu miękkim *(ang. smooth skinning)* [13], które opisuje się następującą zależnością:

$$\begin{bmatrix} v_{local} \\ 1 \end{bmatrix} = \sum_{i}^{n} \omega_i M_{i,local \to global} \frac{1}{B_{i,local \to global}} \begin{bmatrix} v'_{global} \\ 1 \end{bmatrix} . \tag{4.25}$$

W omawianej metodzie mapowania pozycja wierzchołka zależy od $n$ kości, $\omega_i$ jest wagą macierzy transformacji globalnej $M_{i,local \to global}$, natomiast $B_{i,local \to global}$ jest macierzą transformacji kości do współrzędnych lokalnych. Parametry $\omega$ dla każdego wierzchołka muszą spełniać zależność $\sum_{i}^{n} \omega_i = 1$. W opracowanych rozwiązaniach wykorzystywane jest mapowanie sztywne ($n = 1, \omega_1 = 1$).

## Reprezentacja modelu kształtu

W niniejszej pracy wykorzystywane są dwa modele kształtu: model płaski oraz model 3D nazywany modelem siatkowym. Modelem płaskim jest model, w którym każda część ciała jest reprezentowana na płaszczyźnie obrazu przez trapez będący aproksymacją stożka ściętego o zadanych parametrach. Parametrami stożka ściętego reprezentującego daną część ciała są dwa punkty określające środek podstawy górnej koła $t_p$ i podstawy dolnej koła $b_p$ oraz ich promienie podstaw $t_r, b_r$, zob. rys. 4.9. Dodatkowym parametrem jest wartość całkowita $p$, określająca kolor, w którym zamalowany zostanie rzut stożka na płaszczyznę obrazu oraz identyfikator $b$. Identyfikator $b$ służy do określenia pozycji globalnej macierzy transformacji $W$ w wektorze macierzy transformacji globalnych. Tabela 4.6 prezentuje nazwy części ciała, które wykorzystywane są w trakcie renderingu modelu płaskiego. Wspomniana tabela prezentuje także macierze transformacji modelu wykorzystywane do transformacji danej części ciała. Z uwagi na to, że w przyjętym modelu ręce połączone są sztywno z przedramionami, macierze transformacji rąk są tożsame z macierzami transformacji przedramion.

**Tabela 4.6. Reprezentacja części ciała w modelu płaskim**

| Część ciała | Macierz transformacji |
| --- | --- |
| brzuch | $W_{Pelvis}$ |
| klatka piersiowa | $W_{Thorax}$ |
| lewe udo | $W_{LeftHip}$ |
| lewa łydka | $W_{LeftKnee}$ |
| lewa stopa | $W_{LeftKnee}$ |
| prawe udo | $W_{RightHip}$ |
| prawa łydka | $W_{RightKnee}$ |
| prawa stopa | $W_{RightKnee}$ |
| głowa | $W_{Head}$ |
| lewe ramię | $W_{LeftShoulder}$ |
| lewe przedramię | $W_{LeftElbow}$ |
| lewa dłoń | $W_{LeftElbow}$ |
| prawe ramię | $W_{RightShoulder}$ |
| prawe przedramię | $W_{RightElbow}$ |
| prawa ręka | $W_{RightElbow}$ |

Macierz $W$ wykorzystywana jest do transformacji środków podstawy górnej i dolnej do globalnego układu współrzędnych zgodnie z równaniem (4.22).

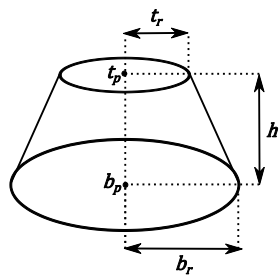

**Rys. 4.9. Parametryzacja stożka ściętego**

Parametry stożków modelu płaskiego służą do wyznaczenia wierzchołków $v_1$, $v_2$, $v_3$, $v_4$ trapezu aproksymującego kształt stożka. Trapez aproksymujący stożek ścięty będzie równoległy do płaszczyzny obrazu, zob. rys. 4.10c. Oznacza to, że wektor normalny płaszczyzny stożka może zostać określony jako $n = \overrightarrow{b_p C}$, gdzie $C$ jest pozycją kamery w przestrzeni. Wektor ten będzie prostopadły do płaszczyzny rozpiętej na wektorach $u = \overrightarrow{b_p t_p}$ i $r = \overrightarrow{b_p v_2}$, zob. rys. 4.10b. Jeśli wektory $u$ i $v$ nie są liniowo zależne, to wektor będący wynikiem iloczynu wektorowego jest prostopadły do płaszczyzny rozpiętej na wektorach $u$ i $r$. Wektor taki nazywany jest wektorem normalnym i oznaczany jest przez $n$ [173], zob. 4.10. Do wyznaczenia wierzchołków trapezu wykorzystany zostanie wektor $r = u \times n$, dzięki któremu można wyznaczyć współrzędne wierzchołków zgodnie z równaniami $v_1 = t_p + t_r r$, $v_2 = b_p + b_r r$, $v_3 = b_p - b_r r$ i $v_4 = t_p + t_r r$.

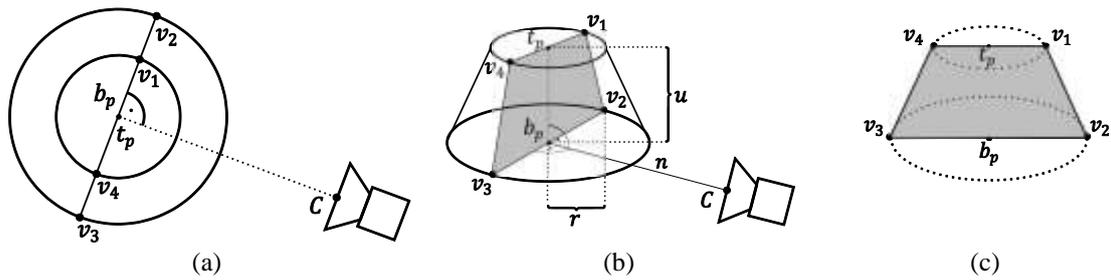

**Rys. 4.10. Wyznaczanie wierzchołków stożka ściętego**

Wyznaczenie współrzędnych wierzchołków $v_1$, $v_2$, $v_3$, $v_4$ odbywa się dla każdej części modelu postaci o zadanej konfiguracji. Wierzchołki trapezu rzutowane są do płaszczyzny obrazu w oparciu o równania przyjętego modelu kamery, który omówiono w podrozdziale 2.2.

W alternatywnym podejściu każda część ciała opisana jest przez zbiór wierzchołków budujących siatkę danej części ciała lub siatkę modelu postaci, który nazywany jest modelem siatkowym. W omawianym modelu każdy wierzchołek składa się z atrybutów określających pozycję wierzchołka $v(x, y, z)$ w przyjętym układzie współrzędnych, a także parametru $p$ określającego kolor wierzchołka i identyfikatora $b$. Podobnie jak w przypadku modelu płaskiego, identyfikator $b$ służy do wyznaczenia macierzy transformacji globalnej, która wykorzystywana jest do transformacji wierzchołka do globalnego układu współrzędnych. Model siatkowy w odróżnieniu od modelu płaskiego oprócz prostego mapowania modelu szkieletowego do modelu kształtu (4.22) może wykorzystywać mapowanie miękkie (4.25). W trakcie mapowania miękkiego wykorzystywany jest zbiór atrybutów $w_1, b_1, w_2, b_2, ..., b_n, w_n$, gdzie $w_n$ jest wagą macierzy o indeksie $b_n$.

## 4.4. Rasteryzacja modelu 3D

Rasteryzacją nazywamy proces przekształcenia modelu 3D o zadanych parametrach i konfiguracji do postaci dwuwymiarowego obrazu prezentującego model zrzutowany zgodnie z przyjętymi parametrami kamery. W niniejszej pracy operacja renderingu realizowana jest w oparciu o rendering programowy oraz rendering sprzętowy. Opracowanie zarówno renderingu programowego, jak i sprzętowego umożliwiło przeprowadzenie badań empirycznych, dzięki którym określono wady i zalety każdej z metod w śledzeniu ruchu 3D w czasie rzeczywistym.

Wynikiem rasteryzacji modelu 3D postaci ludzkiej jest obraz, w którym każdy piksel reprezentowany jest przez dodatnią liczbę całkowitą zapisaną na ośmiu bitach. Na ośmiu bitach koduje się informacje o wystąpieniu krawędzi na obrazie ze zrzutowanym modelem oraz o przynależności piksela do danego segmentu rasteryzowanej sylwetki. Sposób kodowania zależy od przebiegu procesu rasteryzacji oraz od wykorzystywanej funkcji celu.

# Rendering programowy

Rendering programowy dotyczy procesu renderingu, który jest realizowany bez wykorzystania dedykowanych funkcji i układów rasteryzujących. Przyspieszenie renderingu programowego możliwe jest przez wykorzystanie mechanizmów zrównoleglenia obliczeń. Renderowanie odbywa się z wykorzystaniem algorytmów rysowania prostych figur geometrycznych np. trójkątów lub czworokątów. Rendering programowy wykorzystywany jest głównie w przypadku mniej złożonych modeli [67,127,130]. Jednym z ograniczeń omawianej metody jest to, że w niektórych sytuacjach elementy muszą być renderowane sekwencyjnie. Rendering programowy znajduje zastosowanie w rozwiązaniach, w których są niższe wymagania odnośnie do jakości generowanego obrazu. Warto wspomnieć, że w zastosowaniach wymagających większego realizmu generowanej sceny konieczne jest wykorzystanie dodatkowych algorytmów w procesie renderingu, zwykle wprowadzających znaczące narzuty czasowe. Jedną z zalet renderingu programowego jest to, że możliwe jest zdefiniowanie własnych struktur danych oraz realizacja niestandardowych operacji graficznych.

# Rendering z akceleracją sprzętową

Rendering z akceleracją sprzętową w odróżnieniu od renderingu programowego wykorzystuje dedykowane funkcje i układy rasteryzujące. Rendering z akceleracją sprzętową jest zwykle nazywany renderingiem sprzętowym. Wykorzystuje on potok graficzny, którego funkcje i możliwości są z góry określone. W omawianym renderingu dostęp do funkcji rysujących możliwy jest przez interfejs programistyczny API np. OpenGL, DirectX. W odróżnieniu od renderingu programowego, interfejs programistyczny wykorzystywany w renderingu sprzętowym narzuca format struktur wejściowych oraz format generowanego obrazu. Dzięki zrównolegleniu sprzętowemu możliwe jest skrócenie czasu renderingu i uzyskanie lepszego odwzorowania dla bardziej złożonych scen.

## 4.5. Funkcja celu

W przyjętej reprezentacji piksele obrazów sylwetki $s_{x,y}^{(c)}$ i $s_{x,y}^{(c')}$ przyjmują wartości naturalne z zakresu $\langle 1, L \rangle$, gdzie $L$ określa liczbę części ciała. Dzięki takiej reprezentacji wartości te reprezentują etykiety części ciała. Piksele, których wartość jest równa zero, reprezentują tło, natomiast wartości większe od zera identyfikują etykiety poszczególnych segmentów wydzielonej postaci. Segmentacja obrazu [156] jest konieczna w metodach hierarchicznych do śledzenia ruchu postaci ludzkiej [81,126], w pozostałych metodach [12,138] nie ma potrzeby wykorzystywania posegmentowanych obrazów. Wówczas obraz sylwetki ludzkiej reprezentowany jest za pomocą obrazu binarnego, zob. rys. 4.11a. W metodach hierarchicznych przebadano kilka sposobów segmentacji. W najprostszym podejściu wydzielano dwa segmenty. Pierwszy z segmentów reprezentował części ciała o kolorze skóry, drugi zaś pozostałe części cia-ła, zob. rys. 4.11b. W bardziej rozbudowanych podejściach wydzielano także głowę, tułów, kończyny górne i dolne, zob. rys. 4.11c i rys. 4.11d.

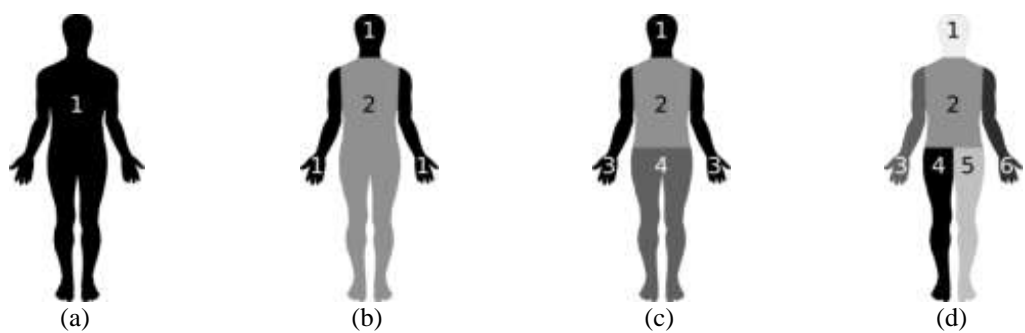

(a)          (b)          (c)          (d)

**Rys. 4.11. Segmentacja sylwetki postaci**

Piksele obrazów $e_{x,y}^{(c)}$ i $e_{x,y}^{(c')}$ reprezentują krawędzie wydzielone na obrazach wejściowych oraz krawędzie na wyrenderowanych obrazach modelu 3D. Piksel może przyjmować wartość jeden, gdy reprezentuje on krawędź oraz zero, w przypadku nie wykrycia krawędzi. Do wydzielania krawędzi wykorzystano metody omówione w podrozdziale 2.1. Natomiast metody tworzenia obrazów krawędzi na podstawie zrzutowanego modelu 3D przybliżone zostały w podrozdziale 5.4.

Binarny obraz krawędzi $E^{(c)}$ wydzielonych w oparciu o obrazy systemu wizyjnego wykorzystywany jest do wygenerowania obrazu mapy odległości $D^{(c)}$. Wartość piksela $d_{x,y}^{(c)}$ obrazu mapy odległości $D^{(c)}$ określa odległość od najbliższej krawędzi. Odległość ta jest wyznaczona zgodnie z metrykami odległości, które zaprezentowano w podrozdziale 2.1. W zaproponowanym rozwiązaniu nie rozróżnia się obszaru zewnętrznego i wewnętrznego postaci ludzkiej, tzn. odległości zewnętrzne i wewnętrzne przyjmują jedynie wartości dodatnie. Przed wykorzystaniem mapy odległości w funkcji celu, wartości pikseli są normalizowane do zakresu $\langle 0,1 \rangle$. Wartość jeden reprezentuje piksele znajdujące się w pobliżu wykrytej krawędzi, natomiast wartość zero symbolizuje piksele znajdujące się w dużej odległości od krawędzi.

W niniejszej pracy zaproponowano trzy funkcje normalizacji wartości piksela $d$ dla mapy odległości:

- funkcję impulsową $n_{impulse}(d)$,
- funkcję proporcjonalną $n_{proportional}(d)$,
- funkcję wykładniczą $n_{exponent}(d)$.

Funkcja impulsowa:

$$n_{impulse}(d) = \begin{cases} 1 & jeśli\ d = 0 \\ 0 & jeśli\ d \neq 0 \end{cases}, \tag{4.26}$$

jest stosowana w algorytmach śledzenia, w których nie jest wykorzystywana mapa odległości. Przykładowo, jeśli jakość krawędzi jest wystarczająca dla zakładanej niezawodności i dokładności śledzenia z jednej strony, z drugiej strony zaś istotny jest czas śledzenia, wówczas śledzenie może być realizowane z pominięciem mapy odległości.

Funkcja proporcjonalna opisywana jest równaniem:

$$n_{proportional}(d) = \begin{cases} 1 & \text{jeśli } d \leq d_{min} \\ 1 - n_{range}\frac{d-d_{min}}{d_{max}-d_{min}} & \text{jeśli } d_{min} < d < d_{max} \\ 0 & \text{jeśli } d \geq d_{max} \end{cases}, \quad (4.27)$$

w którym $d_{max}$, $d_{min}$ i $n_{range}$ są parametrami określającymi kształt funkcji po normalizacji. Wartości $d_{max}$ i $d_{min}$ określają przedział dziedziny mapy odległości, w którym realizowana jest normalizacja do zakresu określanego przez wartość $n_{range}$ (zob. rys. 4.12). Na wspomnianym rysunku funkcja proporcjonalna oznaczona jest etykietą $n_p$, $d_{max} = 100$, $d_{min} = 10$ i $n_{range} = 0{,}25$. Funkcja proporcjonalna dzieli dziedzinę mapy odległości na trzy przedziały: $\langle 0, d_{min} \rangle$, w którym pikselom przypisana zostanie wartość jeden; $\langle d_{max}, \infty \rangle$, w którym pikselom przypisana zostanie wartość zero oraz $(d_{min}, d_{max})$, w którym wartości zostaną proporcjonalne przeskalowane do przedziału $\langle 1, 1 - n_{range} \rangle$. Parametry funkcji proporcjonalnej $d_{min}$, $d_{max}$ i $n_{range}$ dobierane są eksperymentalnie dla danego ustawienia kamery czy też warunków oświetleniowych.

Funkcja eksponencjalna:

$$n_{exp}(d) = \begin{cases} 1 & \text{jeśli } d \leq d_{max} \\ n_{range}\,exp(d \cdot m) + (1 - n_{range}) & \text{jeśli } d_{min} < d < d_{max} \\ 0 & \text{jeśli } d \geq d_{max} \end{cases}, \quad (4.28)$$

parametryzowana jest przez $d_{max}$, $d_{min}$ i $n_{range}$, które określają przedziały podlegające normalizacji oraz jej zakres, parametr $m \in (0,1)$ jest natomiast współczynnikiem określającym stopień dyskryminacji mapy odległości.

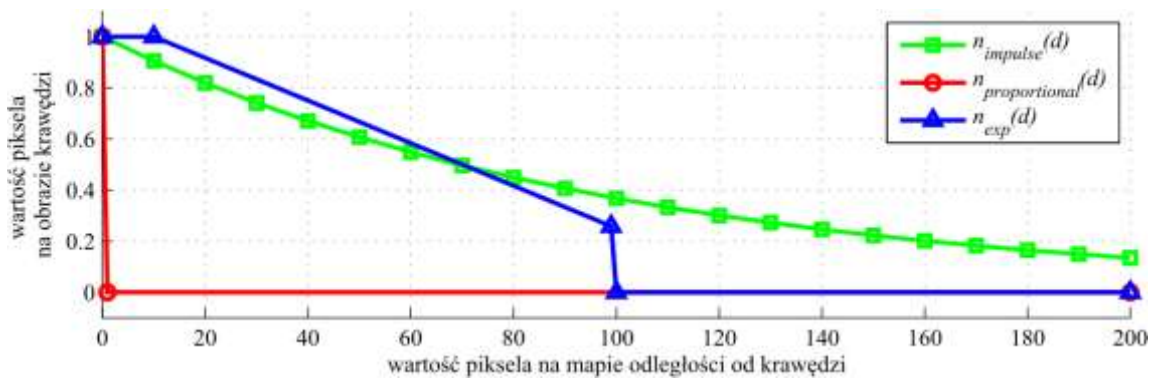

**Rys. 4.12. Funkcje normalizacji mapy odległości**

Funkcja eksponencjalna wykorzystywana była wówczas, gdy wymagana była silniejsza dyskryminacja większych wartości mapy odległości. Funkcja proporcjonalna i eksponencjalna wykorzystywane były wówczas, gdy krawędzie wydzielono algorytmem detekcji krawędzi Canny, natomiast funkcja impulsowa wykorzystywana była wtedy, gdy krawędzie wydzielano algorytmem Sobela. W podrozdziale poświęconym omówieniu funkcji celu przybliżono własności funkcji normalizujących mapy odległości.

Obraz otrzymany w wyniku normalizacji mapy odległości $D^{(c)}$ oznaczany jest przez $\overline{D}^{(c)}$. Obraz ten wykorzystywany jest podczas wyznaczania jednego z komponentów wartości funkcji celu. Jak już wspomniano, funkcja celu wyznaczana jest przez analizę odpowiadających sobie obrazów z systemu wizyjnego oraz wyrenderowanego modelu postaci. W wyniku analizy obrazów wydzielane są cechy opisujące widoczne powierzchnie segmentów oraz powierzchnie pokrywających się segmentów. Wyznaczana jest także liczba pikseli należących do wykrytych krawędzi modelu oraz suma wartości pikseli znormalizowanej mapy odległości w miejscach występowania krawędzi zrzutowanego modelu. Obszarem zainteresowania w kamerze $c$ nazywamy prostokątny wycinek obrazu, który zawiera wszystkie piksele należące do sylwetki postaci ludzkiej. Ze względu na to, że obszar zainteresowania w nowej klatce powinien zawierać wszystkie części ciała, prostokąt reprezentujący osobę jest powiększany w osi poziomej i pionowej o pewną liczbę pikseli. Powiększenie obszaru zainteresowania realizowane jest z uwzględnieniem ruchu postaci, który może mieć miejsce między sąsiednimi klatkami, oraz parametrów trójwymiarowego modelu postaci. Obszar zainteresowania oznaczany jest przez $roi^{(c)}$ dla obrazów z systemu wizyjnego oraz $roi^{(c')}$ dla obrazów reprezentujących wyrenderowany model. W przyjętych oznaczeniach $roi_x$, $roi_y$ oznaczają lewą górną krawędź obszaru zainteresowania, zaś $roi_w$, $roi_h$ oznaczają wysokość i szerokość obszaru zainteresowania.

Powierzchnia wydzielonego segmentu postaci na obrazie sylwetki $S^{(c)}$ wyznaczana jest zgodnie z równaniem (4.29), w którym funkcja $eq_l$ określa przynależność piksela $s_{x,y}^{(c)}$ do segmentu $l$.

$$area_l^{(c)} = \sum_x \sum_y eq_l\left(s_{x,y}^{(c)}\right) \tag{4.29}$$

Powierzchnia segmentu obrazu reprezentującego model w określonej konfiguracji wyznaczana jest w oparciu o poniższe równanie:

$$area_l^{(c')} = \sum_x \sum_y eq_l\left(s_{x,y}^{(c')}\right). \tag{4.30}$$

Funkcja określająca przynależność piksela $s$ do segmentu o etykiecie $l$ opisana jest następującym równaniem:

$$eq_l(s) = \begin{cases} 1 & jeśli\ s = l \\ 0 & jeśli\ s \neq l \end{cases}.$$ (4.31)

Liczba pikseli należących do krawędzi, która reprezentuje segment $l$ na obrazie wyrenderowanego modelu, wyznaczana jest zgodnie z poniższym równaniem:

$$edge_l^{(c')} = \sum_x \sum_y eq_l\left(s_{x,y}^{(c')}\right) \cdot \left\lceil e_{x,y}^{(c')} \right\rceil.$$ (4.32)

Powierzchnia części wspólnej segmentów sylwetki postaci na obrazach $S^{(c)}$ i $S^{(c')}$ należących do segmentu $l$ wyznaczana jest zgodnie z równaniem:

$$overlap_l^{(c')} = \sum_x \sum_y eq_l\left(\sqrt{s_{x,y}^{(c')} \cdot s_{x,y}^{(c)}}\right).$$ (4.33)

Suma wartości znormalizowanej mapy odległości $\bar{D}^{(c)}$, odpowiadającej obrazowi krawędzi $E^{(c')}$ i segmentu $l$ na obrazie sylwetki $S^{(c')}$ wyznaczana jest zgodnie z równaniem:

$$distance_l^{(c')} = \sum_x \sum_y eq_l\left(s_{x,y}^{(c')}\right) \cdot e_{x,y}^{(c')} \cdot \bar{d}_{x,y}^{(c)}.$$ (4.34)

Jak już wspomniano, w przypadku baz danych i sekwencji, w których segmentacja części ciała nie jest możliwa, np. w przypadku baz zawierających wyłącznie obrazy krawędzi i sylwetki postaci [11,153] lub gdy dokładność śledzenia jest zadowalająca, segmentacja postaci może zostać pominięta. Z tego też powodu wykorzystywano postacie funkcji nieuwzględniające informacji o segmentacji. W tym przypadku funkcja (4.31) określająca przynależność piksela o wartości $p$ do sylwetki postaci może być przedstawiona za pomocą następującego równania:

$$eq(p) = min\left(\sum_{l=1}^{L} eq_l(p), 1\right) = \begin{cases} 1 & gdy\ p \neq 0 \\ 0 & w\ przeciwnym\ razie \end{cases}.$$ (4.35)

Wyznaczanie powierzchni sylwetki postaci na obrazie $S^{(c)}$, który nie uwzględnia segmentacji postaci realizowane jest w oparciu o poniższe równanie:

$$area^{(c)} = \sum_l^{L} area_l^{(c)} = \sum_x \sum_y eq\left(s_{x,y}^{(c)}\right).$$ (4.36)

Wyznaczanie powierzchni sylwetki postaci na obrazie $S^{(c')}$, który nie uwzględnia segmentacji postaci realizowane jest w oparciu o poniższe równanie:

$$area^{(c')} = \sum_{l=1}^{L} area_l^{(c')} = \sum_x \sum_y eq\left(s_{x,y}^{(c')}\right). \tag{4.37}$$

Liczba pikseli należących do krawędzi na obrazie wyrenderowanego modelu wyznaczana jest zgodnie z poniższym równaniem:

$$edge^{(c')} = \sum_{l=1}^{L} edge_l^{(c')} = \sum_x \sum_y \left\lfloor e_{x,y}^{(c')}\right\rfloor eq\left(s_{x,y}^{(c')}\right). \tag{4.38}$$

Powierzchnia części wspólnej segmentów sylwetki postaci na obrazach $S^{(c)}$ i $S^{(c')}$ wyznaczana jest zgodnie z równaniem:

$$ovelap^{(c')} = \sum_{l=1}^{L} overlap_l^{(c')} = \sum_x \sum_y eq\left(s_{x,y}^{(c')} \cdot s_{x,y}^{(c)}\right). \tag{4.39}$$

Na rys. 4.13 zilustrowano w sposób poglądowy operację wyznaczania części wspólnej segmentów sylwetki. Na rys. 4.13a znajduje się sylwetka postaci wydzielona przez algorytm omówiony w podrozdziale 2.1, na rys. 4.13b prezentowana jest sylwetka modelu postaci, którą otrzymano w wyniku wyrenderowania modelu 3D, zaś na rys. 4.13c zaprezentowano część wspólną sylwetek. Część wspólna obydwu sylwetek zaznaczona jest kolorem żółtym.

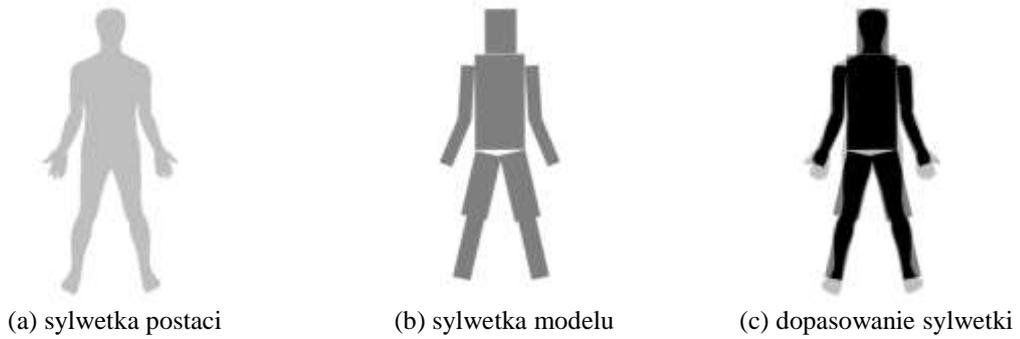

(a) sylwetka postaci      (b) sylwetka modelu      (c) dopasowanie sylwetki

**Rys. 4.13. Sposób wyznaczania części wspólnej segmentów sylwetki**

Suma wartości znormalizowanej mapy odległości $\bar{D}^{(c)}$, odpowiadającej obrazowi krawędzi $E^{(c')}$ i na obrazie sylwetki $S^{(c')}$ wyznaczana jest zgodnie z równaniem:

$$distance^{(c')} = \sum_{l=1}^{L} distance_l^{(c')} = \sum_x \sum_y eq\left(s_{x,y}^{(c')}\right) \cdot e_{x,y}^{(c')} \cdot \bar{d}_{x,y}^{(c)}. \tag{4.40}$$

W proponowanym podejściu funkcja celu składa się z dwóch składników, które reprezentują stopień dopasowania sylwetek oraz wartości wynikającej z rzutowania kra-wędzi modelu na mapę odległości. Wspomniane części składowe funkcji celu przyjmują wartości rzeczywiste i są normalizowane do przedziału $< 0,1 >$, gdzie wartość zero reprezentuje sytuację, w której model i obserwowana sylwetka nie posiadają części wspólnej, zaś wartość jeden odpowiada sytuacji, w której występuje perfekcyjne dopasowanie. Składniki te zaprezentowano bardziej szczegółowo w podrozdziale 4.6, który jest poświęcony omówieniu wykorzystywanych funkcji celu.

Na rys. 4.14 zaprezentowano operację maskowania mapy odległości od krawędzi sylwetki z wykorzystaniem krawędzi zrzutowanego modelu. Na rys. 4.14a znajduje się obraz mapy odległości od krawędzi sylwetki postaci, którą wyznaczono w oparciu o algorytm omówiony w podrozdziale 2.1, na rys. 4.14b zamieszczono obraz krawędzi zrzutowanego modelu 3D postaci, zaś na rys. 4.14c zaprezentowano wynik operacji maskowania. Na omawianym rysunku kolorem żółtym zaznaczono wynik operacji maskowania, która realizowana jest w ramach wyznaczania wartości (4.40).

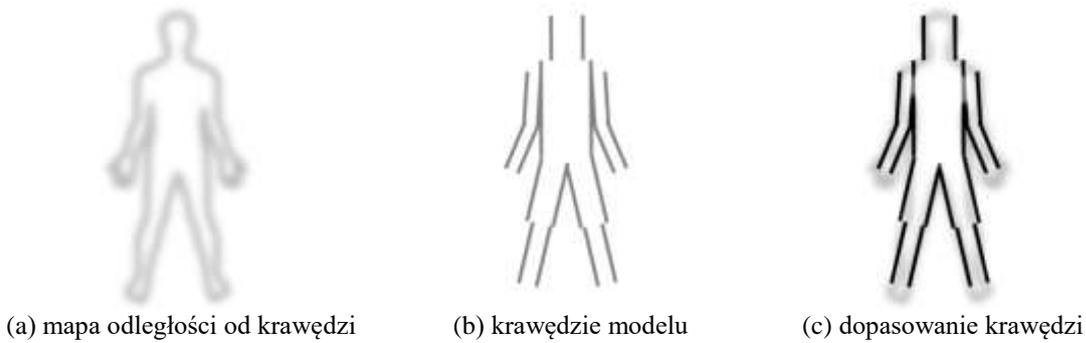

(a) mapa odległości od krawędzi    (b) krawędzie modelu    (c) dopasowanie krawędzi

**Rys. 4.14. Sposób maskowania mapy odległości od krawędzi sylwetki**

## 4.6. Modele obserwacji i funkcje celu

Funkcja określająca stopień dopasowania sylwetki postaci dla kamery $c$ określona jest przez równanie:

$$f_1^{(c)}(x) = \beta \frac{overlap^{(c')}}{area^{(c)}} + (1 - \beta) \frac{overlap^{(c')}}{area^{(c')}}, \qquad (4.41)$$

gdzie parametr $\beta \in\, < 0,1 >$ jest współczynnikiem wagowym dopasowania binarnych sylwetek: od zrzutowanego modelu do obrazu sylwetki oraz od obrazu sylwetki do zrzutowanego modelu.

Funkcja wyznaczająca dopasowanie sylwetki, która uwzględnia obrazy ze wszystkich $C$ kamer systemu wizyjnego, określona jest przez równanie:

$$f_1(x) = \beta \frac{\sum_{c'=1}^{C} overlap^{(c')}}{\sum_{c=1}^{C} area^{(c)}} + (1 - \beta) \frac{\sum_{c'=1}^{C} overlap^{(c')}}{\sum_{c'=1}^{C} area^{(c')}}. \qquad (4.42)$$

W przypadku segmentacji obrazu stopień dopasowania sylwetki wyznaczany jest zgodnie z równaniem (4.43), w którym $w_l$ jest wagą funkcji $f_{1,l}^{(i)}(x)$, wyznaczającej wartość dopasowania segmentu $l$ modelu postaci ludzkiej. Wartość funkcji $f_{1,l}^{(i)}(x)$ wyznaczana jest zgodnie z równaniem (4.41), w którym uwzględnia się jedynie informację o zadanej części ciała. Natomiast wagi $w_l$ spełniają zależność $\sum_l^L w_l = 1$.

$$f_1^{(c)}(x) = \sum_{l=1}^{L} w_l f_{1,l}^{(c)}(x) \tag{4.43}$$

Dopasowanie mapy odległości od krawędzi z kamery $c$ do krawędzi na obrazie z hipotetyczną pozą postaci ludzkiej wyznaczane jest zgodnie z równaniem:

$$f_2^{(c)}(x) = \frac{distance^{(c')}}{edge^{(c')}}. \tag{4.44}$$

Wyznaczanie czynnika reprezentującego dopasowanie krawędzi ze wszystkich kamer systemu wizyjnego zbudowanego z $C$ kamer, realizowane jest zgodnie z równaniem:

$$f_2(x) = \frac{\sum_{c'=1}^{C} distance^{(c')}}{\sum_{c'=1}^{C} edge^{(c')}}. \tag{4.45}$$

Zaprezentowane powyżej funkcje dopasowania wykorzystywane były do budowy funkcji celu. W opracowanym systemie śledzącym przebadano dwie funkcje celu, tzn. ważoną funkcję celu $f_{WS}$ oraz wygładzaną funkcję celu $f_{SP}$, w których wykorzystano wartości dopasowania, wyznaczone zgodnie z równaniami (4.42) i (4.45). Przebadano także trzy funkcje celu, wykorzystujące wartości dopasowania wyznaczonego zgodnie z równaniami (4.43) i (4.44). Opracowane funkcje celu wykorzystującą średnią ważonych dopasowań krawędzi i sylwetki - $f_{AoWS}(x)$, średnią wygładzonego dopasowania krawędzi i sylwetki - $f_{AoPS}(x)$ oraz iloczyn wygładzonych dopasowań krawędzi i sylwetki - $f_{PoPS}(x)$. Wszystkie funkcje celu zwracają wartości z przedziału $< 0, 1 >$, w którym górna wartość reprezentuje idealne dopasowanie między obserwowaną sylwetką i wyrenderowanym modelem.

Ważona funkcja celu opisana jest przez równanie (4.46), gdzie $w_1$ jest wagą dopasowania sylwetki, natomiast $w_2$ jest wagą dopasowania krawędzi. Wagi muszą spełniać zależność $w_1 + w_2 = 1$.

$$f_{WS}(x) = \sum_{i=1}^{2} w_i f_i(x) \tag{4.46}$$

Wygładzona funkcja celu, która zaprezentowana została w równaniu (4.47), charakteryzuje się większą siłą dyskryminującą niż ważona funkcja celu. W omawianej funkcji wygładzanie dopasowania sylwetki realizowane jest dzięki użyciu wykładnika $\omega_1$, natomiast wygładzenie dopasowania krawędzi za pomocą wykładnika $\omega_2$. Wykładniki muszą spełniać zależność $0 \leq \omega_i < 1$.

$$f_{SP}(x) = \prod_{i=1}^{2} \left( f_i(x) \right)^{\omega_i} \tag{4.47}$$

Funkcja celu zbudowana w oparciu o średnią ważonych dopasowań krawędzi i sylwetki kamery opisana jest równaniem (4.48), w którym $C$ jest liczbą kamer składających się na system wizyjny.

$$f_{AoWS}(x) = \frac{1}{C} \sum_{c=1}^{C} \sum_{i=1}^{2} \omega_i f_i^{(c)}(x) \tag{4.48}$$

Funkcja celu wykorzystująca średnią wygładzonych dopasowań krawędzi i sylwetki kamer opisana jest równaniem (4.49). Analogicznie jak w przypadku funkcji (4.48), $C$ oznacza liczbę kamer wchodzących w skład systemu wizyjnego.

$$F_{AoSP}(x) = \frac{1}{C} \sum_{c=1}^{C} \prod_{i=1}^{2} \left( f_i^{(c)}(x) \right)^{\omega_i} \tag{4.49}$$

Ostatnia funkcja celu wykorzystuje iloczyn wygładzonych dopasowań sylwetki oraz dopasowań krawędzi sylwetki. Funkcja ta opisana jest równaniem:

$$F_{PoSP}(x) = \prod_{c=1}^{C} \prod_{i=1}^{2} \left( f_i^{(c)}(x) \right)^{\omega_i}. \tag{4.50}$$

## 4.7. Podsumowanie

W niniejszym rozdziale zaproponowano strukturę modelu 3D do śledzenia ruchu postaci ludzkiej w czasie rzeczywistym. Modelowanie ruchu odbywa się z wykorzystaniem transformacji geometrycznych, natomiast modelowanie kształtu sylwetki odbywa się przez proste figury geometryczne lub za pomocą złożonej siatki. Opracowane rozwiązania umożliwiają wykorzystanie modelu płaskiego zarówno w procesie renderingu programowego, jak i sprzętowego. Dzięki opracowanym rozwiązaniom możliwe jest renderowanie modelu siatkowego z wykorzystaniem akceleracji sprzętowej. Cała struktura zaproponowanego modelu jest elastyczna i może być dostosowana w zależności od przeznaczenia. Zaproponowano kilka wariantów funkcji celu dla śledzenia ruchu 3D w czasie rzeczywistym w oparciu o rendering programowy i sprzętowy modelu 3D.

# Rozdział 5
# Zrównoleglenie funkcji celu

Rozdział składa się z siedmiu podrozdziałów, w których zaprezentowano zrównolegle-
nie funkcji celu. W podrozdziale pierwszym sformułowano problem. W następnym pod-
rozdziale omówiono zaproponowane rozwiązania renderingu programowego modelu
3D z wykorzystaniem CUDA. W kolejnych trzech podrozdziałach omówiono mecha-
nizm renderingu sprzętowego z wykorzystaniem OpenGL, CUDA-OpenGL i OpenCL-
OpenGL. W podrozdziale szóstym zaprezentowano sposób wyznaczania funkcji celu
z wykorzystaniem zaproponowanych rozwiązań w zakresie efektywnego renderingu
modelu 3D. Rozdział zakończony jest podsumowaniem.

## 5.1.  Sformułowanie problemu

W trakcie śledzenia ruchu postaci ludzkiej w oparciu o model 3D najbardziej kosztowną
obliczeniowo operacją jest rendering modelu. W rozwiązaniach opartych na renderingu
zarówno krawędzi, jak i całej sylwetki modelu, czas wyznaczania funkcji celu może
stanowić ponad 90% całego czasu śledzenia. Z kolei w trakcie wyznaczania wartości
funkcji celu najbardziej kosztowną operacją jest rendering modelu 3D. Wysoki koszt
obliczeniowy procesu renderingu wynika z konieczności wygenerowania znaczącej
liczby pikseli, reprezentujących sylwetkę i krawędzie modeli w zadanej liczbie konfigu-
racji. Bez wykorzystania mechanizmów przetwarzania równoległego oraz sprzętowego
renderingu nie byłoby możliwe śledzenie ruchu całej postaci w czasie rzeczywistym.
Wyniki te są zbieżne z wynikami uzyskanymi przez inny zespół badawczy [29]. Wyniki
te wskazują na to, że nakłady obliczeniowe niezbędne do wyznaczania wag cząsteczek
są znacznie większe w porównaniu do pozostałych nakładów obliczeniowych związa-
nych ze śledzeniem ruchu.

## 5.2.  Rendering modelu 3D z wykorzystaniem CUDA

W niniejszym podrozdziale przedstawiono mechanizm renderingu programowego
(ang. *software rendering*) modelu postaci ludzkiej. Algorytm renderingu programowego
służy do wyrysowania płaskich figur geometrycznych na płaszczyźnie obrazu oraz linii
będących krawędziami rysowanych obiektów. Danymi wejściowymi algorytmu rysują-
cego są wierzchołki trójkątów we współrzędnych obrazu cyfrowego, które budują mo-
del postaci ludzkiej. Odległość trójkątów od kamery, nazywana także głębokością
obiektu, nie jest uwzględniania przez algorytm rysujący. Omawiany algorytm nie ko-
rzysta z bufora głębokości, przez co nie ma możliwości ustalenia kolejności przesłonięć.
Algorytm rysujący ignoruje także kierunek wektora normalnego rysowanego trójkąta.
Kierunek wektora normalnego wykorzystywany jest przez mechanizm odrzucania tyl-
nych ścian (ang. *backface culling*) [173]. Przesłonięcia i mechanizm odrzucania tylnych
ścian realizowane są na etapie przetwarzania wstępnego, który poprzedza proces ryso-wania. W każdym kroku algorytmu z parametrów modelu kształtu tworzone są trójkąty, spośród których odrzucane są następnie elementy niewidoczne. Dzięki temu, że przetwarzanie wstępne dokonuje także sortowania rysowanych obiektów według odległości od kamery, uzyskuje się poprawne przesłanianie obiektów. Ostatnim krokiem przetwarzania wstępnego jest rzutowanie współrzędnych wierzchołków modelu do współrzędnych obrazu cyfrowego zgodnie z przedstawionym w podrozdziale 2.2 modelem kamery. Algorytm, w którym obiekty są sortowane i rysowane w kolejności malejącej odległości od kamery określany jest mianem algorytmu malarza (ang. *painter's algorithm*) [60]. W renderingu programowym do wypełniania trójkątów kolorem wykorzystywane są m.in. algorytm rozrostu ziarna (ang. *seed fill*) [28,120], algorytm wypełniania linia po linii (ang. *scanline*) [60,76], algorytm Bresenhama [60] i algorytm sprawdzający przynależność punktu do trójkąta [42]. Algorytm rozrostu ziarna służy do wypełniania zamkniętego obszaru obrazu począwszy od pozycji początkowej, a skończywszy na granicy obiektu lub obrazu. Algorytm rozpoczyna działanie od pozycji początkowej, zob. rys. 5.1a. W trakcie analizy sąsiedztwa wyznaczany jest zbiór pikseli do zamalowania, po czym zamalowywany jest tzw. piksel startowy, zob. rys. 5.1b. Operacja analizy i zamalowywania powtarzana jest dla każdego elementu zbioru do chwili, kiedy zbiór pikseli do zamalowania będzie zbiorem pustym, zob. rys. 5.1c i rys. 5.1d. Implementacja algorytmu rozrostu ziarna wykorzystuje zwykle kolejkę lub stos do przechowywania zbioru pikseli do zamalowania [50,99].

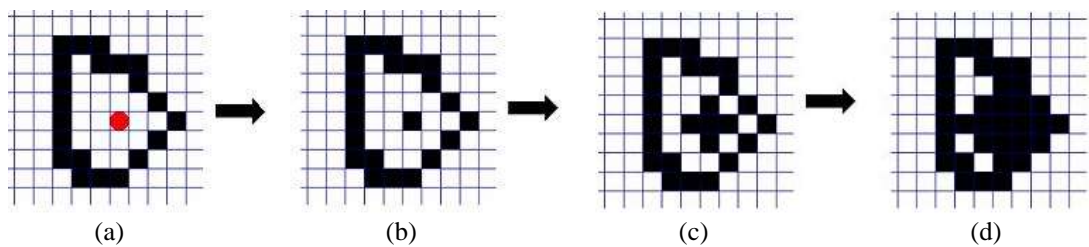

(a)          (b)          (c)          (d)

**Rys. 5.1. Algorytm rozrostu ziarna**

Wadą algorytmu rozrostu ziarna jest konieczność uprzedniego wyrysowania krawędzi modelu. W celu wyrysowania krawędzi wykorzystać można algorytm naiwnego rysowania linii [76]. Naiwny algorytm rysowania linii zamaluje wszystkie piksele obrazu pokrywające się z linią $y = mx + b$ rozpoczynającą się w punkcie $(x_1, y_1)$ i kończącą w punkcie $(x_2, y_2)$. Punkt końcowy i początkowy wykorzystywany jest do wyznaczania parametrów $m = \frac{y_2 - y_1}{x_2 - x_1}$ i $b = y_1 - mx_1$ określających linię. Alternatywnie do wyrysowania linii wykorzystać można algorytmy *digital differential analyzer* [118], algorytm Bresenhama [20] lub algorytm zaproponowany przez Wu [178].

Algorytm rysowania linii wykorzystywany jest, gdy zamalowywany jest trójkąt o wierzchołkach $v_1(x_1, y_1)$, $v_2(x_2, y_2)$, $v_3(x_3, y_3)$, w którym jedna z krawędzi jest równoległa do osi $x$ obrazu cyfrowego.

Gdy wierzchołek $v_1$ znajduje się nad podstawą trójkąta, zob. rys. 5.2a, możliwe jest wyznaczenie współczynników nachylenia dwóch pozostałych ścian trójkąta zgodnie z równaniem:

$$slope_1 = \frac{x_2 - x_1}{y_3 - y_1}, \quad slope_2 = \frac{x_3 - x_1}{y_3 - y_1}. \tag{5.1}$$

Jeśli bok równoległy do osi $x$ obrazu cyfrowego składa się z wierzchołków $v_1$ i $v_2$ oraz wierzchołek $v_3$ znajduje się poniżej podstawy trójkąta (zob. rys. 5.2b), wykorzystywane jest równanie:

$$slope_1 = \frac{x_3 - x_1}{y_3 - y_2}, \quad slope_2 = \frac{x_3 - x_2}{y_3 - y_2}. \tag{5.2}$$

Trójkąt, który nie posiada boku równoległego do osi $x$ obrazu cyfrowego, dzielony jest na dwa mniejsze trójkąty posiadające wspólną krawędź równoległą do osi $x$, zob. rys. 5.2c. W algorytmie wypełniania linia po linii zamalowywanie trójkąta rozpoczyna się od jednej z wartości granicznych trójkąta dla osi $y$. W każdej iteracji algorytm wyznacza graniczne wartości $x$ będące granicami odcinka, wykorzystując w tym celu współczynniki $slope_1$ i $slope_2$.

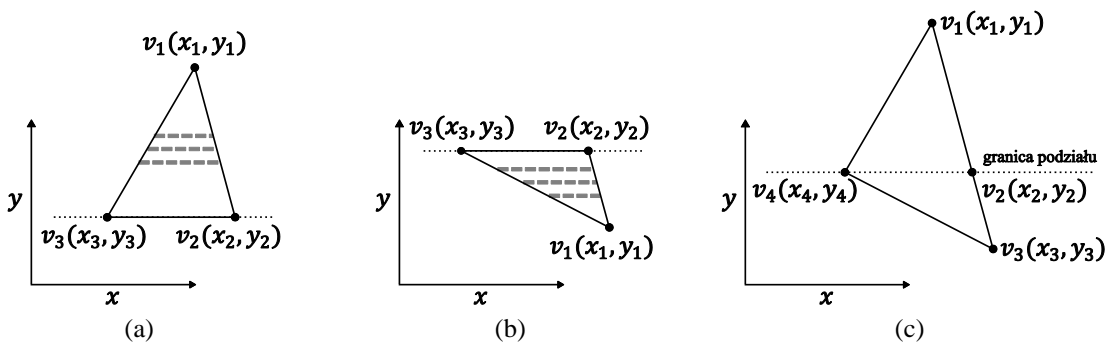

Rys. 5.2. Algorytm wypełniania obiektu linia po linii

Algorytm wypełniania powierzchni Bresenhama różni się od algorytmu wypełniania linia po linii metodą wyznaczania współrzędnych krawędzi rysowanego trójkąta, zob. rys. 5.3. W niniejszej pracy współrzędne krawędzi modelu 3D wyznaczane są za pomocą algorytmu rysowania linii Bresenhama, przedstawionego w pracy [20].

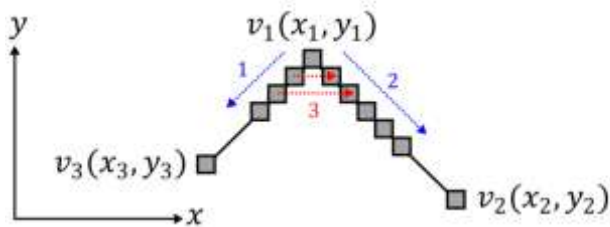

Rys. 5.3. Algorytm Bresenhama

Ostatnim z omawianych algorytmów jest algorytm Barycentric [60]. W omawianym algorytmie wpierw wyznacza się obszar zainteresowania zgodny z osiami obrazu cyfrowego, zob. rys. 5.4, który w grafice komputerowej nazywany jest pudełkiem zorientowanym zgodnie z osiami układu współrzędnych (ang. *axis-aligned bounding box, AABB*) [60,106].

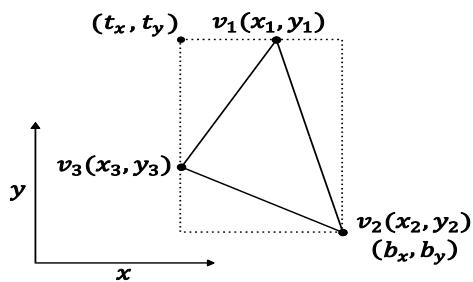

**Rys. 5.4. Obszar zainteresowania w algorytmie Barycentric**

Granice obszaru zainteresowania określane są przez wartości minimalne $(b_x, b_y)$ i maksymalne $(t_x, t_y)$ wierzchołków trójkąta, które wyznaczane są zgodnie z równaniem:

$$t_x = min(x_1, min(x_2, x_3)) \ ,$$
$$t_y = max(y_1, max(y_2, y_3)) \ ,$$
$$b_x = max(x_1, max(x_2, x_3)) \ ,$$
$$b_y = min(y_1, min(y_2, y_3)) \ .$$

(5.3)

W trakcie działania omawianego algorytmu dla wszystkich współrzędnych $p(x_p, y_p)$ pikseli znajdujących się wewnątrz wyznaczonej przestrzeni zainteresowania sprawdza się, czy punkt $p$ znajduje się wewnątrz trójkąta rozpiętego na wierzchołkach $v_1$, $v_2$, $v_3$ zgodnie z równaniem:

$$\frac{\overrightarrow{pv_1} \times \overrightarrow{v_3 v_1}}{\overrightarrow{v_2 v_1} \times \overrightarrow{v_3 v_1}} \geq 0 \ and \ \frac{\overrightarrow{v_2 v_1} \times \overrightarrow{pv_1}}{\overrightarrow{v_2 v_1} \times \overrightarrow{v_3 v_1}} \geq 0 \ and \ \frac{\overrightarrow{pv_1} \times \overrightarrow{v_3 v_1} + \overrightarrow{v_2 v_1} \times \overrightarrow{pv_1}}{\overrightarrow{v_2 v_1} \times \overrightarrow{v_3 v_1}} \leq 1 \ .$$

(5.4)

Alternatywne metody sprawdzania, czy punkt leży wewnątrz trójkąta wykorzystują równania parametryczne funkcji liniowej oraz iloczyn skalarny wektorów. Celem sprawdzenia, czy punkt zawiera się w trójkącie, można wykorzystać równania parametryczne krawędzi rysowanego trójkąta:

$$x(t_1) = t_1(x_2 - x_1) \ ,$$
$$y(t_1) = t_1(y_2 - y_1) \ ,$$
$$x(t_2) = t_2(x_3 - x_1) \ ,$$
$$y(t_2) = t_2(y_3 - y_1) \ .$$

(5.5)

Jeśli parametry $t_1$ i $t_2$ równań (5.5) spełniają zależność określoną w równaniu (5.6), to punkt $p$ znajduje się we wnętrzu trójkąta o wierzchołkach $v_1, v_2, v_3$.

$$0 \leq t_1 \leq 1 \; and \; 0 \leq t_2 \leq 1 \; and \; t_1 + t_2 \leq 1 \qquad (5.6)$$

W metodzie wykorzystującej iloczyn skalarny, aby punkt znajdował się wewnątrz trójkąta, znak każdego z iloczynów skalarnych musi być ujemny, zgodnie z równaniem:

$$0 \leq \overrightarrow{v_1 v_2} \cdot \overrightarrow{p v_1} \; and \; 0 \leq \overrightarrow{v_1 v_2} \cdot \overrightarrow{p v_2} \; and \; 0 \leq \overrightarrow{v_1 v_3} \cdot \overrightarrow{p v_3} \; . \qquad (5.7)$$

Wybór metody zamalowywania trójkąta zależy od środowiska w jakim zadany algorytm będzie działał. W środowiskach równoległych zadanie renderingu dzieli się na podzadania renderujące pojedyncze trójkąty [100]. W architekturach posiadających setki rdzeni zadanie może zostać jeszcze bardziej rozdrobnione przez podzielenie obszaru trójkąta na piksele, które przetwarzane są niezależnie zgodnie z algorytmem Barycentric. Rozdrobnienie zadania w ten sposób, że pojedynczy piksel przetwarzany jest niezależnie, jest rzadko wykorzystywane w architekturach posiadających niewielką liczbę rdzeni. Wynika to m.in. z konieczności realizacji dużej liczby operacji matematycznych w stosunku do algorytmu wypełniania linia po linii i rozrostu ziarna.

W architekturze CUDA przy wykorzystaniu algorytmu Barycentric możliwe jest rozbicie procesu renderingu na tysiące bardzo małych podzadań. Wykorzystanie tysięcy wątków pozwala na implementację liniowego dostępu do pamięci globalnej, dzięki czemu ukrywane są opóźnienia odczytu wartości z pamięci, co w konsekwencji pozwala zwiększyć wydajność metody renderingu programowego. W algorytmie zaimplementowanym w CUDA rendering odbywa się zgodnie z algorytmem malarza, w którym rysowanie elementów realizowane jest od obiektu najdalszego do obiektu najbliższego. Oznacza to, że jeden piksel będzie wielokrotnie wypełniany kolorem – podejście takie określane jest nadmiarowym zamalowywaniem (ang. *overpainting*) [18]. Rysowanie obiektów na scenie w oparciu o algorytm malarza wykorzystuje jedynie operację zapisu pikseli do bufora kolorów umieszczonego w pamięci globalnej urządzenia.

W opracowanym rozwiązaniu rendering programowy realizowany jest w powiązaniu z wyznaczaniem komponentów funkcji celu i stanowi wspólną całość. Metoda renderingu modelu 3D postaci ludzkiej opracowana w niniejszej pracy używa odwrotnego algorytmu malarza [18,60]. W algorytmie tym obiekty rysowane są w kolejności rosnącej odległości od kamery, dzięki czemu nie występuje zjawisko nadmiarowego zamalowywania. Wykorzystanie tego algorytmu powoduje konieczność pobrania wartości pikseli z bufora kolorów przed ich zamalowaniem, jednak w przypadku obliczania wartości komponentów funkcji celu jest to podejście efektywniejsze, zob. podrozdział 6.5. Rendering i wyznaczanie komponentów funkcji celu można rozdzielić, pociąga to jednak za sobą wzrost czasu wykonywania algorytmu.

Rys. 5.5 przedstawia uproszczony potok rysowania programowego, który został zaprojektowany, a następnie skonfigurowany dla algorytmów wykorzystujących CUDA.

W potoku wyszczególnić można funkcje jądra wyznaczające: pełny wektor stanu modelu 3D, lokalne macierze transformacji, globalne macierze transformacji oraz posortowane wierzchołki modelu. Ostatnim elementem potoku jest rendering trójkątów wykorzystujący algorytm malarza oraz procedura sprawdzająca, czy rysowany punkt jest punktem wewnętrznym trójkąta. Powyższe kroki przedstawione zostały na omawianym rysunku jako funkcje jądra technologii CUDA. W opracowanym rozwiązaniu kilka funkcji jądra łączonych jest w całość, aby zmniejszyć liczbę odwołań do pamięci globalnej i wywołań funkcji celu.

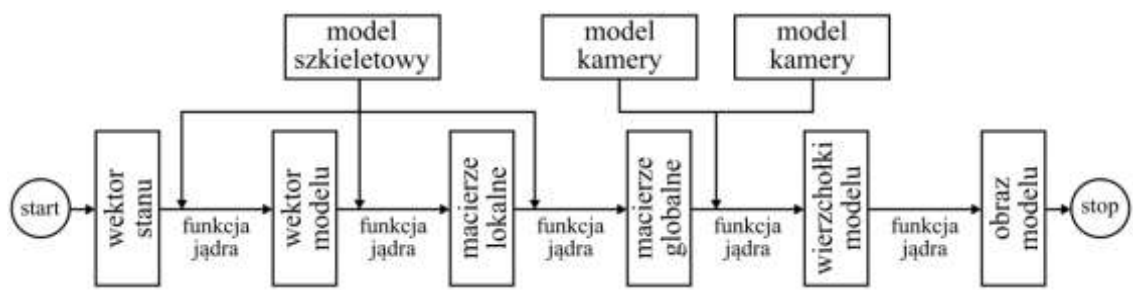

**Rys. 5.5. Potok rysowania programowego w CUDA**

W pierwszym kroku potoku wyznacza się pełny wektor stanu modelu, który omówiono w podrozdziale 4.3. W rozwiązaniu wykorzystującym CUDA zadanie to realizowane jest wielowątkowo i każdy wątek przekształca pojedynczy wektor stanu do pełnego wektora modelu.

## 5.3. Rendering w OpenGL

W niniejszym podrozdziale zaprezentowano rendering sprzętowy (ang. *hardware rendering*) modelu postaci z wykorzystaniem API OpenGL. Opisano w nim także podstawowe zagadnienia związane z przetwarzaniem grafiki komputerowej.

### Potok graficzny

Termin potok graficzny (ang. *graphics pipeline*) odnosi się do sekwencji kroków realizowanych w celu wygenerowania dwuwymiarowego obrazu, będącego cyfrową reprezentacją trójwymiarowej sceny [146,145]. Pierwsze wersje OpenGL udostępniały jedynie stały potok (ang. *fixed function graphics pipeline*), w którym obiekty graficzne przetwarzane były wyłącznie zgodnie z przyjętym modelem API [145]. Wraz z rozwojem układów graficznych, stałe funkcje zastąpiono elementami programowalnymi [114], tworząc programowalny potok graficzny (ang. *programmable graphics pipeline*) [146].

Na rys. 5.6 przedstawiono programowalny potok graficzny, który wykorzystywano w zaproponowanych rozwiązaniach, wspierających rendering sprzętowy. Programowalny potok graficzny składa się z [146]:

- asemblera danych wejściowych (ang. *input assembler*),
- procesora wierzchołków (ang. *vertex processing*),
- funkcji przetwarzania końcowego wierzchołków (ang. *vertex post-processing*),
- funkcji składania prymitywów graficznych (ang. *primitive assembly*),
- funkcji rasteryzacji i interpolacji (ang. *rasterization & interpolation*),
- procesora fragmentów (ang. *fragment processing*),
- funkcji przetwarzania próbek (ang. *per-sample operations*).

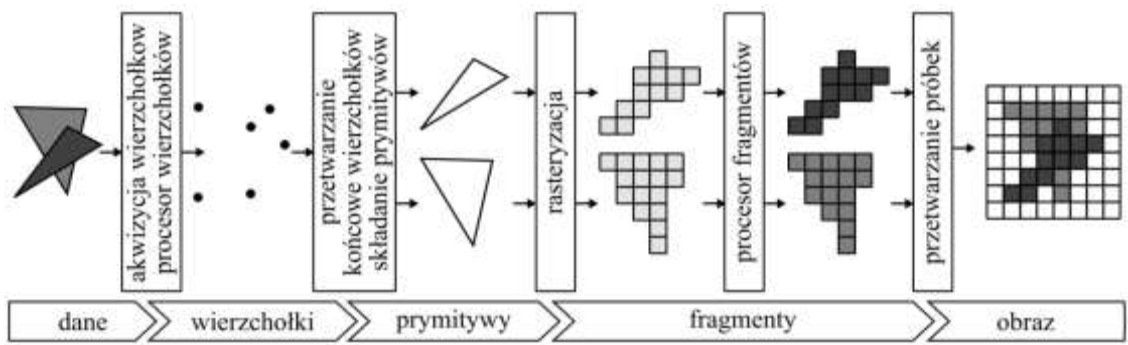

**Rys. 5.6. Programowalny potok graficzny OpenGL 4.4**

Programowalnymi elementami potoku są procesory wierzchołków i fragmentów, realizujące zadania z wykorzystaniem krótkich programów komputerowych nazywanych programami cieniującymi (ang. *shader programs*). Programy cieniujące tworzone są w języku programowania GLSL (ang. *OpenGL Shading Language*), przypominającym składnią język C [131]. GLSL składa się z kilku języków programowania, posiadających część wspólną [73]. Każdy wariant języka GLSL wykorzystywany jest do implementacji określonej funkcjonalności potoku graficznego. Zadana funkcjonalność realizowana jest przez jeden z następujących typów programów cieniujących:

- program cieniowania wierzchołków (ang. *vertex shader*),
- program cieniowania do kontroli teselacji (ang. *tessellation control shader*),
- program cieniowania do ewaluacji teselacji (ang. *tessellation evaluation shader*),
- program cieniowania geometrii (ang. *geometry shader*),
- program cieniowania fragmentów (ang. *fragment shader*),
- program cieniowania do obliczeń ogólnych (ang. *compute shader*).

Program cieniowania wierzchołków wykorzystywany jest do realizacji zestawu indywidualnych operacji na wierzchołkach wejściowych. Teselacja zwiększa liczbę detali przetwarzanych obiektów graficznych przez wykorzystanie metody podziału. Przykładem teselacji może być reprezentacja kuli na scenie. Gdy obiekt znajduje się w dużej odległości od kamery powinien być on reprezentowany przez prosty obiekt graficzny np. sześcian, zob. rys. 5.7a, zaś gdy kamera znajdować się będzie bliżej sześcianu, wykorzystamy podzielenie każdej ściany na cztery mniejsze prymitywy graficzne, zwiększa-jąc tym samym poziom szczegółowości dwukrotnie, zob. rys. 5.7b. Podział kwadratów zrealizowany będzie po raz kolejny, gdy kamera ponownie zbliży się do obiektu, zob. rys. 5.7c, do momentu osiągnięcia satysfakcjonującego stopnia szczegółowości obiektu kuli, zob. rys. 5.7d.

Teselacja składa się z programu cieniowania do kontroli teselacji, którego zadaniem jest wyznaczenie poziomu teselacji obiektu oraz programu cieniującego do ewaluacji teselacji, odpowiedzialnego za wyznaczenie nowych wierzchołków obiektu.

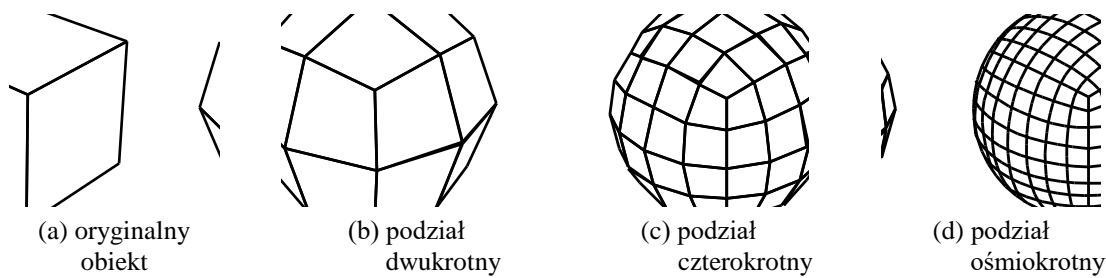

(a) oryginalny
obiekt

(b) podział
dwukrotny

(c) podział
czterokrotny

(d) podział
ośmiokrotny

**Rys. 5.7. Metoda podziału wykorzystywana w teselacji**

Program cieniowania geometrii operuje na obiektach graficznych, generując na ich podstawie nowe obiekty. Podobnie jak teselacja, program cieniowania geometrii może być wykorzystany do zwiększenia szczegółowości obiektu graficznego. Program cieniowania fragmentów przetwarza wycinki obrazu będące reprezentacją obiektów trójwymiarowych, nazywanych fragmentami. Jego zadaniem jest wypełnienie kolorem wszystkich próbek wycinka obrazu. Program cieniowania do obliczeń ogólnych nie jest częścią potoku graficznego, lecz jest elementem dedykowanego potoku obliczeniowego, który wykorzystywany jest do realizacji obliczeń ogólnego przeznaczenia w środowisku OpenGL [146].

W trakcie wykonywania operacji przez potok dane wejściowe pozyskiwane z pamięci układu graficznego przekształcane są na coraz bardziej złożone obiekty, zob. rys. 5.6. Pierwszym krokiem realizowanym przez potok jest akwizycja, a następnie transformacja i rzutowanie wierzchołków do współrzędnych obrazu. Zadanie to realizowane jest przez asembler danych wejściowych, procesor wierzchołków i funkcje przetwarzania końcowego wierzchołków. Ze zrzutowanych wierzchołków budowane są prymitywy graficzne, które po rasteryzacji utworzą fragmenty reprezentujące obrazy prymitywów graficznych. Ostatnim krokiem potoku jest wypełnienie fragmentów kolorem i zapis próbek do obrazu cyfrowego, przechowywanego w pamięci układu graficznego.

Wejściem potoku graficznego jest wyrysowany obiekt graficzny, nazywany prymitywem (ang. *primitive*), który określa kształt rysowanego obiektu i zbiór wierzchołków budujących dany prymityw lub ich serię. Każdy wierzchołek z tego zbioru składa się z identycznej liczby atrybutów określających położenie wierzchołka w przestrzeni oraz innych jego parametrów takich jak np. kolor i współrzędne tekstury. Typy prymitywów można podzielić na pięć kategorii: punkty (zob. rys. 5.8a), linie (zob. rys. 5.8b, c i d), figury geometryczne (zob. rys. 5.8e, f i g), figury geometryczne z sąsiedztwem (zob. rys. 5.9) oraz łaty [146].

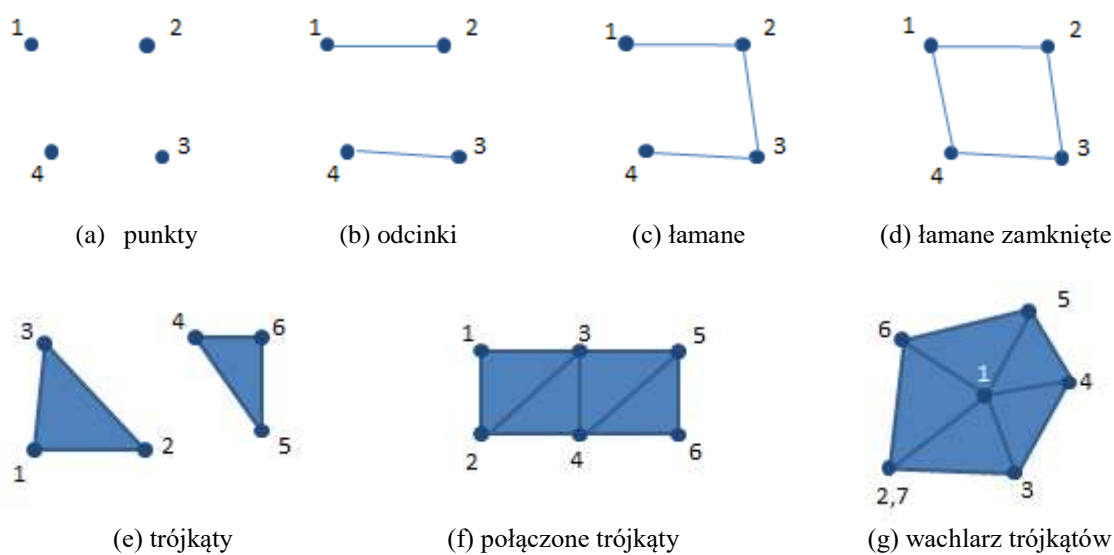

(a) punkty  (b) odcinki  (c) łamane  (d) łamane zamknięte

(e) trójkąty  (f) połączone trójkąty  (g) wachlarz trójkątów

**Rys. 5.8. Podstawowe prymitywy graficzne**

Współczesne układy graficzne optymalizowane są pod kątem rasteryzacji podstawowych typów prymitywów graficznych, tzn. punktów, linii i trójkątów, zob. rys. 5.8. Pozostałe prymitywy graficzne, łaty i figury geometryczne z sąsiedztwem w procesie rasteryzacji przekształcone zostają do prymitywów podstawowych. Z tego powodu prymitywy graficzne z sąsiedztwem, zob. rys. 5.9, wykorzystywane są wyłącznie do analizy obiektów współdzielących wierzchołki lub posiadających wspólną krawędź z wyrysowywanym obiektem. Prymitywy te wykorzystywane są w przypadku stosowania potoku zawierającego program cieniowania geometrii, który przekształca je do prymitywów podstawowych [146], zob. także tabelę 5.1.

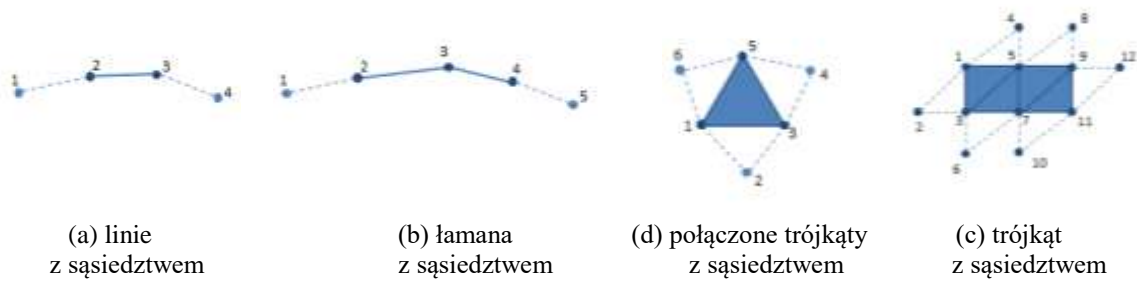

(a) linie  (b) łamana  (d) połączone trójkąty  (c) trójkąt
z sąsiedztwem  z sąsiedztwem  z sąsiedztwem  z sąsiedztwem

**Rys. 5.9. Prymitywy graficzne z sąsiedztwem**

Łaty reprezentują natomiast obiekty graficzne złożone z dowolnej liczby wierz­chołków zdefiniowanych przez twórcę aplikacji. Z tego powodu obiekty zbudowane z łat nie mogą być rasteryzowane bez wykorzystania procesu teselacji, który utworzy renderowalne prymitywy graficzne reprezentujące łaty z określonym poziomem szcze­gółowości [60], zob. tabelę 5.1.

**Tabela 5.1. Wykorzystanie i budowa bufora wierzchołków z $n$ prymitywów**

| Prymityw | Liczba wierzchołków | Wierzchołki prymitywu $i$ ($i \in< 1, n >$) | Program cieniujący |
|---|---|---|---|
| punkty | $n$ | $< i, i >$ | geometrii, fragmentów |
| odcinki (ang. *lines*) | $2n$ | $< 2i - 1, 2i >$ | geometrii, fragmentów |
| łamane (ang. *line strip*) | $n + 1$ | $< i, i + 1 >$ | geometrii, fragmentów |
| łamane zamknięte (ang. *line loop*) | $n + 1$ | $< i, i + 1 > gdy\ i < n$ $< i, 1 > gdy\ i = n$ | geometrii, fragmentów |
| trójkąty (ang. *triangles*) | $3n$ | $< 3i - 2, 3i >$ | geometrii, fragmentów |
| wachlarz trójkątów (ang. *triangle fan*) | $n + 2$ | $< i + 1, i + 2 >$ | geometrii, fragmentów |
| połączone trójkąty (ang. *triangle strip*) | $n + 2$ | $< i, i + 2 >$ | geometrii, fragmentów |
| linia z sąsiedztwem (ang. *line adjacency*) | $4n + 1$ | $< 4i - 2, 4i + 1 >$ | geometrii |
| łamana z sąsiedztwem (ang. *line strip adjacency*) | $n + 2$ | $< i + 1, i + 2 >$ | geometrii |
| trójkąty z sąsiedztwem (ang. *triangles adjacency*) | $6n - 1$ | $< 6i - 5, 6i - 1 >$ | geometrii |
| połączenie trójkątów sąsiedztwem (ang. *triangles strip adjacency*) | $2n - 3$ | $< 2i - 1, 2i - 3 >$ | geometrii |
| łaty (ang. *patches*) zbudowane z $x$ wiechrzołków | $nx$ | $< (i - 1)x + 1, ix >$ | teselacji |

Atrybuty wierzchołków budujących prymitywy przekazywane są do potoku gra­ficznego w buforach wierzchołków (ang. *vertex buffer*) [98]. Bufor, w którym zawarte są wszystkie atrybuty wierzchołków, nazywamy buforem wierzchołków z przeplotem (ang. *interleaved vertex buffer*) [146]. We wspomnianej wcześniej tabeli 5.1 przedsta­wiono liczbę wierzchołków wymaganych do wyrysowania $n$ obiektów określonego typu oraz indeksy wierzchołków należących do zadanego prymitywu graficznego [76]. Tabe­la prezentuje także, jakie programy cieniujące mogą operować na danym prymitywie. Nie uwzględniono w niej programu cieniowania wierzchołków, ponieważ ignoruje on informacje o typie przetwarzanego obiektu graficznego [146]. Jeśli dany prymityw nie może być przetwarzany przez program cieniowania fragmentów, oznacza to, że nie mo­że on zostać wyrasteryzowany bez konwersji do innego typu obiektu graficznego [146], np. trójkąty z sąsiedztwem muszą być zamienione na trójkąty w celu rasteryzacji.

Wynikiem działania potoku graficznego jest dwuwymiarowy obraz reprezentujący trójwymiarową scenę, który przechowywany jest w strukturze określanej mianem bufora ramki (ang. *frame buffer*) [76]. Bufor ramki jest zbiorem dwuwymiarowych buforów identycznego rozmiaru przechowujących informacje o kolorze (ang. *color buffer*), głębi i szablonie (ang. *depth and stencil buffer*) [60]. Bufory te nazywane są także buforami logicznymi [146]. Wspomniane bufory logiczne przechowują w pamięci układu graficznego obrazy, których linie składowane są w kolejności od góry do dołu [98]. Bufory ramki wykorzystywane są zarówno do renderowania obiektów trójwymiarowych na ekranie, jak i renderowania obiektów poza ekranem, tzw. renderingu pozaekranowego (ang. *off-screen rendering*). Tworzenie obiektu bufora ramki odbywa się przez połączenie buforów logicznych z obiektami buforów renderujących (ang. *render buffer*) i obiektami tekstur (ang. *texture*), które służą do składowania pikseli w określonym formacie [146].

Zaletą modułowej budowy współczesnego potoku graficznego jest jej wszechstronność, umożliwiająca m.in. realizację obliczeń ogólnego przeznaczania na GPU [116]. Wszechstronność potoku pozwala także na wykorzystanie modelu kamery [16] w procesie renderingu. W niniejszej pracy wykorzystano tę możliwość do zaimplementowania modelu kamery, który omówiono w podrozdziale 2.2.

# Model kamery w OpenGL

Pozycja wierzchołka na generowanym obrazie wyznaczana jest przez procesor wierzchołków oraz funkcje przetwarzania końcowego wierzchołków. Głównym zadaniem procesora wierzchołków jest wyznaczenie transformacji geometrycznych i projekcja wierzchołka wejściowego $v_{model}(x_{model}, y_{model}, z_{model})$, określonego we współrzędnych modelu (ang. *model space*) do współrzędnych $v_{view}(x_{view}, y_{view}, z_{view}, w_{view})$. Atrybuty wierzchołka $x_{view}, y_{view}, z_{view}$ określają jego pozycję w przestrzeni widoku (ang. *view space*), natomiast atrybut $w_{view}$ służy do skalowania rozmiaru obiektów na obrazie w zależności od odległości do kamery [98]. Wartość atrybutu $w_{view}$ zależy bezpośrednio od wykorzystywanej metody projekcji wierzchołków [120] i w przypadku projekcji ortogonalnej wartość atrybutu jest zawsze równa jeden [98]. Przetwarzanie końcowe wierzchołków przekształca kolejno dane wyjściowe udostępniane na wyjściu programowalnego procesora wierzchołków na odpowiadającą im współrzędną $v_{window}(x_{window}, y_{window}, z_{window})$ okna (ang. *window space*), która składa się z pozycji piksela na obrazie i głębokości piksela. Ponadto przetwarzanie końcowe wierzchołków służy m.in. do określenia kierunku powierzchni i umożliwia zapis przetworzonych przez procesor wierzchołków do pamięci układu graficznego. Do elementów związanych z transformacją wierzchołka do współrzędnych obrazu zaliczamy:

- obcinanie (ang. *clipping*),
- dzielenie perspektywiczne (ang. *perspective divide*),
- transformację widoku (ang. *viewport transform*).

Operacje te realizowane są sekwencyjnie dla każdego z wierzchołków wejściowych. Obcięcie [146] jest procesem polegającym na odrzuceniu wierzchołków obiektu graficznego wychodzących poza dopuszczalny zakres wartości przestrzeni obcięcia (ang. *clip volume*), zob. rys. 5.10. Dopuszczalny zakres wartości definiujący przestrzeń obcięcia może zostać określony przez twórcę aplikacji.

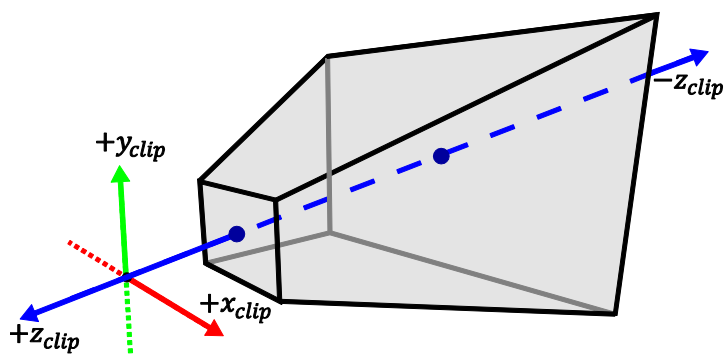

**Rys. 5.10. Przestrzeń obcięcia dla projekcji perspektywicznej**

Wierzchołkami w przestrzeni obcięcia $v_{clip}(x_{clip}, y_{clip}, z_{clip}, w_{clip})$ są wszystkie wierzchołki $v_{view}(x_{view}, y_{view}, z_{view}, w_{view})$ spełniające równanie:

$$-w_{view} \leq x_{view} \leq w_{view} \,,$$
$$-w_{clip} \leq y_{clip} \leq w_{view} \quad, \qquad\qquad (5.8)$$
$$-w_{view} \leq z_{view} \leq w_{view} \,.$$

Wynik procesu obcięcia zależy od typu przetwarzanego prymitywu graficznego [146]. Wszystkie wierzchołki obiektu są usuwane jedynie w przypadku, gdy wszystkie wierzchołki prymitywu nie znajdują się w przestrzeni obcięcia (5.8) [146]. Gdy prymityw leży na krawędzi przestrzeni obcięcia, tworzone są nowe obiekty graficzne, które aproksymują fragment obiektu znajdującego się wewnątrz przestrzeni obcięcia.

Po obcięciu wierzchołki przekształcane są do znormalizowanych współrzędnych urządzenia $v_{ndc}(x_{ndc}, y_{ndc}, z_{ndc})$ [146]. Operacja ta realizowana jest zgodnie z równaniem dzielenia perspektywicznego:

$$\begin{bmatrix} x_{ndc} \\ y_{ndc} \\ z_{ndc} \end{bmatrix} = \begin{bmatrix} \dfrac{x_{clip}}{w_{clip}} \\ \dfrac{y_{clip}}{w_{clip}} \\ \dfrac{z_{clip}}{w_{clip}} \end{bmatrix}. \qquad\qquad (5.9)$$

Przestrzeń znormalizowanych współrzędnych urządzenia reprezentowana jest przez sześcian o długości boku, który jest równy dwóm jednostkom, zob. rys. 5.11.

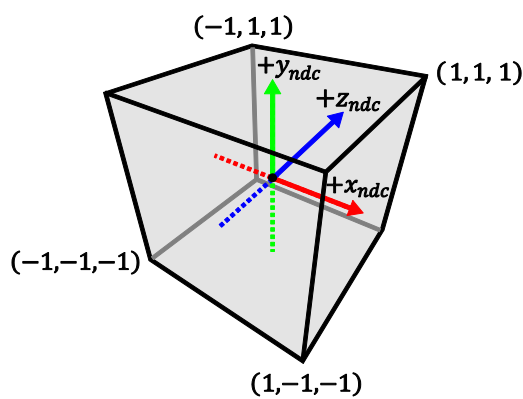

**Rys. 5.11. Przestrzeń znormalizowanych koordynat urządzenia**

Ostatnim krokiem jest przekształcenie znormalizowanych koordynat urządzenia do współrzędnych okna [146], które nazywamy także współrzędnymi ekranu (ang. *screen space*). Koordynaty okna utożsamiane są ze współrzędnymi pikseli na płaszczyźnie bufora ramki wyrażonymi w pikselach $x_{window}$, $y_{window}$ oraz wartością głębi piksela bufora ramki $z_{window}$.

Operacja konwersji znormalizowanych współrzędnych widoku do współrzędnych okna realizowana jest przez transformację widoku, która określona jest równaniem:

$$\begin{bmatrix} x_{window} \\ y_{window} \\ z_{window} \end{bmatrix} = \begin{bmatrix} 0.5p_x x_{ndc} + o_x \\ 0.5p_y y_{ndc} + o_y \\ 0.5(f-n)z_{ndc} + 0.5(n+f) \end{bmatrix}. \tag{5.10}$$

We wspomnianej zależności punkt $o_x$, $o_y$ jest środkiem okna o szerokości $p_x$ i wysokości $p_y$, określającego fragment ekranu o wymiarach $s_x$ na $s_y$ pikseli. Wartości $n$ i $f$ określają minimalną i maksymalną wartość bufora głębi, który stanowi fragment bufora ramki i wykorzystywany jest do realizacji przesłonięć obiektów. Omawiana przestrzeń widoku zaprezentowana została w sposób poglądowy na rys. 5.12. Wyznaczanie przesłonięć realizowane jest z wykorzystaniem algorytmu *z-culling* [60,173].

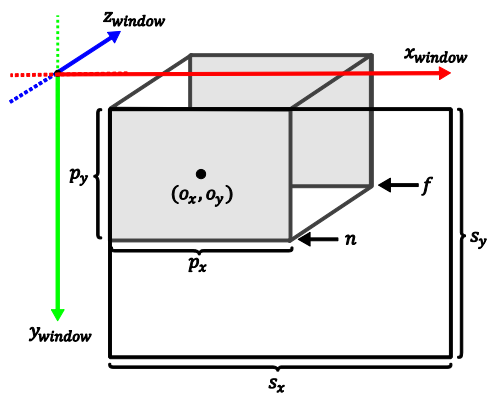

**Rys. 5.12. Przestrzeń widoku w OpenGL**

W OpenGL transformacje geometryczne realizowane są przy pomocy macierzy transformacji modelu, projekcji i widoku, zob. rys. 5.13. Macierz modelu $M_{model}$ (ang. *model matrix*) wykorzystywana jest do transformacji modelu w lokalnym układzie współrzędnych obiektu graficznego, natomiast macierz widoku $M_{view}$ (ang. *view matrix*) służy do transformacji współrzędnych z koordynat modelu do współrzędnych globalnych. Połączenie obydwu transformacji daje nam macierz, która nazywana jest macierzą modelu i widoku $M_{modelview}$ (ang. *model-view matrix*). Macierz projekcji $M_{proj}$ (ang. *projection matrix*), nazywana też macierzą kamery, służy do definiowania metody rzutowania trójwymiarowego wierzchołka do współrzędnych widoku [98]. Schemat transformacji geometrycznych zilustrowano w sposób poglądowy na rys. 5.13.

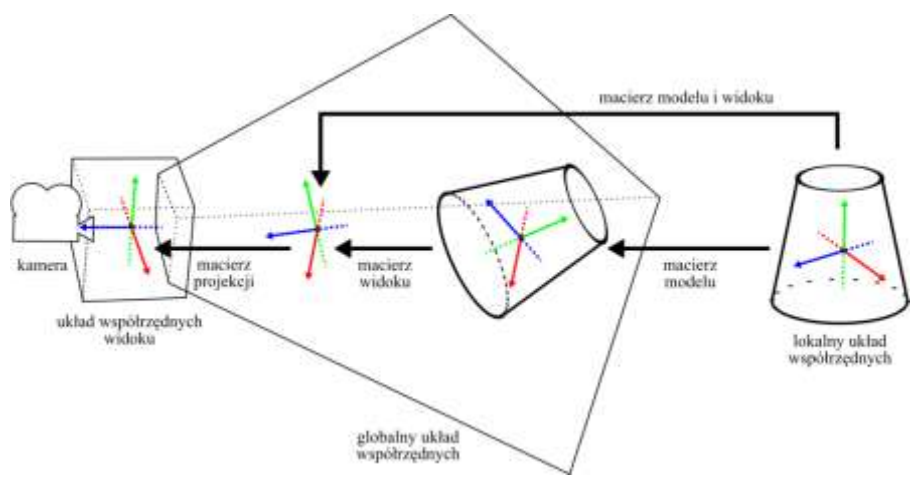

**Rys. 5.13. Transformacje macierzy w OpenGL**

Przekształcenie wierzchołków modelu $v_{model}$ do wierzchołka $v_{view}$ we współrzędnych widoku realizowane jest zgodnie z następującym równaniem:

$$v_{view} = M_{proj}\left( (M_{view} M_{model}) \begin{bmatrix} v_{object} \\ 1 \end{bmatrix} \right) = M_{proj}\left( M_{modelview} \begin{bmatrix} v_{object} \\ 1 \end{bmatrix} \right). \qquad (5.11)$$

W obecnej wersji OpenGL [146] wyznaczanie współrzędnych $v_{view}$ realizowane jest przez program cieniowania wierzchołków, w którym definiuje się zmienne jednolite (ang. *uniform*) typu mat4 przechowujące macierze $M_{modelview}$ i $M_{proj}$, które dostępne są dla wszystkich wątków zadania. Każdy wątek przetwarza jeden wierzchołek wejściowy pobrany z bufora wierzchołków. Koordynaty wyjściowe wierzchołka zapisywane są do predefiniowanej zmiennej, która przekazywana jest do następnego elementu potoku graficznego.

Wykorzystując zależność pomiędzy wierzchołkiem $v_{view}$ generowanym przez procesor wierzchołków, a współrzędnymi $v_{window}$ wyznaczanymi w procesie przetwarzania końcowego, można określić związek pomiędzy wierzchołkiem $v_{view}$, a współrzędnymi $u, v$ obrazu cyfrowego wyznaczonymi zgodnie z modelem kamery [16].

Omawiana zależność przyjmuje postać równania (5.12), w której $i_x$, $i_y$ określają wysokość i szerokość obrazu cyfrowego, natomiast $d$ jest odległością wierzchołka $v_{view}$ od kamery.

$$v_{view} = \begin{bmatrix} x_{view} \\ y_{view} \\ z_{view} \\ w_{view} \end{bmatrix} = \begin{bmatrix} \dfrac{u}{i_x} \\ 1 - \dfrac{v}{i_y} \\ d \\ 1 \end{bmatrix} \tag{5.12}$$

Ponieważ w OpenGL oś $y$ we współrzędnych widoku skierowana jest do góry [113], zob. rys. 5.10, natomiast po realizacji transformacji widoku (5.10) oś $y$ współrzędnych okna jest skierowana w dół, zob. rys. 5.12, obraz w buforze ramki zapisywany jest w kolejności linii od dołu do góry. Z tego powodu realizowane jest odbicie pionowe wierzchołków we współrzędnych widoku (5.12), a co za tym idzie obraz po zapisie do bufora ramki jest przechowywany w kolejności linii od góry do dołu.

## Zaproponowana metoda rasteryzacji modelu

Celem rasteryzacji modelu jest wygenerowanie obrazu cyfrowego reprezentującego model w zadanej pozie wraz z wyrenderowanymi krawędziami. Obraz cyfrowy modelu i krawędzi rasteryzowany jest zgodne z przyjętym modelem kamery [16]. Obraz rasteryzowany jest dla każdej kamery systemu wizyjnego i składa się z pikseli zapisanych na ośmiu bitach. Kolor wyrysowanego obiektu reprezentowany jest przez wartość całkowitą $p_{ub}$ kodowaną na pierwszych siedmiu bitach oraz flagi określającej wystąpienie krawędzi, która zapisana jest na ostatnim bicie. W przyjętej reprezentacji fragmenty modelu można wyrysować w 127 unikalnych kolorach. Tabela 5.2 prezentuje przyjęty sposób kodowania.

**Tabela 5.2. Piksel generowanego modelu**

| Bit | 0 | 1 | 2 | 3 | 4 | 5 | 6 | 7 |
|---|---|---|---|---|---|---|---|---|
| **Wartość** | kolor piksela $p_{ub} \in\; <0, 2^7 - 1>$ | | | | | | | flaga krawędzi |

Atrybut koloru wierzchołka przekazywanego do potoku zapisany jest w postaci znormalizowanej liczby rzeczywistej $p$ wyznaczonej zgodnie z równaniem (5.13), w którym $p_{ub}$ określa kolor piksela zapisany na siedmiu bitach.

$$p = \frac{max(p_{ub}, 127.0)}{255.0} \tag{5.13}$$

W języku GLSL kolory reprezentowane są w postaci wektora zbudowanego z czterech znormalizowanych liczb zmiennoprzecinkowych [131]. Wartość piksela zapisywanego w buforze ramki niekoniecznie musi przechowywać wszystkie komponenty wektora koloru wykorzystywanego przez GLSL. Co więcej, dane komponenty wektora mogą być przechowywane w buforze ramki jako wartości zmiennoprzecinkowe lub wartości całkowite [146]. Konwersja koloru z postaci wektora znormalizowanego do formatu zapisanego w buforze ramki realizowana jest automatycznie przez OpenGL.

Proces rasteryzacji modelu sylwetki realizowany jest w dwóch krokach. W pierwszym z nich model kształtu przekształcany jest do prymitywów graficznych, których powierzchnia zostanie zamalowana kolorem o wartości $p$. W drugim kroku wyznaczane i wyrysowywane są linie stanowiące krawędzie modelu zbudowanego z prymitywów graficznych. Oba kroki realizowane są przez potoki graficzne. Jednak z powodów praktycznych implementacja potoków graficznych zależy od przyjętej reprezentacji modelu kształtu. W opracowanych rozwiązaniach wykorzystywane są dwa potoki graficzne: potok renderowania bezpośredniego i potok renderowania pośredniego prymitywów graficznych. Potok renderowania bezpośredniego, zob. rys. 5.14, wykorzystuje program cieniowania wierzchołków i cieniowania fragmentów do wyrysowania na buforze ramki prymitywów wejściowych, składających się z wierzchołków zapisanych w buforze wierzchołków.

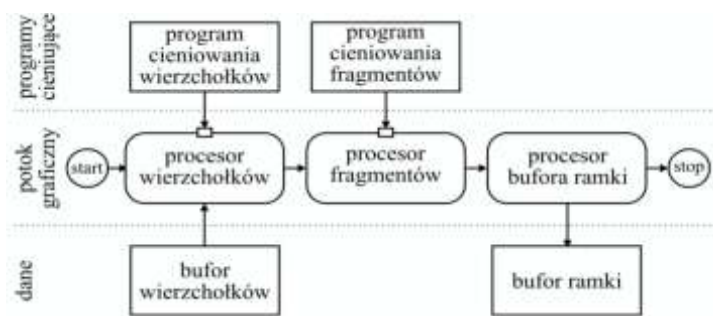

**Rys. 5.14. Potok renderingu bezpośredniego prymitywów graficznych**

Potok rysowania pośredniego, zob. rys. 5.15, składa się z programu cieniowania wierzchołków, programu cieniowania geometrii oraz programu cieniowania fragmentów. Ponieważ w potoku wykorzystywany jest program cieniowania geometrii, prymitywy wejściowe zostaną przekształcone na inne obiekty graficzne. Z tego powodu potok ten nazywamy potokiem renderowania pośredniego.

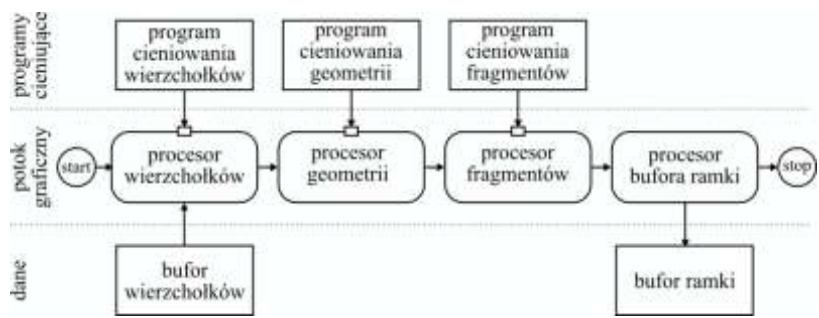

**Rys. 5.15. Potok renderowania pośredniego prymitywów graficznych**

Sposób wykorzystania zaprezentowanych powyżej potoków zostanie przybliżony poniżej na przykładzie procesu renderingu modelu kształtu.

## Wyrysowanie płaskiego modelu kształtu

Operacja wyrysowania płaskiego modelu kształtu wykorzystuje potok graficzny renderowania bezpośredniego, zob. rys. 5.15. Zastosowanie programu cieniowania geometrii jest konieczne, ponieważ dla każdej konfiguracji modelu kształtu generowane będą wierzchołki trapezu zgodnie z metodą przedstawioną w podrozdziale 4.3. Oznacza to, że danymi wejściowymi potoku będą parametry stożków ściętych wykorzystanych w modelu, zob. tabela 4.6 i rys. 5.16a. W trakcie działania potoku program cieniowania geometrii posługuje się parametrami stożków ściętych i wyznacza wierzchołki trapezów, zob. rys. 5.16b, które renderowane będą w formie trójkątów, zob. rys. 5.16c, natomiast krawędzie modelu wyrysowane będą w postaci linii łączących wierzchołki wyznaczonych trapezów, zob. rys. 5.16d.

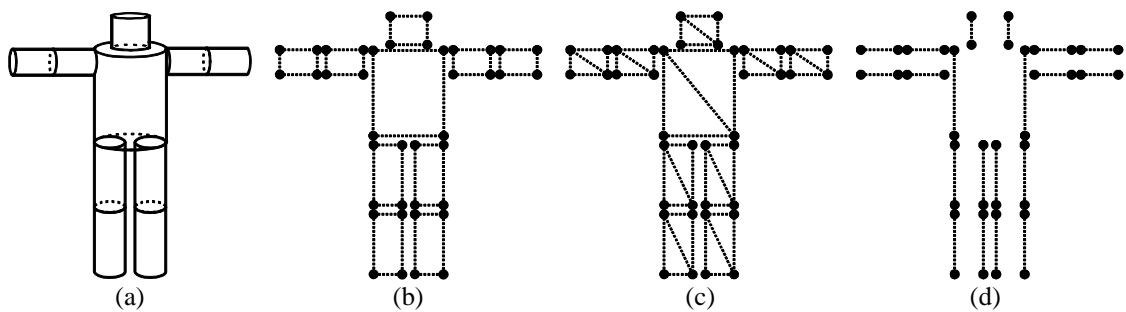

(a)      (b)      (c)      (d)

**Rys. 5.16. Rendering modelu płaskiego w OpenGL**

Parametry stożków ściętych przekazywane są do potoku w formie wierzchołków prymitywu trójkąta, którego wierzchołki składają się z czterech zmiennoprzecinkowych atrybutów. W atrybutach wierzchołków trójkąta zakodowane są parametry stożka ściętego reprezentującego dany element modelu. W odróżnieniu od programu cieniowania wierzchołków, program cieniowania geometrii: (i) posiada dostęp do wierzchołków przetwarzanego podzbioru prymitywów wraz z ich atrybutami, (ii) przetwarza kolejne podzbiory wierzchołków spośród wszystkich wierzchołków. Kodowanie parametrów wejściowych stożka ściętego w atrybutach wejściowych wierzchołków realizowane jest przez umieszczenie w pierwszym wierzchołku współrzędnych środka i promienia koła budującego podstawę górną stożka ściętego $v^t(t_x, t_y, t_z, t_r)$. Drugi wierzchołek, podobnie jak pierwszy, przechowuje informację o podstawie dolnej $v^b(b_x, b_y, b_z, b_r)$. Ostatni wierzchołek zawiera informację o kolorze, w jakim wyrysowywany będzie stożek oraz indeks macierzy transformacji globalnej $v^{pb}(p, b, 0, 0)$. Ze względu na to, że każdy wierzchołek musi składać się ze stałej liczby atrybutów, ostatnie dwa atrybuty trzeciego wierzchołka są wypełnione zerami. Tabela 5.3 przedstawia ułożenie danych w pamięci dla modelu 3D postaci ludzkiej, który zaprezentowano w podrozdziale 4.3.

**Tabela 5.3. Opis modelu płaskiego za pomocą wierzchołków**

| Wierzchołek | 0 | 1 | 2 | 3 | 4 | 5 | ... | 39 | 40 | 41 |
|---|---|---|---|---|---|---|---|---|---|---|
| Część modelu | 0 | | | 1 | | | ... | 14 | | |
| Parametry części modelu | $v_0^t$ | $v_0^b$ | $v_0^{pb}$ | $v_1^t$ | $v_1^b$ | $v_1^{pb}$ | ... | $v_{14}^t$ | $v_{14}^b$ | $v_{14}^{pb}$ |

Wierzchołki generowane przez program cieniowania geometrii budują dwa trójkąty reprezentujące trapez skierowany w stronę kamery. Wyznaczenie wierzchołków odbywa się zgodnie z metodą tworzenia bilbordu [5]. Zastosowana metoda tworzenia bilbordu wykorzystuje bufor danych współdzielonych z programem cieniującym (ang. *Shared Storage Buffer Object, SSBO*), który umożliwia dostęp do macierzy transformacji globalnych modelu szkieletowego postaci ludzkiej.

Do generowania wierzchołków może być wykorzystana także alternatywna metoda zaprezentowana w pracy [140]. Wierzchołek wygenerowany przez program cieniowania geometrii składa się ze współrzędnych wierzchołka, wyznaczonego zgodnie z równaniem (5.12) oraz koloru wykorzystywanego podczas renderingu modelu. Kolor reprezentowany jest przez cztery komponenty składowe RGBA [73]. Na ostatnim komponencie składowym przechowywana jest informacja o przezroczystości (ang. *alpha information*).

# Model siatkowy

Jak już wspomniano w podrozdziale 4.2, do reprezentacji sylwetki wykorzystywany jest model siatkowy. Wyrysowanie modelu siatkowego wymaga wykorzystania dwóch różnych potoków graficznych. Zadaniem pierwszego potoku jest zamalowanie powierzchni modelu siatkowego. Potok ten wykorzystuje program cieniowania wierzchołków, program cieniowania fragmentów, zob. rys. 5.14, oraz rysowanie indeksowe (ang. *indexed draw*). Rysowanie indeksowe obejmuje przekazanie do potoku dwóch buforów, z których pierwszy zawiera wierzchołki modelu, natomiast drugi zawiera indeksy bufora wierzchołków i wykorzystywany jest podczas składania prymitywów graficznych. Siatka o takiej budowie nazywana jest siatką indeksowaną (ang. *indexed mesh*) [60]. W modelu siatkowym wierzchołki wejściowe są opisane przez współrzędne wierzchołka $v_{model}(x_{model}, y_{model}, z_{model})$, kolor wierzchołka $p$ i indeks $b$ macierzy transformacji globalnej. Zadaniem programu cieniowania wierzchołków jest transformacja wierzchołka ze współrzędnych modelu do współrzędnych widoku oraz komponentów koloru RGBA.

Wyrysowanie krawędzi modelu realizowane było w oparciu o metodę zaproponowaną w pracy [5]. Potok graficzny odpowiedzialny za wyznaczenie i wyrysowanie krawędzi modelu składa się z programu cieniowania wierzchołków, geometrii i fragmentów, zob. rys. 5.15. Potok graficzny przyjmuje na wejściu strukturę, która opisuje wierzchołki typu prymityw trójkąta z sąsiedztwem. Struktura ta składa się z sześciu wierzchołków opisujących trójkąt i trzy przylegające do niego trójkąty. Dzięki takiej strukturze wyznaczyć można załamania powierzchni, które tworzą krawędzie modelu.

Załamaniem powierzchni nazywamy krawędź pomiędzy dwoma prymitywami, których wektory normalne skierowane są w przeciwne strony [98,131].

# Optymalizacja potoków graficznych

Rendering modelu postaci w środowisku OpenGL realizowany jest przez dwa potoki graficzne uruchamiane sekwencyjnie. Pierwszy z nich służy do wyrysowania sylwetki modelu kształtu, natomiast drugi wykorzystywany jest do wyrysowania krawędzi modelu. Wspomniana kolejność uruchamiania potoków zapewnia poprawną realizację przesłonięć sylwetki modelu i jego krawędzi. Analogiczna sytuacja występuje w potokach wykorzystywanych do renderingu modelu siatki, które korzystają z identycznego programu cieniowania wierzchołków odpowiedzialnego za ich rzutowanie do powierzchni obrazu. Porównanie potoków graficznych renderingu modelu kształtu i krawędzi modelu zaprezentowano na rys. 5.17. Na wspomnianym rysunku można zauważyć, że operacja rzutowania wierzchołków w oparciu o przyjęty model kamery realizowana jest dwukrotnie. Pozostałe operacje, tzn. rasteryzacja, przetwarzanie fragmentów i przetwarzanie próbek wykorzystywane są do wypełniania powierzchni modelu oraz jego krawędzi zadanym kolorem, a ich realizacja i implementacja zależy od wykorzystywanego modelu kształtu. Celem zmniejszenia nakładów obliczeniowych wynikających z podwójnego rzutowania wierzchołków, część wspólna, która wykorzystuje macierze transformacji globalnych oraz parametry kamery, została oddzielona od potoków renderingu sylwetki modelu i krawędzi modelu. Wydzielony fragment tworzy nowy potok graficzny, który jest odpowiedzialny za wyznaczenie zrzutowanych wierzchołków modelu kształtu dla zadanej konfiguracji modelu szkieletowego i ich zapis w pamięci układu graficznego. Omawiany potok graficzny wykorzystuje funkcję sprzężenia zwrotnego wierzchołków, która realizowana jest w bloku przetwarzania końcowego wierzchołków, zob. rys. 5.18.

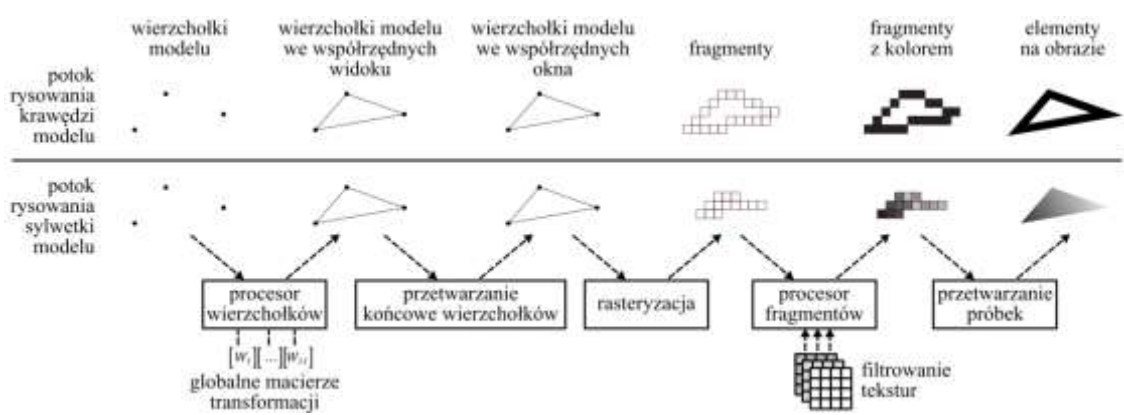

**Rys. 5.17. Postać ogólna operacji potoku graficznego**

Sprzężenie zwrotne umożliwia zapis wszystkich atrybutów wierzchołków do pamięci układu graficznego. Schemat potoku graficznego wykorzystującego mechanizm sprzężenia zwrotnego przedstawiony został na rys. 5.18. Potok sprzężenia zwrotnego nie wykorzystuje procesora fragmentów i nie modyfikuje zawartości bufora ramki. Oznacza to, że potok zakończy pracę po zakończeniu przetwarzania końcowego wierzchołków.

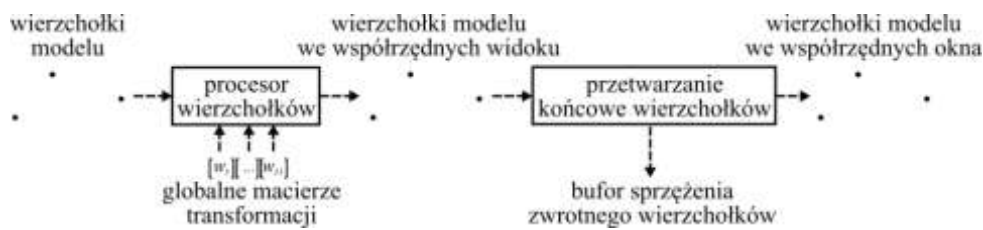

**Rys. 5.18. Potok sprzężenia zwrotnego**

Zapisane w pamięci wierzchołki składają się z czterech atrybutów, określających pozycję wierzchołka $v_{view}(x_{view}, y_{view}, z_{view})$ we współrzędnych widoku (5.12) oraz kolor wierzchołka $p$ (5.13). Tabela 5.4 przedstawia rozmieszczenie atrybutów wierzchołków w pamięci bufora sprzężenia zwrotnego. Liczba elementów przechowywanych w buforze zależy od wykorzystywanego typu modelu.

**Tabela 5.4. Rozmieszczenie atrybutów w buforze sprzężenia zwrotnego**

| Wierzchołek | 0 | | | | 1 | | | | ... |
|---|---|---|---|---|---|---|---|---|---|
| **Atrybut** | $x_{view}^0$ | $y_{view}^0$ | $z_{view}^0$ | $p^0$ | $x_{view}^1$ | $y_{view}^1$ | $z_{view}^1$ | $p^1$ | ... |

Wierzchołki zapisane w buforze sprzężenia zwrotnego wykorzystywane są jako dane wejściowe potoków realizujących rasteryzację modelu i jego krawędzi. Ponieważ format danych zapisanych przez potok sprzężenia zwrotnego różni się od formatu potoków rysujących, wymagane jest przystosowanie programów cieniujących tych potoków do nowego formatu danych wejściowych. Gdy stosowany jest model płaski, program cieniowania geometrii nie jest wykorzystywany, ponieważ wierzchołki modelu są generowane przez potok sprzężenia zwrotnego. Ten wykorzystywany jest także do określenia pozy modeli dla wszystkich hipotez generowanych przez filtr cząsteczkowy lub algorytm optymalizacji w oparciu o rój cząsteczek. Kształt sylwetki określany jest przez położenie wierzchołków składających się na model. Wyznaczanie położeń wierzchołków realizowane jest jednocześnie dla wszystkich kamer. Wspomniana operacja realizowana jest dzięki rysowaniu instancyjnemu (ang. *instanced draw*) [146], które umożliwia wielokrotne przetworzenie obiektu przez jeden potok graficzny. W grafice komputerowej rysowanie instancyjne wykorzystywane jest m.in. do renderingu tłumów [39]. Każda instancja przetwarzanego obiektu jest identyfikowana przez wartość $id \in \langle 0, ID \rangle$, gdzie $ID$ jest liczbą instancji obiektu. W opracowanym systemie liczba instancji wyznaczana jest zgodnie z równaniem (5.14), gdzie $C$ reprezentuje liczbę kamer w systemie wizyjnym, zaś $N$ oznacza liczbę konfiguracji modeli do przetworzenia.

$$ID = C \cdot N \tag{5.14}$$

Wierzchołki modelu o zadanej konfiguracji rzutowane do współrzędnych danej kamery zapisane są obok siebie w buforze sprzężenia zwrotnego. Do rzutowania wierzchołków konieczne jest wyznaczenie identyfikatora kamery oraz identyfikatora konfiguracji przetwarzanego modelu. Dla zadanej instancji $id$ obiektu, identyfikator kamery $c_{id}$ wyznaczany jest zgodnie z równaniem:

$$c_{id} = mod(id, C) \,. \tag{5.15}$$

Wartość $c_{id}$ wykorzystywana jest do pozyskania parametrów kalibracji kamery, dla której rzutowany będzie model. Parametry kamer przechowywane są w jednorodnym obiekcie programu cieniowania (ang. *Uniform Buffer Object, UBO*), który utożsamiany jest z tablicą przechowującą wartości zadanego typu. Natomiast identyfikator konfiguracji $n_{id}$ wyznaczany jest zgodnie z równaniem:

$$n_{id} = \frac{id - c_{id}}{C} \,. \tag{5.16}$$

W oparciu o identyfikator konfiguracji modelu $n_{id}$ i identyfikator macierzy transformacji globalnej $b$ przetwarzanego wierzchołka wyznaczyć można identyfikator macierzy transformacji globalnej dla zadanej konfiguracji modelu:

$$b_{id} = b + B \cdot n_{id} \,, \tag{5.17}$$

gdzie $B$ jest liczbą macierzy transformacji globalnych modelu szkieletowego. Liczba macierzy transformacji globalnej modelu jest równa liczbie kości modelu (zob. tabela 4.4).

Jednoczesna rasteryzacja wielu obrazów modeli możliwa jest przez utworzenie obiektu bufora ramki, którego wymiary będą wielokrotnością rozdzielczości obrazu cyfrowego reprezentującego model. Maksymalny rozmiar bufora ramki zależy od wykorzystanego układu graficznego i dostępnej pamięci, jednak w najgorszym wypadku rozmiar bufora ramki nie powinien być mniejszy niż 16384x16384 pikseli [145]. Wymiary bufora ramki nie są uzależnione od reprezentacji piksela w pamięci. Przez wzgląd na to, że GLSL przechowuje wartości kolorów w postaci wektora złożonego z czterech znormalizowanych liczb rzeczywistych, wykorzystywany załącznik kolorów bufora ramki zmodyfikowany został tak, aby możliwe było przechowywanie wszystkich komponentów składowych RGBA koloru. Komponenty składowe koloru przechowywane są w postaci ośmiobitowej wartości całkowitej. Umożliwia to umieszczenie w każdym z nich jednego obrazu modelu w odcieniach szarości. Operacja rysowania zadanego obrazu na wybranym komponencie kolorów wymaga wykorzystania funkcji mieszania kolorów (ang. *blending*), która realizowana jest w kroku przetwarzania próbek potoku graficznego, zob. rys. 5.6. Funkcja mieszania kolorów umożliwia wyznaczenie wartości koloru, który zapisywany jest w buforze ramki na podstawie wartości aktualnie prze-chowywanej oraz koloru zwróconego przez procesor fragmentów. OpenGL udostępnia kilka sposobów mieszania kolorów [146], jednak w opracowanym systemie wykorzy-stywana jest zależność określająca wartość maksymalną, która wynika z równania (5.18). W omawianym równaniu $p_r(r_r, g_r, b_r, a_r)$ jest wartością koloru, który zapisany zosta-nie w buforze ramki, $p_s(r_s, g_s, b_s, a_s)$ jest kolorem zwróconym przez procesor fragmen-tów, nazywanym także kolorem źródłowym, natomiast $p_d(r_d, g_d, b_d, a_d)$ jest kolorem aktualnie przechowywanym w buforze ramki, zwanym również kolorem docelowym. Ponieważ potok wyrysowania krawędzi jest uruchamiany po potoku wyrysowania syl-wetki, kolor krawędzi musi być reprezentowany przez wartość większą od koloru repre-zentującego sylwetkę.

$$\begin{bmatrix} r_r \\ g_r \\ b_r \\ a_r \end{bmatrix} = \begin{bmatrix} max(r_s, r_d) \\ max(g_s, g_d) \\ max(b_s, b_s) \\ max(a_s, a_s) \end{bmatrix}$$ (5.18)

Mieszanie kolorów realizowane jest jedynie w przypadku, gdy test głębi (ang. *depth test*), wykorzystywany w algorytmie *z-culling*, zakończy się pomyślnie. Po-dobnie jak miało to miejsce w przypadku mieszania kolorów, OpenGL definiuje kilka możliwych metod testu głębi [145]. W opracowanym systemie wykorzystywany jest test głębi opisany przez równanie:

$$z_s \leq z_d \,,$$ (5.19)

gdzie $z_d$ określa wartość przechowywaną w buforze głębi, natomiast wartość $z_s$ określa wartość głębi rysowanego piksela.

Wykorzystanie funkcji testu głębi umożliwia wyrysowanie oddzielnych obrazów na każdym z komponentów bufora ramki. Najprostszą metodą wyrysowania oddzielnych obrazów jest wyrysowanie modeli i krawędzi na zadanym komponencie kolorów, a następnie wyczyszczenie bufora głębi. W niniejszej pracy rozwiązanie to nie jest wy-korzystywane, ponieważ operacja czyszczenia bufora głębi wiąże się z dodatkowymi nakładami obliczeniowymi. Alternatywną metodą jest podzielenie bufora głębi na czte-ry części, które wykorzystywane są przez obiekty rysowane na zadanych komponentach koloru. Wykorzystanie takiego rozwiązania umożliwia wyrysowanie modeli na zada-nych komponentach bez konieczności czyszczenia zawartości bufora ramki. Wykorzy-stanie wspomnianego podejścia wymaga jednak innej metody wyznaczania kolorów, które wykorzystywane są do zamalowania krawędzi oraz sylwetki modelu. W tym celu na podstawie równania (5.20) określana jest wartość $l$:

$$l = mod(n_{id}, 4) \,,$$ (5.20)

która reprezentuje komponent, na którym realizowane jest rysowanie. Wartość zero reprezentuje składową czerwoną, wartość jeden zieloną, wartość dwa niebieską, natomiast wartość trzy reprezentuje komponent opisujący przezroczystość piksela.

Wartość $l$ wykorzystana jest do modyfikacji koloru $p$ wierzchołka zgodnie z równaniem:

$$p' = 256^{l-3}p\,. \tag{5.21}$$

Z kolei wartość $p'$ przekształcona jest przez programy cieniowania wierzchołków na wektor komponentów RGBA zgodnie z równaniem (5.22), gdzie $p'_{rgba}(p'_r, p'_g, p'_b, p'_a) = (256^3 p', 256^2 p', 256 p', p')$.

$$\begin{bmatrix} p_r \\ p_g \\ p_b \\ p_a \end{bmatrix} = \left( \left( \begin{bmatrix} p'_r \\ p'_g \\ p'_b \\ p'_a \end{bmatrix} - \begin{bmatrix} \lfloor p'_r \rfloor \\ \lfloor p'_g \rfloor \\ \lfloor p'_b \rfloor \\ \lfloor p'_a \rfloor \end{bmatrix} \right) - \left( \begin{bmatrix} p'_r \\ p'_r \\ p'_g \\ p'_b \end{bmatrix} - \begin{bmatrix} \lfloor p'_r \rfloor \\ \lfloor p'_r \rfloor \\ \lfloor p'_g \rfloor \\ \lfloor p'_b \rfloor \end{bmatrix} \right) \right) \begin{bmatrix} 0 \\ 256^{-1} \\ 256^{-1} \\ 256^{-1} \end{bmatrix} \tag{5.22}$$

Wymagana jest także modyfikacja głębokości danego wierzchołka w zależności od komponentu koloru, na którym jest on rysowany. Aktualna głębokość wierzchołka $d_{id}$ wyznaczana jest zgodnie z równaniem (5.23), gdzie $d$ określa odległość wierzchołka od kamery:

$$d_{id} = \frac{(3-l)+d}{4}\,. \tag{5.23}$$

Ostatnią operacją umożliwiającą wyrysowanie oddzielnych obrazów modeli na składowych komponentach RGBA, która musi być realizowana w programie cieniowania wierzchołków, jest operacja wyznaczania pozycji wierzchołka we współrzędnych widoku. Operacja ta realizowana jest zgodnie z równaniem:

$$x_{view} = \frac{2\left\lfloor \frac{F_x}{i_x} \right\rfloor}{i_x} \left( max\left(0, min\left(1, \frac{u}{i_x}\right)\right) + mod\left(\frac{Nc_{id} + n_{id}}{4}, \left\lfloor \frac{F_x}{i_x} \right\rfloor\right) \right) \quad ,$$

$$y_{view} = 1 - \frac{2\left\lfloor \frac{F_y}{i_y} \right\rfloor}{i_y} \left( max\left(0, min\left(1, \frac{v}{i_y}\right)\right) + mod\left(\left\lfloor \frac{Nc_{id} + n_{id}}{4 i_x} \right\rfloor, \left\lfloor \frac{F_y}{i_y} \right\rfloor\right) \right) \quad , \tag{5.24}$$

w którym $F_x$ i $F_y$ określają rozdzielczość bufora ramki, natomiast $i_x$ i $i_y$ określają rozdzielczość obrazu modelu. Rozdzielczość bufora ramki musi umożliwiać wyrysowanie wszystkich obrazów modeli jednocześnie. Oznacza to, że rozdzielczość bufora ramki musi spełniać zależność $\lfloor F_y / i_y \rfloor \cdot \lfloor F_x / i_x \rfloor \cdot 4 \geq C \cdot N$.

Istotnym aspektem renderingu modelu postaci jest synchronizacja potoków graficznych. Ponieważ cały proces renderingu odbywa się na pozaekranowych buforach ramki, podczas synchronizacji potoków ignorowany jest czas wymiany (ang. *swap interval*) [146].

Wspomniany czas ma wpływ na utrzymanie synchronizacji pomiędzy urządzeniem wyjściowym pracującym z daną częstotliwością a poleceniami interfejsu graficznego. Jeśli czas wymiany jest uwzględniany, synchronizacja odbywa się w interwałach czasowych równych częstotliwości odświeżania urządzenia wyświetlającego, nawet w przypadku renderingu pozaekranowego, zob. rys. 5.19. Zignorowanie czasu wymiany umożliwia skrócenie czasu wymaganego na synchronizację poleceń interfejsu OpenGL. Wspomniana synchronizacja wymagana jest m.in. przy współpracy interfejsu OpenGL z technologiami CUDA i OpenCL. W kolejnych podrozdziałach zaprezentowano wykorzystanie obu technologii w procesie renderingu modeli 3D postaci.

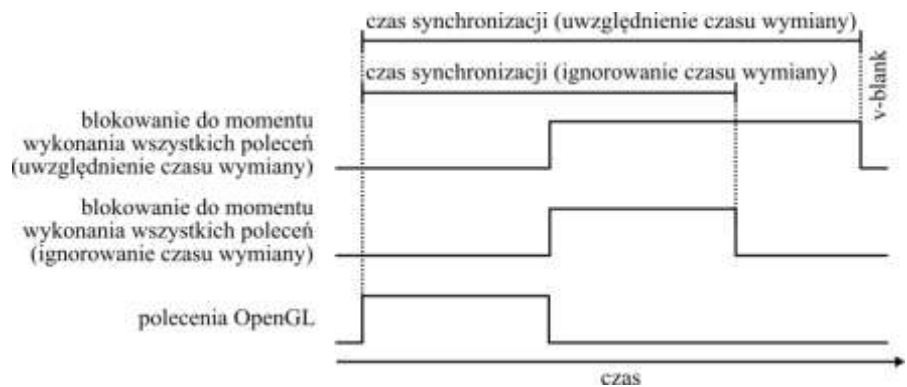

**Rys. 5.19. Wpływ czasu wymiany na synchronizację instrukcji OpenGL**

## 5.4. Rendering modelu 3D z wykorzystaniem CUDA-OpenGL

Interfejs programowania aplikacji OpenGL umożliwia wykorzystanie akceleracji sprzętowej do tworzenia fotorealistycznych obrazów generowanych przez narzędzia grafiki komputerowej. Sprzętowa akceleracja grafiki komputerowej realizowana jest zwykle przez układy graficzne. Dzięki narzędziom programowym takim jak Mesa [100] możliwe jest także wykorzystanie procesorów centralnych. Rendering modelu kształtu z wykorzystaniem technologii CUDA i OpenGL sprowadza się do połączenia metod sprzętowej akceleracji rysowania grafiki trójwymiarowej i możliwości obliczeniowych technologii CUDA do generowania obrazów modeli 3D. Rendering modelu wymaga wyznaczania macierzy transformacji globalnych dla wszystkich konfiguracji modelu, które przechowywane będą w buforze danych współdzielonych z programem cieniującym. W typowym podejściu do renderingu, za wyznaczanie macierzy transformacji odpowiada procesor centralny. Oznacza to, że macierze transformacji muszą być przesyłane z pamięci układu centralnego do pamięci układu graficznego. W technologii CUDA, operacja transmisji pamięci do układu graficznego nie jest konieczna, ponieważ dane przechowywane są w pamięci układu graficznego. OpenGL nie umożliwia wymiany informacji z innymi technologiami programistycznymi z wyjątkiem API DirectX. Z tego też powodu cały proces komunikacji realizowany jest po stronie CUDA. Przestrzeń adresowa pamięci OpenGL nie jest bezpośrednio dostępna w technologii CUDA. Dostęp do pamięci OpenGL odbywa się na zasadzie wzajemnego wykluczania dostępu, zob. rys. 5.20. Oznacza to, że w danej chwili z zasobu nie może korzystać jednocześnie CUDA i OpenGL. W celu uzyskania dostępu do buforów graficznych przez CUDA wymagane jest utworzenie zasobu na podstawie istniejącego obiektu OpenGL (zob. blok rejestracja zasobu CUDA na rys. 5.20). Utworzony zasób wykorzystywany jest do akwizycji aktualnej kopii pamięci bufora OpenGL, na którym realizowane będą operacje w CUDA. Kopiowanie pamięci realizowane jest w trakcie mapowania bufora OpenGL do zasobu CUDA (zob. blok mapowanie bufora OpenGL na rys. 5.20). Po zakończeniu pracy z obiektem zawartość pamięci synchronizowana jest z oryginalnym buforem OpenGL, przy czym operacja ta realizowana jest w trakcie odłączania współdzielonego zasobu (zob. blok odłączenie zasobu OpenGL na 5.20). Tworzenie kopii obiektu graficznego przez CUDA ma za zadanie optymalizację dostępu do pamięci bufora. Oznacza to, że w zależności od przeznaczenia zasobu, CUDA może utworzyć więcej niż jedną kopię bufora graficznego.

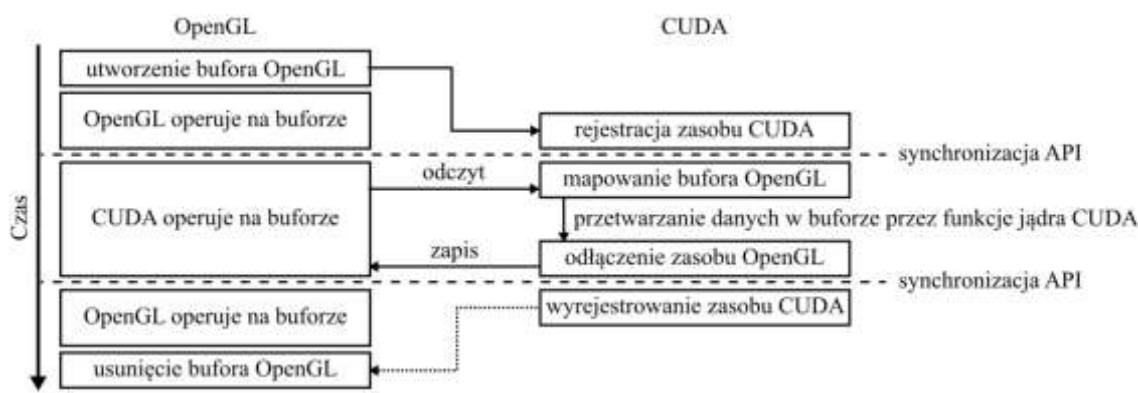

**Rys. 5.20. Współpraca pomiędzy OpenGL i CUDA**

W trakcie procesu renderingu, CUDA i OpenGL współdzielą bufor zawierający macierze transformacji globalnych oraz załączony do bufora ramki bufor koloru, który przechowuje wyrysowane obrazy modeli. Wyznaczanie wektora macierzy globalnych odbywa się zgodnie z metodą przedstawioną w podrozdziale 5.2. Natomiast bufor ramki używany jest w procesie wyznaczania komponentów składowych funkcji celu, które wykorzystywane są w trakcie wyznaczania wartości funkcji celu (zob. podrozdział 4.5). Niestety, gdy stosowany będzie więcej niż jeden układ graficzny, połączenie technologii CUDA i OpenGL nie umożliwi efektywnego wykorzystania wszystkich zasobów sprzętowych. Powodem tego stanu rzeczy jest sterownik układu graficznego Nvidia, który wysyła komendy OpenGL na wszystkie układy graficznie [40] lub korzysta z technologii SLI [15], w której klatki są sekwencyjnie renderowane na kolejnych układach graficznych. Przy wykorzystaniu kilku układów występuje konieczność transmisji zawartości buforów z układu, na którym odbywa się rendering, do układu, na którym utworzony został kontekst CUDA. Dzięki wykorzystaniu rozszerzenia NV_gpu_affinity [93] dostępnego w kartach z serii Nvidia Quadro, możliwe jest utworzenie kontekstu OpenGL na układzie, na którym operuje kontekst CUDA i tym samym wyeliminowanie transmisji zawartości pamięci pomiędzy układami graficznymi. Warto wspomnieć, że metoda renderingu sprzętowego CUDA przedstawiona w podrozdziale 5.2 nie wymaga transmisji danych pomiędzy układami graficznymi i dlatego możliwe jest wykorzystanie zasobów sprzętowych wielu układów graficznych.

## 5.5.   Rendering modelu 3D z wykorzystaniem OpenCL-OpenGL

Rendering z wykorzystaniem technologii OpenCL i OpenGL realizowany jest zgodnie ze schematem wzajemnego wykluczania dostępu, który omówiono w podrozdziale 5.4. Oznacza to, że podczas renderingu OpenGL współdzieli z OpenCL bufor zawierający macierze transformacji globalnych i załącznik kolorów bufora ramki, który wykorzystywany jest do wyznaczania wartości funkcji celu, zob. podrozdział 6.5. Wyznaczanie macierzy transformacji lokalnej i globalnej realizowane jest w sposób analogiczny do sposobu przedstawionego w podrozdziale 5.2.

Istotną przewagą technologii OpenCL nad technologią CUDA jest przenośność i skalowalność przygotowanej aplikacji na inne architektury sprzętowe [74]. Dzięki temu przygotowana aplikacja może być uruchamiana zarówno na układach graficznych, jak i centralnych jednostkach CPU. API OpenCL nie pozwala na dostęp do pamięci przez wskaźniki. Zarządzanie pamięcią w OpenCL odbywa się przez typ cl_mem, nazywany obiektem pamięci, który jest numerycznym identyfikatorem zaalokowanej pamięci. Dostęp do buforów graficznych w OpenCL odbywa się przez obiekt pamięci, który tworzony jest na podstawie istniejącego bufora OpenGL, zob. blok utworzenie bufora OpenCL na rys. 5.21. Obiekt pamięci wykorzystywany jest do przechowywania kopii pamięci bufora OpenGL, tworzonej podczas akwizycji bufora przez OpenCL, zob. blok przejęcie bufora OpenGL na rys. 5.21. Zawartość obiektu pamięci synchronizowana jest z buforem graficznym po zakończeniu operacji na obiekcie pamięci OpenCL, zob. blok zwolnienie bufora OpenGL na rys. 5.21. Mechanizm wymiany danych jest analogiczny do mechanizmu wymiany danych CUDA i OpenGL.

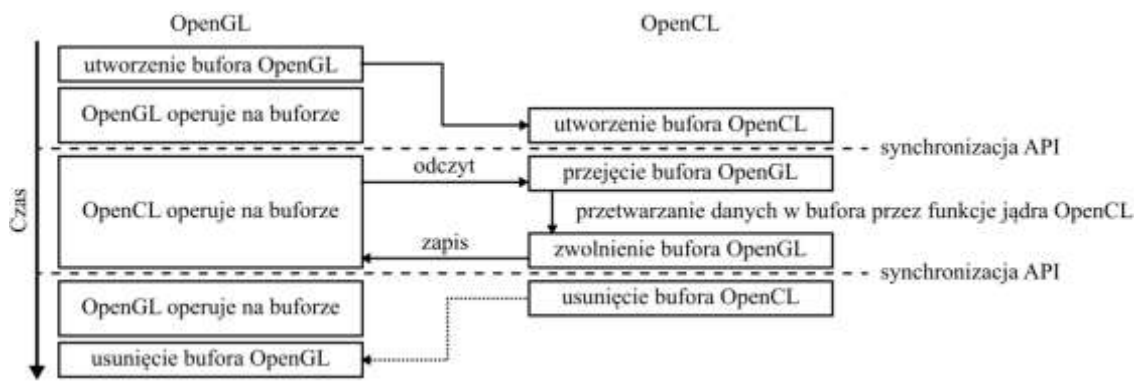

**Rys. 5.21. Współpraca pomiędzy OpenGL i OpenCL**

## 5.6. Wykorzystanie renderingu w równoległej funkcji celu

### Wyznaczanie wartości komponentów składowych funkcji celu

Przed wyznaczeniem wartości komponentów składowych funkcji celu realizowana jest operacja renderingu, która generuje obraz modelu. Z kolei sposób wyznaczania wartości komponentów składowych uzależniony jest od wykorzystywanej metody renderingu modelu 3D. W dalszej części podrozdziału przedstawione zostaną metody wyznaczania komponentów składowych funkcji celu dla renderingu sprzętowego i programowego.

### Wyznaczanie komponentów składowych funkcji celu w oparciu o rendering sprzętowy

Przy wykorzystaniu akceleracji sprzętowej OpenGL, piksele renderowanego obrazu kodowane są w buforze ramki w sposób przedstawiony w podrozdziale 5.3. Tabela 5.5 przedstawia sposób kodowania obrazu renderowanego w OpenGL oraz obrazów krawędzi i sylwetki. Jak można zauważyć, pojedynczy piksel każdego z obrazów kodowany jest na ośmiu bitach. Etykieta sylwetki modelu kodowana jest na pierwszych siedmiu bitach, natomiast informacja o wystąpieniu krawędzi kodowana jest na ostatnim bicie. Odległość od najbliższej krawędzi normalizowana jest do wartości całkowitej i zapisywana jest na ośmiu bitach.

**Tabela 5.5. Zestawienie kodowania wartości piksela obrazu modelu, sylwetki i krawędzi**

| Bit obrazu | 0 | 1 | 2 | 3 | 4 | 5 | 6 | 7 |
|---|---|---|---|---|---|---|---|---|
| **Obraz modelu** | etykieta modelu | | | | | | flaga krawędzi | |
| **Obraz sylwetki** | etykieta modelu | | | | | | flaga krawędzi | |
| **Obraz krawędzi** | znormalizowana odległość od najbliższej krawędzi | | | | | | | |

Wyznaczanie wartości komponentów składowych funkcji celu realizowane jest w zdefiniowanym obszarze zainteresowania obiektu. Obszar zainteresowania określany jest na podstawie czterech największych obiektów widocznych na obrazie sylwetki. Pole powierzchni obiektów, które wykorzystywane są do wyznaczania obszaru zainteresowania musi być większe niż przyjęta dobrana eksperymentalnie wartość graniczna. W opracowanym systemie wykorzystywane są dwa warianty funkcji celu, które określone są zależnością (4.46) i (4.47). Omawiane funkcje celu umożliwiają wyznaczenie komponentów składowych dla poszczególnych segmentów postaci oraz wyznaczenie komponentów składowych funkcji celu dla całej sylwetki. W przypadku, gdy wykorzystywana jest segmentacja postaci, wartości komponentów wyznaczane są oddzielnie dla każdej części ciała. Implementacja funkcji celu wykorzystującej segmentacje modelu w technologii CUDA nie jest rozwiązaniem efektywnym, ponieważ tablice wykorzystywane do przechowywania wartości komponentów składowych funkcji celu powinny być przechowywane w pamięci o krótkim czasie dostępu. Efektywniejsze implementacje funkcji celu w technologii CUDA wyznaczają komponenty składowe funkcji celu dla całej sylwetki postaci bez etykietowania. Tabela 5.6 przedstawia sposób kodowania obrazu sylwetki, obrazu krawędzi oraz obrazu modelu, których piksele reprezentowane są przez ośmiobitowe wartości całkowite. Piksel zakodowanego obrazu kandydata na pierwszych siedmiu bitach koduje kolor modelu kształtu, natomiast ostatni bit informuje, czy w danym pikselu wykryta został krawędź modelu kształtu. Ponieważ segmentacja osoby nie jest wykorzystywana, informacja o sylwetce jest przechowywana na ostatnim bicie zakodowanego obrazu modelu, natomiast pierwsze siedem bitów przechowuje zdyskretyzowaną wartość krawędzi obrazu referencyjnego.

**Tabela 5.6. Sposoby kodowania piksela w algorytmach niewykorzystujących segmentacji sylwetki**

| Bit | 0 | 1 | 2 | 3 | 4 | 5 | 6 | 7 |
|---|---|---|---|---|---|---|---|---|
| Obraz kandydata sylwetki i krawędzi | flaga modelu | | | | | | | flaga krawędzi |
| Obraz referencyjny | odległość od najbliższej krawędzi | | | | | | flaga sylwetki | |

Ponieważ piksele obrazu reprezentowane są przez wartości ośmiobitowe, możliwe jest upakowanie czterech pikseli w jednej wartości 32 bitowej, na której realizowane będą operacje bitowe. Oznacza to, że operacja wyznaczania komponentów składowych funkcji celu wykorzystuje zrównoleglenie bitowe w celu porównania czterech sąsiadujących pikseli obrazu modelu, obrazu sylwetki i krawędzi. Ponadto w implementacji CUDA (zob. podrozdział 5.4), zadanie wyznaczenia komponentów składowych jest dekomponowane na bloki wątków podzielonych na grupy, zob. rys. 5.22. Każda z grup złożona jest z pewnej liczby wątków, składających się na kilka grup *warp*. Grupa wątków współpracuje przez wykorzystanie pamięci współdzielonej i odpowiada za wyznaczenie komponentów funkcji celu dla czterech obrazów sylwetki i krawędzi. Liczba grup w bloku zależy od liczby rdzeni i wieloprocesorów dostępnych na wykorzystywanym układzie graficznym, jak również liczby obrazów modeli, dla których należy wyznaczyć komponenty składowe funkcji celu.

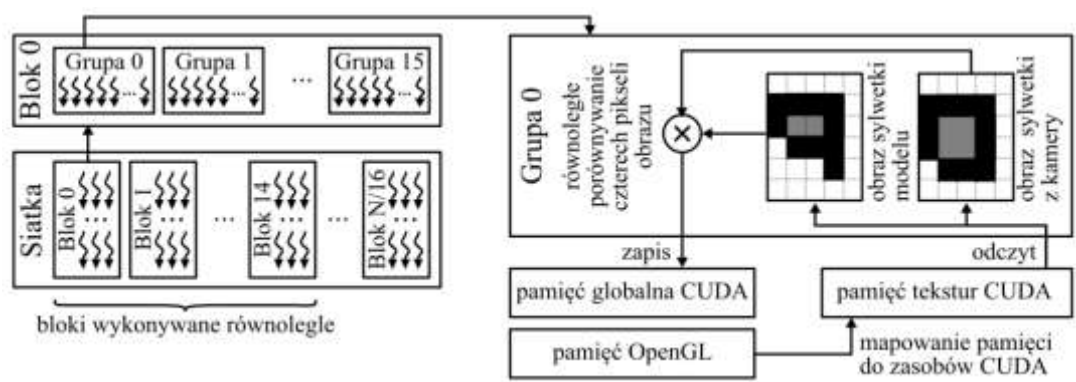

**Rys. 5.22. Wyznaczanie wartości funkcji celu w CUDA-OpenGL**

W opracowanej metodzie śledzenia ruchu 3D, obrazy sylwetki i krawędzi przechowywane są w szybkiej pamięci stałej, zob. podrozdział 3.7, natomiast obrazy z wyrenderowanymi postaciami w hipotetycznych pozach przechowywane są w teksturze zmapowanej z zasobem OpenGL, na którym realizowany był rendering, zob. podrozdział 5.5.

# Wyznaczanie komponentów składowych w oparciu o rendering programowy

Mechanizmy renderingu programowego, które opisano w podrozdziale 5.2, można wykorzystać do generowania obrazów kandydatów. Podejście takie umożliwia wyznaczenie komponentów funkcji celu w sposób analogiczny do metody wykorzystywanej przy renderingu sprzętowym. Podejście takie jest jednak nieefektywne z powodu dużej liczby odwołań do pamięci globalnej podczas renderowania modelu 3D. Efektywniejszym rozwiązaniem jest wyznaczanie wartości komponentów funkcji celu w trakcie procesu rasteryzacji programowej. Oznacza to, że obraz powstały w procesie rasteryzacji nie jest wykorzystywany w dalszym procesie wyznaczania wartości funkcji celu. Celem wyrysowania modelu kształtu każdy element modelu aproksymowany jest przez dwa trójkąty budujące trapez, którego wektor normalny skierowany jest w stronę kamery. Dla każdego trójkąta wyznaczany jest prostokątny obszar zainteresowania AABB, w którym realizowane będzie rysowanie, zob. rys. 5.23. Kolejność rysowania trójkątów zależy od ich odległości od kamery. W omawianym podejściu obiekty znajdujące się najbliżej kamery będą wyrysowane jako pierwsze.

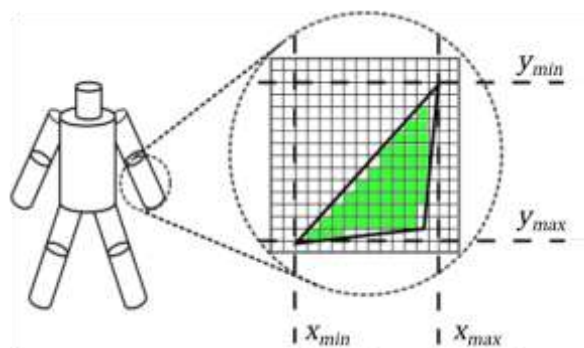

**Rys. 5.23. Wyznaczanie obszaru zainteresowania dla trójkąta**

W CUDA każdy obraz będzie wyrenderowany przez grupę wątków pracujących w obrębie tego samego bloku, zaś liczba bloków zależy od liczby modeli do wyrysowania oraz od liczby grup przypadających na wątek. Każdy wątek w bloku służy do zamalowania kolumny pikseli budujących trójkąt. W kolumnie zamalowane zostaną jedynie piksele, które nie były wcześniej zamalowane. Jeśli szerokość trójkąta jest większa niż liczba współpracujących wątków, cykl zamalowywania zostanie powtórzony, zob. rys 5.24. Ponieważ elementy wyrysowywane są począwszy od trójkątów znajdujących się najbliżej kamery, a skończywszy na elementach znajdujących się najdalej – oznacza to, że każdy piksel zostanie zamalowany tylko raz. Dzięki temu możliwe jest wyznaczenie wartości komponentów składowych funkcji celu.

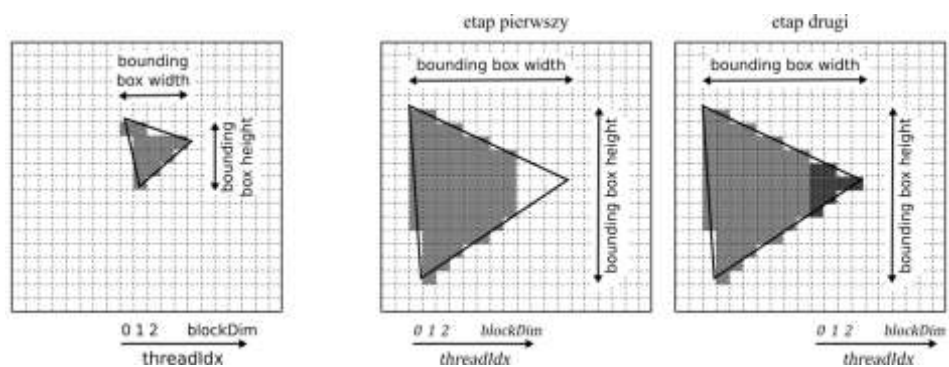

**Rys. 5.24. Wypełnianie trójkątów z wykorzystaniem wątków CUDA**

Podobnie jak miało to miejsce w przypadku wcześniej omówionych metod wyznaczania komponentów funkcji celu, obraz referencyjny przechowywany jest w pamięci stałej, natomiast obraz, na którym realizowany jest rendering programowy, przechowywany jest w pamięci globalnej układu graficznego. W odróżnieniu od wcześniej omówionych metod, dla przetwarzanego obrazu nie jest określany obszar zainteresowania. Oznacza to, że przetworzone zostaną wszystkie zamalowane piksele budujące obraz modelu kształtu.

## 5.7. Podsumowanie

W niniejszej części pracy opracowano efektywne metody zrównoleglenia funkcji celu. Opracowano i przebadano rendering programowy modelu 3D z wykorzystaniem CPU i CUDA. Zaproponowano strukturę programowalnych potoków graficznych na potrzeby renderingu sprzętowego w oparciu o OpenGL. Rozwiązania oparto o język GLSL, stworzono pakiet programów do cieniowania wierzchołków, pakiet programów cieniowania geometrii i pakiet programów cieniowania fragmentów. Dzięki wspomnianym rozwiązaniom uzyskano przenośność i skalowalność rozwiązań. Przygotowano pakiet programów do rzutowania modelu 3D do przestrzeni obrazu w oparciu o model kamery Tsai na potrzeby śledzenia ruchu 3D w czasie rzeczywistym. Opracowane rozwiązania optymalizowano pod kątem renderingu modelu płaskiego i modelu siatkowego. Poszukiwania dotyczyły efektywnego wykorzystania pamięci, a w szczególności jednoczesnego renderingu wielu modeli w sposób umożliwiający szybkie wyznaczenie składowych funkcji celu. Przygotowano mechanizmy renderingu modelu 3D z wykorzystaniem CUDA-OpenGL, a w szczególności mechanizmy umożliwiające efektywną współpracę obydwu technologii. Przygotowano także mechanizmy renderingu modelu 3D z wykorzystaniem OpenCL-OpenGL. Opracowano metody umożliwiające wykorzystanie renderingu modelu 3D w równoległych funkcjach celu.

# Rozdział 6

# Badania eksperymentalne

Na rys. 6.1 zilustrowano ideę metody śledzenia ruchu w systemie wielokamerowym z wykorzystaniem modelu 3D. W oparciu o wydzielone obrazy sylwetki postaci ludzkiej, zob. dolna część rysunku, a także $N$ wyrenderowanych obrazów dla każdej z kamer, zob. górna część rysunku, następuje wyznaczenie funkcji dopasowania, zob. środkowa część rysunku. Główna część nakładów obliczeniowych wiąże się z wyrenderowaniem sylwetek w $N$ hipotetycznych pozach. W niniejszej pracy, rendering postaci w zadanej pozie, zob. blok rendering na rys. 6.1, realizowany był programowo i sprzętowo. W niniejszym rozdziale przedstawiono wyniki badań eksperymentalnych dla opracowanych algorytmów śledzenia ruchu postaci ludzkiej, a w szczególności wyniki badań dotyczące renderingu obrazów, wyznaczania funkcji dopasowania, a także śledzenia ruchu całej postaci ludzkiej w czasie rzeczywistym. Omówiono także mechanizmy i funkcjonalność zaproponowanego systemu do śledzenia ruchu postaci ludzkiej.

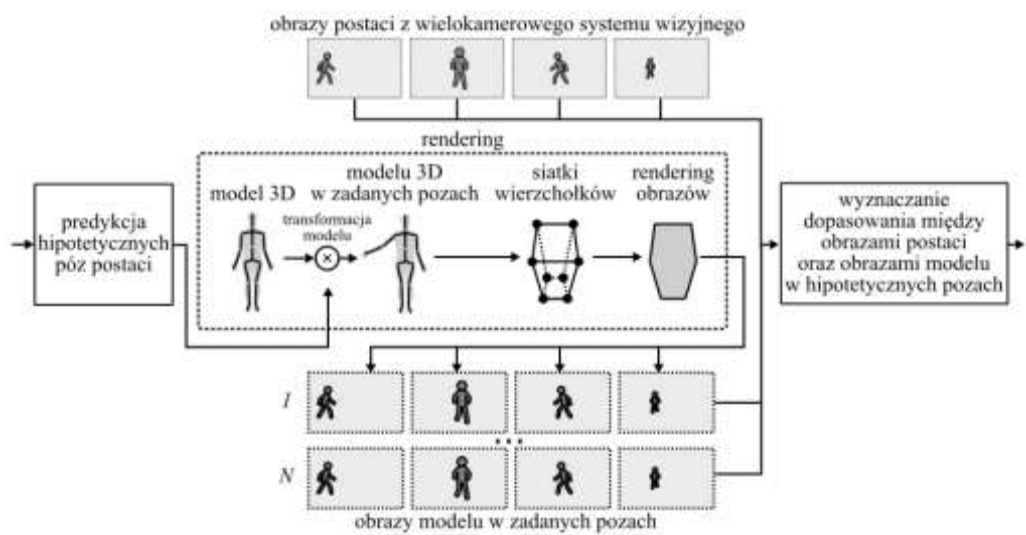

**Rys. 6.1. Śledzenie ruchu 3D w systemie wielokamerowym w oparciu o model 3D**

Rozdział składa się z dziewięciu podrozdziałów. Podrozdział pierwszy przedstawia zagadnienia związane z określaniem dokładności śledzenia ruchu 3D z wykorzystaniem danych odniesienia mocap. W podrozdziale drugim zaproponowano sposób dekompozycji równoległego algorytmu do śledzenia ruchu 3D na GPU. Podrozdział trzeci zawiera omówienie zaproponowanych rozwiązań do śledzenia ruchu 3D z wykorzystaniem CUDA. W podrozdziale czwartym omówiono uzyskane wyniki badań eksperymentalnych dla algorytmu śledzącego zaimplementowanego w CUDA. W podrozdziale piątym omówiono zaproponowane rozwiązania dla śledzenia ruchu 3D w CUDA i OpenGL.

Podrozdział szósty prezentuje wyniki badań eksperymentalnych dla wspomnianego wcześniej rozwiązania opartego na CUDA i OpenGL. W podrozdziale siódmym dokonano ewaluacji zaproponowanych metod na potrzeby śledzenia ruchu 3D w czasie rzeczywistym. W przedostatnim podrozdziale omówiono rozwiązania dla śledzenia ruchu w czasie rzeczywistym, a także zaprezentowano uzyskane wyniki. Rozdział zamyka podsumowanie.

## 6.1. Badanie dokładności śledzenia

Weryfikacja poprawności śledzenia realizowana była poprzez wykorzystanie danych odniesienia z systemu mocap. Dokładność śledzenia określano w oparciu o dane przechowywane w formacie C3D [14]. Wspomniane pliki C3D zawierają pozycję markerów umieszczonych na aktorze, które zarejestrowane zostały przez system mocap. Pliki C3D przechowują także informacje o konfiguracji aktora, budowie szkieletu i konfiguracji markerów.

W literaturze związanej ze śledzeniem ruchu postaci ludzkiej wykorzystuje się kilka miar dokładności śledzenia ruchu 3D postaci. Pierwszą z nich jest miara błędu konfiguracji, która oparta jest o różnicę póz (kątów) i pozycji, w jakich znajduje się model i postać [84,111]. Drugą z nich jest miara błędu, która wykorzystuje różnicę w pozycji 3D markerów systemu mocap i wirtualnych markerów na modelu postaci ludzkiej [82,140]. W niniejszej pracy określenie błędów realizowano w oparciu o drugą z wymienionych metod.

Jak już wspomniano, w niniejszej pracy dokładność śledzenia określana jest poprzez porównanie pozycji fizycznych markerów śledzonych przez system mocap z pozycją wirtualnych markerów. Pozycja wirtualnych markerów wyznaczana jest na podstawie estymat pozy z systemu śledzącego. Błędem $m$-tego markera określana jest odległość euklidesowa pomiędzy pozycją wirtualnego markera $p_m(\hat{x})$ i pozycją fizycznego markera $p_m(x)$:

$$\bar{e}_m = \|p_m(x) - p_m(\hat{x})\|,\qquad(6.1)$$

gdzie rzeczywisty stan modelu określany jest przez $x$, zaś estymata pozy określana jest przez $\hat{x}$. Rzeczywisty stan modelu może być określany przez system mocap, jednak wymaga to zastosowania identycznego modelu w systemie mocap i w systemie śledzącym. W przypadku, gdy plik mocap nie przechowuje stanu modelu lub stan przechowywanego modelu nie odpowiada modelowi wykorzystanemu do śledzenia, wykorzystywane są pozycje markerów przechowywane w formacie C3D. W sytuacji, w której częstotliwość pracy systemu mocap różni się od częstotliwości pracy systemu wizyjnego, wykorzystuje się uśrednienie stanu modelu dla danej częstotliwości bądź wykorzystuje się najbliższy stan modelu systemu mocap. Błąd śledzenia markera $m$ dla sekwencji obrazów złożonej z $K$ klatek wyznacza się na podstawie zależności:

$$\bar{E}_m = \frac{1}{K}\sum_{i=1}^{K} \bar{e}_m = \frac{1}{K}\sum_{i=1}^{K} \left\| p_m^{(i)}(x) - p_m^{(i)}(\hat{x}) \right\| . \tag{6.2}$$

Średni błąd dla wszystkich markerów w całej sekwencji wyznacza się na podstawie zależności:

$$\bar{E} = \frac{1}{M}\sum_{i=1}^{M} \bar{E}_i , \tag{6.3}$$

gdzie $M$ jest równe liczbie markerów.

# Wybór funkcji celu

Celem przebadania dokładności śledzenia dla funkcji celu zaproponowanych w podrozdziale 4.6, zrealizowano badania eksperymentalne mające na celu porównanie błędów śledzenia. W tabeli 6.1 zamieszczono średnie błędy śledzenia ruchu postaci ludzkiej dla przebadanych empirycznie funkcji celu (4.46)–(4.50). Śledzenie ruchu postaci ludzkiej realizowano z wykorzystaniem algorytmu PSO, w którym zmieniano liczbę cząsteczek oraz liczbę iteracji. Prezentowane wyniki dotyczą średnich wartości błędu dla całych sekwencji, które z kolei uśredniono dla 10-krotnego powtórzenia eksperymentu.

**Tabela 6.1. Średni błąd śledzenia w oparciu o zaproponowane funkcje celu**

| Sekwen-cja | Konfiguracja PSO | | Średni błąd śledzenia [mm] w zależności od postaci funkcji celu | | | | |
|---|---|---|---|---|---|---|---|
| | Liczba cząsteczek | Liczba iteracji | WS 4 (4.46) | SP 8 (4.47) | AoSP 4 (4.49) | PoPS 7 (4.50) | AoWS 4 (4.48) |
| **P1S** | 96 | 10 | 72.7±44.5 | 74.7±53.5 | 65.4±40.9 | 72.9±46.3 | 74.7±47.4 |
| | | 15 | 62.3±35.8 | 60.6±34.4 | 66.1±43.7 | 65.6±40.3 | 71.1±39.4 |
| | | 20 | 67.3±41.3 | 56.7±32.6 | 60.4±40.2 | 60.8±37.5 | 66.2±37.8 |
| | 304 | 10 | 60.2±39.9 | 65.3±39.0 | 53.5±29.9 | 56.2±32.9 | 59.1±32.8 |
| | | 15 | 55.8±32.1 | 54.8±30.1 | 55.0±31.2 | 55.7±33.3 | 60.5±33.2 |
| | | 20 | 55.7±33.8 | 56.7±34.6 | 51.0±28.7 | 49.9±29.3 | 53.0±29.9 |
| | 496 | 10 | 57.2±30.9 | 76.6±49.9 | 57.8±35.9 | 56.0±38.1 | 56.3±38.8 |
| | | 15 | 56.0±31.6 | 57.9±32.3 | 53.8±32.8 | 55.6±31.4 | 53.4±31.1 |
| | | 20 | 53.8±31.5 | 57.2±30.2 | 51.8±30.5 | 55.0±30.3 | 51.6±30.3 |
| **P1D** | 96 | 10 | 57.5±35.0 | 67.1±48.4 | 67.9±54.2 | 68.2±54.4 | 74.9±53.8 |
| | | 15 | 57.1±39.1 | 62.9±46.4 | 64.4±49.9 | 55.8±36.5 | 57.5±36.2 |
| | | 20 | 58.0±33.7 | 49.6±25.0 | 58.1±42.9 | 54.8±36.5 | 55.7±35.9 |
| | 304 | 10 | 50.3±24.7 | 49.7±25.9 | 53.6±33.0 | 49.7±27.1 | 52.8±27.6 |
| | | 15 | 48.8±24.1 | 47.9±25.6 | 48.5±28.6 | 49.8±30.8 | 52.1±30.4 |
| | | 20 | 48.1±25.3 | 60.9±40.5 | 54.2±37.8 | 58.0±40.4 | 50.9±39.8 |
| | 496 | 10 | 55.8±33.9 | 56.3±32.4 | 56.4±37.9 | 51.8±31.9 | 56.9±32.7 |
| | | 15 | 47.9±23.9 | 46.7±23.2 | 49.9±34.3 | 49.1±28.9 | 52.0±28.7 |
| | | 20 | 46.3±23.7 | 48.4±25.3 | 48.0±30.1 | 51.8±35.1 | 51.5±35.3 |

**Tabela 6.1 (kontynuacja). Średni błąd śledzenia w oparciu o zaproponowane funkcje celu**

| Sekwencja | Konfiguracja PSO | | Średni błąd śledzenia [mm] w zależności od postaci funkcji celu | | | | |
|---|---|---|---|---|---|---|---|
| | Liczba cząsteczek | Liczba iteracji | WS 4 (4.46) | SP 8 (4.47) | AoSP 4 (4.49) | PoPS 7 (4.50) | AoWS 4 (4.48) |
| LeeWalk | 96 | 10 | 49.6±17.3 | 49.8±15.5 | 49.1±14.3 | 49.1±15.1 | 53.5±15.3 |
| | | 15 | 50.0±15.0 | 48.3±14.0 | 46.2±13.5 | 48.9±15.5 | 49.7±15.2 |
| | | 20 | 49.2±15.6 | 45.7±13.1 | 46.1±13.3 | 44.3±13.3 | 48.0±13.1 |
| | 304 | 10 | 44.0±12.8 | 46.1±12.9 | 46.2±14.0 | 47.2±14.8 | 51.0±15.1 |
| | | 15 | 44.3±14.2 | 45.4±13.0 | 45.3±14.1 | 43.7±13.9 | 44.2±13.8 |
| | | 20 | 42.6±12.7 | 41.6±11.5 | 43.8±14.0 | 43.9±13.5 | 41.2±13.7 |
| | 496 | 10 | 46.3±13.7 | 43.6±12.5 | 47.3±14.8 | 42.9±11.8 | 44.8±11.8 |
| | | 15 | 44.3±12.1 | 41.6±11.5 | 45.2±13.8 | 45.7±14.8 | 43.0±14.9 |
| | | 20 | 48.1±19.4 | 40.5±11.3 | 42.7±13.0 | 42.9±13.9 | 40.1±13.7 |

W tabeli 6.1 po nazwie wykorzystywanej funkcji znajdują się liczby ilustrujące, w ilu przypadkach dany algorytm okazał się najlepszy. Jak można zauważyć, w przekroju rozpatrywanych sekwencji funkcje celu SP i PoSP umożliwiają uzyskanie najlepszych wyników. Funkcja celu SP określona zależnością (4.47) jest nieznacznie lepsza i dlatego wykorzystywana ona będzie w dalszej części rozdziału. Warto podkreślić, że postać funkcji celu nie ma znaczącego wpływu na czasy przetwarzania i efektywność obliczeń równoległych.

W trakcie badań dokładność śledzenia była także szacowana wizualnie w oparciu o analizę dopasowania zrzutowanego modelu 3D na sylwetki śledzonej postaci. Na rys. 717.2 zilustrowano w sposób poglądowy wyniki śledzenia ruchu na sekwencji LeeWalk dla PSO składającego się z 300 cząsteczek i wykonującego 10 iteracji. Model w najlepszej pozie rzutowany jest do przestrzeni 2D i nakładany na obrazy sylwetki danej kamery. W kolejnych wierszach zamieszczono obrazy z poszczególnych kamer dla co piętnastej klatki sekwencji LeeWalk. Jak można zauważyć, algorytm PSO dość dobrze śledzi postać mimo znaczących zmian sylwetki postaci.

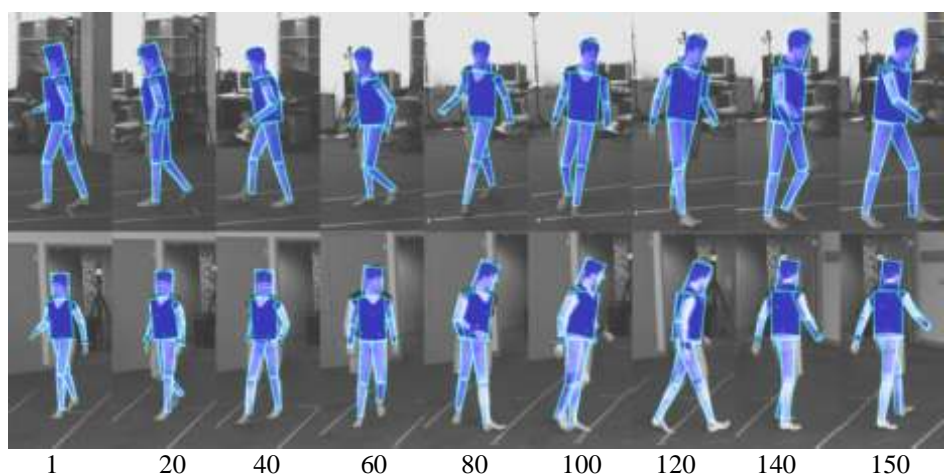

**Rys. 6.2. Śledzenie ruchu 3D postaci w sekwencji LeeWalk**

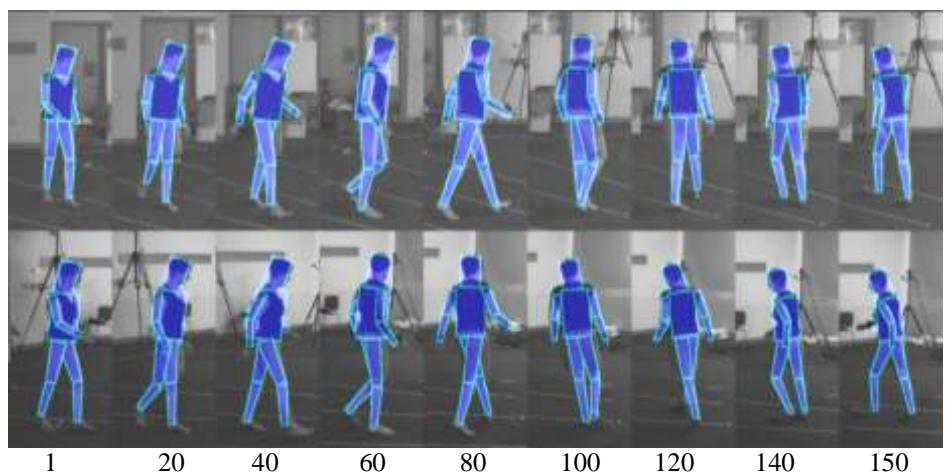

| 1 | 20 | 40 | 60 | 80 | 100 | 120 | 140 | 150 |

**Rys. 6.2 (kontynuacja). Śledzenie ruchu 3D postaci w sekwencji LeeWalk**

## 6.2. Równoległy algorytm śledzenia ruchu postaci ludzkiej

### Dekompozycja algorytmu optymalizacji w oparciu o rój cząsteczek

Na potrzeby śledzenia ruchu 3D postaci ludzkiej z wykorzystaniem GPU dokonano dekompozycji algorytmu PSO. W trakcie śledzenia ruchu postaci ludzkiej w oparciu o algorytm PSO wyróżnić można 5 faz, mianowicie: fazę inicjalizacji cząsteczek, fazę zmiany pozy modelu, fazę renderingu modelu 3D i wyznaczania komponentów składowych funkcji celu, fazę ewaluacji wartości funkcji celu oraz fazę, w której odbywa się ruch cząsteczek. Na rys. 6.3 zamieszczono schemat blokowy ilustrujący sposób dekompozycji algorytmu na potrzeby obliczeń równoległych na GPU. Faza inicjalizacji PSO realizowana jest w oparciu o bloki generowanie liczb losowych i inicjalizacja cząsteczek, faza ruchu modelu realizowana jest przez blok ruch modelu, faza renderingu modelu i wyznaczania komponentów składowych funkcji celu realizowana jest przez blok rendering, faza ewaluacji wartości funkcji celu realizowana jest przez blok funkcja celu, zaś faza ruchu cząsteczek realizowana jest przez blok aktualizacja cząsteczek. Na wspomnianym rysunku zamieszczono także blok decyzyjny, w którym podejmowana jest decyzja o zatrzymaniu algorytmu. W dalszej części podrozdziału omówiono bloki initialization i RNG, zaś pozostałe bloki omówiono w następnych podrozdziałach. W niniejszej pracy rendering w zadanej pozie realizowany był z wykorzystaniem CPU, CUDA oraz CUDA-OpenGL (OpenCL-OpenGL). W zależności od wykorzystywanej implementacji algorytmu śledzącego, rendering realizowany jest programowo (CPU, CUDA) bądź sprzętowo (CUDA-OpenGL, OpenCL-OpenGL). W niniejszej pracy uwagę skupiono na równoległych algorytmach śledzenia ruchu postaci ludzkiej i dlatego implementacja CPU nie jest bliżej opisywana. Omawiany algorytm został szerzej opisany w pracach [140,142]. W algorytmach CUDA, CUDA-OpenGL, OpenCL-OpenGL funkcje składające się na bloki zilustrowane na rys. 6.3 realizowane są na GPU.

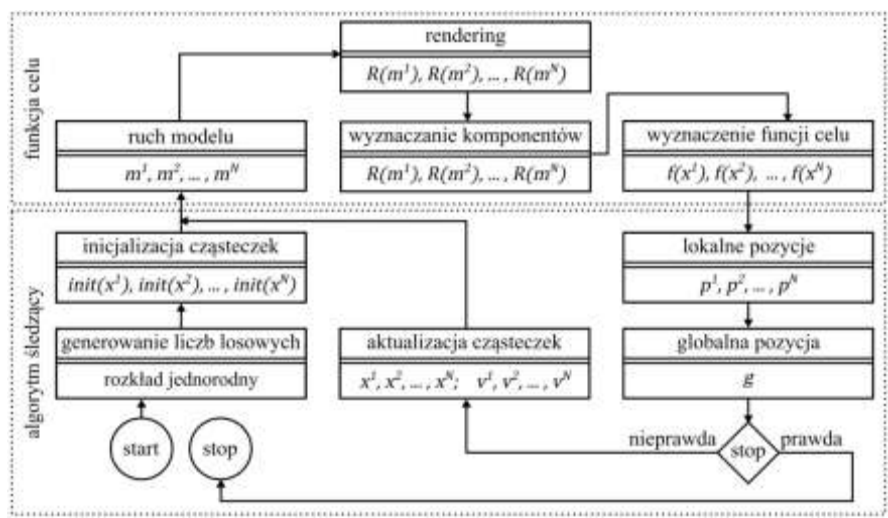

**Rys. 6.3. Dekompozycja algorytmu PSO do obliczeń na GPU dla jednej klatki obrazów**

Jak już wspomniano, śledzenie ruchu w oparciu o optymalizację wartości funkcji celu algorytmu PSO wymaga wygenerowania liczb losowych, w szczególności liczb losowych o rozkładzie normalnym oraz liczb losowych o rozkładzie jednorodnym, zob. rys. 6.3. W opracowanym algorytmie liczby losowe o rozkładzie normalnym generowane są na podstawie liczb losowych o rozkładzie jednorodnym. Generowanie liczb o rozkładzie normalnym odbywa się w oparciu o algorytm Box Mueller [17]. Do wygenerowania pary liczb losowych z rozkładem normalnym wymagane są dwie liczby losowe o rozkładzie jednorodnym. Liczby o rozkładzie jednorodnym generowane są na początku algorytmu w oparciu o algorytm Mersenne Twister [97]. Algorytm ten dostępny jest w bibliotece cuRand dołączonej do CUDA SDK [30]. Celem optymalnego wykorzystania zasobów sprzętowych algorytm generuje całą pulę liczb losowych o rozkładzie jednorodnym dla inicjalizacji wszystkich cząsteczek oraz wszystkich iteracji algorytmu dla pojedynczej klatki obrazów. Mając na względzie to, że układ posiada znaczącą liczbę jednostek arytmetycznych, generowanie wszystkich liczb losowych w jednym kroku algorytmu umożliwia lepsze wykorzystanie wspomnianych jednostek arytmetycznych GPU. W pracy [168] zamieszczono przegląd implementacji algorytmów wykorzystujących inteligencję roju w zadaniach optymalizacji na GPU, zaś w pracy [142] omówiono równoległe algorytmy śledzenia ruchu w oparciu o rój cząsteczek.

## Dekompozycja wyznaczania macierzy transformacji globalnych

W implementacjach CUDA, CUDA-OpenGL oraz OpenCL-OpenGL blok ruch modelu realizowany jest na GPU. Na rys. 6.4 zaprezentowano schemat blokowy algorytmu wyznaczającego macierze transformacji modelu dla algorytmów CUDA i CUDA-OpenGL. Dla algorytmu OpenCL-OpenGL wyznaczanie macierzy transformacji realizowane jest w sposób analogiczny. Wyznaczanie macierzy transformacji odbywa się w oparciu o metody zaprezentowane w podrozdziale 4.4. Macierze transformacji w implementacji

GPU wyznaczane są jednocześnie dla wszystkich cząsteczek. Obliczenia dla zadanej hipotezy realizowane są w pojedynczym wątku, zob. blok wątek na rys. 6.4. Każdy z wątków wczytuje wpierw strukturę modelu z pamięci globalnej CUDA oraz wektor stanu modelu 3D, zob. tabela 4.1. W oparciu o powyższe dane budowane są macierze transformacji lokalnej, zob. blok wyznaczanie macierzy transformacji lokalnej na rys. 6.4, a następnie tworzone są macierze transformacji globalnej modelu, zob. blok wyznaczanie macierzy transformacji globalnej na omawianym rysunku. Macierze transformacji świata zapisywane są do pamięci globalnej GPU. Wyznaczanie macierzy lokalnych realizowane jest w oparciu o zależności (4.11, 4.12, 4.18) z uwzględnieniem struktury modelu. W zależności od wykorzystywanej struktury modelu szkieletowego (liczby stopni swobody danej kości) macierz transformacji lokalnej kości modelu przyjmuje jedną z postaci przedstawionych w tabeli 4.4. Natomiast wyznaczanie macierzy transformacji globalnej realizowane jest w oparciu o zależności (4.2, 4.21). Macierze transformacji globalnej kości dla przyjętej struktury modelu określane są na podstawie tabeli 4.5.

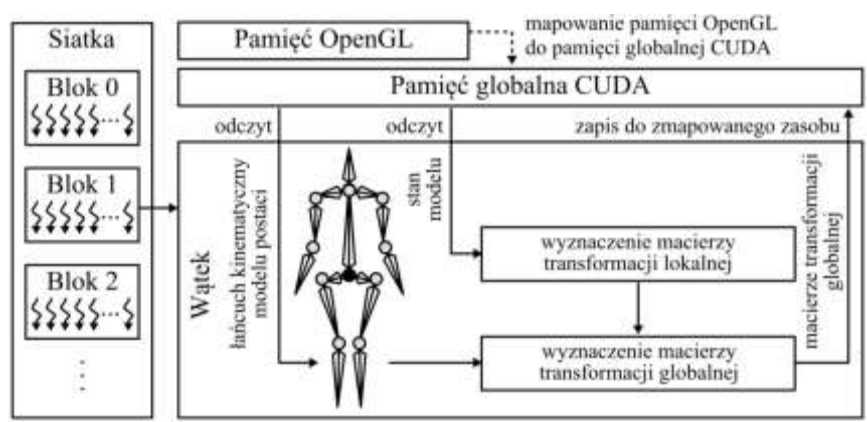

**Rys. 6.4. Wyznaczanie macierzy transformacji modelu 3D na GPU**

Liczba uruchamianych wątków jest równa liczbie cząsteczek algorytmu PSO lub PF. Jak już wspomniano w podrozdziale 3.7, w technologii CUDA wątki grupowane są w bloki, natomiast w API OpenCL dzielone są na grupy robocze.

Wykorzystywany układ graficzny GTX 780Ti posiada 12 wieloprocesorów strumieniujących, z których każdy może uruchomić $T_{max}$ wątków na 240 rdzeniach. Liczba wątków uruchamianych w bloku określana jest na podstawie zależności:

$$X_w = \begin{cases} N & \text{jeśli } N < T_{max} \\ 240 & \text{jeśli } N < 12 \cdot 240 \\ \frac{N}{12} & \text{jeśli } \frac{N}{12} < T_{max} \\ T_{max} & \text{w przeciwnym razie} \end{cases} , \qquad (6.4)$$

gdzie $N$ oznacza liczbę cząsteczek. Omawiana zależność umożliwia optymalne wykorzystanie rdzeni dostępnych w układzie GPU. Liczba bloków uruchamianych na układzie GPU określana jest na podstawie zależności:

$$X_b = \left\lceil \frac{N}{X_w} \right\rceil,$$ (6.5)

Powyższe zależności dotyczą jednowymiarowej siatki wątków i bloków CUDA, dla której $T_{max} = 1024$.

W OpenCL rozmiar grupy roboczej $W_x$ jest równy $X_w$, gdzie $T_{max} = 512$. Natomiast liczba uruchamianych wątków jest określana na podstawie zależności:

$$T_x = W_x \left\lceil \frac{N}{W_x} \right\rceil.$$ (6.6)

# Dekompozycja zadania do wyznaczania pozycji i rzutowania wierzchołków modelu 3D

Macierze transformacji globalnych modelu 3D wykorzystywane są do wyznaczania położenia wierzchołków modelu w przestrzeni 3D. Wyznaczanie pozycji wierzchołków realizowane jest w oparciu o zależności (4.24) i (4.25), natomiast rzutowanie wierzchołków do przestrzeni obrazu realizowane jest w oparciu o model kamery Tsai (zob. podrozdział 2.2). Wyznaczanie pozycji i rzutowanie wierzchołków zilustrowano na rys. 6.5. Wspomniane operacje realizowane są w oparciu o macierze transformacji wczytywane z pamięci globalnej i parametry modelu kształtu, które przechowywane są w pamięci stałej. Każdy z uruchomionych wątków wyznacza pozycję wierzchołków dla pojedynczej części modelu i zapisuje je w pamięci globalnej. Dzięki możliwości zapisu całych bloków pamięci jednocześnie możliwe jest ukrycie czasu dostępu do pamięci globalnej.

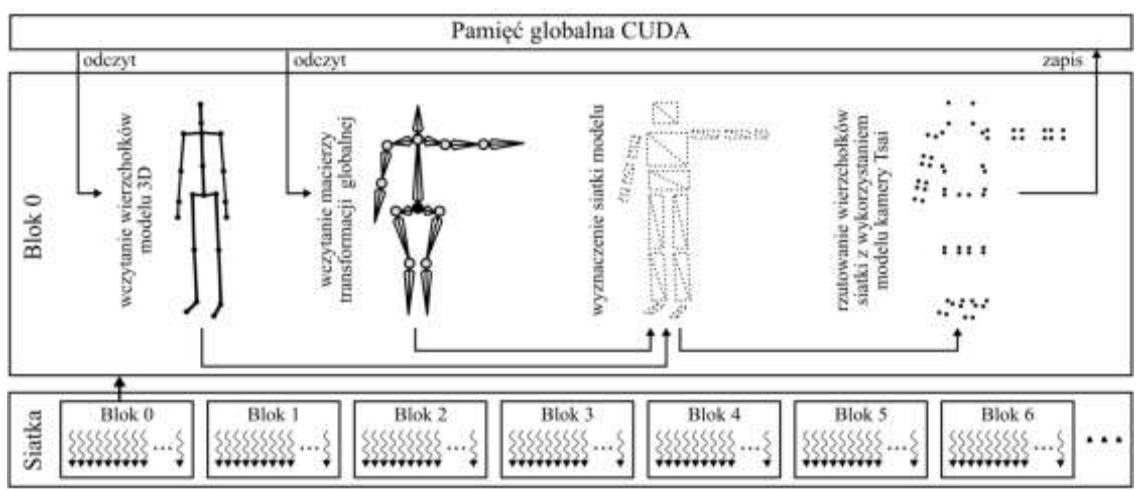

**Rys. 6.5. Wyznaczanie i rzutowanie wierzchołków modelu 3D**

W implementacji CUDA wykorzystywany jest płaski model kształtu, który omówiono w podrozdziale 4.3. Dla przyjętej struktury modelu, która składa się z 16 trapezów modelujących główne części ciała, zob. tabela 4.6, wymagane jest wyznaczenie czterech wierzchołków. Dla każdej części ciała w przyjętej strukturze modelu, zob. tabela 4.6, wymagane jest wyznaczenie pozycji czterech wierzchołków. Mając na względzie to, że model całej postaci ludzkiej składa się z 16 części ciała, wymagane jest wyznaczenie pozycji 64 wierzchołków. Wyznaczanie pozycji wierzchołków dla każdej części ciała modelu realizowane jest przez jeden wątek. Oznacza to, że wyznaczenie pozycji wierzchołków dla całego modelu wymaga uruchomienia 16 wątków. Omawiane zadanie jest odwzorowane na dwuwymiarowy blok wątków CUDA i jednowymiarową siatkę bloków. Liczba wątków w bloku CUDA określana jest na podstawie zależności:

$$X_w = 16 \qquad ,$$

$$Y_w = \begin{cases} \left\lceil \dfrac{N}{X_w} \right\rceil & \text{jeśli } N < \left\lceil \dfrac{T_{max}}{X_w} \right\rceil \\[2ex] \dfrac{240}{X_w} & \text{jeśli } N < \left\lceil \dfrac{12 \cdot 240}{X_w} \right\rceil \\[2ex] \dfrac{N}{12 X_w} & \text{jeśli } \dfrac{N}{12} < \dfrac{T_{max}}{X_w} \\[2ex] T_{max} & \text{jeśli } \dfrac{N}{12} \geq \dfrac{T_{max}}{X_w} \end{cases} \qquad , \tag{6.7}$$

gdzie $T_{max} = 1024$, $N$ określa liczbę cząsteczek, natomiast stałe 12 i 240 określają kolejno liczbę wieloprocesorów strumieniujących i liczbę rdzeni dla układu GTX 780Ti. Liczba bloków określana jest na podstawie zależności:

$$X_b = \left\lceil \dfrac{N}{Y_w} \right\rceil . \tag{6.8}$$

## Dekompozycja wyznaczania komponentów składowych funkcji celu

Jak już wspomniano w podrozdziale 4.5, funkcja celu składa się z pięciu komponentów które określane są na podstawie równań (4.36–4.40) Wyznaczanie komponentów składowych funkcji celu realizowane jest w oparciu o wyrenderowane obrazy modelu oraz sylwetki postaci wydzielonej na obrazach pobranych z kamery. Rendering obrazu modelu realizowany jest programowo lub sprzętowo w zależności od wykorzystywanego algorytmu, tj. CPU, CUDA, CUDA-OpenGL, OpenCL-OpenGL.

Rendering programowy przedstawiony został w podrozdziale 5.2, natomiast rendering sprzętowy zaprezentowano w podrozdziale 4.4. W dalszej części niniejszego rozdziału przedstawiono szczegóły związane z realizacją renderingu programowego bądź sprzętowego na wykorzystywanym GPU.

Wyznaczanie poszczególnych komponentów składowych funkcji celu realizowane jest równolegle. W przyjętym rozwiązaniu obraz dzielony jest na bloki obrazu. Każdy blok obrazu przetwarzany jest przez pojedynczy wątek, który wyznacza komponenty składowe funkcji celu dla przetwarzanego bloku obrazu. W celu wyznaczenia wartości komponentów składowych funkcji celu, wątki operujące na blokach danego obrazu wykorzystują metodę redukcji równoległej [30,112], która realizowana jest po zakończeniu przetwarzania wszystkich pikseli w bloku obrazu, zob. rys. 6.6. Dzięki zastosowaniu wspomnianej redukcji równoległej możliwe jest równoległe zsumowanie komponentów funkcji celu dla całego obrazu. Jak można zauważyć w kolejnych iteracjach, liczba aktywnych wątków zmniejszana jest o połowę aż do zsumowania wszystkich wartości składowych.

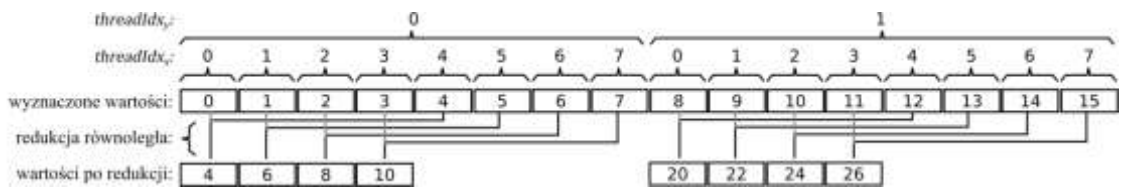

**Rys. 6.6**. Metoda redukcji równoległej

## Dekompozycja wyznaczania wartości funkcji celu

Wartości komponentów składowych funkcji celu wykorzystywane są do wyznaczenia wartości funkcji celu, zob. podrozdział 4.6. Dekompozycja wyznaczania wartości funkcji celu realizowana jest w analogiczny sposób do dekompozycji wyznaczania macierzy transformacji globalnych. W przyjętym rozwiązaniu każdy wątek wyznacza wartość funkcji celu dla pojedynczej cząsteczki i zapisuje wynik w pamięci globalnej. Zadanie realizowane jest przy wykorzystaniu jednowymiarowego bloku i jednowymiarowej siatki, których wymiary określone są przez równania (6.4) i (6.5).

## 6.3. Śledzenie ruchu 3D z wykorzystaniem CUDA

Celem uniknięcia zbędnych operacji na pamięci globalnej, rendering i ruch modelu z rys. 6.3 połączono w jeden wspólny blok. Na rys. 6.7 zamieszczono schemat operacji realizowanych przez wspomniany blok. Wyznaczanie wartości funkcji celu dla każdej cząsteczki realizowane jest przez współpracujące ze sobą wątki. Liczba uruchomionych wątków jest proporcjonalna do liczby cząsteczek, natomiast liczba współpracujących wątków, które są zaangażowane w wyznaczanie wartości funkcji celu dla jednej cząsteczki zależy od przetwarzanej sekwencji i może zostać dobrana eksperymentalnie.

W przyjętym rozwiązaniu wątki wczytują kolejno kości danego modelu kształtu postaci i odpowiadające im macierze transformacji globalnej. W następnej kolejności wyznaczane są pozycje wierzchołków danej kości modelu i ich rzuty na płaszczyznę obrazu. Omawiana operacja realizowana jest analogicznie do operacji omówionej w podrozdziale 6.2, zob. także rys. 6.5. W oparciu o wyznaczone tym sposobem pozycje zrzutowanych wierzchołków realizowana jest operacja renderingu programowego modelu 3D, zob. rys. 5.24. W trakcie renderingu renderowane są poszczególne trójkąty oraz uaktualniane są wartości komponentów składowych funkcji celu. Renderowany obraz przechowywany jest w pamięci globalnej, zaś obraz sylwetki wykorzystywany przy wyznaczaniu komponentów składowych przechowywany jest w pamięci tekstury, zob. rys. 6.7. Dzięki przechowywaniu obrazu sylwetki w pamięci tekstury uzyskuje się szybszy dostęp do danych i tym samym skrócenie czasu wyznaczania wartości funkcji celu w porównaniu do rozwiązania, w którym dane przechowywane byłyby w pamięci globalnej. W trakcie przetwarzania kolejnych trójkątów algorytm pomija piksele, które są już zamalowane, zob. także algorytm przedstawiony w podrozdziale 5.2. Przed zapisem do pamięci, nowo zamalowane piksele wykorzystywane są do aktualizacji wartości komponentów składowych funkcji celu. Po przetworzeniu wszystkich części modelu, każdy wątek przechowuje wartości komponentów składowych funkcji celu dla przetwarzanych przez niego pikseli. Aby wyznaczyć wartość funkcji celu, wykorzystywany jest algorytm redukcji równoległej, zob. rys. 6.6, po czym wynik zapisywany jest w pamięci globalnej.

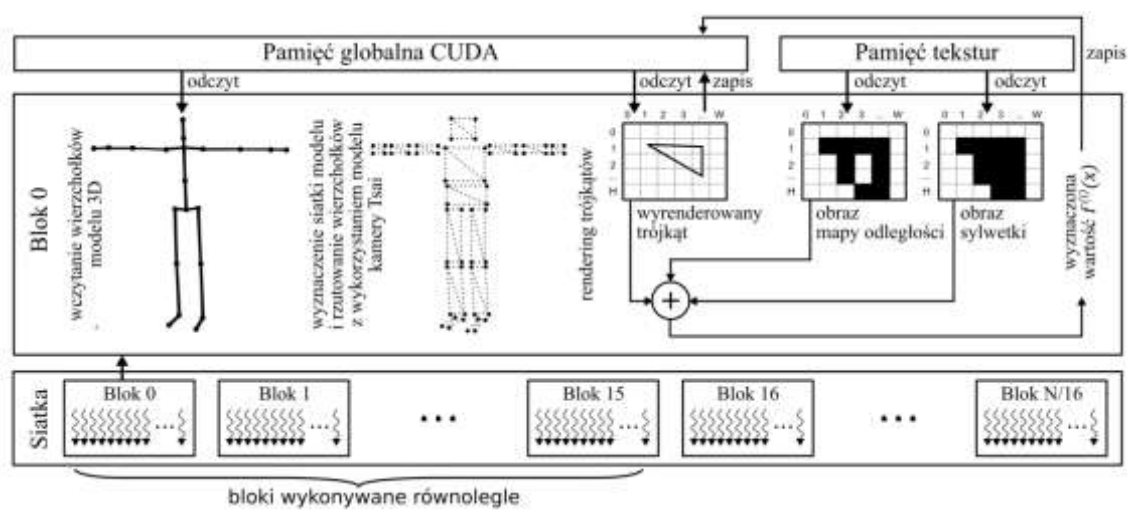

**Rys. 6.7. Wyznaczanie wartości funkcji celu CUDA**

Jak już wspomniano, liczba wątków $X_w$ wyznaczających wartość funkcji celu dla pojedynczej cząsteczki dobierana jest eksperymentalnie. Liczba wątków określa liczbę jednocześnie przetwarzanych pikseli, zob. rys. 5.24. Wspomniana liczba wątków nie powinna być większa od szerokości największej części modelu 3D. W trakcie wyznaczania wartości funkcji celu wykorzystywany jest trójwymiarowy blok i jednowymiarowa siatka wątków. Liczba wątków dla wymiaru $Y_w$ bloku jest równa liczbie kamer $C$, natomiast liczba wątków dla wymiaru $Z_w$ bloku wyznaczana jest zgodnie z równaniem:

$$Z_w = \begin{cases} \left\lceil \frac{240}{X_w Y_w} \right\rceil & \text{jeśli } N < 12 \\ \left\lceil \frac{1024}{X_w Y_w} \right\rceil & \text{jeśli } N \geq 12 \end{cases} . \qquad (6.9)$$

Liczba bloków określana jest na podstawie zależności:

$$X_b = \left\lceil \frac{N}{Z_w} \right\rceil \quad . \tag{6.10}$$

Dzięki wykorzystaniu zależności (6.9) i (6.10) możliwe jest efektywne wykorzystanie dostępnych zasobów układu graficznego.

Zaproponowana metoda renderingu programowego wykorzystuje uproszczony mechanizm przycięcia trójkątów do obszaru zainteresowania, który obejmuje śledzoną postać. Sposób wyznaczania obszaru zainteresowania przybliżono w podrozdziale 2.1. Jeśli renderowany trójkąt wykracza poza obszar zainteresowania tak, jak zilustrowano to na rys. 6.8a, wówczas realizowane jest przycięcie. W klasycznym podejściu przycięcie realizowane jest przez metodę zaprezentowaną na rys. 6.8c. Celem skrócenia procesu renderingu wykorzystywana jest metoda uproszczona, która została zaprezentowana w sposób schematyczny na rys. 6.8b.

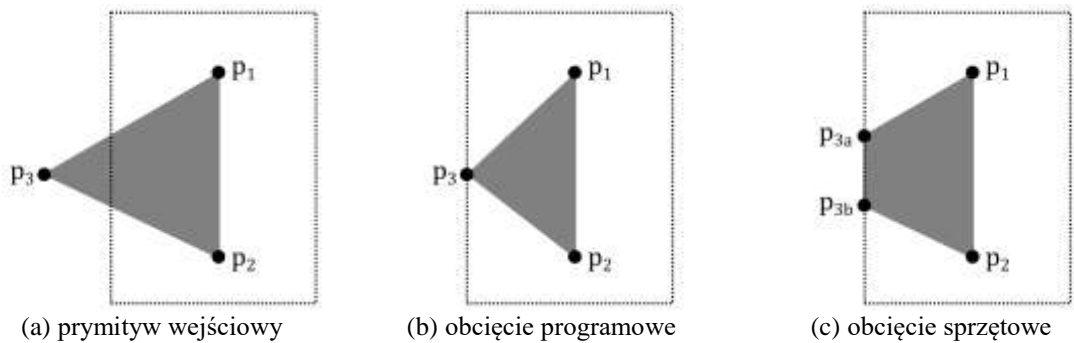

(a) prymityw wejściowy     (b) obcięcie programowe     (c) obcięcie sprzętowe

**Rys. 6.8. Obcinanie prymitywów graficznych**

## 6.4. Wyniki badań eksperymentalnych dla algorytmu CUDA

W tabeli 6.2 zamieszczono średnie czasy wyznaczania pozy dla pojedynczej klatki z sekwencji P1S. Omawiane czasy uzyskano dla algorytmu PSO składającego się z 100, 300, 1000 cząsteczek oraz realizującego 10 iteracji. Dla porównania zamieszczono także czasy uzyskane w oparciu o algorytm CPU [140]. Omawiane czasy obliczeń uzyskano na komputerze stacjonarnym wyposażonym w sześciordzeniowy procesor Intel Xeon X5690 3,46 GHz, 16 GB pamięci RAM i dwie karty graficzne Nvidia GTX 590. Karta graficzna Nvidia GTX 590 składa się z dwóch układów graficznych i każdy układ zawiera osiem wieloprocesorów strumieniowych z 64 rdzeniami na każdym wieloprocesorze. W drugiej konfiguracji komputer PC został wyposażony w kartę Nvidia GTX 780Ti. Karta graficzna Nvidia GTX 780Ti składa się z 12 wieloprocesorów strumieniowych z 240 rdzeniami na każdy wieloprocesor. Oznacza to, że w pierwszej konfiguracji mamy do dyspozycji 2048 rdzeni CUDA, natomiast w drugiej konfiguracji mamy do dyspozycji 2880 rdzeni CUDA. Dostępna moc obliczeniowa GPU w pierwszej konfiguracji wynosi 4976 GFLOPS, natomiast moc obliczeniowa GPU dostępna w drugiej konfiguracji wynosi 5045 GFLOPS. Na CPU obliczenia były realizowane z wykorzystaniem czterech rdzeni, natomiast w obliczeniach na GPU wykorzystywane były jednostki obliczeniowe na obydwu kartach. Jak można zauważyć w tabeli 6.2, dzięki wykorzystaniu GPU do śledzenia ruchu uzyskano znaczące skrócenie średniego czasu wyznaczania pozy w pojedynczej klatce. Prezentowane wyniki uzyskano dla 10 uruchomień algorytmu dla rożnych wartości początkowych. Dla PSO składającego się ze 100 cząsteczek i wykonującego 10 iteracji przyspieszenie jest nieco mniejsze niż 4, zaś dla konfiguracji składającej się z 1000 cząsteczek i wykonującego 10 iteracji przyspieszenie jest znacząco lepsze – przekracza 12. Uzyskanie większego przyspieszenia dla 1000 cząsteczek w algorytmie PF w porównaniu ze 100 cząsteczkami w algorytmie PSO wynika z lepszego wykorzystania zasobów kart graficznych GTX 590. W dalszej części pracy wykorzystywano kartę GTX 780Ti, ponieważ realizacja algorytmów śledzących na wielu układach wymaga dłuższego czasu komunikacji i synchronizacji.

**Tabela 6.2. Średni czas wyznaczania pozy oraz przyspieszenia obliczeń dla pojedynczej klatki z sekwencji P1S w oparciu o algorytm PSO**

| Liczba cząsteczek | Liczba iteracji | CPU [ms] | 2xGTX 590 [ms] | CPU / 2xGTX 590 | GTX 780Ti [ms] | CPU / GTX 780Ti |
|---|---|---|---|---|---|---|
| 100 | 10 | 170.8±18.6 | 45.4±11.5 | 3.8 | 62.0±5.2 | 2.8 |
| 300 | 10 | 443.4±54.8 | 61.7±6.9 | 7.2 | 67.5±6.7 | 6.6 |
| 1000 | 10 | 1386.2±158.1 | 110.8±10.2 | 12.5 | 81.7±5.3 | 16.7 |

Obliczenia wielowątkowe na CPU realizowane były ze wsparciem API OpenMP. Dzięki zdekomponowaniu obliczeń w ten sposób, że na każdym rdzeniu realizowane były obliczenia dla każdej kamery, uniknięto zbędnej komunikacji i synchronizacji między wątkami. Mając na względzie to, że karta graficzna Nvidia GTX 590 zawiera dwa układy GPU, obliczenia realizowane były na czterech układach GPU. Celem maksymalnego wykorzystania zasobów sprzętowych bloki ruch modelu, rendering i wyznaczenie komponentów z rys. 6.3 realizowane były na wszystkich dostępnych układach, zaś każdy układ realizował obliczenia dla części cząsteczek. Operacje składające się na pozostałe bloki z rys. 6.3 realizowane były na jednym układzie. Wspomniany zabieg możliwy był dzięki temu, że omawiane bloki nie wymagają znaczących nakładów obliczeniowych. Innym czynnikiem przemawiającym za taką dekompozycją zadania jest to, że operacje realizowane są wspomnianych blokach są prostymi operacjami.

W tabeli 6.3 zamieszczono czasy i przyspieszenia dla wyznaczania pozy w jednej klatce dla dwóch sekwencji P1S i P1D. Dla porównania zamieszczono także czasy uzyskane z wykorzystaniem algorytmu PF. Jak można zauważyć, uzyskane przyspieszenia obliczeń dla obydwu sekwencji są podobne. Z kolei czasy wyznaczania pozy postaci ludzkiej przez PSO są dłuższe w porównaniu z odpowiadającymi im czasami PF. Jest to spowodowane tym, że dla algorytmu PSO wymagane są dodatkowe synchronizacje po zakończeniu każdej iteracji. Z kolei czas wyznaczania konfiguracji postaci ludzkiej przez algorytm PSO dla danej liczby wywołań funkcji celu umożliwia uzyskanie lepszej dokładności śledzenia. Przykładowo, konfiguracji algorytmu PSO złożonego ze 100 cząsteczek i realizującego 10 iteracji odpowiada algorytm PF złożony z 1000 cząsteczek. Jak można zauważyć w tabeli 6.3, czas śledzenia pojedynczej klatki przez algorytm PF jest krótszy w porównaniu do algorytmu PSO. Wyniki ilustrujące dokładność śledzenia uzyskaną w oparciu o algorytmy PF i PSO zamieszczono w końcowej części niniejszego podrozdziału.

**Tabela 6.3. Przyspieszenie obliczeń dla estymacji pozy modelu 3D z wykorzystaniem CUDA na układzie GPU Nvidia GTX 590**

| Algorytm | Liczba cząsteczek | Sekwencja P1S | | | Sekwencja P1D | | |
| --- | --- | --- | --- | --- | --- | --- | --- |
| | | CPU [ms] | GTX 590 [ms] | CPU / GTX 590 | CPU [ms] | GTX 590 [ms] | CPU / GTX 590 |
| PF | 1000 | 143.1 | 14.5 | 9.9 | 141.6 | 15.2 | 9.3 |
| | 2000 | 264.6 | 26.7 | 9.9 | 263.0 | 26.3 | 10.0 |
| | 3000 | 387.1 | 40.0 | 9.8 | 388.5 | 39.6 | 9.8 |
| | 4000 | 515.8 | 52.2 | 9.9 | 517.7 | 54.5 | 9.5 |
| PSO (10 iteracji) | 100 | 138.7 | 30.7 | 4.5 | 139.0 | 30.5 | 4.6 |
| | 200 | 253.3 | 41.5 | 6.1 | 254.2 | 42.3 | 6.0 |
| | 300 | 368.0 | 58.5 | 6.3 | 370.0 | 58.8 | 6.3 |
| | 400 | 484.0 | 70.8 | 6.8 | 482.7 | 72.1 | 6.7 |

W tabeli 6.4 zamieszczono średni błąd śledzenia sekwencji P1S z wykorzystaniem algorytmów PSO i PF. Jak można zauważyć, dla czterokrotnie większej liczby wywołań funkcji celu algorytm PF zwraca większe błędy w porównaniu z algorytmem PSO. Mając na względzie wyniki przedstawione w tabeli 6.3, zauważyć można, że w dwukrotnie krótszym czasie, algorytm PF nie uzyska lepszej dokładności śledzenia w porównaniu do algorytmu PSO.

**Tabela 6.4. Średni błąd śledzenia sekwencji P1S w oparciu o algorytm PF i PSO**

| Algorytm | Średni błąd śledzenia [mm] ± odchylenie standardowe [mm] | | | |
| --- | --- | --- | --- | --- |
| | 960 wywołań funkcji celu | 2040 wywołań funkcji celu | 3040 wywołań funkcji celu | 5120 wywołań funkcji celu |
| PF | 112.4 ±42.2 | 93.1 ±37.9 | 90.1±36.7 | 90.7 ±37.3 |
| PSO (10 iteracji) | 48.7 ±13.0 | 43.6 ±11.4 | 43.3 ±12.0 | 40.0 ±10.7 |

Na rys. 6.9 zamieszczono wykresy ilustrujące dokładność śledzenia dla 5, 10, 15 i 20 iteracji PSO w zależności od liczby cząsteczek. Jak można zauważyć, dla PSO wykonującego pięć iteracji średni błąd śledzenia nie spada znacząco wraz ze wzrostem liczby cząsteczek. Spowodowane jest to tym, że pięć iteracji nie gwarantuje zbieżności algorytmu PSO. Dla pozostałych konfiguracji PSO następuje spadek błędu wraz ze wzrostem liczby cząsteczek. Jak można zauważyć na wykresach zamieszczonych na rys 6.9, najmniejszy błąd wynosi około 50 mm. Wraz ze wzrostem liczby cząsteczek dla algorytmu PSO wykonującego 10 iteracji następuje dość wolny spadek błędu śledzenia i powolne zbliżanie się do granicy 50 mm. Z kolei algorytmy realizujące 15 i 20 iteracji znacznie szybciej zmniejszają błąd śledzenia wraz ze wzrostem liczby cząsteczek.

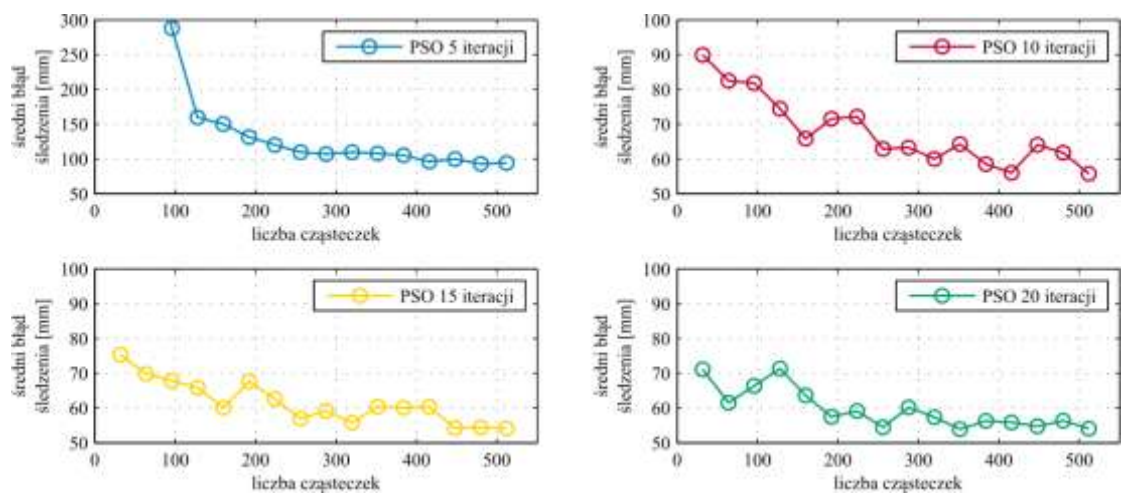

**Rys. 6.9. Dokładność śledzenia ruchu 3D w zależności od liczby cząsteczek**

Na rys. 6.10 zamieszczono wykresy ilustrujące zależność liczby przetwarzanych klatek na sekundę od liczby cząsteczek. Jak można zauważyć, dla PSO wykonującego pięć iteracji, liczba przetwarzanych klatek na sekundę wynosi 15 i nie zależy znacząco od liczby cząsteczek. Jednak jak już wspomniano, taka liczba iteracji nie gwarantuje uzyskania dokładności śledzenia zbliżonej do 50 mm. Dla algorytmu PSO realizującego 10, 15 i 20 iteracji liczba przetwarzanych klatek na sekundę maleje wraz ze wzrostem liczby cząsteczek.

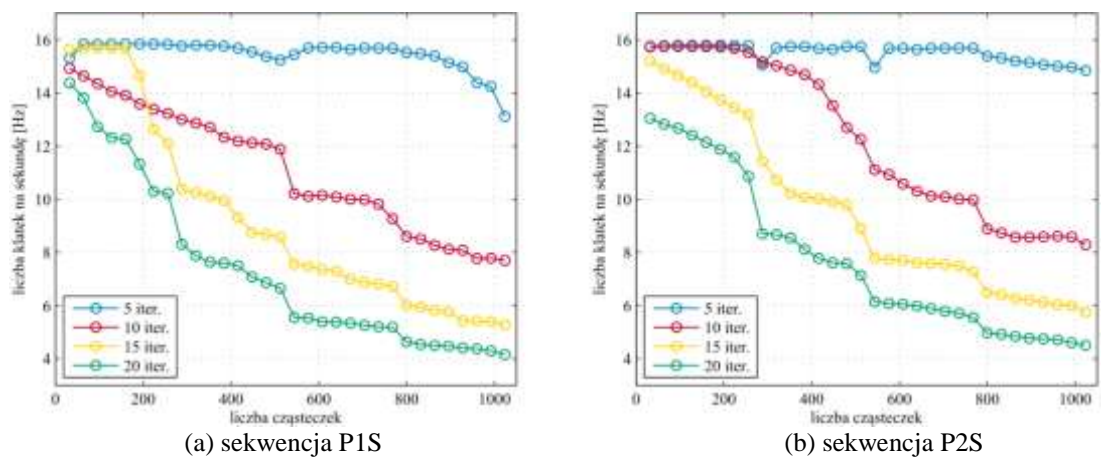

(a) sekwencja P1S  (b) sekwencja P2S

**Rys. 6.10. Liczba przetwarzanych klatek na sekundę w zależności od liczby cząsteczek**

Na rys. 6.11 zamieszczono wykres ilustrujący procentowy udział czasu niezbędnego do wyznaczenia wartości funkcji celu i pozostałych operacji w zależności od liczby cząsteczek. Jak można zauważyć na wspomnianym rysunku, dla małej liczby cząsteczek procentowy udział czasu wymaganego do wyznaczenia funkcji celu wynosi około 87%.

Wraz ze wzrostem liczby cząsteczek wspomniany procentowy udział rośnie i osiąga wartość 96% dla 1024 cząsteczek. Omawianą zależność uzyskano dla sekwencji Lee-Walk, dla 10-krotnego powtórzenia eksperymentów. Wspomniany wykres silnie przemawia za użyciem renderingu sprzętowego. W następnych podrozdziałach przedstawiono rozwiązania CUDA-OpenGL i OpenCL-OpenGL, które umożliwiły znaczące skrócenie czasu śledzenia ruchu dla pojedynczej klatki.

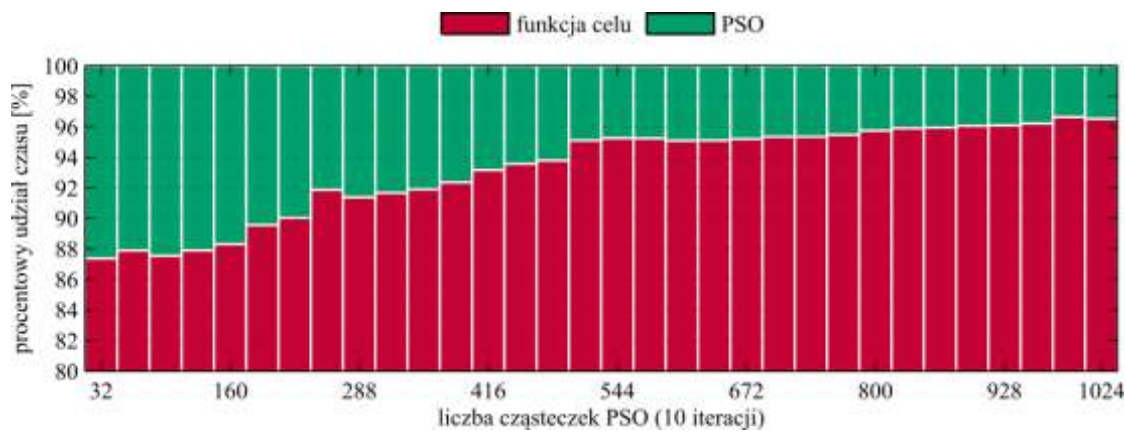

**Rys. 6.11. Procentowy udział czasu wyznaczania funkcji celu i algorytmu śledzącego dla CUDA dla estymacji pozy w pojedynczej klatce**

## 6.5. Śledzenie ruchu 3D z wykorzystaniem CUDA i OpenGL

Mając na względzie to, że zasadnicza część nakładów obliczeniowych w śledzeniu ruchu postaci ludzkiej dotyczy wyznaczania funkcji celu, w niniejszej pracy opracowano efektywne mechanizmy renderingu z wykorzystaniem CUDA i OpenGL. W niniejszym podrozdziale zestawiono wyniki badań eksperymentalnych, które uzyskano z wykorzystaniem mechanizmów współpracy CUDA-OpenGL zaprezentowanych w podrozdziale 3.6. W trakcie badań wykorzystywano mechanizmy renderingu zaprezentowane w podrozdziale 4.4. Jak już wspomniano, w technologii CUDA-OpenGL rendering i wyznaczanie funkcji celu realizowane jest w dwóch różnych technologiach. Z tego powodu wymagane jest zastosowanie dodatkowych mechanizmów synchronizacji i wymiany danych pomiędzy technologiami. W odróżnieniu od wyznaczania wartości funkcji celu w CUDA, format wyrenderowanych obrazów narzucany jest przez mechanizmy OpenGL. Z tego względu zachodzi konieczność zastosowania innej dekompozycji zadania, uwzględniającej ograniczenia i zasoby sprzętowe wykorzystywanego układu GPU.

W OpenGL wyrenderowane obrazy modeli zapisywane są w buforze ramki. Wykorzystywana karta graficzna pozwala na utworzenie bufora ramki o rozmiarach 16384x16384 pikseli, w którym każdy piksel składa się z czterech komponentów. W omawianym buforze składowane są jednocześnie obrazy dla pewnego podzbioru cząsteczek. Maksymalna liczba cząsteczek, dla których możliwe jest zapisanie wyren-derowanych obrazów wynika z wymiaru renderowanych obrazów, który jest równy rozmiarom obrazów z wydzielonymi sylwetkami. Wyznaczanie wartości komponentów składowych funkcji celu w CUDA realizowane jest przez wątki zgrupowane w trójwymiarowe bloki wątków oraz dwuwymiarową siatkę bloków. Na rys. 6.12 zilustrowano odwzorowanie podobrazów bufora ramki na dwuwymiarową siatkę bloków i trójwymiarowe bloki. Na omawianym rysunku zilustrowano mapowanie trójwymiarowego bloku wątków do dwóch podobrazów wyrenderowanych przez OpenGL, zob. element Blok. Z kolei mapowanie wątków w bloku do pikseli obrazu zilustrowano na rys. 6.13. Celem uproszczenia wizualizacji, na rys. 6.12 rozpatrzono funkcję celu wyznaczającą wartości dla 32 hipotetycznych konfiguracji modelu widocznych w czterech kamerach. Każdy z podobrazów zapisanych w buforze ramki przechowuje obrazy modelu wyrenderowane dla czterech hipotetycznych konfiguracji modelu 3D. Liczba bloków $X_b$ jest równa liczbie wykorzystanych kamer. Liczba bloków $Y_b$ i liczba wątków $X_w$ wynikają z liczby cząsteczek, zaś ich iloczyn jest równy $N/4$. Wątki, których $z_w = 0$, przetwarzają pierwszy podobraz, natomiast wątki, których $z_w = 1$, przetwarzają drugi podobraz.

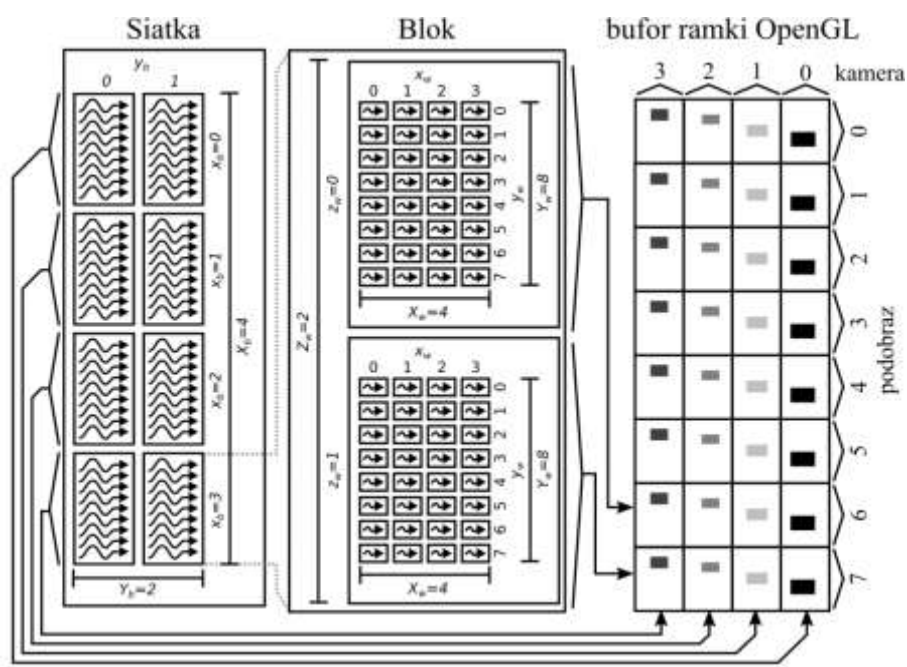

**Rys. 6.12. Mapowanie podobrazów bufora ramki na bloki wątków CUDA**

Na rys. 6.13 zaprezentowano wspomniane wcześniej mapowanie wątków bloku do pikseli podobrazu bufora ramki. Wątki odpowiedzialne są za wyznaczanie komponentów składowych funkcji celu dla czterech sylwetek zapisanych na przetwarzanym podobrazie. W odróżnieniu od systemu wykorzystującego rendering programowy CUDA, w omawianym systemie wykorzystywane jest dwuwymiarowe indeksowanie wątków operujących na wyrenderowanych obrazach. Oznacza to, że w danej chwili przetworzo-ny zostanie wycinek obrazu o wymiarach $X_w \times Y_w$ pikseli, gdzie $X_w$ i $Y_w$ określają wymiar bloku wątków. Na omawianym rysunku zilustrowano proces wyznaczania komponentów składowych funkcji celu w obszarze zainteresowania o rozmiarze $roi_x^{(c)} \times roi_y^{(c)}$. Komponenty funkcji celu w omawianym obszarze wyznaczane będą przez blok składający się z $X_w \times Y_w$ wątków. W sytuacji, gdy wymiar bloku wątków nie jest wielokrotnością wysokości lub szerokości obszaru zainteresowania, część wątków będzie nieaktywna przy przetwarzaniu pikseli granicznych obszaru zainteresowania. Zarówno przetwarzany podobraz, jak i obraz sylwetki przechowywane są w pamięci tekstury. Wynika to z faktu, że w odróżnianiu od metody wyznaczania wartości funkcji celu z wykorzystaniem renderingu programowego CUDA, obraz sylwetki jest wyrenderowany przez OpenGL i nie wymaga zmian.

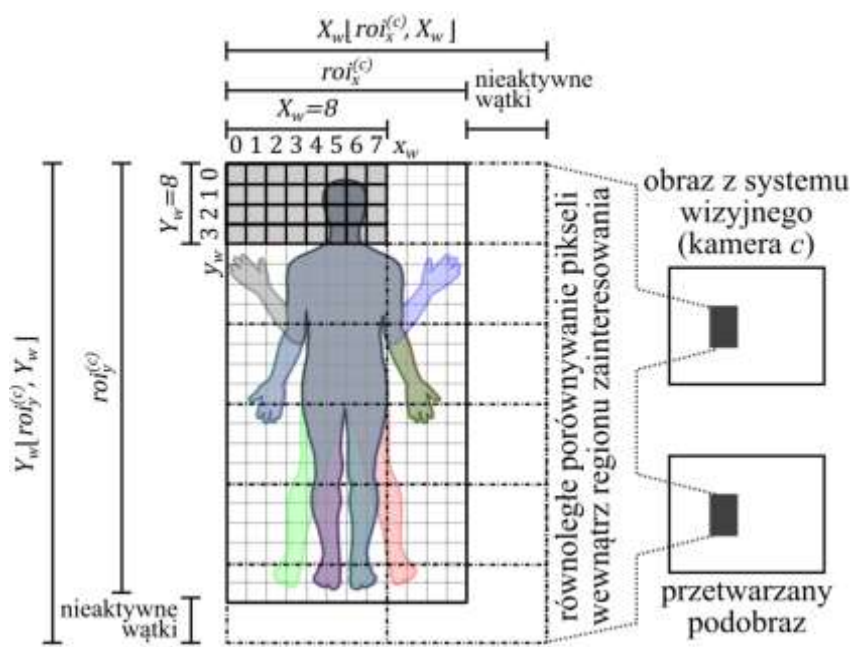

**Rys. 6.13**. **Wielowątkowe wyznaczanie komponentów składowych funkcji celu**

Jak już wspomniano, każdy kanał piksela bufora ramki przechowuje piksele obrazu kodowane na ośmiu bitach. Ze względu na to, że w CUDA możliwy jest jedynie odczyt 32-bitowych wartości z tekstury przechowującej obraz bufora ramki, wątki operują na 32-bitowych wartościach całkowitych. Celem wyeliminowania operacji na 8-bitowych kanałach, cztery kolejne piksele bufora ramki mapowane są do formatu, w którym 8-bitowe piksele wyrenderowanej sylwetki znajdują się obok siebie. Omawiany proces mapowania przedstawiony został na rys. 6.14. Celem przyspieszenia wspomnianego mapowania wykorzystano instrukcje *shuffle*, które są dostępne jedynie na nowszych układach CUDA [30,112]. Dzięki temu, że wspomniana funkcja operuje na rejestrach, czas mapowania jest znacząco krótszy w porównaniu do tradycyjnych funkcji, operujących na pamięci współdzielonej. Wspomniane instrukcje nie są dostępne w OpenCL i muszą być realizowane z wykorzystaniem pamięci wspólnej.

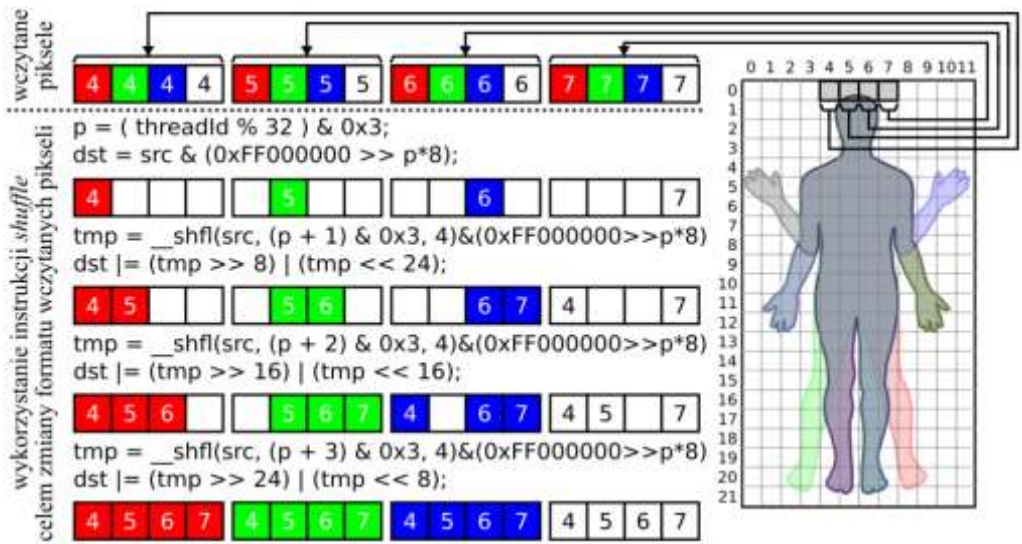

**Rys. 6.14. Wyznaczanie dopasowania dla poszczególnych pikseli**

Na rys. 6.15 zilustrowano, w jaki sposób pamięć przechowująca wyrenderowany podobraz jest widoczna dla wątków po operacji opisanej powyżej. Jak można zauważyć, dzięki operacji mapowania, wątki operują wyłącznie na pikselach należących do obrazu jednego modelu. Dzięki temu wyznaczenie komponentów składowych dla danego obrazu modelu wymaga mniejszej liczby operacji na pamięci współdzielonej. Wspomniana operacja wyznaczania komponentów składowych funkcji celu może być zrealizowana przy wykorzystaniu algorytmu redukcji równoległej.

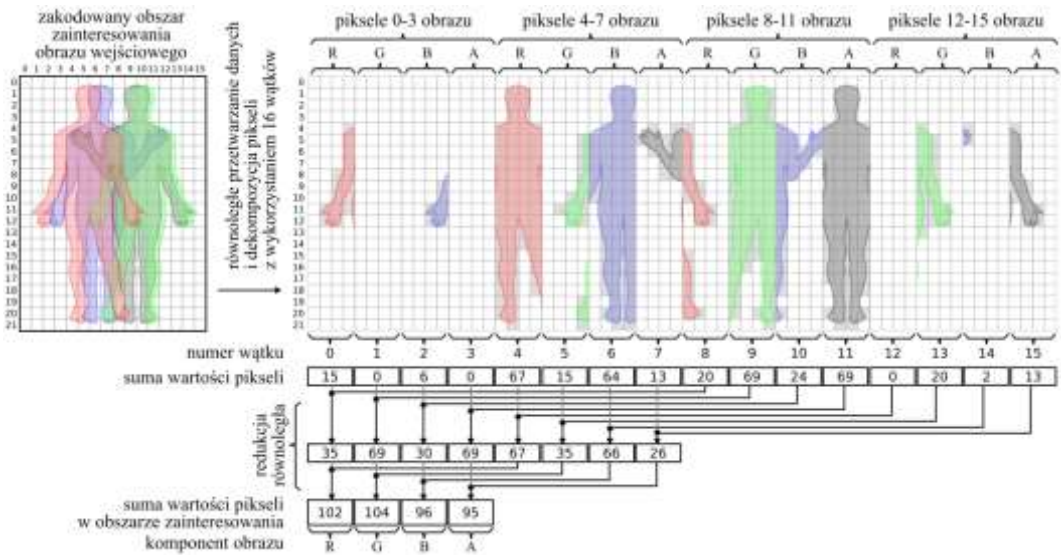

**Rys. 6.15. Dekompozycja zadannia dla algorytmu OpenCL-CUDA**

## 6.6. Wyniki badań eksperymentalnych dla algorytmu CUDA-OpenGL

W tabeli 6.5 zamieszczono średnie czasy estymacji pozy i średni błąd śledzenia dla pojedynczej klatki dla sekwencji LeeWalk dla przebadanych modeli 3D. Przebadano model płaski, model siatkowy, w którym główne części ciała reprezentowane są przez ostrosłupy ścięte o podstawie kwadratu oraz model siatkowy, w którym główne części ciała reprezentowane są przez ostrosłupy ścięte o podstawie elipsy. Jak można zauważyć, najlepsze wyniki uzyskano dla modelu płaskiego, w którym główne części ciała reprezentowane są przez trapezy parametryzowane położeniem czterech wierzchołków. Wspomniany model umożliwia uzyskanie najlepszej dokładności śledzenia oraz najkrótszego czasu śledzenia. Co istotne, wraz ze wzrostem złożoności modelu nie uzyskano znaczącego zmniejszenia błędu śledzenia. Omawiane wyniki uzyskano dla algorytmu PSO wykonującego 10 iteracji i zbudowanego z 96 i 304 cząsteczek. Mając na względzie powyższe wyniki, w dalszej części rozdziału wykorzystywano model płaski.

**Tabela 6.5. Średni czas estymacji pozy i średni błąd śledzenia sekwencji LeeWalk dla przebadanych modeli 3D**

| Reprezentacja modelu i liczba wierzchołków na część ciała | | Konfiguracja algorytmu PSO (10 iteracji) | | | |
|---|---|---|---|---|---|
| | | 96 cząsteczek | | 304 cząsteczki | |
| | | średni czas estymacji [ms] | średni błąd [mm] | średni czas estymacji [ms] | średni błąd [mm] |
| Model płaski: trapez | 4 | 75.7±5.9 | 47.9±11.7 | 90.9±7.2 | 44.1±11.8 |
| Model siatkowy: ostrosłup ścięty o podstawie kwadratu | 8 | 82.5±7.0 | 59.1±21.6 | 97.8±5.6 | 51.8±18.3 |
| Model siatkowy: ostrosłup ścięty o podstawie elipsy | 18 | 79.1±7.2 | 50.6±14.3 | 103.5±5.3 | 47.9±15.2 |
| | 34 | 86.7±7.0 | 49.2±14.9 | 111.4±4.9 | 47.0±14.8 |
| | 66 | 90.7±6.0 | 53.0±15.6 | 128.8±5.0 | 46.6±13.4 |
| | 130 | 104.2±6.2 | 51.7±16.1 | 158.3±5.3 | 45.6±13.1 |
| | 258 | 131.6±7.6 | 50.3±15.2 | 237.4±5.8 | 46.7±13.3 |
| | 514 | 170.5±6.1 | 52.4±17.2 | 389.6±5.9 | 46.7±14.3 |

W tabeli 6.6 zaprezentowano czasy obliczeń, które uzyskano na karcie Nvidia 780Ti dla algorytmu śledzącego CUDA-OpenGL. Jak można zauważyć, dla algorytmu PSO składającego się z 400 cząsteczek przyspieszenie obliczeń algorytmu CUDA-OpenGL w stosunku do algorytmu wykonującego się na CPU jest bliskie 13. Dla PSO złożonego z 100 cząsteczek przyspieszenie to również jest znaczące i wynosi 5. Wartości zaprezentowane w tabeli 6.3 wskazują na to, że czasy obliczeń na układach GTX590 dla algorytmu CUDA są około dwukrotnie krótsze niż na układzie GTX780Ti. Powodem tego jest fakt, że algorytm CUDA realizuje większość operacji na pamięci globalnej. Ponieważ w obliczeniach realizowanych na układach GTX590 wykorzystywane są dwie karty, a co za tym idzie dwa niezależne układy sterujące dostępem do pamięci, uzyskiwana przepustowość jest dwukrotnie większa. Ze względu na brak możliwości renderingu OpenGL na obydwu kartach graficznych GTX590 jednocześnie, algorytm CUDA-OpenGL zaimplementowano jedynie na karcie GTX780Ti.

**Tabela 6.6. Przyspieszenie obliczeń dla estymacji konfiguracji modelu z wykorzystaniem CUDA-OpenGL oraz układu GTX 780Ti.**

| Algorytm | Liczba cząste-czek | Sekwencja P1S | | | Sekwencja P2S | | |
|---|---|---|---|---|---|---|---|
| | | CPU [ms] | GPU [ms] | CPU/GPU | CPU [ms] | GPU [ms] | CPU/GPU |
| PSO CUDA (10 iteracji) | 100 | 139.0 | 110.8 | 1.2 | 143.8 | 115.3 | 1.2 |
| | 200 | 254.2 | 115.6 | 2.2 | 273.5 | 123.1 | 2.2 |
| | 300 | 370.0 | 120.6 | 3.1 | 388.2 | 136.0 | 2.9 |
| | 400 | 482.7 | 138.0 | 3.5 | 495.3 | 147.7 | 3.6 |
| PSO CUDA-OpenGL (10 iteracji) | 100 | 138.7 | 27.9 | 5.0 | 143.3 | 29.1 | 4.9 |
| | 200 | 253.3 | 29.0 | 8.7 | 271.8 | 32.7 | 8.3 |
| | 300 | 368.0 | 35.2 | 10.5 | 373.1 | 39.3 | 9.5 |
| | 400 | 484.0 | 37.6 | 12.9 | 502.9 | 42.2 | 11.9 |

Dzięki znaczącym przyspieszeniom obliczeń, w szczególności dla 400 cząsteczek, możliwe było wyznaczanie pozy modelu 3D z częstotliwością 27 klatek na sekundę przy zadowalającej dokładności. Wyniki ilustrujące dokładność śledzenia zaprezentowano w dalszej części rozdziału.

Na rys.6.16 zaprezentowano średnie czasy niezbędne do wyznaczenia wartości funkcji celu w pojedynczej klatce sekwencji LeeWalk. Śledzenie realizowano w oparciu o algorytm PSO zbudowany z 512 cząsteczek i wykonujący 10 iteracji. Na omawianym rysunku zamieszczono wykresy dla różnych wymiarów bloków funkcji wyznaczającej komponenty składowe funkcji celu. Jak można zauważyć czas wyznaczania wartości funkcji celu zależy nie tylko od przetwarzanego obrazu, ale także od wymiarów uruchamianego bloku. Najlepsze wyniki uzyskano dla konfiguracji $X_w = Y_w = 32$, $Z_w = 1$. Warto wspomnieć, że w omawianej konfiguracji przetwarzane są wycinki po-dobrazów o wymiarach 32x32 pikseli.

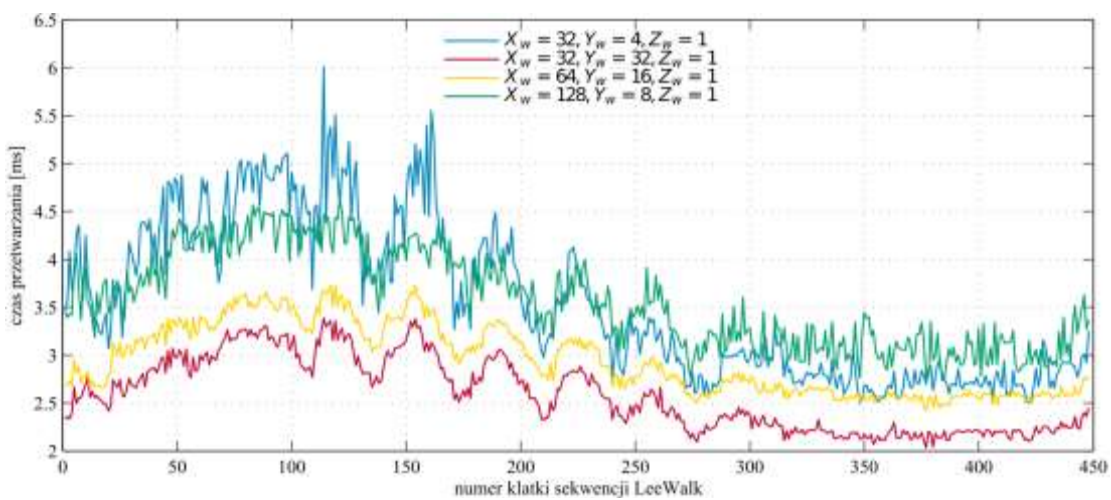

**Rys. 6.16. Czas wyznaczania wartości funkcji celu dla poszczególnych klatek sekwencji LeeWalk**

Na rys. 6.17 zaprezentowano procentowy udział czasu wyznaczania funkcji celu i algorytmu śledzącego dla implementacji algorytmu CUDA-OpenGL w trakcie rekonstrukcji pozy w pojedynczej klatce sekwencji LeeWalk. Prezentowane wyniki reprezentują średnie udziały z całej sekwencji w wyniku 10-krotnego powtórzenia eksperymentu. Jak można zaobserwować, dzięki wykorzystaniu renderingu sprzętowego procentowy udział czasu wyznaczania funkcji celu jest mniejszy w porównaniu do systemu śledzącego wykorzystującego rendering programowy, zob. rys. 6.11.

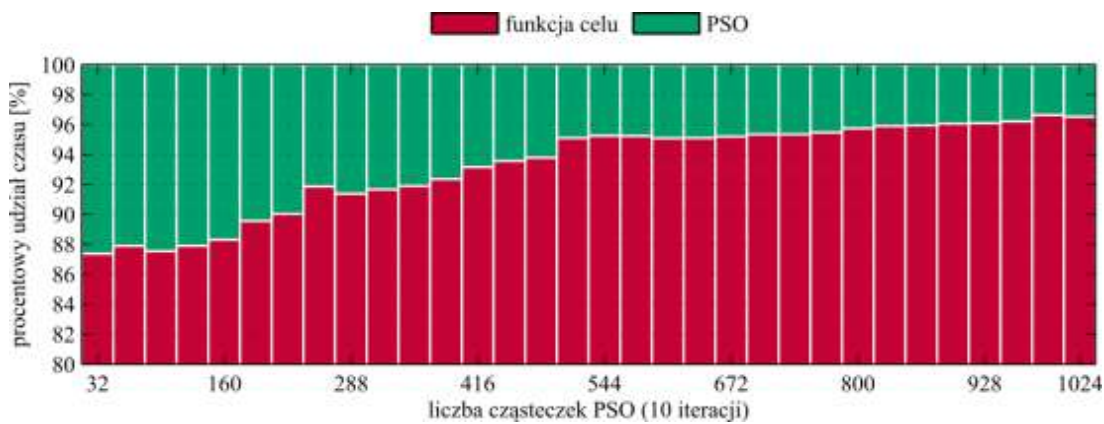

Rys. 6.17. Procentowy udział czasu wyznaczania funkcji celu i algorytmu śledzącego dla CUDA-OpenGL dla rekonstrukcji pozy w pojedynczej klatce

## 6.7. Ewaluacja metod na potrzeby śledzenia ruchu 3D w czasie rzeczywistym

W celu wyznaczenia wąskich gardeł (ang. *bottlenecks*), tzn. narzutów czasowych wprowadzonych przez poszczególne składniki funkcji celu, w różnych konfiguracjach algorytmu, zmierzono czasy operacji i transmisji dla funkcji celu opartej o rendering programowy i sprzętowy. Dla renderingu programowego przebadano implementacje CPU i implementacje CUDA. Natomiast dla renderingu sprzętowego opartego o OpenGL przebadano implementację CPU-OpenGL ze wsparciem instrukcji SSE, implementację CUDA-OpenGL oraz implementację OpenCL-OpenGL.

Celem zbadania narzutów czasowych wprowadzanych przez poszczególne bloki składające się na funkcję celu, zmierzono czasy obliczeń, posługując się precyzyjnymi zegarami na CPU i GPU [30,74,146]. W ramach badań mierzono czasy operacji wymaganych do zmiany pozy modelu, zob. blok ruch modelu na rys. 6.3, czasy transmisji danych i renderingu, zob. blok rendering na wspomnianym rysunku oraz czas wyznaczania wartości komponentów składowych i funkcji celu, zob. blok wyznaczanie komponentów i wyznaczanie funkcji celu na rys. 6.3.

W tabeli 6.7 zebrano czasy realizacji operacji związanych z wyznaczaniem funkcji celu dla renderingu programowego i sprzętowego. Do śledzenia wykorzystano algorytm PSO składający się z 10 iteracji i 96, 304 i 512 cząsteczek. W trakcie badań eksperymentalnych realizowano śledzenie dla całej sekwencji LeeWalk oraz sumowano czasy poszczególnych operacji. Uśrednione czasy poszczególnych operacji zaprezentowano we wspomnianej tabeli. Obliczenia CPU realizowano na wątkach OpenMP z wykorzystaniem mechanizmu dynamicznego zarządzania zadaniami [115]. Liczba wątków na CPU równa była liczbie kamer wykorzystanej sekwencji. Obliczenia na GPU realizowano na karcie Nvidia GTX 780Ti.

W trakcie badań eksperymentalnych zmierzono czasy: wyznaczania macierzy lokalnych i globalnych, transmisji wyznaczonych macierzy do pamięci OpenGL, renderingu w OpenGL, transmisji bufora ramki do pamięci układu, na którym realizowane jest wyznaczanie wartości funkcji celu, wyznaczania wartości komponentów składowych i funkcji celu oraz wymiany danych (transmisji wektora stanu i wartości funkcji celu). W oparciu o wspomniane czasy określono sumaryczny czas potrzebny do wyznaczenia wartości funkcji celu w pojedynczej iteracji. Oprócz omawianego czasu wyznaczono także sumaryczny czas wyznaczania wartości funkcji celu bez uwzględnienia czasów transmisji, renderingu i wymiany danych, tzn. czasu obliczeń.

Z wyników zamieszczonych w tabeli 6.7 wynikają następujące wnioski:

- w implementacjach CPU-OpenGL i CPUSSE-OpenGL czasy transmisji bufora ramki są wąskim gardłem rozwiązania, por. wiersz d z wierszem g,
- w implementacjach CUDA-OpenGL i OpenCL-OpenGL, czas renderingu jest dłuższy od czasu wyznaczania komponentów i wartości funkcji celu, por. wiersz c i wiersz e,
- dla implementacji CPU i CUDA wąskim gardłem jest wyznaczenie komponentów i wartości funkcji celu stanowi, por. wiersz e z wierszami a–d i f.

Jak można zauważyć, średni czas wyznaczania funkcji celu w pojedynczej iteracji, zob. wiersz g w tabeli 6.7, jest najkrótszy dla implementacji GPU opartej na renderingu sprzętowym, zob. kolumny CUDA-OpenGL i OpenCL-OpenGL w tabeli 6.7. Zadowalający czas osiągany jest także dla implementacji CUDA. Dla liczby cząsteczek równej 96 i 100 zadowalające są także czasy dla implementacji CPU, zob. wiersz g, kolumna CPU w tabeli 6.7.

W wierszu h zamieszczono czasy obliczeń, które są sumą czasów przedstawionych w wierszach a, c, e i f. Czasy te przygotowano z myślą o obliczeniach na układach, które posiadałyby wspólną magistralę pamięci, a co za tym idzie czasy transmisji danych do i z OpenGL byłyby znacząco mniejsze. W takim przypadku wykorzystanie implementacji CPU-OpenGL i CPUSSE-OpenGL byłoby zasadne. Co więcej, przygotowanie oprogramowania byłoby znacznie prostsze, gdyż nie byłoby wymagane przygotowanie oprogramowania w CUDA.

Mając na względzie czasy przetwarzania uzyskane w implementacjach CUDA-OpenGL i OpenCL-OpenGL, zauważyć można, że całkowity czas wyznaczania wartości funkcji celu w oparciu o OpenCL jest dłuższy w porównaniu do czasu śledzenia w oparciu o CUDA. Warto jednak podkreślić, że czas wyznaczania macierzy transformacji, zob. czasy zamieszczone w wierszu a tabeli 6.7 oraz czasy transmisji danych

z OpenGL, zob. czasy zamieszczone w wierszu d tabeli 6.7, są krótsze dla implementacji OpenCL-OpenGL. Czasy pozostałych operacji są dłuższe. Główną przyczyną jest to, że OpenCL jest narzędziem bardziej uniwersalnym i co za tym idzie nie wykorzystuje w pełni możliwości obliczeniowych układu GPU. Warto podkreślić, że przygotowany kod w OpenCL może być wykorzystany na innych platformach, w szczególności na procesorach DSP oraz FPGA.

**Tabela 6.7. Zestawienie czasów operacji związanych z wyznaczaniem funkcji celu dla renderingu programowego i sprzętowego**

| Rendering | | Programowy | | Sprzętowy | | | |
|---|---|---|---|---|---|---|---|
| Operacja | Liczba cząste-czek | CPU [ms] | CUDA [ms] | CPU-OpenGL [ms] | CPU SSE-OpenGL [ms] | CUDA-OpenGL [ms] | OpenCL-OpenGL [ms] |
| a) wyznaczenie macierzy | 96 | 0.21 | 0.85 | 0.21 | 0.17 | 0.83 | 0.61 |
| | 304 | 0.52 | 0.92 | 0.52 | 0.45 | 0.77 | 0.65 |
| | 512 | 0.91 | 0.93 | 0.91 | 0.76 | 0.82 | 0.60 |
| b) czas transmisji danych do OpenGL | 96 | - | - | 0.26 | 0.25 | 0.26 | 0.38 |
| | 304 | - | - | 0.31 | 0.30 | 0.28 | 0.36 |
| | 512 | - | - | 0.34 | 0.31 | 0.30 | 0.39 |
| c) rendering w OpenGL | 96 | - | - | 1.48 | 1.48 | 1.40 | 1.42 |
| | 304 | - | - | 2.81 | 2.74 | 2.71 | 2.68 |
| | 512 | - | - | 3.92 | 3.89 | 3.81 | 3.79 |
| d) czas transmisji danych z OpenGL | 96 | - | - | 63.37 | 62.88 | 0.40 | 0.25 |
| | 304 | - | - | 201.47 | 212.91 | 0.34 | 0.28 |
| | 512 | - | - | 309.53 | 318.27 | 0.38 | 0.29 |
| e) wyznaczenie wartości funkcji celu | 96 | 25.78 | 13.05 | 14.58 | 6.41 | 0.94 | 1.30 |
| | 304 | 91.65 | 21.54 | 49.79 | 18.66 | 1.66 | 2.22 |
| | 512 | 175.85 | 36.07 | 78.97 | 30.11 | 2.39 | 3.75 |
| f) narzut czasowy | 96 | 0.35 | 0.82 | 0.28 | 0.31 | 0.56 | 1.15 |
| | 304 | 0.38 | 0.90 | 0.27 | 0.30 | 0.56 | 1.19 |
| | 512 | 0.47 | 0.91 | 0.26 | 0.29 | 0.53 | 1.22 |
| g) sumaryczny czas (wiersze a-f) | 96 | 26.34 | 14.72 | 80.18 | 71.5 | 4.40 | 5.11 |
| | 304 | 92.55 | 23.36 | 255.17 | 235.36 | 6.30 | 7.38 |
| | 512 | 177.23 | 37.91 | 393.93 | 353.63 | 8.23 | 10.04 |
| h) sumaryczny czas (wiersze a i e) | 96 | 26.34 | 14.72 | 16.55 | 8.37 | 3.73 | 4.48 |
| | 304 | 92.55 | 23.36 | 53.39 | 22.15 | 5.7 | 6.74 |
| | 512 | 177.23 | 37.91 | 84.06 | 35.05 | 7.55 | 9.36 |

Na rys. 6.18 zamieszczono średnie czasy śledzenia pojedynczej klatki sekwencji LeeWalk w zależności od liczby cząsteczek. Prezentowane czasy uzyskano dla algorytmu PSO wykonującego 10 iteracji. Na wykresie zamieszczono czasy dla implementacji CUDA i implementacji CUDA-OpenGL. Jak można zauważyć, dla liczby cząsteczek mniejszej od 450 następuje niewielki wzrost czasu obliczeń dla zwiększającej się liczby cząsteczek w implementacji CUDA-OpenGL. Dla implementacji CUDA zależność czasu śledzenia od liczby cząsteczek może być dość dobrze aproksymowana zależnością liniową. Dla implementacji CUDA-OpenGL stosunkowo niewielki przyrost czasu dla liczby cząsteczek zwiększonej do 450 jest skutkiem efektywniejszego wykorzystania zasobów GPU. Dla liczby cząsteczek większej od 450 dla implementacji CUDA następuje liniowy przyrost czasu obliczeń, gdyż zaczyna brakować zasobów do realizacji obliczeń, przez co efektywność obliczeń równoległych jest mniejsza. Zauważyć można także znaczący wzrost czasu obliczeń dla implementacji CUDA-OpenGL dla liczby cząsteczek większej od 800. Jest to spowodowane tym, że obliczenia mogą być realizowane maksymalnie dla 800 cząsteczek jednocześnie, a co za tym idzie konieczne jest wykonanie zadania w kilku sesjach.

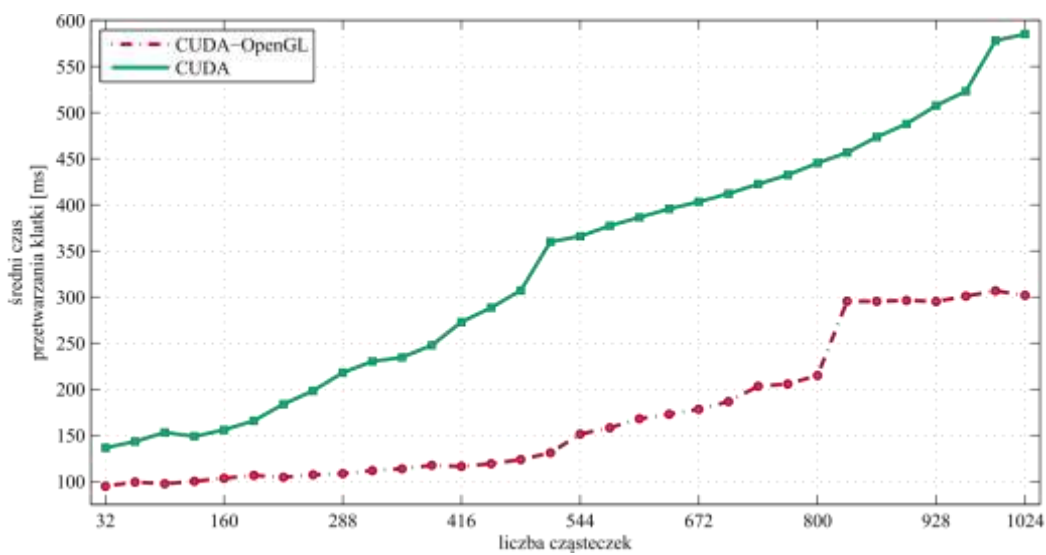

**Rys. 6.18. Porównanie średniego czasu śledzenia pozy w pojedynczej klatce LeeWalk z wykorzystaniem renderingu sprzętowego i programowego**

W tabeli 6.8 zestawiono średnie błędy śledzenia ruchu 3D w sekwencjach P1S, P1D, P2S, P2D i LeeWalk z wykorzystaniem algorytmu PSO i implementacji CUDA oraz CUDA-OpenGL. Jak można zaobserwować, pomimo wykorzystania tej samej postaci funkcji celu dokładność śledzenia w implementacji CUDA-OpenGL jest lepsza. Wynika to z faktu, że rendering sprzętowy umożliwia wyrenderowanie sylwetek z uwzględnieniem przesłonięć poszczególnych części ciała. Jak już wspomniano w podrozdziale 2.4, sekwencje P1S, P1D, P2S, P2D różnią się znacząco od sekwencji LeeWalk. Zasadnicze różnice dotyczą wielkości obrazów, wielkości sylwetki, a także częstotliwości z jaką sekwencje były nagrywane. Sekwencja LeeWalk nagrywana była z częstotliwością 60 Hz, podczas gdy pozostałe sekwencje nagrywane były z częstotliwością 25 Hz. Sekwencja LeeWalk nagrywana była za pomocą kamer monochromatycznych, podczas gdy pozostałe sekwencje nagrywane były kamerami kolorowymi. Mając na względzie wspomniane czynniki, a także czynniki wymienione w podrozdziale 2.1, zaobserwowano, że dokładność śledzenia ruchu w sekwencji LeeWalk jest znacząco lepsza niż w pozostałych. Na sekwencjach P1S, P1D, P2S, P2D możliwe jest śledzenie ruchu z dokładnością 50–60 mm.

**Tabela 6.8. Zestawienie średnich błędów śledzenia ruchu modelu 3D z wykorzystaniem implementacji CUDA i CUDA-OpenGL**

| Algorytm | liczba cząsteczek | Sekwencja P1S średni błąd [mm] | Sekwencja P1D średni błąd [mm] | Sekwencja P2S średni błąd [mm] | Sekwencja P2D średni błąd [mm] | Sekwencja LeeWalk średni błąd [mm] |
|---|---|---|---|---|---|---|
| **PSO (10 iteracji)** | 100 | 66.6±46.9 | 60.6±23.7 | 82.8±53.0 | 77.9±54.8 | 45.7±12.8 |
| **implementacja** | 300 | 60.6±40.1 | 58.3±22.3 | 69.0±46.4 | 66.8±43.5 | 41.4±11.3 |
| **CUDA** | 500 | 54.9±36.7 | 51.6±15.5 | 64.9±40.3 | 66.4±43.8 | 41.7±10.4 |
| **PSO (10 iteracji)** | 100 | 53.9±39.9 | 52.3±18.6 | 72.8±49.8 | 65.8±40.1 | 41.3±9.8 |
| **implementacja** | 300 | 43.1±23.7 | 44.5±11.2 | 60.3±37.1 | 57.7±29.2 | 38.8±8.2 |
| **CUDA-OpenGL** | 500 | 49.5±36.2 | 43.2±9.4 | 61.7±45.4 | 57.2±28.0 | 37.6±7.6 |

Na rys. 6.19 i rys. 6.20 i zamieszczono wykresy ilustrujące średni błąd śledzenia ruchu postaci ludzkiej w zależności od liczby cząsteczek. Na sekwencji P1S osoba poruszała się na wprost dwóch kamer, podczas gdy w sekwencji P1D osoba poruszała się po przekątnej sceny, zob. rys. 1 w pracy [133]. Omawiane wykresy sporządzono w oparciu o błędy śledzenia uzyskane przez algorytm PSO wykonujący zadaną liczbę iteracji w implementacji CUDA-OpenGL. Jak można zauważyć, dla algorytmu wykonującego 10 iteracji algorytm PSO złożony z 200 cząsteczek na sekwencjach Pxx osiąga dokładność 50 mm. Na sekwencji LeeWalk osiągnięto dokładność lepszą niż 40 mm. Warto wspomnieć, że w większości prac błąd śledzenia jest większy od 40 mm [11], a jedynie nieliczne prace uzyskały błąd mniejszy od 40 mm [44]. Dla większej liczby iteracji algorytm PSO osiąga wspomnianą dokładność dla mniejszej liczby cząsteczek. Jak można zaobserwować, dla algorytmu PSO wykonującego 20 iteracji wariancje są najmniejsze.

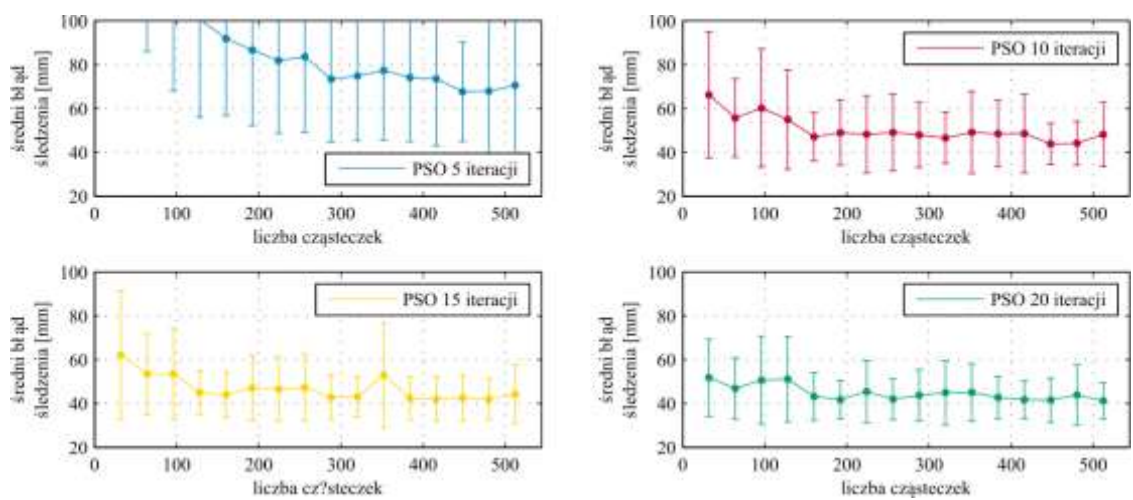

**Rys. 6.19. Średni błąd śledzenia ruchu postaci dla P1S**

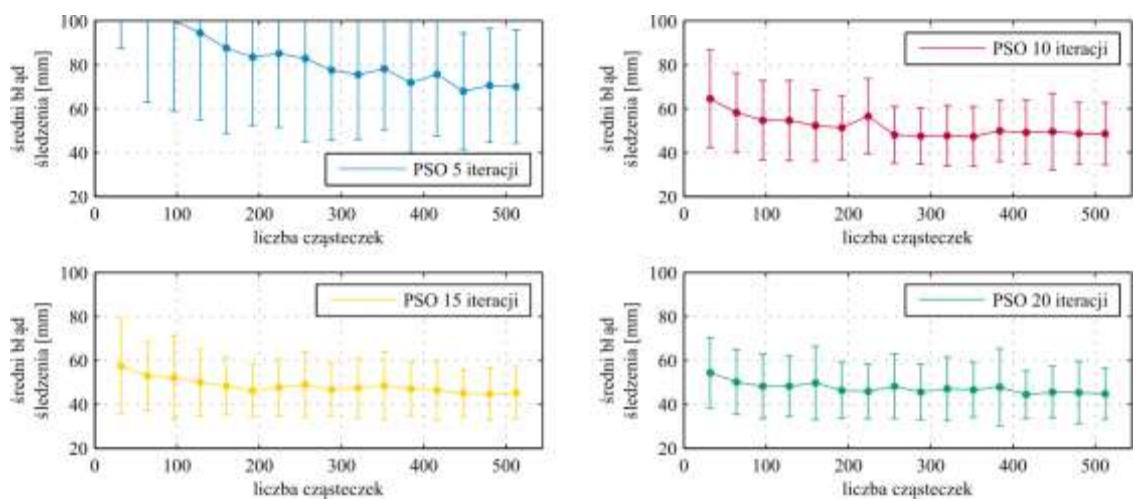

**Rys. 6.20. Średni błąd śledzenia ruchu postaci dla P1D**

Na rys. 6.21 zaprezentowano średni błąd śledzenia dla sekwencji LeeWalk w zależności od liczby cząsteczek. Śledzenie realizowane było w oparciu o algorytm PSO wykonujący zadaną liczbę iteracji w implementacji CUDA-OpenGL. Jak można zauważyć, dla 10 iteracji i dla 200 cząsteczek, błąd śledzenia jest zbliżony do 40 mm. Dla większej liczby iteracji wspomniany błąd osiągany jest dla 128 cząsteczek.

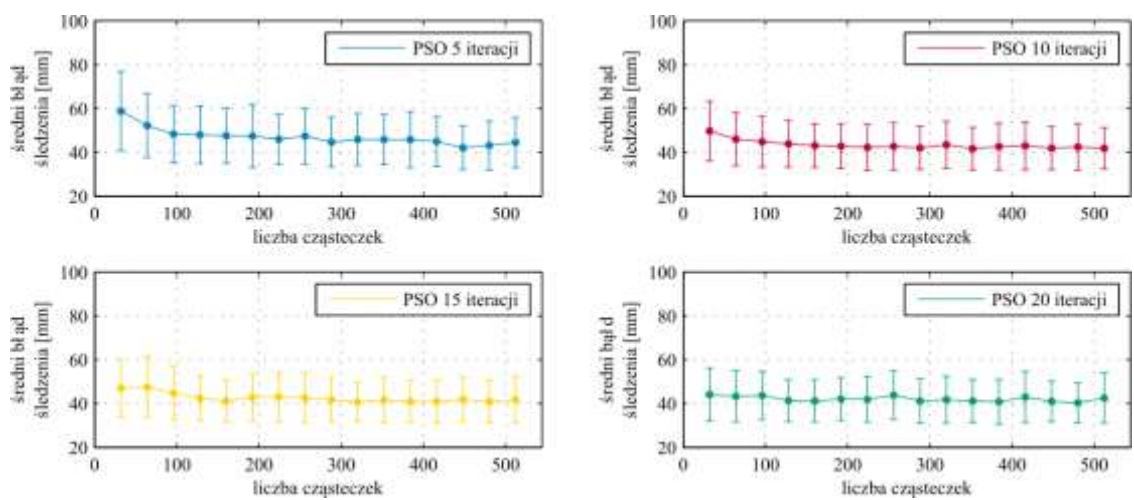

**Rys. 6.21. Średni błąd śledzenia ruchu dla sekwencji LeeWalk**

Omówione wyniki uzyskano na sekwencjach obrazów, które były nagrane ze stałą częstotliwością 25 Hz (Pxx) i 60 Hz (LeeWalk). Prezentowane dokładności śledzenia dotyczą zatem sytuacji, w której śledzenie byłoby realizowane na bieżąco. Zasadnicza część prac [11,44] prezentuje wyniki śledzenia na sekwencjach nagranych zawczasu, gdzie śledzenie jest realizowane bez pominięcia klatek, tzn. przy założeniu że system śledzący ruch jest zdolny do przetworzenia każdej pojawiającej się na bieżąco klatki. W przypadku pominięcia klatek, a co za tym idzie symulowaniu pracy systemu z mniejszą liczbą klatek na sekundę, znacząco wzrastają błędy śledzenia, gdyż zmiany pozy między przetwarzanymi klatkami są większe. Większe zmiany pozy wymagają z kolei przeszukania większych obszarów, a co za tym idzie większych wariancji podczas inicjalizacji cząsteczek. W celu minimalizacji błędu niezbędne jest zatem znalezienie kompromisu między liczbą cząsteczek i liczbą iteracji, od których zależy liczba przetwarzanych klatek na sekundę, a w konsekwencji błąd śledzenia. Określenie wspomnianych zależności nie jest zadaniem trywialnym ze względu na charakter realizowanych obliczeń równoległych CUDA-OpenGL, które, jak wykazano, umożliwiają uzyskanie najkrótszych czasów, zob. tabela 6.7 i najlepszych dokładności, zob. tabela 6.8. W następnym podrozdziale zaprezentowano rozwiązania, a także wyniki uzyskane w trakcie śledzenia w czasie rzeczywistym, tzn. w sytuacji, gdy obrazy pobierane są z częstotliwością wynikającą z częstotliwości pracy systemu śledzącego ruch 3D.

## 6.8. Śledzenie ruchu 3D w czasie rzeczywistym

W niniejszym podrozdziale omówiony zostanie system do śledzenia pozy postaci w czasie rzeczywistym. Zaproponowany system służy do symulacji rzeczywistego systemu zbudowanego z zadanej liczby kamer, pobierającego obrazy z częstotliwością wynikającą z uzyskiwanego czasu śledzenia dla badanej liczby cząsteczek i liczby iteracji algorytmu PSO. Jak już wspomniano w poprzednim podrozdziale, dokładność układu śledzącego ruch w czasie rzeczywistym zależy nie tylko od wielkości obrazów, wielkości osoby itp., ale także od liczby klatek obrazów pobieranych i przetwarzanych przez system w czasie jednej sekundy.

Opracowany system do śledzenia ruchu w czasie rzeczywistym składa się z modułów: akwizycji obrazów, przetwarzania obrazów oraz modułu śledzącego, które realizowane są w oddzielnych wątkach. Na rys. 6.22 zilustrowano schemat systemu śledzącego ruch w oparciu o obrazy pobierane z czterech kamer.

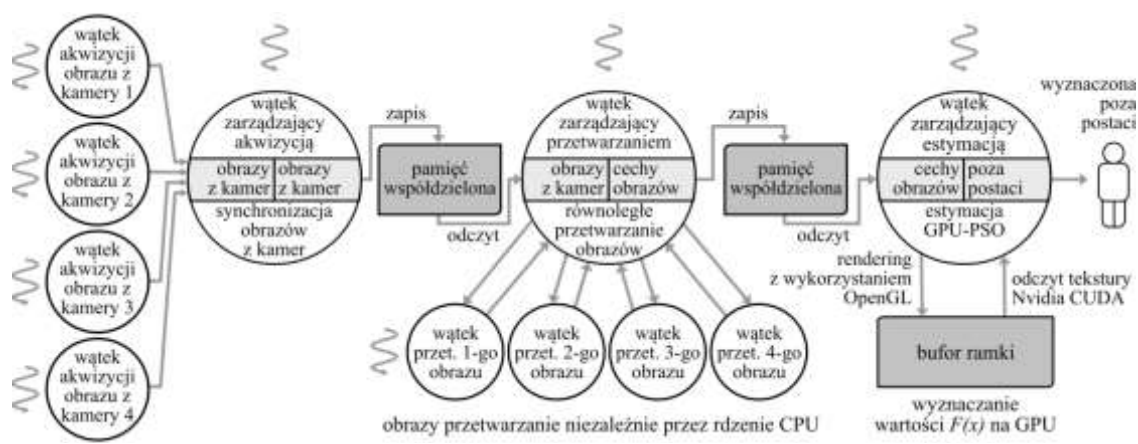

**Rys. 6.22. Schemat systemu do śledzenia ruchu 3D w czasie rzeczywistym w oparciu o CUDA-OpenGL**

W module przetwarzania obrazów i module śledzącym ruch 3D wykorzystano omówione wcześniej algorytmy. Moduły akwizycji i przetwarzania obrazów realizowane są na CPU i wykorzystują cztery wątki, z których każdy pozyskuje i przetwarza obrazy z jednej kamery. Komunikacja pomiędzy modułem akwizycji i przetwarzania, jak i pomiędzy modułem przetwarzania i śledzenia ruchu odbywa się dzięki współdzieleniu pamięci. Moduł śledzenia ruchu nadzoruje zarówno kontekst CUDA, jak i OpenGL. Kontekst CUDA nadzorowany jest dzięki CUDA Runtime API, natomiast kontekst OpenGL nadzorowany jest przez biblioteki GLEW i GLES [76]. Komunikacja pomiędzy CUDA i OpenGL, w szczególności dostęp do bufora ramki, realizowana jest w oparciu o metody zaprezentowane w podrozdziale 5.4.

Na rys. 6.23 przedstawiono błędy śledzenia ruchu postaci ludzkiej dla sekwencji P1S. Śledzenie zrealizowano w oparciu o algorytm PSO i rendering CUDA-OpenGL. Wyniki prezentowane na wykresach ukazują średni błąd śledzenia ruchu, który uzyskano dla 10-krotnego powtórzenia eksperymentu. Omawiane wykresy ilustrują błąd śledzenia ruchu offline, tzn. gdy śledzenie jest realizowane dla wszystkich klatek z danej sekwencji oraz błąd śledzenia ruchu w czasie rzeczywistym, tzn. gdy system na bieżąco pobiera i przetwarza klatki. Jak można zauważyć, system offline uzyskuje lepszą dokładność śledzenia ruchu 3D. Ze względu na niemożność przetworzenia wszystkich klatek w określonych limitach czasowych, system online pomija część klatek. Z powodu pominięcia części klatek, zmiany pozy postaci pomiędzy przetwarzanymi i sąsiednimi klatkami są większe, a w konsekwencji system ma większe trudności w określeniu aktualnej pozy 3D.

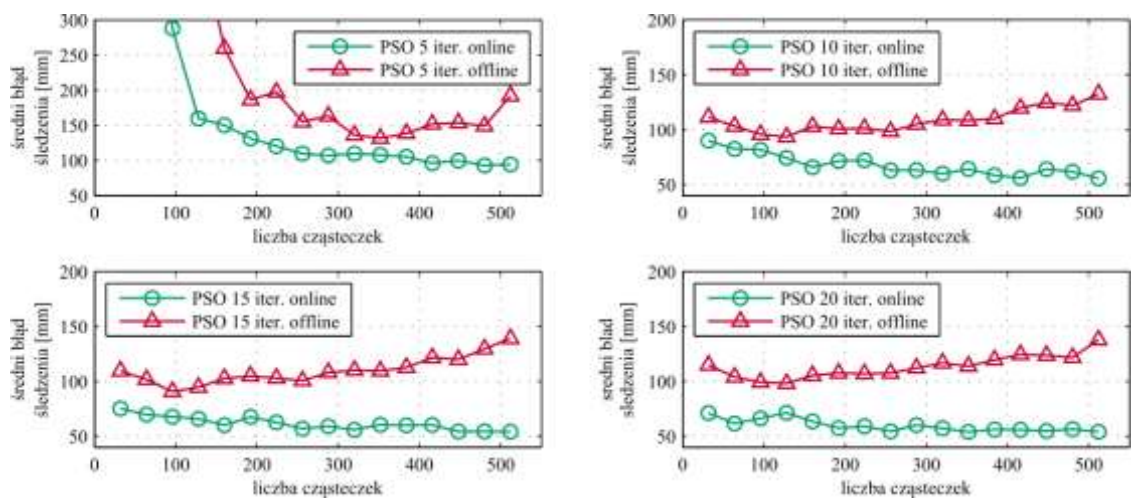

**Rys. 6.23**. Dokładność śledzenia postaci na sekwencji P1S w zależności od liczby cząsteczek

Na rys. 6.24 przedstawiono diagram czasowy ilustrujący działanie systemu do śledzenia ruchu 3D w czasie rzeczywistym dla konfiguracji PSO pozwalającej na osiągnie najmniejszego błędu śledzenia w czasie rzeczywistym. Na omawianym rysunku przedstawiono diagram czasu dla śledzenia ruchu na sekwencji P1S. Jak można zauważyć na omawianym rysunku, moduł akwizycji pracuje z częstotliwością 25 Hz. Dzięki dostępnej mocy obliczeniowej procesora możliwe jest przetwarzanie obrazów na bieżąco. Ze względu na to, że czas śledzenia ruchu jest większy od 40 ms, przetwarzane są jedynie klatki, które dostępne są tuż po zakończeniu śledzenia poprzedniej klatki. Diagram prezentuje śledzenie ruchu z wykorzystaniem algorytmu PSO zbudowanego z 15 iteracji i 200 cząsteczek. Wspomniane wartości zapewniają osiągnięcie najmniejszego błędu w trakcie śledzenia w czasie rzeczywistym. Dla wspomnianej konfiguracji PSO możliwe jest przetwarzanie 11 klatek na sekundę sekwencji P1S z błędem śledzenia ruchu nie przekraczającym 100 mm. Częstotliwość pracy systemu jest równa 11 klatkom na sekundę i wynika z czasu śledzenia na GPU. Czas opóźnienia, tzn. czas, po którym dostępny jest wynik (licząc od czasu pojawienia się klatki) jest równy sumie czasu akwizycji, ekstrakcji cech oraz śledzenia i wynosi 130 ms.

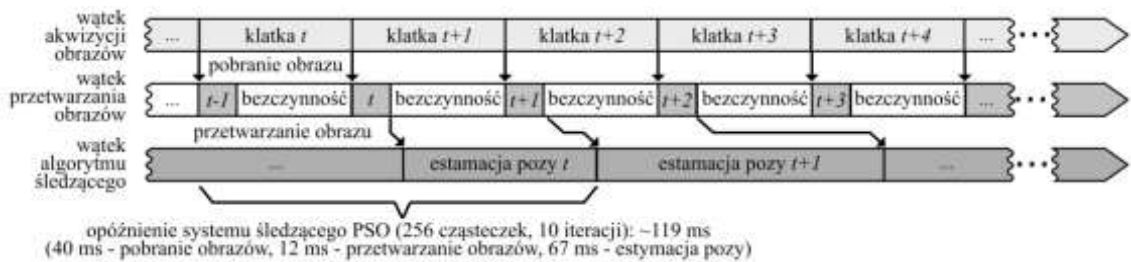

**Rys. 6.24. Diagram czasowy śledzenia ruchu 3D w czasie rzeczywistym dla sekwencji P1S**

Dzięki opracowanemu systemowi symulacyjnemu możliwe jest podanie parametrów pracy systemu śledzącego ruch 3D w czasie rzeczywistym, tzn. dokładności śledzenia postaci oraz liczby przetwarzanych klatek na sekundę.

## 6.9. Podsumowanie

W niniejszym rozdziale zaprezentowano sposób określania dokładności śledzenia ruchu 3D w czasie rzeczywistym. Przebadano eksperymentalnie funkcje celu pod kątem dokładności śledzenia. Przedstawiono także równoległy algorytm śledzenia ruchu postaci ludzkiej, a w szczególności dekompozycję wykorzystywanego algorytmu śledzącego, mając na względzie dostępne zasoby GPU. Zaprezentowano dokładności śledzenia ruchu 3D, które uzyskano w oparciu o algorytmy śledzące zaimplementowane w CUDA, CUDA-OpenGL i OpenCL-OpenGL. Dokonano także ewaluacji metod na potrzeby śledzenia ruchu w czasie rzeczywistym. Podano parametry systemu śledzącego ruch 3D w czasie rzeczywistym.

# Podsumowanie

W pracy opracowano i przebadano komputerowe algorytmy śledzenia ruchu 3D człowieka w czasie rzeczywistym. Zagadnienie to nie należy do łatwych, lecz jest często podejmowane przez zespoły badawcze ze względu na zapotrzebowanie na tanie i efektywne systemy do śledzenia ruchu 3D postaci ludzkiej. Istnieje znaczące zapotrzebowanie na takie systemy ze strony producentów gier, lekarzy i rehabilitantów, a także projektantów systemów wspomagających interakcję człowiek-komputer. W niniejszej pracy uwagę skupiono na bezmarkerowym śledzeniu ruchu 3D postaci ludzkiej w oparciu o system wielokamerowy. Jest to problematyka bardzo aktualna ze względu na wspomniane wyżej zastosowania. Ze względu na znaczące wymagania obliczeniowe istnieje zapotrzebowanie na efektywne obliczeniowo algorytmy umożliwiające śledzenie ruchu 3D w czasie rzeczywistym.

W pracy szczególną uwagę poświęcono opracowaniu i przebadaniu równoległych algorytmów do śledzenia ruchu w oparciu o model 3D. Badania realizowano pod kątem uzyskania zadowalających dokładności śledzenia ruchu 3D w czasie rzeczywistym. Mając na względzie to, że w podejściu opartym o model 3D zasadnicza część nakładów obliczeniowych dotyczy renderingu modelu w zadanej pozie, opracowano efektywne algorytmy renderingu modelu 3D na CPU i GPU. Opracowano i przebadano efektywne metody renderingu modelu 3D z wykorzystaniem metod sprzętowych i programowych. Śledzenie ruchu realizowano z wykorzystaniem algorytmów optymalizacji w oparciu o rój cząsteczek i algorytmy filtru cząsteczkowego. Zaproponowano i opracowano równoległe implementacje wspomnianych algorytmów. Zrealizowano badania eksperymentalne, w których przebadano i określono wąskie gardła obliczeń równoległych, a w szczególności określono dokładności śledzenia ruchu 3D. Podano parametry pracy systemu śledzącego ruch 3D w czasie rzeczywistym.

Do najważniejszych osiągnięć można zaliczyć:

- opracowanie i przebadanie równoległej implementacji algorytmu śledzenia ruchu 3D w oparciu o rój cząsteczek i algorytmu kondensacji stanu,
- zaproponowanie konfigurowalnego modelu 3D, wykorzystywanego do modelowania sylwetki i ruchu postaci ludzkiej,
- przygotowanie i przebadanie programowych metod renderingu modelu 3D,
- opracowanie dedykowanych potoków OpenGL do renderingu modelu 3D na potrzeby śledzenia ruchu,
- zaproponowanie i przebadanie kilku funkcji celu,
- opracowanie i przebadanie implementacji algorytmu śledzącego na CPU, CPUSSE-OpenGL, CUDA, CUDA-OpenGL, OpenCL-OpenGL,
- opracowanie i przebadanie systemu umożliwiającego śledzenie ruchu w czasie rzeczywistym.

Dalsze prace skupią się na rozwoju systemu śledzenia ruchu 3D postaci ludzkiej w czasie rzeczywistym. Planuje się opracowanie rozwiązań umożliwiających śledzenie w czasie rzeczywistym w oparciu o obrazy o większych rozdzielczościach. W tym kontekście planowane jest wykorzystanie modelu siatkowego, który na obrazach o większej rozdzielczości powinien umożliwić zmniejszenie błędów śledzenia. Celem zwiększenia częstotliwości śledzenia planuje się wykorzystanie kilku kart graficznych z serii Nvidia GTX Titan.

# Bibliografia

# Spis ilustracji